\documentclass{article}

\PassOptionsToPackage{numbers}{natbib}

 \usepackage[eandd, final]{neurips_2026}

\usepackage[utf8]{inputenc} 
\usepackage[T1]{fontenc}    
\usepackage{hyperref}       
\usepackage{url}            
\usepackage{booktabs}       
\usepackage{amsfonts}       
\usepackage{nicefrac}       
\usepackage{microtype}      
\usepackage{xcolor}         
\usepackage{graphicx}
\usepackage{multirow} 
\usepackage{amsmath} 
\usepackage{soul}
\usepackage{wrapfig}
\usepackage{makecell}
\usepackage[normalem]{ulem}
\usepackage{titletoc}

\definecolor{darkgreen}{HTML}{006400}
\definecolor{correcthighlight}{HTML}{34674B} 
\definecolor{hallhighlight}{HTML}{BB3450}
\definecolor{darkblue}{HTML}{B5C0C8}
\definecolor{lightblue}{HTML}{DDF0F7}

\newcommand{\figref}[1]{Fig.~\ref{#1}}
\newcommand{\tabref}[1]{Table~\ref{#1}}
\newcommand{\secref}[1]{\S~\ref{#1}}
\newcommand{\appendixref}[1]{\S ~\ref{#1}}

\newcommand{\nll}{\texttt{NLL}}
\newcommand{\ent}{\texttt{Ent}}
\newcommand{\ic}{\texttt{IC}}
\newcommand{\cl}{\texttt{CL}}
\newcommand{\glsim}{\texttt{GLSim}}
\newcommand{\svar}{\texttt{SVAR}}
\newcommand{\pas}{\texttt{PAS}}

\newcommand{\llavasm}{LLaVA-1.5-7B}
\newcommand{\llavalg}{LLaVA-1.5-13B}
\newcommand{\instructblip}{InstructBLIP-Vicuna-7B}
\newcommand{\qwen}{Qwen3VL-8B-Instruct}
\newcommand{\internvl}{InternVL2-8B}
\newcommand{\internvlnew}{InternVL3.5-8B}
\newcommand{\qwennew}{Qwen3.5-9B}
\newcommand{\gemma}{Gemma-4-E4B-it}

\newcommand{\thetitle}{Robustness of Internal Signals for Hallucination Detection in Vision-Language Models}

\title{\thetitle}

\author{
  Allison Chen \\
  allisonchen@princeton.edu
  \And
  William Yang \\
  williamyang@princeton.edu
  \And
  Salma Abdel Magid \\
  sabdelmagid@princeton.edu
  \And
  Jonathan Williams \\
  jw4199@princeton.edu
  \And
  Olga Russakovsky \\
  olgarus@princeton.edu
  \And
  Esin Tureci \\
  esin.tureci@princeton.edu
  \AND
  \\
  Department of Computer Science \\ Princeton University, Princeton, NJ 08544
  \\
}

\begin{document}

\maketitle
\begin{abstract}
    When given an image and a text prompt, vision-language models (VLMs) can produce seemingly sophisticated text responses.
    However, they are also prone to hallucinate objects in their output, which can have detrimental consequences in safety-critical applications.
    To detect these hallucinations, certain methods interestingly leverage internal signals from the VLM as it generates the text response; potentially, these methods use the same signals that led to a hallucination to detect it.
    We conduct an in-depth analysis to understand the robustness of these detection methods.
    First, we show that a naive baseline of a generated word's \emph{token position} in the VLM text response achieves competitive performance.
    Further, we reveal how some detection methods \emph{rely} on token position and use this heuristic as a shortcut.
    Second, we explore detection methods' sensitivity to language and vision modalities. 
    Our experiments suggest that strong detection performance is often driven by detecting language-based hallucinations, and when methods rely on token position, they tend to be less visually grounded and exhibit reduced robustness when hallucinations occur earlier in a text response.
    Lastly, we demonstrate that different detection methods can carry complementary signals; when ensembling methods together, adding weaker methods to an ensemble of higher performing ones still offers performance improvements.
    Altogether, while there may be useful internal VLM signals to detect hallucinations, our investigation reveals important limitations of the current methods that try to leverage them.
    \looseness=-1
\end{abstract}

\section{Introduction}
\label{sec:intro}
While vision-language models (VLMs) are increasingly used for a wide variety of visual understanding tasks \citep{driess_palme_2023, liu_mmbench_2025}, they are known to hallucinate,
generating plausible sounding text responses that are incorrect or inconsistent with the given image \citep{liu_survey_2024, chen_survey_2026}. 
These hallucinations are especially problematic when VLMs are deployed in high-stakes contexts (e.g., healthcare, law, or autonomous driving) \citep{nazi_large_2024, westermann_analyzing_2024, tian_drivevlm_nodate, nath_vilam3_2025}, spurring the need for solutions that can detect hallucinations. 

While there are many approaches for detecting hallucinations including using external evaluators \citep{kaul_throne_2024, ge_visual_2024, park_halloc_2025, tong_faithscan_2026} or sampling uncertainty estimates \citep{liao_visionamplified_2026, zhang_vluncertainty_2024, farquhar_detecting_2024},  a more computationally lightweight approach is to leverage signals from the VLM as it generates text \citep{park_glsim_2025, jiang_devils_2025, jiang_interpreting_2025, phukan_logit_2025, hoang-xuan_pas_2025}.
These methods may be preferred for real-time applications because unlike the others, they do not require additional memory or computation beyond the VLM's forward pass. 
However, since these methods appear to be using the same signals that lead to hallucinations in the first place, it is important to understand their robustness beyond what standard evaluations reveal. \looseness=-1

In this work, we aim to understand whether internal signal-based detection methods use reliable cues that enable robust generalization or simply depend on surface-level heuristics.
Our work goes one step beyond the standard evaluation paradigm of reporting metrics (e.g., AUROC or AUPRC) on large datasets \citep{fieback_metatoken_2025, park_glsim_2025, park_halloc_2025, fu_mme_nodate} to understand the underlying mechanisms of these methods.
To provide a comprehensive analysis, we study both ``white box'' methods that leverage internal signals \citep{park_glsim_2025, jiang_devils_2025, jiang_interpreting_2025, phukan_logit_2025}, such as representation embeddings or attention matrices, alongside ``black box'' methods that rely solely on output logits \citep{zhou_analyzing_2024, malinin_uncertainty_2021, chen_survey_2026}.
Specifically, our analyses center on the original setting the methods were designed for: detecting object hallucinations in image-captioning tasks. \looseness=-1

Concretely, our work primarily reveals that the token position of a word in the VLM-generated caption can play a surprising role in detection methods leveraging internal VLM signals.
We first introduce a \textbf{token position heuristic as a competitive hallucination detector} (\secref{sec:token_pos_baseline}) and show that \textbf{many white box methods \emph{rely} on token position for detection to varying degrees} (\secref{sec:remove_replace}).
Second, we explore how detection methods leverage different modalities (i.e., language and vision). 
We find that \textbf{white box methods perform well by detecting language-based hallucinations} (\secref{sec:partition}), but \textbf{highly token position-dependent methods can be independent of visual signals}, suggesting they may make predictions agnostic to the image (\secref{sec:visual_signals}).
We then demonstrate the brittleness of token position-dependent methods by showing a concrete setting where \textbf{token position-dependent methods are not reliable detectors} (\secref{sec:tp_case}) and performance drops to near random chance. 
Lastly, to encourage future work improving these methods, we use progressive ensembles to demonstrate that \textbf{black box detection methods can provide complementary signals to higher performing white box methods}, contributing a relative $7.6\%$ increase in an ensemble's AUPRC performance (\secref{sec:ensemble}).
This suggests that one possible avenue for future work to overcome token position biases may be to combine signals from multiple methods.
Altogether, our work reveals insights regarding the robustness of hallucination detection methods leveraging a VLM's internal signals beyond standard benchmarks.
We hope this work encourages further in-depth analyses and careful development of error detection tools to ensure high robustness before being deployed. \looseness=-1

\section{Related Work}

\subsection{Hallucinations in Vision Language Models}

Broadly, a VLM is said to hallucinate when it generates text that is inaccurate or inconsistent with a provided image \citep{liu_survey_2024, chen_survey_2026, jiang_interpreting_2025, zhou_analyzing_2024, guan_hallusionbench_2024}. 
Initial VLM hallucination works primarily focused on object hallucinations where \emph{objects} are not in the image but mentioned in the caption \citep{rohrbach_object_2018, jiang_devils_2025, lovenia_negative_2024, park_glsim_2025}.
More recent works have expanded to include a variety of hallucinations, such as attributes, relations between objects \citep{zhou_analyzing_2024, yu_hallucidoctor_2024, zheng_reefknot_2025, wang_amber_2024, wu_evaluating_2024}, or broadly, any type of inaccuracy \citep{phukan_logit_2025, park_halloc_2025, wang_evaluation_2023}.

To further study hallucinations, it is common to develop novel benchmarks, expanding the scope of evaluation settings.
For example, some works evaluate different visual tasks such as image captioning \citep{rohrbach_object_2018, jing_faithscore_2024, huang_opera_2024, wang_evaluation_2023, petryk_aloha_2024, ben-kish_mitigating_2024} or visual question-answering \citep{guan_hallusionbench_2024, cao_visdiahalbench_2024, tu_ode_2025, gunjal_detecting_2024, villa_magic_2025, huang_visual_2024}.
Others focus on evaluating VLM hallucinations in different response types, such as open-ended \citep{farquhar_detecting_2024, tu_ode_2025, rohrbach_object_2018, liu_automatically_2024}, binary or multiple choice questions \citep{guan_hallusionbench_2024, fu_mme_nodate, li_evaluating_2023}, and even conversational dialogue \citep{park_mitigating_2024, cao_visdiahalbench_2024}.
In this work, we focus on object hallucinations in image captioning settings because it provides the clearest ground-truth labels and aligns with prior work on the detection methods we study.
\looseness=-1

\subsection{Detecting Hallucinations in VLMs} \label{sec:related_detection}

While both detecting and mitigating hallucinations is important, we focus on detection methods because hallucinations are unlikely to be mitigated altogether, given the statistical nature of VLMs \citep{banerjee_llms_2025}.
We elaborate on prior works in detecting hallucinations here and describe mitigation works in \appendixref{app:mitigation}. \looseness=-1

In our work, we focus on methods that detect hallucinations using the VLM's signals as it generates text. 
These detection methods are advantageous for practical applications because they do not require external evaluators, training, or multiple passes of the VLM; in short, they are computationally lightweight.
They compute hallucination scores for every generated text token and can be grouped into two categories: ``black box'' and ``white box'' \citep{chen_survey_2026}.
Black box methods rely on the VLM's text or logit (i.e., pre-softmax) outputs \citep{zhou_analyzing_2024, malinin_uncertainty_2021} while white box methods require internal image and/or text representations \citep{phukan_logit_2025, jiang_interpreting_2025, park_glsim_2025} or attention patterns \citep{jiang_devils_2025, hoang-xuan_pas_2025}. 
While black box methods do not require access to model internals, white box methods tend to perform better, presumably due to their access to richer signals. 
\looseness=-1

Beyond these single-pass internal signal-based methods, others tend to require external evaluators or multiple passes. 
For example, external evaluators can detect hallucinations in captions directly \citep{kaul_throne_2024, ge_visual_2024, yu_hallucidoctor_2024, jing_faithscore_2024, chen_unified_2024, wang_evaluation_2023, wu_logical_2024, wada_zina_2026} or compare the consistency of responses across multiple text generations \citep{liao_visionamplified_2026, farquhar_detecting_2024, yu_hallucidoctor_2024}. Some require fine-tuning large vision or language models \citep{park_halloc_2025, gunjal_detecting_2024, fieback_metatoken_2025, xiao_detecting_2025, chen_detecting_2026} or training a lightweight classifier on VLM signals \citep{tong_faithscan_2026, zhang_dhcp_2025, azachi_leveraging_2025, nath_hallushift_2025}.
Methods requiring multiple passes include contrastive decoding \citep{leng_mitigating_2024, wang_mitigating_2024, zhao_mitigate_2025} which requires two forward passes to calculate logit differences between the original and modified inputs and semantic entropy which computes scores based on multiple sampling trajectories \citep{liao_visionamplified_2026, zhang_vluncertainty_2024, farquhar_detecting_2024}.
We exclude these methods from our analysis because they do not solely rely on a VLM's internal signals as it generates the caption. \looseness=-1

Developing practical and effective hallucination detection methods requires a comprehensive understanding of their behavior, but the predominant evaluation paradigm can only offer limited insight.
Works tend to report standard detection performance metrics (e.g., AUROC, AUPRC) over a large dataset \citep{park_glsim_2025, jiang_interpreting_2025, jiang_devils_2025, zhang_vluncertainty_2024, tong_faithscan_2026, park_halloc_2025, liao_visionamplified_2026} where  white box methods are presented as a clear forerunner over black box \citep{park_glsim_2025, jiang_interpreting_2025, jiang_devils_2025, hoang-xuan_pas_2025}. 
However, this setup does not consider what signals may \emph{drive} white box methods' performance.
Our work aims to provide a more holistic and thorough understanding regarding how white box detection methods work; we reveal novel insights that standard evaluations conceal, such as dependencies on token position heuristics. \looseness=-1
\section{Experimental Setup} \label{setup}

In this section, we first describe the seven hallucination detection methods we study: two are black box, relying only on the caption-generating VLM's output logits, and the remaining five are white box, accessing the model's internal signals.
Then, we provide an overview of our experimental setting, namely the framework, metrics, and VLMs analyzed.

\subsection{Hallucination Detection Methods}

All the selected detection methods can output a hallucination score for every text token generated by the VLM.
We directionally align methods such that higher scores correspond to higher likelihood of hallucination, compute scores for multi-token objects by using the first token \citep{jiang_devils_2025, park_glsim_2025}, and only evaluate the first occurrence of a word in the caption \citep{park_glsim_2025, phukan_logit_2025}.
Here we describe each method at a high level; for details and mathematical formulations, see \appendixref{app:hd_details} or the original works. \looseness=-1

\noindent \textbf{Black box methods.} The two black box methods we evaluate are Negative Log Likelihood (\nll) \citep{zhou_analyzing_2024} and Entropy (\ent) \citep{malinin_uncertainty_2021}.
\nll \ is the negative log likelihood of a generated token and \ent \ is the entropy of the token's generation output probability distribution. \looseness=-1

\textbf{White box methods.} The five white box methods are Internal Confidence (\ic) \citep{jiang_interpreting_2025}, Contextual Lens (\cl) \citep{phukan_logit_2025}, Global-Local Similarity (\glsim) \citep{park_glsim_2025}, Summed Visual Attention Ratio (\svar) \citep{jiang_devils_2025}, and Prelim Attention Score (\pas) \citep{hoang-xuan_pas_2025}. 
\ic \ \citep{jiang_interpreting_2025} is based on how well the generated object can be predicted from just the image patch representations with Logit Lens \citep{nostalgebraist_interpreting_2020}. 
\cl \ \citep{phukan_logit_2025} identifies hallucinations using cosine similarity between the image patches and a generated text token's representations. 
\glsim \ also leverages cosine similarity between representations, but computes a weighted average 
of a generated token's ``global'' similarity with the last prompt token and ``local'' similarity with image patches.
Lastly, \svar \ and \pas \ leverage attention patterns; \svar \ sums the attention from a generated token to the image patches and \pas \ sums attention from a generated token to previously generated caption tokens. \looseness=-1

\subsection{Framework, Metrics, and Models}\label{sec:exp_setting}

We evaluate methods on their ability to detect object hallucinations in image captioning tasks. 
This is to reproduce prior methods and fairly compare them in the settings they were developed in \citep{park_glsim_2025, jiang_devils_2025, jiang_interpreting_2025, chen_survey_2026, bai_hallucination_2025}.
To do so, we use the CHAIR framework \citep{rohrbach_object_2018}.
Given a caption and an image with object-level annotations, CHAIR uses string-based text matching to determine what objects in the caption are hallucinated.
While this framework is widely used, it has limitations such as being context-independent (further described in \appendixref{app:chair_lims}).
To avoid known systematic noise in our main results, we filter out objects that were consistently flagged incorrectly (i.e., ``orange'' mis-flagged as an object when it frequently refers to the color).
However, we show that our results are robust to CHAIR labeling noise in \appendixref{app:orange_results} and \appendixref{app:non_objects}. 
Our analyses report threshold-less metrics of AUROC (Area Under the Receiver Operating Characteristic Curve) and AUPRC (Area Under the Precision-Recall Curve) which is particularly informative under large class imbalances, as with hallucinations. 
While these are similar metrics to standard evaluation settings, our analyses importantly differs by revealing previously inaccessible properties about the analyzed detection methods. \looseness=-1

Our primary analysis is conducted on $5,000$ randomly selected images from the MSCOCO 2014 validation set \citep{lin_microsoft_2014}, following prior work \citep{jiang_interpreting_2025, jiang_devils_2025, park_glsim_2025, hoang-xuan_pas_2025}.
We expand existing works \citep{jiang_devils_2025, jiang_interpreting_2025, park_glsim_2025, hoang-xuan_pas_2025} by conducting our analyses over six diverse open-source VLMs: LLaVA-1.5-7B \citep{liu_improved_2024}, LLaVA-1.5-13B \citep{liu_improved_2024}, InstructBLIP-7B \citep{dai_instructblip_nodate}, Qwen3VL-8B-Instruct \citep{bai_qwen3vl_2025}, InternVL2-8B \citep{chen_how_2024}, and \gemma \ \citep{team_gemma_2026}.
See \appendixref{app:hparams} for further setup and hyperparameter details.
In our main paper, we primarily report results with \llavalg; results for other VLMs can be found in \appendixref{app:exp_details_results}.
To demonstrate the robustness of our results, we additionally conduct generalization experiments in \appendixref{app:generalization}, such as with newer VLMs (Qwen3.5-9B \citep{qwenteam_qwen35_2026} and InternVL3.5-8B \citep{wang_internvl35_2025}), and on a second dataset, Objects365 \citep{shao_objects365_2019}. \looseness=-1
\section{The Role of Token Position in Hallucination Detection} 
\label{sec:token_pos}
In uncovering the driving signals behind internal signal-based detection methods, we reveal three key observations regarding the role of an object's token position in a VLM's caption.
First, we find that token position, derived separately from the content of both the reference image and generated caption, serves as a surprisingly competitive baseline. 
Second, we observe that many top-performing white box methods exhibit a strong correlation with token position, calling into question whether they \emph{rely} on this positional cue.
Lastly, we demonstrate that in fact a number of white box methods depend on token position shortcuts, suggesting they may not leverage reliable internal signals or grounded visual reasoning. 
We further present how these results are generalizable across datasets, prompts, seeds, and labeling noise in (\appendixref{app:generalization}).
\looseness=-1

\subsection{Token Position is a Surprisingly Strong Baseline}\label{sec:token_pos_baseline}

Previous work revealed that hallucinations by VLMs are more likely to occur in later token positions \citep{zhou_analyzing_2024, li_hidden_2025, sun_exploring_2025, favero_multimodal_2024, liu_countering_2026, wang_amber_2024}, which we also show in \appendixref{app:token_pos}. 
One potential explanation is that VLMs begin to rely on the generated objects earlier in the caption to infer the existence of other, but incorrect, objects.
To quantify how strongly this positional bias correlates with hallucination likelihood, we first establish a baseline detection method using token position alone.
Concretely, for each object of interest, this heuristic is defined as the object's first token position, beginning at the caption start.

\begin{wrapfigure}{r}{0.5\textwidth}
  \begin{center}
    \vspace{-0.25in}
    \includegraphics[width=\linewidth, trim={0.15in 0.5in 0.2in 0.5in}, clip]{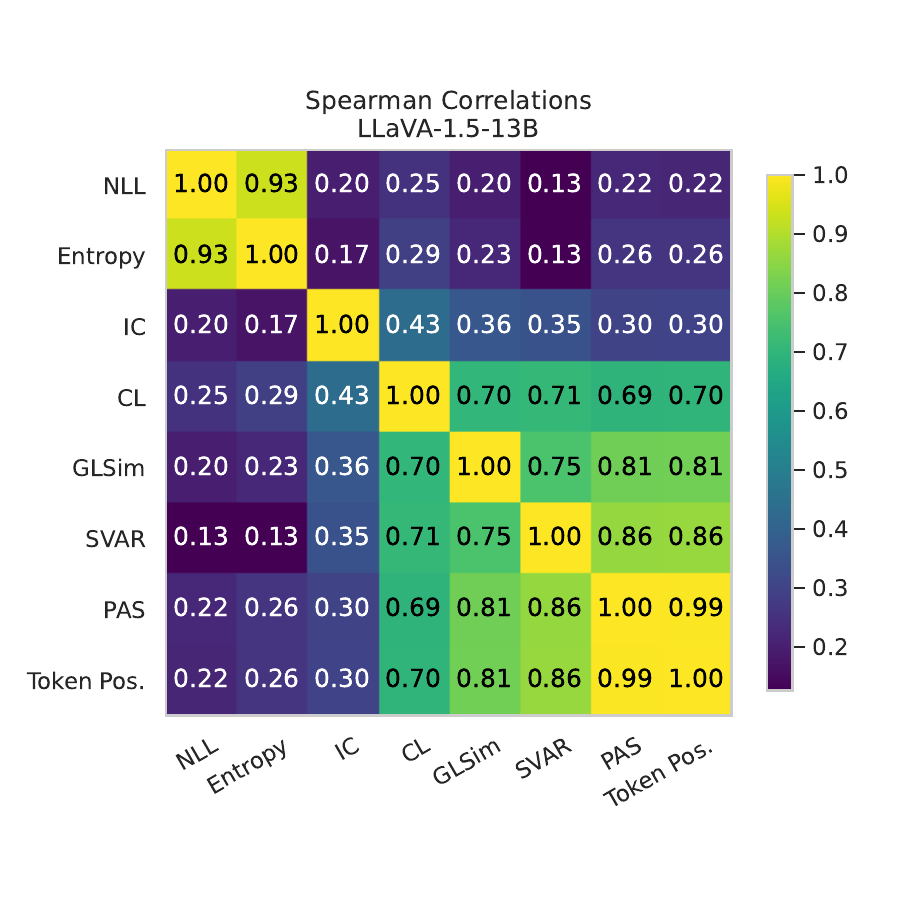}    
    \end{center}
    \vspace{-0.2in}
  \caption{\textbf{\cl, \glsim, \svar, and \pas \ are highly correlated with token position and each other.}}
  \vspace{-0.2in}
  \label{fig:correlations}
\end{wrapfigure}

Surprisingly, we find that \textbf{token position yields evaluation scores on par with state of the art white box methods}, and normalizing by the caption length can push these metrics even higher, especially for LLaVA VLMs (\tabref{tab:overall_aucs}).
These results suggest that standardizing this ``blind'' baseline is crucial for calibrating the true difficulty of detecting hallucinations for a particular VLM.
It can reveal whether a benchmark success requires deep internal signals or merely aligns with a positional bias.
As a result, these findings urges caution when interpreting reported metrics that exclude this naive baseline. 
\looseness=-1


\begin{table}
    \caption{\textbf{Token position achieves evaluation scores on par with white box detection methods, despite being agnostic to the image and caption content.} 
    Across many VLMs, the token position heuristic achieves comparable AUROC (top) and AUPRC (bottom) to established detection methods, even achieving \textbf{best} or \underline{second best} performance for certain VLMs.
  Token position is an object's first token position in the caption, and normalized token position divides by the caption's length.}
  \centering
  \resizebox{\textwidth}{!}{
  \begin{tabular}{ll c cc ccccc cc}
    \toprule
    \multirow{2}{*}{\textbf{Metric}} & \multirow{2}{*}{\textbf{Model}} & & \multicolumn{2}{c}{\textbf{Black Box}} & \multicolumn{5}{c}{\textbf{White Box}} & \multicolumn{2}{c}{\textbf{Heuristic}} \\
    \cmidrule(lr){4-5} \cmidrule(lr){6-10} \cmidrule(lr){11-12}
    & & Random & NLL & Entropy & IC & CL & GLSim & SVAR & PAS & Token Pos. & \makecell{Token Pos.\\(norm.)} \\

    \midrule
    \multirow{7}{*}{AUROC} 
    & \llavalg        & \multirow{5}{*}{50.0} & 72.2 & 73.6 & 74.8 & 82.3 & 82.8 & 81.4 & \underline{83.4} & 83.2 & \textbf{84.6} \\
    & \llavasm        & & 72.4 & 73.4 & 78.6 & 79.3 & 83.5 & 81.1 & \underline{84.0} & 83.4 & \textbf{84.7} \\
    & InstructBLIP-7B & & 81.1 & 82.5 & 88.1 & 86.7 & 86.3 & \underline{88.7} & \textbf{89.1} & 88.5 & 88.7 \\
    & Qwen3VL-8B      & & 66.5 & 68.5 & 69.1 & 67.4 & 64.7 & \textbf{79.3} & \underline{77.1} & 73.9 & 74.2 \\
    & \internvl       & & 64.3 & 65.8 & 77.1 & 69.6 & \underline{79.2} & \textbf{81.0} & 78.8 & 77.0 & 77.8 \\
    & \gemma & & 65.4 & 67.3 & 69.8 & 53.9 & \textbf{80.2} & \underline{76.2} & 74.9 & 72.2 & 72.7 \\

    \cmidrule{2-12}
    & \textit{Average} & \textit{50.0} & \textit{70.3} & \textit{71.9} & \textit{76.2} & \textit{73.2} & \textit{79.5} & \textit{\textbf{81.3}} & \textit{\underline{81.2}} & \textit{79.7} & \textit{80.5}\\
    \midrule
    \multirow{7}{*}{AUPRC} 
    & \llavalg        & 22.8 & 44.7 & 45.1 & 50.4 & 53.6 & \underline{54.4} & 49.3 & 51.4 & 50.9 & \textbf{56.3} \\
    & \llavasm        & 25.1 & 47.2 & 46.4 & 57.5 & 55.5 & \textbf{60.7} & 51.9 & 56.5 & 54.7 & \underline{59.8} \\
    & InstructBLIP-7B & 26.2 & 61.0 & 62.1 & \textbf{73.1} & 64.1 & 64.9 & 68.7 & \underline{69.3} & 66.3 & 67.9 \\
    & Qwen3VL-8B      & 20.8 & 35.7 & 38.9 & 38.2 & 33.3 & 30.4 & \textbf{50.5} & \underline{44.0} & 37.2 & 39.1 \\
    & \internvl       & 18.3 & 30.4 & 31.2 & 43.5 & 34.6 & \underline{44.9} & \textbf{48.1} & 39.8 & 37.0 & 41.2 \\
    & \gemma & 15.1 & 27.0 & 29.7 & 25.6 & 19.2 & \textbf{40.5} & \underline{33.7} & 31.1 & 27.2 & 29.7 \\
    
    \cmidrule{2-12}
    & \textit{Average} & \textit{21.4} & \textit{41.0} & \textit{42.2} & \textit{48.1} & \textit{43.4} & \textit{\underline{49.3}} & \textit{\textbf{50.4}} & \textit{48.7} & \textit{45.5} & \textit{49.0}\\

    \bottomrule
  \end{tabular}
  }
  \label{tab:overall_aucs}
\end{table}

Next we investigate the performance trends of detection methods and the token position heuristic by computing the correlation of object-wise hallucination scores.
As demonstrated in \figref{fig:correlations}, four of the white box methods (\cl, \glsim, \svar, and \pas) exhibit high correlation with token position ($ \rho \geq 0.70$, all $p < 0.001$) and with one another ($\rho \geq 0.69$, all $p < 0.001$), but not with \ic \ or black box methods ($\rho \leq 0.43$, all $p < 0.001$) (see \appendixref{app:correlations} for additional results with other VLMs and qualitative examples).
The high correlation of many white box methods with token position raises the question: are these methods simply using token position as a shortcut? \looseness=-1

\subsection{Reliance of Detection Methods on Token Position}
\label{sec:remove_replace}

To investigate the degree to which detection methods rely on token position as shortcut signal, we conduct a pair of interventions on the generated captions.
The interventions aim to isolate the effect of an object's token position on a detection method's score while controlling for semantic influences.
In the first intervention, for each object in a caption, we modify its token position by \emph{removing} all complete sentences preceding it, recompute hallucination scores, and calculate the difference in scores from the original.
This intervention effectively reduces the object's token position; however, it also removes the semantic content of the preceding sentences.
To control for this potential confound, we perform a similar intervention but \emph{replace} the sentences with image-agnostic but task-relevant text and re-compute the difference in hallucination scores from the original.
In short, both interventions remove or reduce semantic information but only the former changes token position significantly.
If a detection method relies on token position, it will behave similarly to a token position heuristic in these interventions: hallucination scores decrease when the preceding text is removed and only change negligibly when it is replaced with uninformative text. 
See \appendixref{app:remove_replace} for replacement sentences and further details.
\looseness=-1

\begin{figure}[t]
    \centering
    \includegraphics[width=1.0\linewidth, trim={0 0.1in 0 0.08in}, clip]{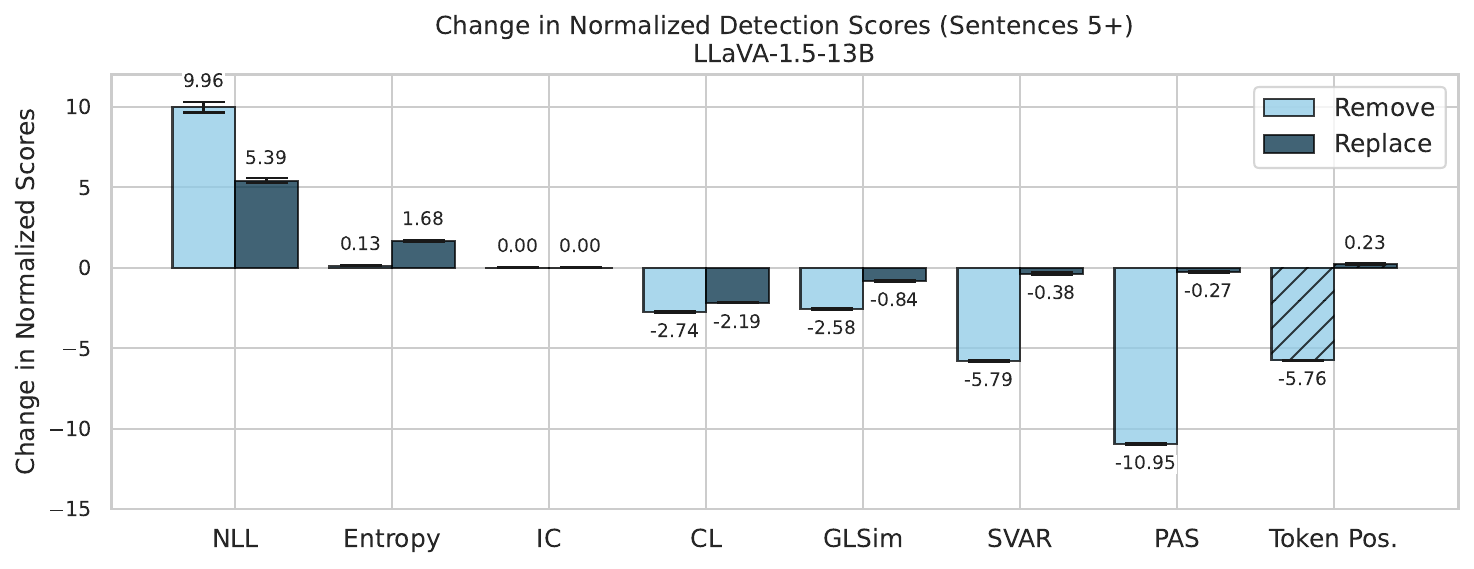}
    \caption{\textbf{Our experiments can reveal when white box methods (\cl, \glsim, \svar, \pas) demonstrate a reliance on token position.} 
    These four methods resemble behavior of the token position heuristic, especially attention-based methods: \svar \ and \pas. 
    On the other hand, black box methods tend to behave in the opposite direction. 
    We only examine objects in sentence 5+ of the caption to ensure sufficiently long preceding context.
    Scores are normalized by mean and standard deviation to be on the same scale. 
    Error bars represent 95\% confidence intervals. \looseness=-1
    }
    \label{fig:remove_replace}
    \vspace{-0.1in}
\end{figure}

\noindent\textbf{White Box Methods.} We find that the \textbf{white box methods that highly correlate with token position also depend on it}, especially attention-based \svar \ and \pas. 
As shown in \figref{fig:remove_replace}, when preceding sentences of the caption are removed, white box scores decrease, resembling the change in token position score.
This holds for all white box methods evaluated except \ic, as expected since it leverages only image representations and is agnostic to text representations.
For other white box methods, after accounting for the effect of preceding semantic information (i.e., the change in score with the replacement intervention), token position accounts for approximately 20\%, 67\%, 93\% and 98\% of \cl, \glsim, \svar, and \pas's decrease in scores respectively.
These results suggest that most white box methods rely on token position cues to some degree, revealing robustness limitations that current evaluation settings conceal. 
Conducting this analysis with other VLMs in \appendixref{app:remove_replace} shows similar trends for white box methods, especially attention-based methods \svar \ and \pas.
We further demonstrate robustness of these results across newer VLMs, datasets, prompts, and context-aware labeling in \appendixref{app:generalization}. \looseness=-1

\noindent\textbf{Black Box Methods.} 
We observe that token position has an inverse effect on black box methods: when an object's token position decreases, black box methods either do not change or score it as \textit{more} likely to be hallucinated. 
While \ent \ seems to be more sensitive to information than token position, roughly 46\%  
of \nll's increase in mean hallucination score can be attributed to token position alone.
We hypothesize the inverse effect can be attributed to sensitivity to training objectives like perplexity that measure how well a token ``fits'' the preceding context
(see \appendixref{app:remove_replace} for more details).
Although black box methods do not seem to \textit{rely} on token position like white box methods, these results suggest they still exhibit some sensitivity to language-based cues. \looseness=-1

\section{Token Position-dependent Methods are not Robust Predictors}

Having established how certain white box detection methods highly rely on a token position shortcut,
we explore how this reliance is reflected in a method's visual grounding abilities and robustness to varying object hallucination patterns.
First, we analyze how the hallucination modality source affects detection performance and find that white box methods' competitive advantage arises from detecting language-prior induced hallucinations (\secref{sec:partition}).
Next, we explore the extent to which white box methods are grounded in image signals (\secref{sec:visual_signals}).
We find that highly token position-dependent methods exhibit minimal grounding in image-based signals, suggesting they may depend on this heuristic \textit{in lieu} of visual signals. 
Lastly, we demonstrate that token position is not a robust signal by presenting an example prompt setting that can lead token position-dependent methods to perform near random chance (\secref{sec:tp_case}). \looseness=-1

\subsection{White Box Methods' Competitive Advantage is from Language-based Hallucinations} \label{sec:partition}

First, we aim to explore the degree to which the performance of white and black box methods are dependent on the source modality of the hallucination (i.e., language-based or not).
We hypothesize that white box methods gain their competitive advantage over black box methods from detecting language-based hallucinations but not necessarily other hallucinations (e.g., an object is misrepresented in the image embeddings).
We could see this behavior whether white box methods leverage rich visual signals or rely on language patterns like token position heuristics. 
Visually grounded methods could fail on non-language hallucinations because corrupted image signals would mislead detection methods. 
On the other hand, language heuristics would simply fail to detect hallucinations that do not arise from language generation or modeling patterns.
\looseness=-1


To explore this hypothesis, we partition hallucinations into whether they are language-based or not and compare performance between white and black box methods on each partition separately.
Following prior work, we determine which partition a hallucinated object belongs to by feeding the generated caption with a \textit{blank} gray image into the VLM and examining whether the new predicted token for each hallucinated object is the same or not \citep{zhang_mathverse_2025, zhao_mitigate_2025, favero_multimodal_2024, xie_vdpo_2024}.
If the hallucination persists even when there is no image, the hallucination goes in the language-based partition because it is not dependent on the image and likely arises from language factors \citep{zhang_mathverse_2025, zhao_mitigate_2025, favero_multimodal_2024, xie_vdpo_2024}.
Otherwise, the hallucination cannot be strictly isolated to language sources.
The other partition contains hallucinations sensitive to the image, and it primarily encompasses hallucinations from visual deficiencies (e.g., ambiguous visual representations or misalignment between vision and language \citep{liu_survey_2024, chen_survey_2026}).
While the image-sensitive partition may contain some language hallucinations that are also influenced by the image, any performance differences between the partitions is likely to be driven by the introduction of vision-based errors.
For further elaboration, see \appendixref{app:partition}. \looseness=-1

\begin{figure}[t]
  \centering
  \includegraphics[width=1.0\linewidth]{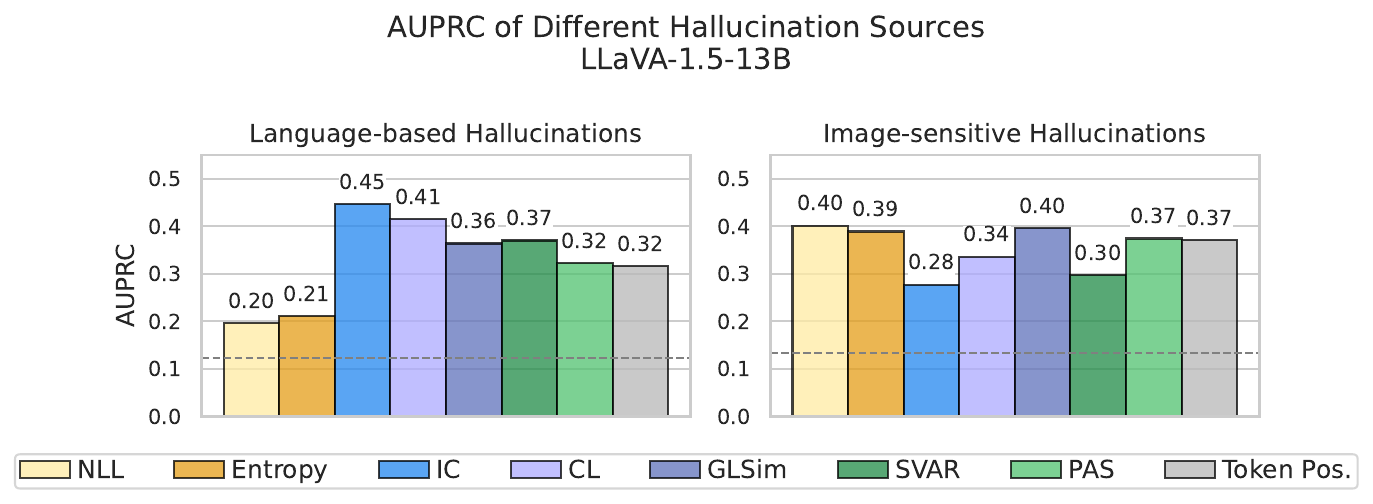}
  \caption{\textbf{The competitive advantage of white box methods stems from detecting language-based hallucinations.}
  White box methods outperform black box methods on language-based hallucinations (left) but not on image-sensitive hallucinations (right).
  We report AUPRC as it is more sensitive than AUROC to performance on hallucinations. 
  Black box methods are \nll \ and \ent, token position is in gray for reference, and horizontal dashed lines represent the random baseline, i.e., hallucination rate. \looseness=-1
  }
  \label{fig:partition_llava}
\end{figure}

As shown in \figref{fig:partition_llava}, \textbf{white box methods consistently outperform black box methods on language-based hallucinations but not on image-sensitive hallucinations}. 
Specifically, on language-based hallucinations, all white box methods outperform black box methods by at least $+0.12$ and up to $+0.24$ on AUPRC.
Interestingly, we also observe that token position can outperform black box methods, suggesting that detecting some language-based hallucinations may not always require access to image representations.
On the other hand, on image-sensitive hallucinations, black box methods perform comparably to white box methods, outperforming some white box methods by up to $+0.12$ in AUPRC and even matching the performance of the best white box detection method.
We observe similar trends across VLMs, as demonstrated in \appendixref{app:partition}. 
These results suggest that white box methods are leveraging a signal that black box methods do not in order to detect language-based hallucinations. 
In the next section, we explore the extent to which this signal is visually grounded. \looseness=-1

\subsection{Token Position-dependent Methods Lack Visual Grounding} \label{sec:visual_signals}

One possible explanation for white box methods achieving strong performance on language-based hallucinations is that they appropriately leverage visual signals. 
This could enable them to cross validate the presence of an object in the image.
However, given the reliance on token position shortcuts of some white box methods, it is unclear whether this is the case. 
\looseness=-1

We set out to examine the degree to which white box methods are grounded in image signals; to do so, we use the blank image replacement setup from \secref{sec:partition}.
This time, we evaluate each method's detection performance with the original image compared to a blank one of the same size. 
Importantly, the captions and ground truth labels are derived from the original image to ensure consistency across the two settings.
If a detection method is faithfully relying on image-based signals, then with the blank image, it should score all objects as likely to be hallucinated and performance should approach random chance (i.e., the hallucination rate).
On the other hand, if a method ignores image-based signals, its performance would remain high when given a blank image. \looseness=-1

\begin{figure}[t]
  \centering
  \includegraphics[width=1.0\linewidth]{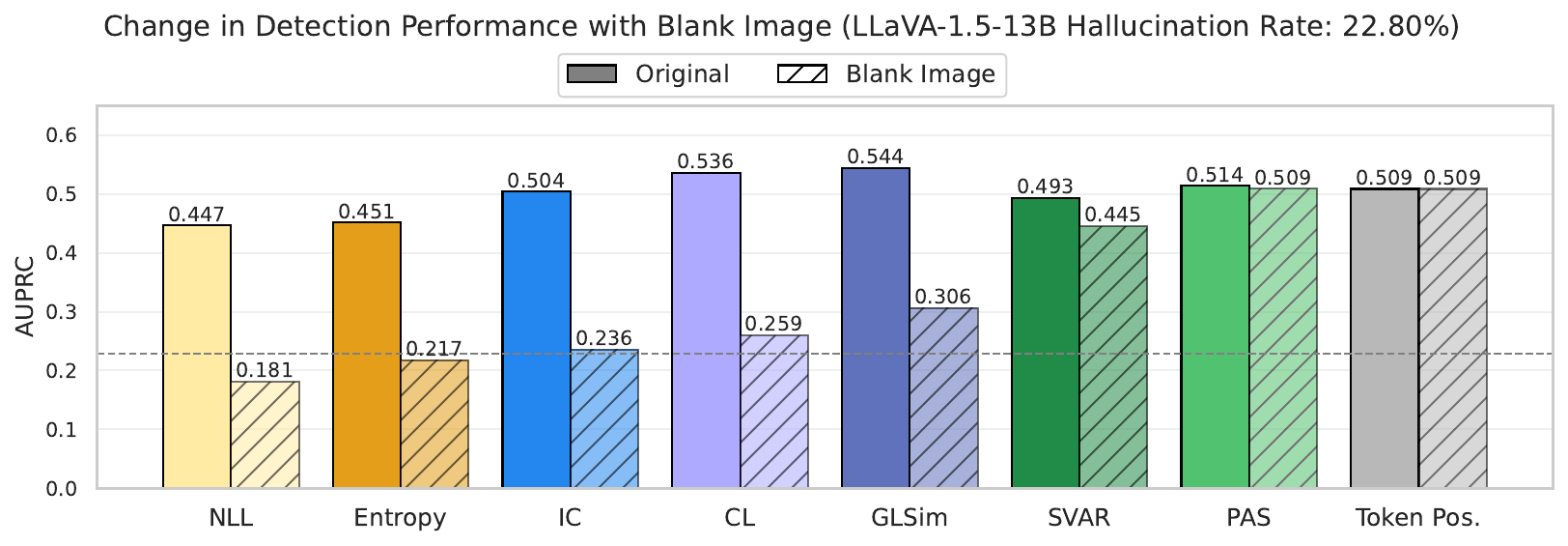}
  \caption{\textbf{Detection methods that depend on token position (i.e., \svar, \pas \ for \llavalg) exhibit less visual grounding than other detection methods.} 
  We compare detection method performance in AUPRC when using the original image (solid) compared to replacing the image with a blank gray image but retaining the original caption (hatched). 
  A greater drop suggests greater grounding in visual signals, while negligible drops indicate little to no grounding in the image.
  }
  \vspace{-0.1in}
  \label{fig:blank_image}
\end{figure}

We observe in \figref{fig:blank_image} that \textbf{the detection methods that are dependent on token position tend to exhibit very little grounding in image-based signals}, suggesting these methods are not visually grounded.
Specifically, methods that are more dependent on token position (\svar, \pas) retain the most performance in the blank image setting, whereas black box methods (\nll, \ent) and \ic \ drop to or below random chance.
Methods based on text and image similarity (\cl, \glsim) fall somewhere in between. 
Without the image, \nll, \ent, and \ic \ are no longer able to complete this hallucination detection task, suggesting these methods are responsive to changes in the image and to some extent, are desirably visually grounded.
Similarly, \cl \ and \glsim \ performance drops to just above random chance, suggesting a degree of visual grounding.
On the other hand, the minimal change in \svar \ and \pas's performance suggests that these attention-based methods can depend on token position heuristics \textit{in lieu} of image-based signals. 
In summary, white box methods that are less dependent on token position seem to leverage visual signals more than token position-dependent methods.
While results generally hold for other VLMs (\appendixref{app:visual}), we observe few exceptions that suggest that depending on the VLM, some methods can be sensitive to both token position and visual signals simultaneously.
The lack of visual grounding suggests that token position-dependent methods may not be robust predictors, which we further explore in the next section. 
\looseness=-1

\subsection{Object Biases Lead Token Position to be an Unreliable Predictor} \label{sec:tp_case}

If reliance on token position heuristics instead of visual signals leads to non-robust detectors, this raises a natural question: why can these methods still perform well across a variety of captioning settings, as in \appendixref{app:generalization}? 
One hypothesis is that their success arises in part from a correlation between object salience and a VLM's generation order.
When asked to describe an image, VLMs may be more likely to mention larger and more visually salient objects earlier.
Meanwhile smaller, less salient objects are mentioned later, and because they are more difficult to recognize, they are more prone to hallucination. 
This induces a token-position signal that can serve as a reliable shortcut under standard captioning settings across datasets, generation seeds, and variations of the prompt (\appendixref{app:generalization}). 

However, VLMs are general-purpose models rather than dedicated image captioners, and the generated object ordering may vary substantially across downstream applications. 
We therefore test whether token position remains reliable when this ordering is altered by prompting the VLM to identify the least obvious, smallest, or hardest-to-see objects first, as may be useful in applications such as assistive visual description  \citep{gurari_vizwiz_2018, xie_visual_2025}. 
Concretely, we prompt the VLM to ``Describe the least obvious, smallest, or hardest-to-see elements of the image first. Then describe the most prominent objects.'' 
To attribute changes in performance to token position alone, we primarily evaluate over the subset of objects that appear in the captions in both the original and prompting small objects first settings.
See \appendixref{app:unreliable} for further details. \looseness=-1

\begin{figure}[t]
  \centering
  \includegraphics[width=0.9\linewidth, trim={0in, 0.1in, 0in, 0in}, clip]{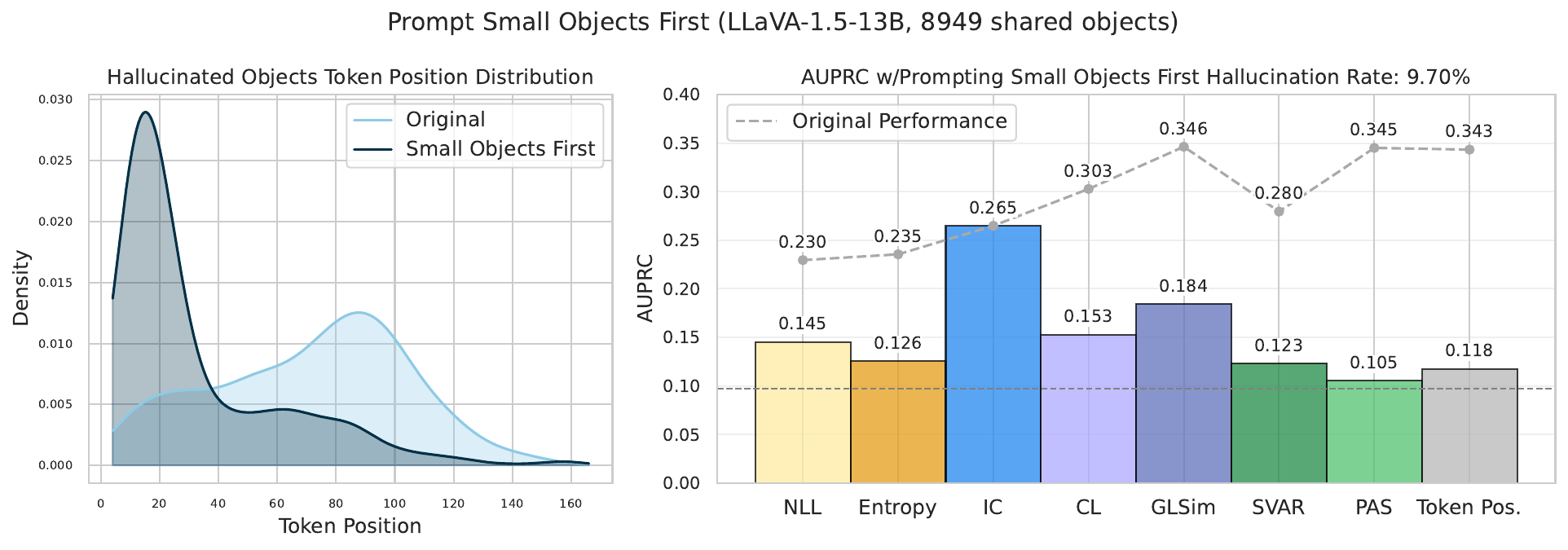}
  \vspace{-0.1in}
  \caption{\textbf{Token position and position-dependent methods (e.g., \svar, \pas) are not always reliable hallucination detectors.} 
  \textbf{Left}: While hallucinated objects tend to appear later in the caption for typical settings (\sethlcolor{lightblue}\hl{light blue}), prompting for small objects first can lead VLMs to hallucinate earlier in the caption (\sethlcolor{darkblue}\hl{dark blue}).
  \textbf{Right}: When \llavalg \ is asked to describe small objects first, \svar \ and \pas \ perform near random chance. 
  To ensure fair comparisons, we plot AUPRC over the objects shared in captions from prompting for small objects first (bars) and in the original setting (dashed lines) for a given image. \looseness=-1
  }
  \label{fig:real_world}
\end{figure}

Our results in \figref{fig:real_world} show that hallucinations can occur much earlier in the caption with the alteration.
As a result, the token position heuristic performs no better than black box methods at detecting hallucinations.
Further, we see how attention-based methods that depend on token position (\svar, \pas) drop in relative performance as well, now falling behind representation-based methods of \ic, \cl, and \glsim.
While the same prompt may induce this effect to varying degrees for different VLMs (see \appendixref{app:unreliable}), 
it is conceivable that there are other prompts, applications, and tasks would similarly render token position-dependent methods less usable, such as multiple choice or short answer VQA \citep{li_evaluating_2023, guan_hallusionbench_2024, fu_mme_nodate}.
Taken together, in contrast with standard evaluation settings that suggest attention-based methods like \svar \ and \pas \ to be strong contenders for detecting hallucinations from VLM-generated captions, our analyses suggest that they can be more likely to depend on unreliable token position heuristics that may not generalize beyond the standard captioning settings. \looseness=-1

\section{Evidence of Complementary Signals Beyond Token Position}\label{sec:ensemble}

So far, we have revealed limitations of certain hallucination detection methods leveraging internal signals, leaving open the question of how to improve these methods beyond token position.
To begin answering this question, we observe in \figref{fig:correlations} that while certain methods exhibit high correlation with token position, others do not.
This leads us to hypothesize that methods may carry complementary signals to be leveraged. 
We test this hypothesis by building progressive ensembles, starting with token position, adding groups of detection methods, and observing performance gains with each addition. \looseness=-1

\begin{figure}[t]
    \centering
    \includegraphics[width=1.0\linewidth, trim={0 1in 0 0.075in}, clip]{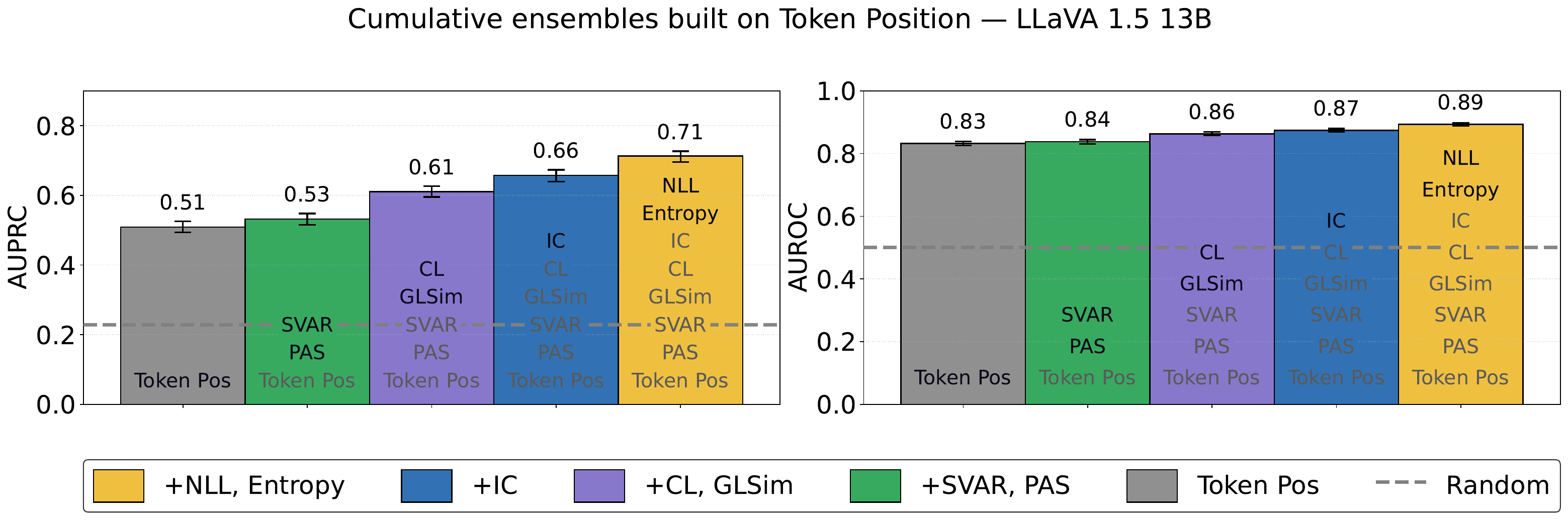}
    \vspace{-0.3in}
    \caption{\textbf{Detection methods can contain complementary signals to one another.}
    We cumulatively build logistic-regression ensembles on top of token position adding methods in order of decreasing correlation and dependence with token position: (1) ~+{\svar, \pas}; (2) ~+{\cl, \glsim}; (3) ~+\ic; and (4) black box methods ~+{\nll, \ent}.
    Except for (~+{\svar, \pas}), each additional group of detection methods improves the ensemble's performance, suggesting the groups carry non-overlapping signal to one another.
    Error bars are $95\%$ confidence intervals computed via percentile bootstrap of validation predictions. \looseness=-1} 
    \label{fig:ensemble}
    \vspace{-0.1in}
\end{figure}

Concretely, to explore whether the methods capture complementary or overlapping signal, we first partition them based on correlation with (\figref{fig:correlations}) and dependence on (\figref{fig:remove_replace}) token position.
Then we train logistic-regression classifiers, progressively adding one group at a time.
Detection methods are partitioned into four groups: (1) white box methods that are highly correlated with and dependent on token position (\svar, \pas), (2) white box methods that are moderately correlated and dependent on token position (\cl, \glsim), (3) the last white box method that is weakly correlated and independent of token position (\ic), and (4) black box methods (\nll, \ent) which are weakly correlated with token position and most white box methods.
For each ensemble, we train logistic regression classifiers using the scores from the included hallucination detection methods on 5,000 randomly selected images from MSCOCO training split.
We primarily report each progressive ensemble's performance on 5,000 validation images using AUPRC as this metric considers class imbalances, and we additionally include AUROC in \figref{fig:ensemble}. 
For additional details, results on additional VLMs, and further  experiments on signal complementarity, see \appendixref{app:ensemble}-\appendixref{app:cross-model}. \looseness=-1

We observe that for \llavalg, while \svar \ and \pas \ do not contribute informative signal above token position, \textbf{the remaining white box methods can contribute additional predictive power, and black box methods can further offer complementary non-redundant information}.
This can be seen by the monotonic structure in \figref{fig:ensemble}. 
Specifically, adding \cl \ and \glsim \ on top of \svar, \pas, and token position raises AUPRC from $0.53$ to $0.61$ ($+0.08$). 
This gain suggests \cl \ and \glsim \ can offer complementary signals to improve performance beyond what token position and token position-dependent methods can offer.
Next, incorporating \ic \ further pushes AUPRC to $0.66$ ($+0.05$) over the other four white box methods, indicating that the Logit-Lens-based \citep{nostalgebraist_interpreting_2020} signal can supply discriminative information not captured by the embedding similarity (\glsim, \cl) or attention-based (\svar, \pas) methods.
Lastly, adding black box methods (\nll, \ent) further yields an additional improvement of $+0.05$ to reach AUPRC of $0.71$, notable since these scores are computed without internal model access. 
\looseness=-1 

Overall, the combined ensemble performs the best, implying that methods can contribute unique discriminative power to the combined detector.
On top of token position alone, using all methods results in AUPRC of $0.71$ ($+0.20$ above token position alone) and AUROC of $0.89$ ($+0.06$), suggesting that developing a hallucination detection method that leverages multiple types of detection methods can outperform any existing method alone.
We further include results with other VLMs in \appendixref{app:ensemble}, which demonstrate that while the full ensemble consistently performs best, the degree of complementarity between each group of methods may vary per VLM.
Taken together, these results reveal that different methods can carry non-redundant but complementary signals, suggesting that future detection methods could improve performance and reliability by combining different mechanisms from different methods. \looseness=-1

\section{Conclusion}

In this work, we conduct an in-depth analysis to understand the robustness of hallucination detection methods that leverage internal signals as a VLM generates text. 
We reveal insights inaccessible through typical benchmarks, and in light of our findings, we recommend three practical suggestions for researchers and practitioners working with VLM hallucination detection. \looseness=-1

\textbf{Standardize a ``blind'' positional baseline.} We demonstrate that token position is a competitive detection baseline in standard evaluation settings (\secref{sec:token_pos_baseline}), and that certain white box methods exhibit some degree of dependency on token position (\secref{sec:remove_replace}). 
Consequently, we advocate for all new detection methods to be compared to a token position baseline: simply the position of the word's first token in the generated caption.
By standardizing this baseline, we obtain a better assessment on the progress of hallucination detection that cannot be explained away by a simple heuristic based on the structure of the generated output.
\looseness=-1

\textbf{Don't forget the ``V'' in VLM.} 
In revealing the practical limitations of detection methods, we find that white box methods' competitive advantage tends to be from detecting language-based hallucinations (\secref{sec:partition}), and when methods rely on token position, they tend to not be visually grounded (\secref{sec:visual_signals}).
Consequently, they can exhibit biases towards an object's position in the caption (\secref{sec:tp_case}). 
Hallucination detection methods must exhibit visual grounding which in its simplest form, can be operationalized as sensitivity to the image. 
Without it, they can fail beyond the standard image captioning evaluation. 
We advocate for researchers to validate and quantify their method's image sensitivity to ensure it uses the image rather than heuristic cues. 
Without this validation, we risk a ``vision-blind'' research trajectory where detecting hallucinations reflects detecting artifacts of the evaluation setting rather than building a more robust detection that generalizes across many real-world situations. \looseness=-1

\textbf{Leverage strengths of different methods.} 
Current evaluations of hallucination detection methods encourage discounting methods that achieve lower performance, such as black box methods. 
However, our ensemble experiments demonstrate that despite this weakness, black box methods can carry \emph{complementary} signals to white box methods (\secref{sec:ensemble}). 
In other words, black box methods can detect certain hallucinations that white box methods do not.
To further improve hallucination detection, we encourage future work to explore how to leverage these complementary strengths.

\begin{ack}
We would like to thank all those who provided thoughtful feedback and discussion, especially members of the Princeton Visual AI Lab and the anonymous reviewers. 
We gratefully acknowledge support from the NSF Graduate Research Fellowship Program (A.C., J.W.), Princeton University President’s Fellows (J.W.), Princeton Presidential Postdoctoral Fellowship (S.A.) and from Solidigm AI SW. \looseness=-1

\end{ack}

\bibliographystyle{plainnat}
\bibliography{references}

\appendix
\newpage

\clearpage
\appendix


{
    \centering
    \Large
    \textbf{\thetitle}\\
    \vspace{0.5em}
    Appendix\\
    \vspace{1.0em}
}

\hypertarget{toc}{}
\startcontents[sections]
\printcontents[sections]{l}{1}{\setcounter{tocdepth}{2}}

\newpage

\section{Implementation details}\label{app:impl_details}

This section covers details regarding the hallucination detection methods analyzed and token position baseline (\secref{app:hd_details}), and details on experimental setup and implementation (e.g., hyperparameters) (\secref{app:hparams}).

\subsection{Hallucination detection method details}\label{app:hd_details}

In this section we provide greater detail and mathematical formulations of the hallucination detection problem in general (\secref{app:math_hd}) and of each hallucination detection method (\secref{app:specific_hd}).

First, the detection method must rely on signals from the VLM generating the caption, either internal representations like embeddings or attention (white box) or external logits or probabilities (black box). 
This excludes methods that detect hallucinations using auxiliary models, such as LLMs or lightweight classifiers \citep{kaul_throne_2024, ge_visual_2024, yu_hallucidoctor_2024, jing_faithscore_2024, chen_unified_2024, park_halloc_2025, fieback_metatoken_2025}.
Second, we selected detection methods that could output a token-wise hallucination score with a single pass, excluding methods based on contrastive decoding \citep{leng_mitigating_2024, wang_mitigating_2024, zhao_mitigate_2025} or that require multiple passes \citep{liao_visionamplified_2026, zhang_vluncertainty_2024, farquhar_detecting_2024}.

\subsubsection{Mathematical formulation of hallucination detection methods.} \label{app:math_hd}

Let $x$ denote the input image and instruction prompt ``Describe this image in detail'' and $f_\theta$ the VLM. 
$f_\theta$ consists of a vision encoder $f_\phi$ that represents the image using $N$ image embeddings, language model decoder $f_\pi$ that consists of $L$ transformer layers, and a projector $\Phi$ that projects image embeddings into the language space.

Given the inputs $x$, $f_\theta$ then generates a token sequence of length $T$, $y=(y_1, \dots, y_T)$ autoregressively with the conditional distribution $p_\theta(y_t \mid x, y_{<t})$.
Let $\mathcal{M}=\{m_1,\dots,m_7\}$ denote the set of methods, where some are black box, accessing only output probabilities or logits $p_\theta(\cdot \mid x, y_{<t})$, and others which are white box, additionally leveraging internal model signals.
Each detection method $m \in \mathcal{M}$ assigns a per-token hallucination score as $f_\theta$ generates each token autoregressively:
\[
s_m(y_t; x, y_{<t}) \in \mathbb{R},
\]
For a multi-token object $o=(y_a,\dots,y_b)$, we define the per-object score using the first token, following prior work \citep{jiang_devils_2025, park_glsim_2025}:
\[
s_m(o) = s_m(y_a; x, y_{<a}).
\]

We do not modify the original methods, except to directionally align scores such that higher always corresponds to a higher likelihood of hallucination.
Following prior work, we always consider the first token of each object \citep{jiang_devils_2025, park_glsim_2025}. 
Additionally, if an object is mentioned multiple times in the caption, we only compute the hallucination scores for the first time it is mentioned. 
This is to have one hallucination score for each object mentioned and the first instance is when the object would be likely to have higher hallucination scores.

\subsubsection{Equations of specific methods.} \label{app:specific_hd}

\ic \ \citep{jiang_interpreting_2025} computes scores based on how likely the object can be predicted from the image representations. Concretely, it applies Logit Lens \citep{nostalgebraist_interpreting_2020} on all $N$ image embeddings $I$ from all $L$ LLM layers and takes the maximum value for the generated token, indexed by $o$:

\vspace{-0.5em}
$$s_{ic} = 1 - \mathrm{max}_{\substack{1\leq l \leq L \\ 1 \leq i \leq N}} \{\mathrm{softmax}(\mathrm{LogitLens}(I_{li}))_o\}$$
\vspace{-0.5em}

\cl \ \citep{phukan_logit_2025} identifies hallucinations based on similarity between image and text representations. Specifically, \cl \ computes the maximum cosine similarity between the generated token's text embedding $t$ from all $N$ image embeddings $I$, crossing between layers: 

\vspace{-1em}
$$s_{cl} = 1 - \mathrm{max}_{\substack{1\leq l \leq L \\ 1 \leq i \leq N}} \{\mathrm{CosineSim}(I_{li}, t_l)\}$$
\vspace{-0.5em}

\glsim \ \citep{park_glsim_2025} also leverages cosine similarity between representations, but computes a weighted average of ``local'' similarity between the token's text embedding $t$ with $k$ image embeddings $I_{1:k}$ and a ``global'' similarity between the token's embedding with the last token embedding of the prompt, $r$. Layers for embeddings $l$ and $l'$ are hyperparameters, along with the weight term $w$ and number of image patches $k$. See \citep{park_glsim_2025} for original formulations. 

Here we describe further details about \glsim \ that were omitted in the main text for brevity.

Recall that \glsim, or Global-Local Similarity computes a weighted average of two similarity terms, and is defined as:

$$s_{glsim} = 1 - \left(w \cdot \mathrm{GlobalSim}(r_l, t_{l'})+ (1-w) \cdot \mathrm{LocalSim}(I_{l, 1:k}, t_{l'}) \right)$$

where $r$ is the last token of the prompt embedding, $t$ is the embedding of the object token $o$, and $I$ represents image patch embeddings. $w$, $k$, $l$, and $l'$ are hyperparameters. 
More concretely, $\mathrm{GlobalSim}=\mathrm{CosineSim}(r_l, t_{l'})$.
Local similarity first requires selecting the top-$k$ image patches by applying Logit Lens \citep{nostalgebraist_interpreting_2020} on all $N$ image patches, and selecting the $k$ that output the highest probability for the current object token $o$. Mathematically, the top-$k$ image embeddings at layer $l$ are 

$$I_{l, 1:k} = \mathrm{TopK_{I \in \textbf{I}}}(\mathrm{softmax}(\mathrm{LogitLens}(I))_o)$$

Once the top-$k$ image patches are selected, the local similarity is defined as:

$$\mathrm{LocalSim}=\frac{1}{k}\sum_{I_l \in I_{l, 1:k} }\mathrm{CosineSim}(I_l, t_{l'})$$

\svar \ \citep{jiang_devils_2025} leverages attention instead of representation embeddings like the previous methods. Specifically, it sums the attention from the generated token to all image patches, aggregating across layers $l_s$ to $l_e$ and averaging across $H$ heads. 
By default, SVAR uses layers 5 to 18, but we conduct hyperparameter searches for each VLM.
Because this sum is unbounded, we direction-align \svar \ by multiplying it by $-1$.

\vspace{-1em}
$$s_{svar} = -1 * \frac{1}{H} \sum_{l=l_s}^{l_e} \sum_{h=1}^H\sum_{i=1}^N A_t^{(l, h)}(t, I_i)$$
\vspace{-0.5em}

\pas \ \citep{hoang-xuan_pas_2025} also leverages attention, like \svar, but sums attention from the generated token to previously generated caption tokens. 
The intuition is that if there is more attention distributed to caption tokens, less is distributed to the image which is more likely to occur with hallucinations.
By default, \pas \ uses attention from the first layer and averages across $H$ attention heads, but we conduct a hyperparameter search for layer selection. 
In the equation below, $m$ is the number of prompt tokens, $t$ is the current token's index, and $j$ represents indices of all prior tokens in the caption.

\vspace{-1em}
$$s_{pas} = \frac{1}{H}\sum_{h=1}^H\sum_{j=m+1}^{t-1} A_t^{(0, h)}(t, j)$$
\vspace{-0.5em}

\subsubsection{Token position baseline}

The token position baseline uses the position of the first token of a word in the caption, beginning at the start of the caption, i.e., ignoring the image and prompt instructions tokenization.
Most of the VLMs consistently allocate the same number of tokens per image because they resize them, but \qwen \ does not. We also tested if accounting for the image's token count with Qwen resulted in a better token position baseline, but it did not. 

\subsection{Reproducibility details, word-matching, \& hyperparameters}\label{app:hparams}

In this section, we provide more details regarding reproducibility of results including experimental settings, our word-matching algorithm, and hyperparameters we used. Code is publicly available at \url{https://github.com/princetonvisualai/hallucination-detection-robustness}

\textbf{Setup.} Following prior work \citep{park_glsim_2025, jiang_interpreting_2025, jiang_devils_2025}, we randomly select 5,000 images from the MSCOCO 2014 validation set to evaluate the detection methods. Concretely, we selected these with \texttt{seed=0}.
For all VLMs, we set the following generation hyperparameters: \texttt{sampling=False}, \texttt{temp=0}, \texttt{max\_new\_tokens=512}, and \texttt{num\_beams=1}.
We additionally fixed all \texttt{seed=0} for deterministic outputs and reproducibility.
All experiments were conducted in internal compute clusters with either 1 RTX3090 GPU with 24GB of RAM or 1 L40 GPU with 48 GB of RAM for larger VLMs (e.g., \qwen \ or \llavalg).
All experiments were allocated 8 CPUs and while most experiments took 3-4 GPU-hours, no experiment exceeded 48 GPU-hours. \qwen \ tended to take the longest because its captions are the longest.
Total compute time is estimated to be roughly 120 GPU-hours. This does not include preliminary and exploratory experiments that did not make it into the final draft, which we estimate to be roughly 72 GPU hours.

Our setup in determining which token and word instances to use were based on prior work \citep{park_glsim_2025, jiang_devils_2025}. When a word spans multiple tokens, we always use the hallucination scores for the first token \citep{park_glsim_2025, jiang_devils_2025}, and when a word appears multiple times in the caption, we also use the first occurrence, as in \citet{park_glsim_2025}. 
This assumes that applications of hallucination detection will be more interested in catching hallucinations of objects, not necessarily instances of objects. Thus, we focus on whether a method can detect whether an object is hallucinated to begin with by using the first instance.

We selected the six VLMs in order to have a diversity of architectures and methods. 
For example, \llavasm, \llavalg, and \instructblip \ all use Vicuna as an LLM backbone while \qwen \ and \internvl \ do not.
\qwen \ performs adaptive image sizing while other VLMs use fixed image sizes.
\instructblip \ considers the instruction prompt when computing the image representations and \internvl \ uses a pre-trained language model as a vision-language adaptor such that the image representations are shaped by language before they enter the LLM backbone. 
Other VLMs, however, do not consider language in the visual representation and use lightweight adaptors.

MSCOCO \citep{lin_microsoft_2014} was accessed via \url{https://cocodataset.org/#download}, and we use the 2014 splits.
For robustness experiments, data for Objects365 was accessed via OpenDataLab at \url{https://opendatalab.com/OpenDataLab/Objects365_v1}.
All VLMs were accessed via HuggingFace at the following URLs:
\begin{itemize}
    \item \llavasm: \url{https://huggingface.co/llava-hf/llava-1.5-7b-hf}
    \item \llavalg: \url{https://huggingface.co/llava-hf/llava-1.5-13b-hf}
    \item \instructblip: \url{https://huggingface.co/Salesforce/instructblip-vicuna-7b}
    \item \qwen: \url{https://huggingface.co/Qwen/Qwen3-VL-8B-Instruct}
    \item \internvl: \url{https://huggingface.co/OpenGVLab/InternVL2-8B}
    \item \gemma: \url{https://huggingface.co/google/gemma-4-E4B-it}
\end{itemize}

\textbf{Word-Matching.} In this work we meticulously developed a robust word-matching system to account for edge cases when identifying which tokens corresponded to which word.
While prior work \citep{park_glsim_2025, jiang_devils_2025} simply obtained the first instance that the desired token appeared in the output, we found that there were edge cases where this would give a false match.
For example, if the word was ``table,'' depending on the tokenizer, if the word ``tablecloth'' appeared before ``table'' in the sentence, this would be the word that was returned, resulting in a false match.
To mitigate this, we included additional checks to ensure the word was a standalone word, such as ensuring that the word was preceded and followed by a space or punctuation, accounting for plural forms.

Another difficulty was if a tokenizer does not detokenize with spaces, such that sometimes two words when placed next to each other appear like a potential object. For example ``a calm man'' becomes ``acalman'' and the object ``man'' becomes a false match.
To mitigate this issue, we perform checks on the preceding and following words in the original and decoded captions, if applicable, and if it is a false match, we skip this instance and check the next occurrence of the word.

While this level of granularity in word matching may not have a large effect in these evaluation settings, if these methods are to be deployed in real-world settings, greater care must be taken to ensure proper token matching. Even with our additional checks, we do observe there are bound to be cases that are not caught.

\textbf{Addressing CHAIR's context-insensitivity.} One of the limitations of CHAIR is that it uses text-based string matching and thus is insensitive to the context. For example, if a VLM generates the word ``orange'' to describe the color, CHAIR will flag this as the object orange being mentioned. 
To exclude this systematic issue from our results, we omit the word ``orange'' from all evaluations presented in the main text.
However, we addit

\textbf{Hyperparameter Selection.} 
For methods with hyperparameters (\glsim, \svar, \cl, and \pas) we aim to stay as faithful as possible to the original work. 
If a VLM was reported in the original paper (\llavasm \ and \llavalg), we use the reported hyperparameters.
If not, we start with the hyperparameters reported for \llavasm \ or \llavalg \ and then conduct a hyperparameter search by evaluating over $500$ samples from the 2014 training split of MSCOCO \citep{lin_microsoft_2014}. 
We choose the hyperparameter configuration that jointly maximizes AUROC and AUPRC on the $500$ samples.
If there is a tie, we prioritize maximizing AUPRC, as it is a better metric for unbalanced classes.
We report the hyperparameters for MSCOCO in \tabref{tab:hparams_mscoco} and for Objects365 in \tabref{tab:hparams_o365}. \looseness=-1

\glsim \ has four hyperparameters: the weight parameter $w$, the number of image patches $k$, the layer of the image and prompt embeddings $l$, and the layer of the object token embedding $l'$. 
\svar \ has two hyperparameters: the start ($s$) and end ($e$) layers that attention is extracted from. 
\cl \ and \pas \ each have one hyperparameter: the layer ($l$) at which we extract representations or attention patterns.

\begin{table}[h]
    \centering
    \caption{Hyperparameter values for \glsim, \svar, \cl, and \pas \ on all VLMs with MSCOCO. VLMs under the horizontal line are different versions of evaluated VLMs}
    \resizebox{0.8\textwidth}{!}{
        \begin{tabular}{lllll|ll|l|l}
        & \multicolumn{4}{c|}{\glsim} & \multicolumn{2}{c|}{\svar} & \multicolumn{1}{c|}{\cl} & \multicolumn{1}{c}{\pas}\\
        VLM & $w$ & $k$ & $l$ & $l'$ & $s$ & $e$ & $l$ & $l$ \\
        \midrule
        \llavasm      & 0.6 & 32 & 32 & 31 & 5 & 18 & 30 & 0 \\
        \llavalg      & 0.6 & 32 & 40 & 38 & 5 & 18 & 25 & 0 \\
        \instructblip & 0.8 & 4 & 32 & 32 & 1 & 7  & 30 & 0 \\
        \qwen         & 0.3 & 8 & 35 & 35 & 2 & 21 & 20 & 14 \\
        \internvl     & 0.8 & 8 & 23 & 32 & 1 & 10 & 10 & 9 \\
        \gemma  & 0.7 & 16 & 42 & 33 & 1 & 20 & 10 & 9 \\

        \midrule
        InternVL3.5-8B & 0.7 & 8 & 33 & 36 & 1 & 18 & 15 & 35 \\
        Qwen3.5-9B    & 0.7 & 32 & 31 & 31 & 2 & 5 & 15 & 2 \\
        \bottomrule
        \end{tabular}
    }
    \label{tab:hparams_mscoco}
\end{table}

\begin{table}[h]
    \centering
    \caption{Hyperparameter values for \glsim, \svar, \cl, and \pas \ on all VLMs with Objects365. NA indicates using the maximum across all layers yields best results.}
    \resizebox{0.8\textwidth}{!}{
        \begin{tabular}{lllll|ll|l|l}
        & \multicolumn{4}{c|}{\glsim} & \multicolumn{2}{c|}{\svar} & \multicolumn{1}{c|}{\cl} & \multicolumn{1}{c}{\pas}\\
        VLM & $w$ & $k$ & $l$ & $l'$ & $s$ & $e$ & $l$ & $l$ \\
        \midrule
        \llavasm      & 0.6 & 32 & 32 & 31 & 5 & 10 & NA & 0 \\
        \llavalg      & 0.5 & 8 & 37 & 40 & 5 & 18 & 25 & 0 \\
        \instructblip & 0.7 & 8 & 32 & 32 & 1 & 15 & 20 & 0 \\
        \qwen         & 0.1 & 8 & 35 & 35 & 10 & 35 & 30 & 34 \\
        \internvl     & 0.6 & 8 & 20 & 32 & 5 & 18 & NA & 9 \\
        Gemma-4-E4B-it    & 0.1 & 8 & 42 & 30 & 1 & 20 & 10 & 9 \\
        \bottomrule
        \end{tabular}
    }
    \label{tab:hparams_o365}
\end{table}

\section{Limitations}\label{app:limitations}

In this section, we non-exhaustively elaborate on limitations of the CHAIR evaluation setting (\secref{app:chair_lims}) and of the limited scope of our analyses (\secref{app:analysis_lims}).

\subsection{Limitations of CHAIR evaluation}\label{app:chair_lims}

Although CHAIR \citep{rohrbach_object_2018} is commonly used for labeling hallucinations, we acknowledge that it has limitations, as other evaluation settings try to address \citep{ben-kish_mitigating_2024, li_evaluating_2023}.
Namely, the limitations come from the method using direct string matching to label objects and their synonyms without considering the context which recent work aims to mitigate \citep{ben-kish_mitigating_2024}.
For example, if there is an image of an empty park and the caption consists of a hypothetical phrase such as ``this park looks like a great place for people to enjoy a sunny afternoon,'' CHAIR would identify ``people'' as a hallucination. 
However, in the full context of the caption, the model is not actually hallucinating.
CHAIR faces similar limitations for other phrases such as ``the shirt says `cat in the house'''. 
Further, with CHAIR, it is not possible to distinguish between multiple instances of the same object class. For example, if an image contains one person on the left, but a caption says ``there is one person on the left and one person on the right,'' this is a hallucination that CHAIR is unable to detect.  

Lastly, CHAIR cannot differentiate between different parts of speech, such as when the word ``orange'' appears as the fruit or the color because it does not take into account contextual cues. 
Because of this, we noticed that some VLMs, primarily \qwen \ and \internvl, frequently hallucinated the word ``orange,'' but manual investigation demonstrated that these instances were from references to the color rather than the object. 
Because CHAIR mislabels these as hallucinations, we omitted all instances of the word ``orange'' from all analyses. However, as we show in \secref{app:orange_results}, the results with and without the object ``orange'' do not lead to different conclusions. 

\subsection{Limited scope of analyses}\label{app:analysis_lims}

While our analyses present interesting findings, we acknowledge a few limitations here, such as only considering object hallucinations, a subset of hallucination detection methods.
First, our analyses were limited to detecting object hallucinations in image captioning settings and do not consider attribute or relational hallucinations nor various settings like multiple choice question-answering or dialogues.
Second, in order to operate in a well-constrained environment, we only considered hallucination detection methods that were training-free and calculated hallucination scores based on a single pass of the VLM's generation.
However, there are many other types of hallucination detection methods such as those that require multiple passes \citep{leng_mitigating_2024, wang_mitigating_2024, zhao_mitigate_2025, liao_visionamplified_2026, zhang_vluncertainty_2024, farquhar_detecting_2024}, or that utilize auxiliary detector models \citep{kaul_throne_2024, ge_visual_2024, yu_hallucidoctor_2024, jing_faithscore_2024, chen_unified_2024}.
Third, while we tested on a variety of VLMs with different language model backbones, we were limited to smaller open-source models due to computational constraints.
These may exhibit different hallucination rates than frontier models.
Lastly, while we report results on two independently curated datasets, MSCOCO \citep{lin_microsoft_2014} and Objects365\citep{shao_objects365_2019}, these datasets may not be fully representative of the real-world.

Future work should consider expanding upon these limitations, considering a wider variety of hallucination types, VLM response settings, a broader suite of hallucination detection methods, and various VLMs, and additional datasets. 
It is important to conduct experiments in a variety of settings to evaluate whether the findings hold in diverse conditions or not.

\subsection{Potential impacts}\label{app:potential_impacts}

Potential positive impacts of our work include motivating future work to contribute analyses of different aspects of hallucination detection methods, more thorough evaluation of hallucination detection methods, and responsible deployment of hallucination detection tools.
While the positive potential impacts outweigh the negative, potential negative impacts could include overconfidence in particular detection methods and developing user-facing applications before robustly analyzing the methods. 
However, because our work is aimed to encourage researchers and developers to take more care when evaluating hallucination detection methods, we do not foresee large negative societal impact. 
\section{Experimental details \& additional results}\label{app:exp_details_results}

In this section, we go through each analysis in the  order presented in the main paper and elaborate on experimental details and present results from other VLMs not reported in the main paper. Because figures tend to be quite large, we place them at the end of this section (\appendixref{app:exp_details_results}).

\subsection{Token position distribution of real vs. hallucinated objects}\label{app:token_pos}

\begin{figure}[h!]
    \centering
    \includegraphics[width=0.48\linewidth, trim={0 0.1in 0 0.08in}, clip]{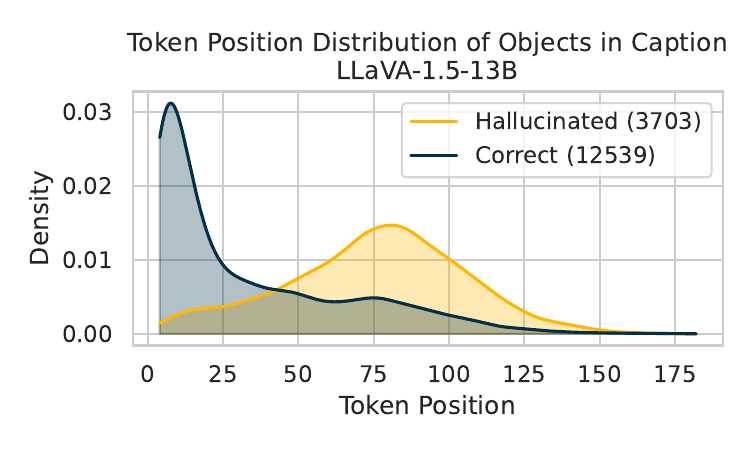}
    \includegraphics[width=0.48\linewidth, trim={0 0.1in 0 0.08in}, clip]{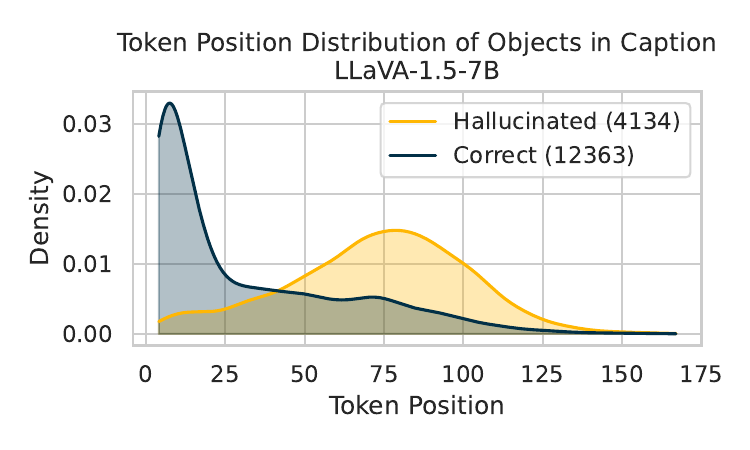}
    \includegraphics[width=0.48\linewidth, trim={0 0.1in 0 0.08in}, clip]{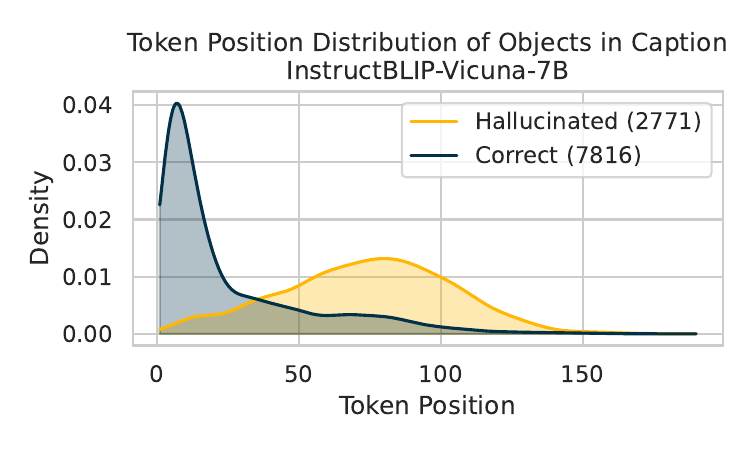}
    \includegraphics[width=0.48\linewidth, trim={0 0.1in 0 0.08in}, clip]{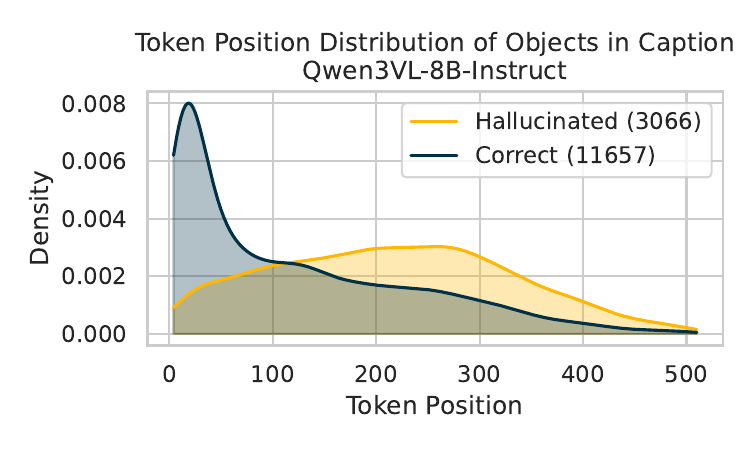}
    \includegraphics[width=0.48\linewidth, trim={0 0.1in 0 0.08in}, clip]{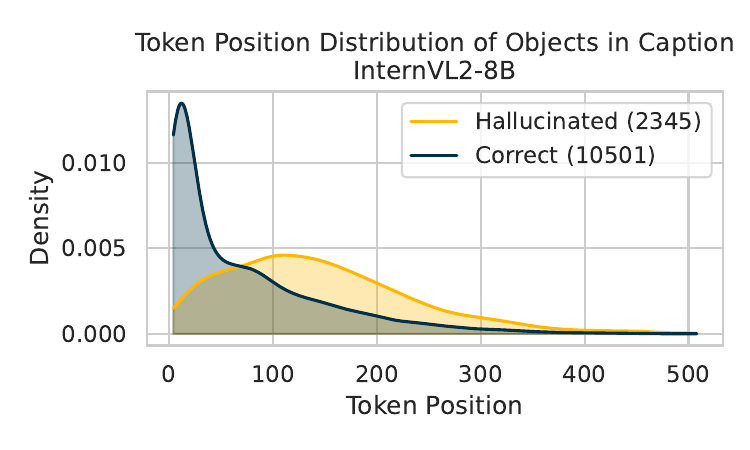}
    \includegraphics[width=0.48\linewidth, trim={0 0.1in 0 0.08in}, clip]{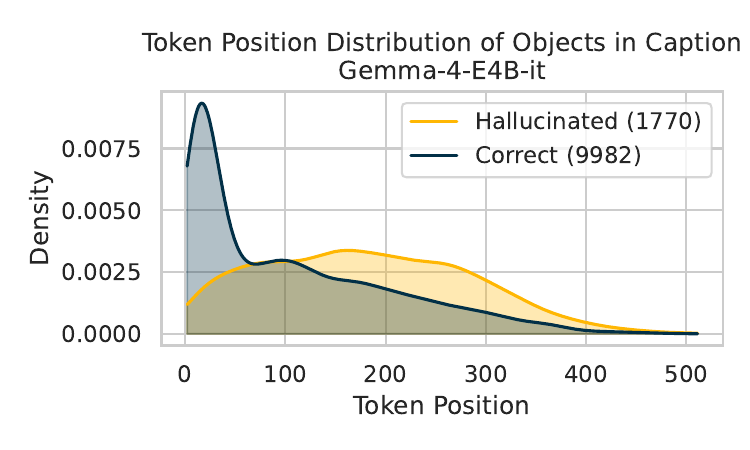}
    \caption{\textbf{Objects mentioned at later positions in the caption are more likely to be hallucinated.} Plots demonstrate kernel density functions of token position distributions of hallucinated and correct (or real) objects. The effect is especially pronounced for LLaVA and InstructBLIP, which leverage Vicuna \citep{team_vicuna_2023} backbones. While less pronounced for non-Vicuna VLMs, the trend still persists. Number of hallucinated and correct objects for each VLM is labeled in the legend.}
    \label{fig:token_pos_dist}
\end{figure}

This is more of a preliminary analysis to demonstrate that for all VLMs, later token positions are more likely to be hallucinations. This observation has been alluded to or demonstrated in much of prior work as well \citep{vo_vision_2026, mckenna_sources_2023, zhou_analyzing_2024, li_hidden_2025, sun_exploring_2025, favero_multimodal_2024}. 
For this analysis, we simply partition the objects into real and hallucinated and plot the distribution of token position. Results can be found in \figref{fig:token_pos_dist}.

\subsection{Correlation analysis between detection methods and token position}\label{app:correlations}

Here we supply additional figures and graphs to demonstrate the correlation between hallucination score and token position.

\begin{figure}[h!]
    \centering
    \includegraphics[width=\linewidth, trim={1.5in 1.25in 1.5in 1.25in}, clip]{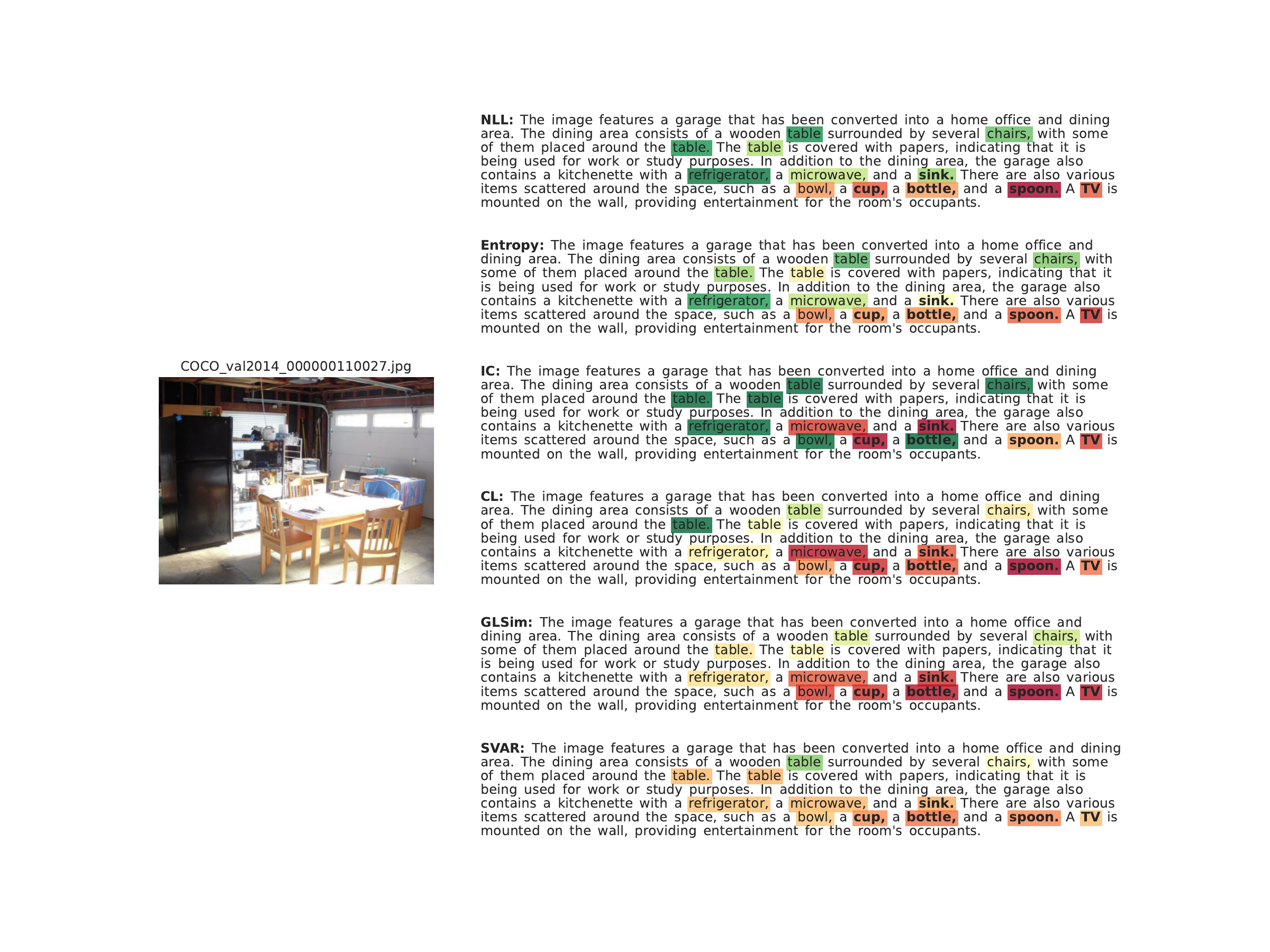}
    \caption{\textbf{Qualitative example of hallucination score and token position correlation (1/4)}. \svar, \glsim, and \cl \ tend to score later words as more likely to be hallucinated, \emph{which sometimes is consistent with the hallucination patterns of the VLM}. Captions are repeated to demonstrate scores from each detection method, \textbf{bold} indicates hallucinated object, \sethlcolor{correcthighlight}\hl{green} indicates low hallucination score, and \sethlcolor{hallhighlight}\hl{red} indicates high. Image caption generated is by \llavalg.}
    \label{fig:qualitative_results1}
\end{figure}

\begin{figure}[h!]
    \includegraphics[width=\linewidth, trim={1.5in 1.25in 1.5in 1.25in}, clip]{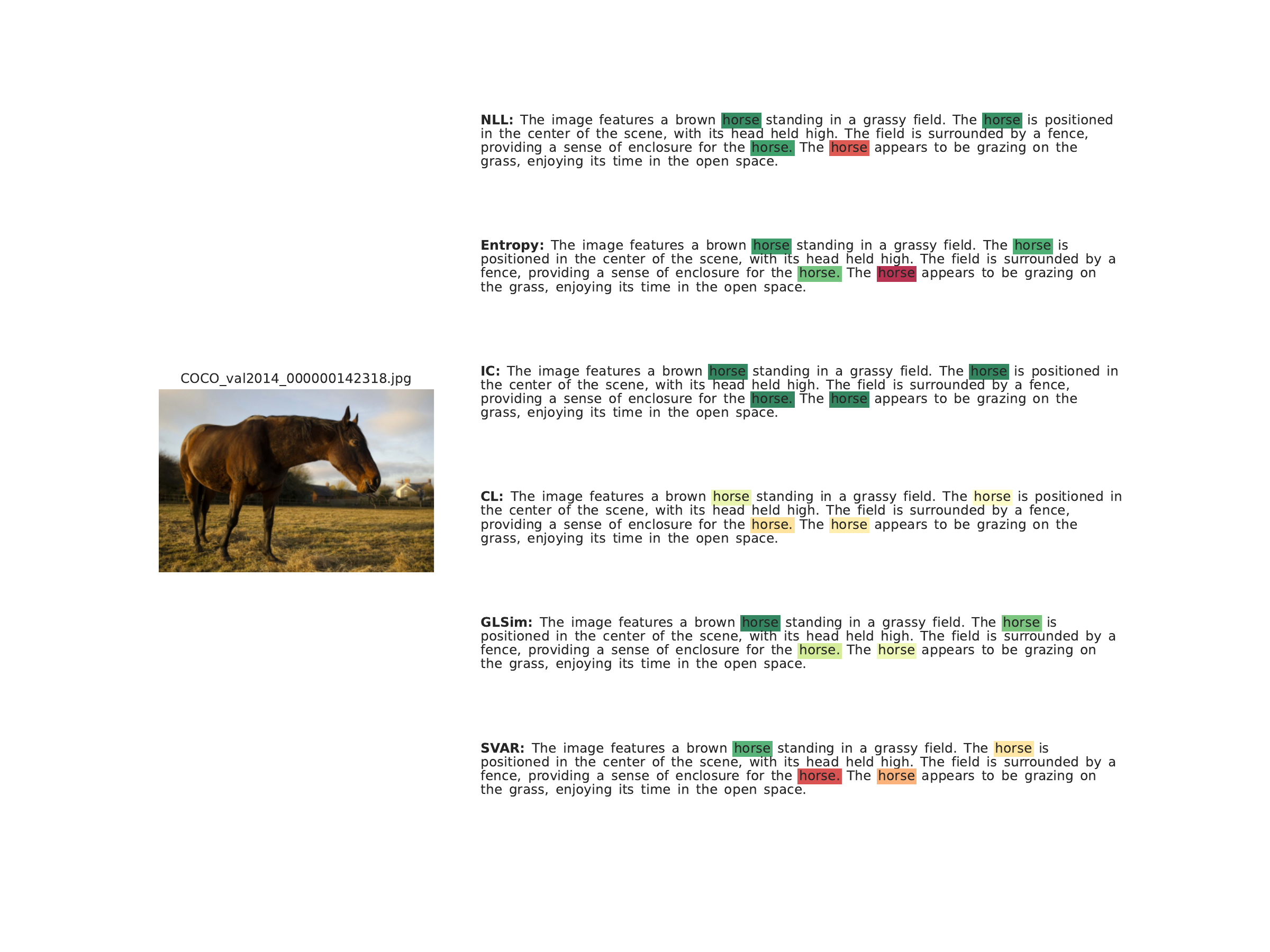}
    \caption{\textbf{Qualitative example of hallucination score and token position correlation (2/4)}. \svar, \glsim, and \cl \ tend to score later words as more likely to be hallucinated, \emph{even if the object is the same}. Captions are repeated to demonstrate scores from each detection method, \textbf{bold} indicates hallucinated object, \sethlcolor{correcthighlight}\hl{green} indicates low hallucination score, and \sethlcolor{hallhighlight}\hl{red} indicates high. Image caption generated is by \llavalg.}
    \label{fig:qualitative_results2}
\end{figure}

\begin{figure}[h!]
    \centering
    \includegraphics[width=\linewidth, trim={1.5in 1.25in 1.5in 1.25in}, clip]{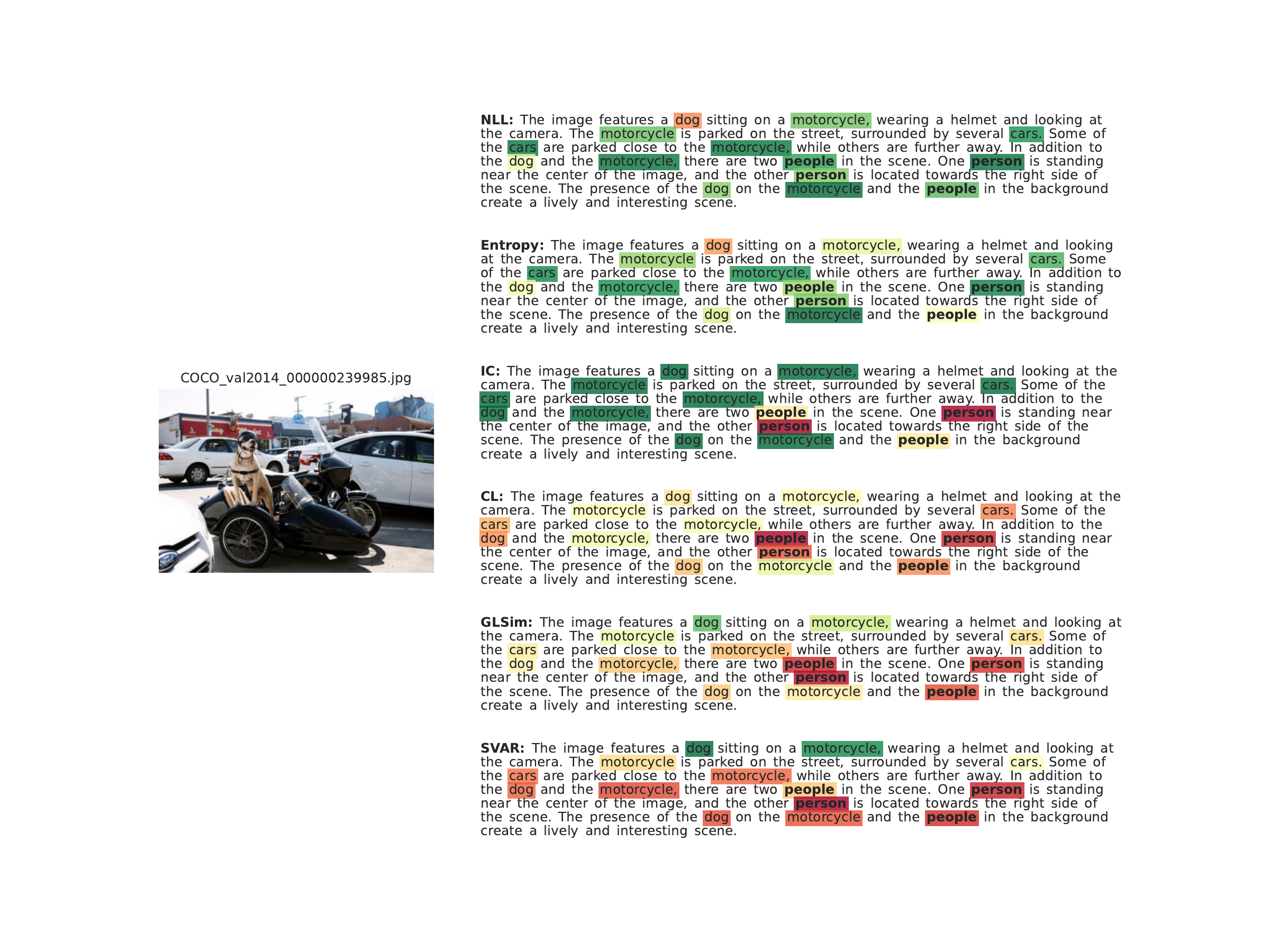}
    \caption{\textbf{Qualitative example of hallucination score and token position correlation (3/4)}. \svar, \glsim, and \cl \ tend to score later words as more likely to be hallucinated, \emph{whereas \nll \ and \ent \ are less likely to do so}. Captions are repeated to demonstrate scores from each detection method, \textbf{bold} indicates hallucinated object, \sethlcolor{correcthighlight}\hl{green} indicates low hallucination score, and \sethlcolor{hallhighlight}\hl{red} indicates high. Image caption generated is by \llavalg.}
    \label{fig:qualitative_results3}
\end{figure}

\begin{figure}[h!]
    \centering
    \includegraphics[width=\linewidth, trim={1.5in 1.25in 1.5in 1.25in}, clip]{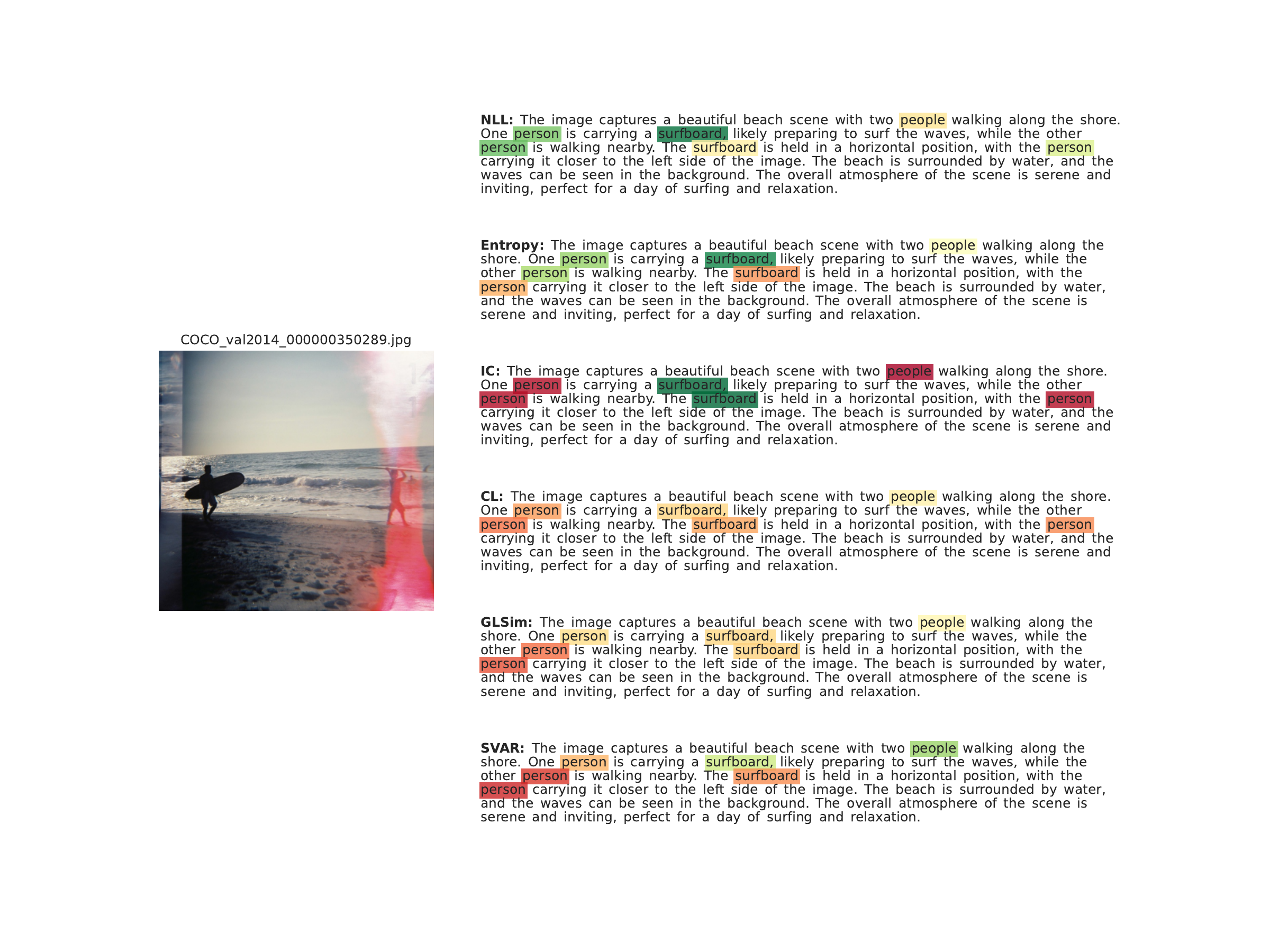}
    \caption{\textbf{Qualitative example of hallucination score and token position correlation (4/4)}. \svar, \glsim, and \cl \ tend to score later words as more likely to be hallucinated, \emph{even for correct objects}. Captions are repeated to demonstrate scores from each detection method, \textbf{bold} indicates hallucinated object, \sethlcolor{correcthighlight}\hl{green} indicates low hallucination score, and \sethlcolor{hallhighlight}\hl{red} indicates high. Image caption generated is by \llavalg.}
    \label{fig:qualitative_results4}
\end{figure}

In \figref{fig:qualitative_results1}-~\ref{fig:qualitative_results4}, we present qualitative examples of the correlations in captions generated by \llavalg. 
As demonstrated by the examples, \svar, \glsim, \ and to a certain degree \cl \ tend to score objects later in the caption as more likely to be hallucinated, even if the same object was deemed as less likely to be hallucinated earlier in the caption. 
While our analyses only evaluates the \emph{first} occurrence of each word in a caption, we visualize all instances to show how a word's hallucination score can change depending on its location in the caption.
\textbf{Bolded text} indicates the object is a hallucination according to CHAIR \citep{rohrbach_object_2018}. \sethlcolor{green}\hl{Green} indicates the detection method scored this word as a low probability of hallucination while \sethlcolor{red}\hl{red} indicates a higher probability of hallucination.

\begin{figure}[h!]
    \centering
    \includegraphics[width=0.48\linewidth, trim={0 0.1in 0 0.08in}, clip]{figures/llava1.5-13b/hd_spearman_correlations.pdf}
    \includegraphics[width=0.48\linewidth, trim={0 0.1in 0 0.08in}, clip]{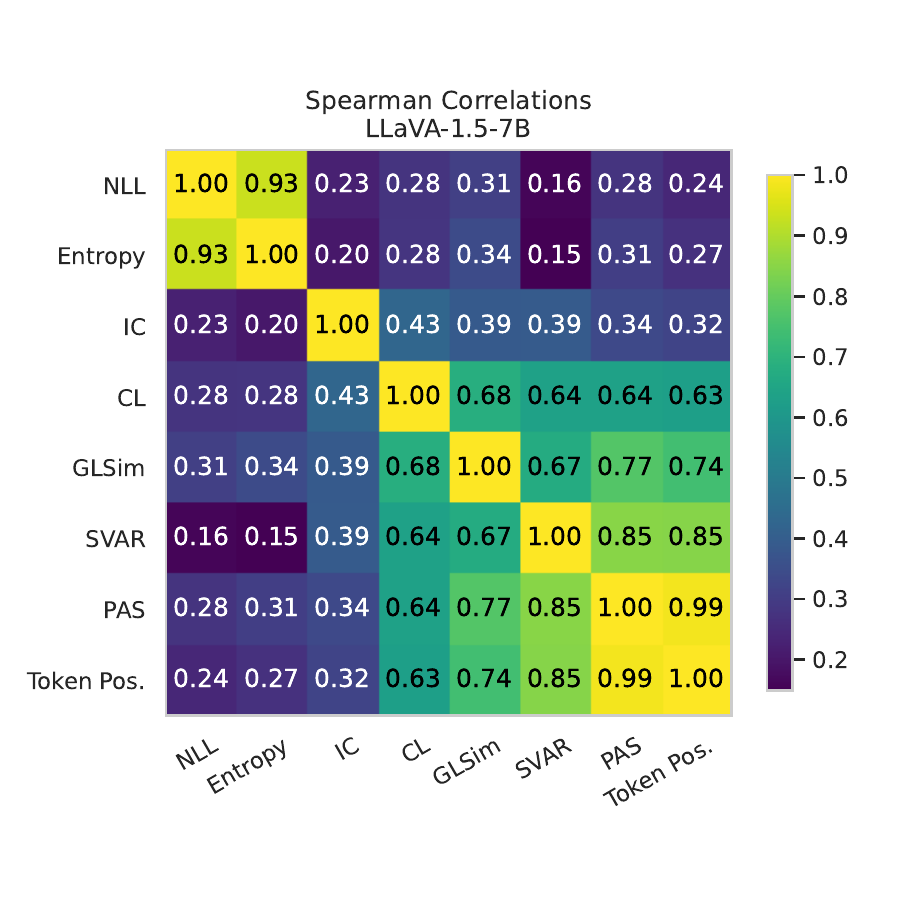}
    \includegraphics[width=0.48\linewidth, trim={0 0.1in 0 0.08in}, clip]{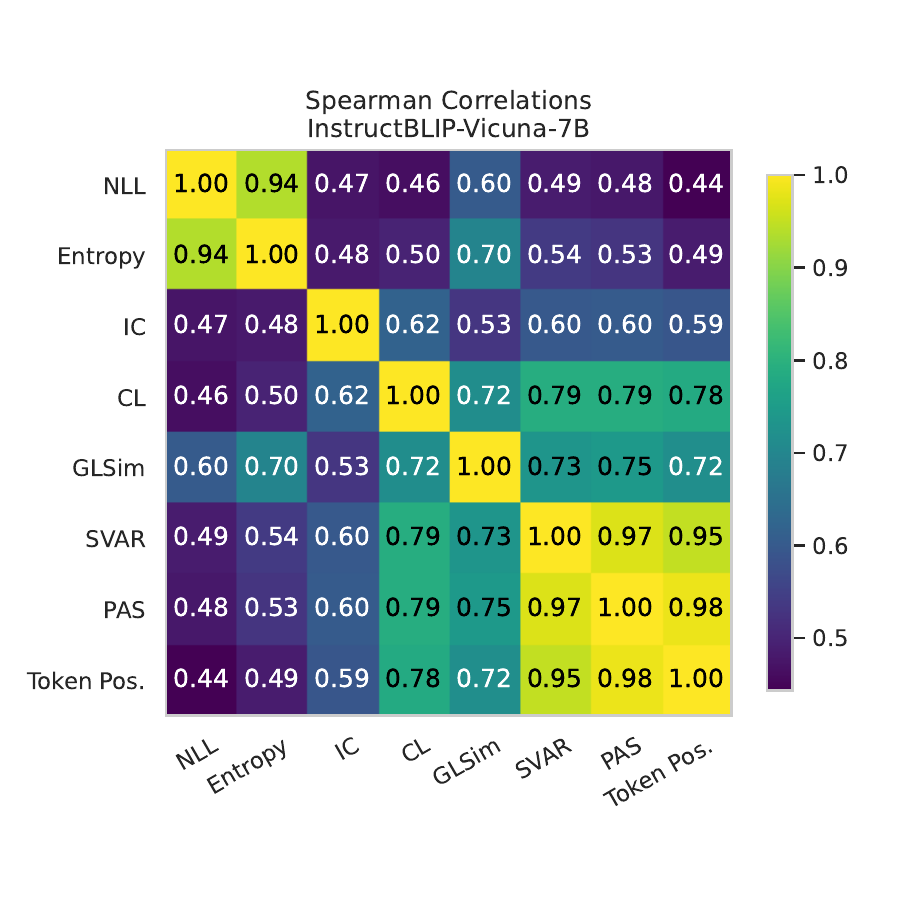}
    \includegraphics[width=0.48\linewidth, trim={0 0.1in 0 0.08in}, clip]{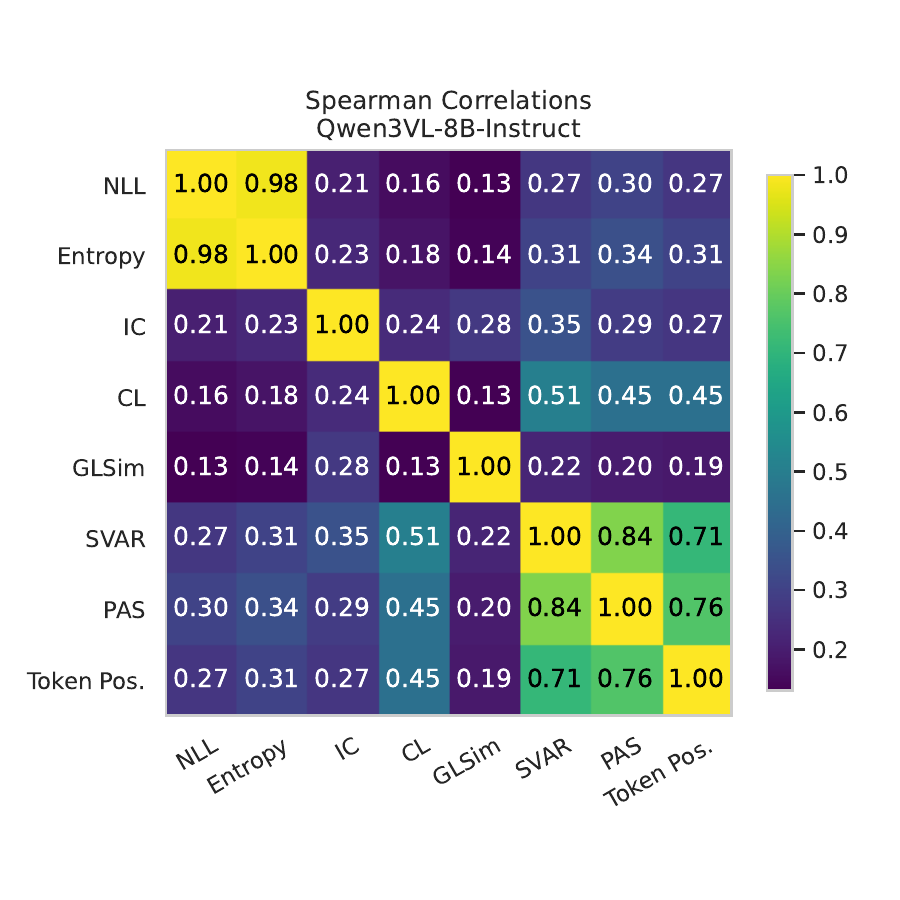}
    \includegraphics[width=0.48\linewidth, trim={0 0.1in 0 0.08in}, clip]{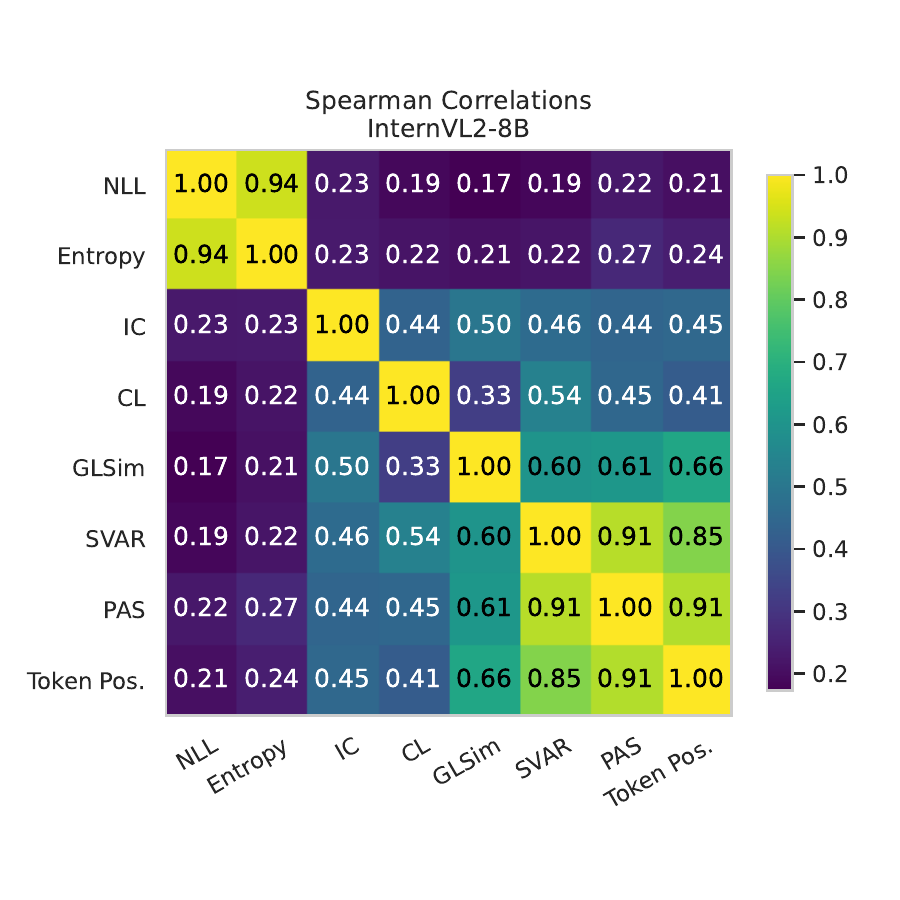}
    \includegraphics[width=0.48\linewidth, trim={0 0.1in 0 0.08in}, clip]{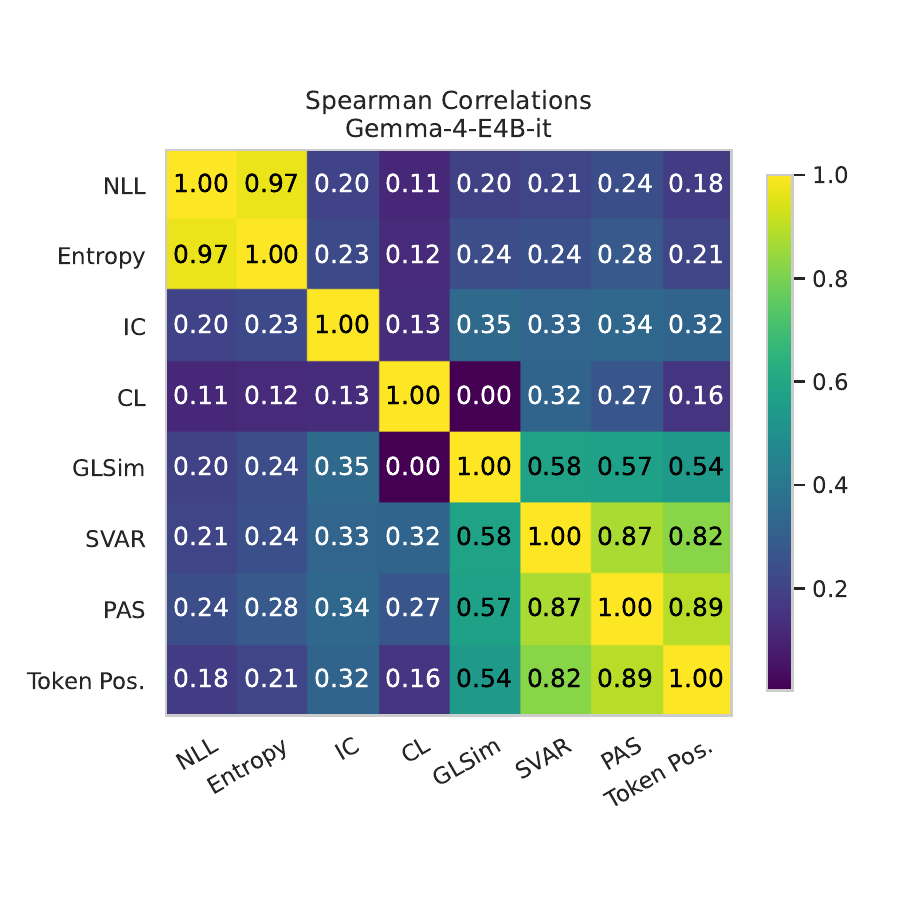}
    \caption{\textbf{Correlations matrices between all methods and token position for all VLMs.} The high correlation between \svar \ and token position and between black box methods remains consistent across all VLMs. Other correlations appear to be more VLM-dependent.}
    \label{fig:addit_correlations}
\end{figure}

Further, in \figref{fig:addit_correlations}, we show correlation matrices for all VLMs, including \llavalg \ again for reference.
We computed pairwise Spearman's rank correlation between the object-wise scores for all methods and the object's token position.
We chose Spearman's over Pearson's because the former does not assume linear relationships between the methods. 

As shown in the figure, the correlations between \svar \ and token position and between black box methods remain consistent across different VLM architectures. 
However, for other pairwise correlations, especially with white box methods, we observe variation across different VLMs, which we provide some potential explanations for here.
For \instructblip \ and \internvl, the correlations between IC and other white box methods are higher than for the other four VLMs. 
We hypothesize this is due to the influence of language in shaping the image representations \textit{before} they enter the LLM backbone. 
In other VLMs, the image representation is less affected by text, especially since the image is always before the instruction text, which could lead \ic \ to perform more differently in these cases.
We also note that \gemma \ exhibits very low correlation between \cl \ and \glsim, which is surprising, given the similarity of the methods' underlying computations.
All correlations were statistically significant ($p < 0.001$) even when accounting for multiple comparisons.

\subsection{Reliance of detection on token position}\label{app:remove_replace}

\subsubsection{Experimental details}

To replace sentences with image-agnostic but task-relevant text, we generated a pool of sentences with varying lengths using Gemini \citep{team_gemini_2025} with the following prompt: ``I need to replace sentences one at a time in image captions generated by a vision-language model, such as Qwen. I am replacing sentences with image-agnostic text, starting at the beginning of the caption. I need a variety of sentences that are different lengths, ranging from 10-50 tokens when using the llava-1.5-7b tokenizer. Starting at 10 tokens long and increasing by 5 tokens until you reach 50 tokens, give me a sentence that could replace a sentence in an image caption.''

Then we manually inspected them and checked the token lengths according to the different tokenizers. Specifically, we removed phrases that seemed to describe staged images and made small edits to adjust incorrect token lengths. Below are the pool of sentences followed by the number of tokens (using \llavasm \ tokenizer) and the number of words.

\begin{itemize}
    \item The overall composition of the image is quite balanced. (11, 9)
    \item Objects are arranged throughout the scene in a specific way, naturally or deliberately. (17, 13)
    \item Lighting plays a crucial role in defining the textures and colors that are visible across the entire frame. (22, 18)
    \item The perspective captured here offers a unique viewpoint that emphasizes the important relationship between the foreground elements and the background details. (25, 21)
    \item It is important to consider how the spatial distribution of these items contributes to the narrative being told by the visual information presented within this specific context. (30, 27)
    \item The way the light interacts with various surfaces creates a subtle range of highlights and soft shadows that defines the physical presence of everything within the area without highlighting one specific subject. (38, 32)
    \item Various elements across the entire area seem to fall into a pattern, where the distribution of shapes and colors allows the gaze to wander freely through the space without any single point dominating the overall visual experience. (43, 37)
    \item The overall atmosphere of this particular setting is characterized by a blend of various textures and shapes that come together to form a cohesive view, providing a sense of depth and perspective that defines the entire area being observed here. (47, 40)
\end{itemize}

When replacing preceding sentences, we replaced one sentence at a time by computing the token length of the sentence and replacing it with the one from the pool that was closest in token length. 
We acknowledge this is not perfect as it could result in sentences being repeated, but we selected to do this for simplicity yet maintaining readability.
Additionally, it is arguable that there is still \emph{some} information in these sentences and that to completely remove information, one should fill it with spaces or other filler text. 
We decided to use the sentence-wise replacement in order for the caption to still resemble natural language. 
However, this experiment could be done in future work.

Other than being solely dependent on token position (i.e., scores decrease significantly in the remove intervention but do not change in the replace intervention), we could observe a few other behaviors.
In some cases, the detection method could be dependent on both token position and semantic content. 
If this is the case, then its score may fluctuate in the replace intervention but we would still expect the hallucination scores to decrease when text is removed compared to when text is replaced.
However, in these latter cases, we cannot make as precise claims about the degree of dependency on only token position.
In other cases, detection methods could be modeling the congruity of the object in preceding context, as we hypothesize is the case with \nll \ and \ent. 

\subsubsection{Results for reliance on token position for all VLMs}

\begin{figure}[h!]
    \centering
    \includegraphics[width=0.7\linewidth, trim={0 0.1in 0 0.08in}, clip]{figures/llava1.5-13b/remove_replace.pdf}
    \includegraphics[width=0.7\linewidth, trim={0 0.1in 0 0.08in}, clip]{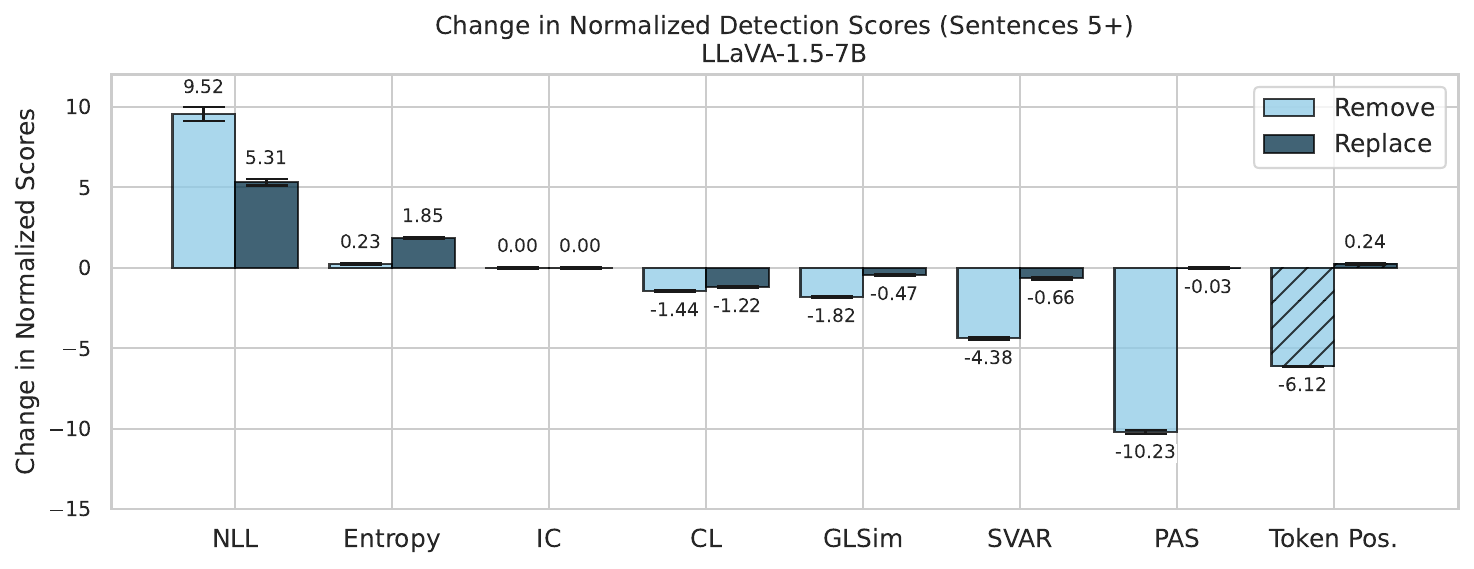}
    \includegraphics[width=0.7\linewidth, trim={0 0.1in 0 0.08in}, clip]{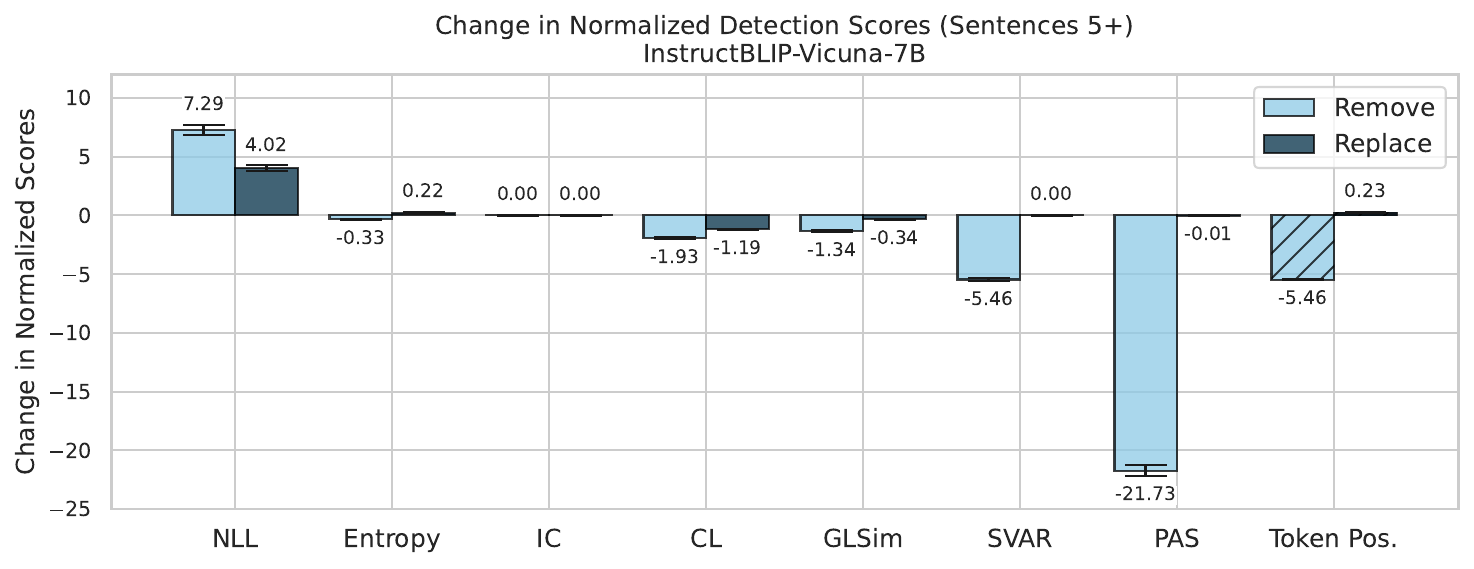}
    \includegraphics[width=0.7\linewidth, trim={0 0.1in 0 0.08in}, clip]{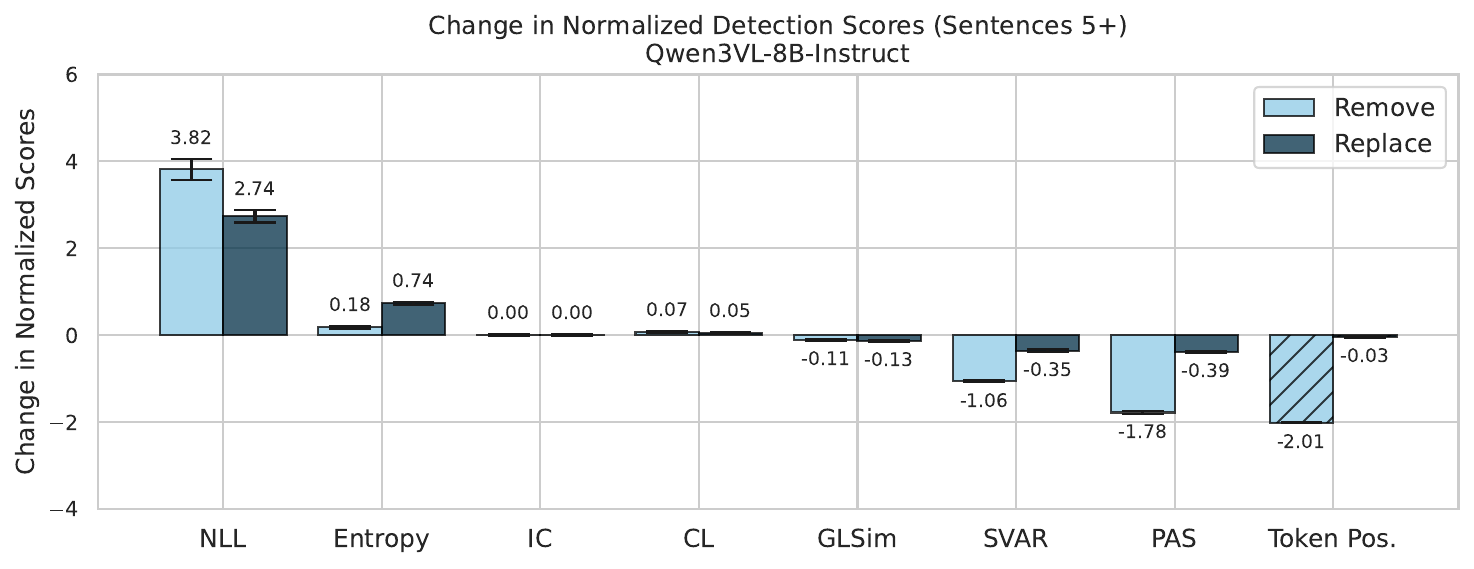}
    \includegraphics[width=0.7\linewidth, trim={0 0.1in 0 0.08in}, clip]{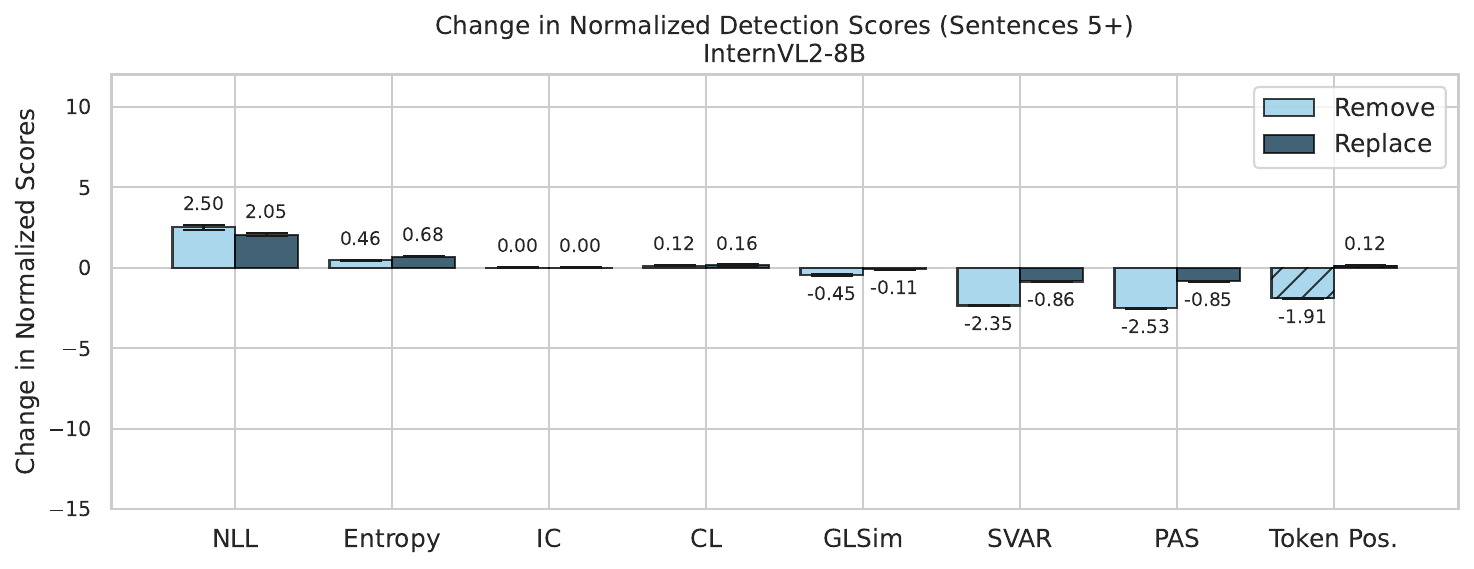}
    \includegraphics[width=0.7\linewidth, trim={0 0.1in 0 0.08in}, clip]{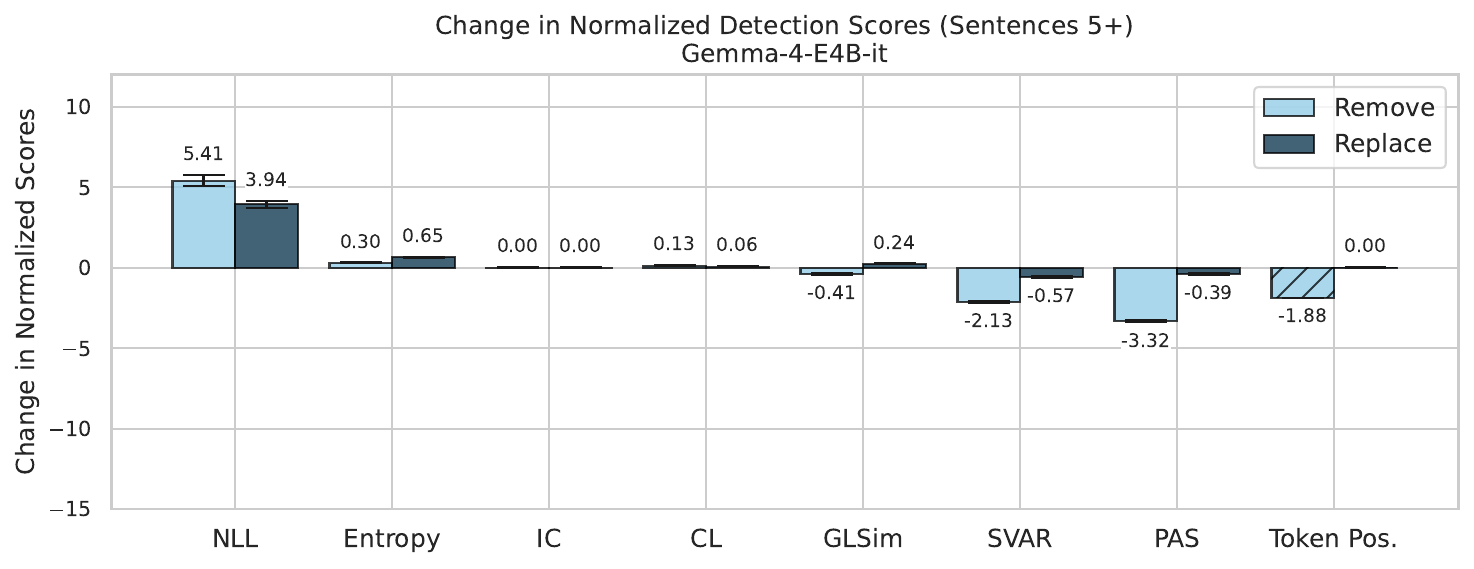}
    \caption{\textbf{Change in normalized hallucination scores when removing and replacing preceding context for all VLMs.} Trends for all methods hold across VLMs, but we note \cl \ and \glsim seem to be less text sensitive on certain VLMs. \looseness=-1}
    \label{fig:addit_remove_replace}
\end{figure}

In \figref{fig:addit_remove_replace}, we show how hallucination scores change for all detection methods and VLMs reported in \tabref{tab:overall_aucs}.
The trends described in the main paper still reliably hold. 
Black box methods increase hallucination scores (i.e., judging objects to be more likely hallucinated) while white box methods, especially attention-based \svar \ and \pas \ are more likely to decrease scores or remain the same.

We observe for newer VLMs (i.e., \qwen, \internvl, and \gemma), there is reduced direct token position dependence across white box methods, demonstrated by the relative size of the bars in the ``remove'' setting to be more comparable between white box methods and token position. 
This suggests that internal signals in newer models may be more robust to this heuristic. 
It additionally suggests that a single detection method can be leveraging different types of signals depending on the VLM; in other words, two VLMs' representations or attention patterns can carry different signals which would shape the detection methods' performance.

For other VLMs, we observe that either most of \cl \ and \glsim's change in score in the remove intervention is accounted for by the replace experiment or there is very little change for \cl.
This suggests that for some VLMs, \cl may be more sensitive to prior context, not just token position, and for others it is insensitive to either.
We further observe that the behavior of attention-based methods (\svar \ and \pas) also remains consistent across VLMs, suggesting these methods are consistently dependent on token position and  less so on context.

In \secref{sec:remove_replace}, we briefly mention a hypothesis of the mechanisms causing black box method scores to increase when removing or replacing preceding context. 
We elaborate further here, and this also applies to our interpretation of \pas \ with some of the VLMs.
One possible explanation for the behavior observed with black box methods is that these methods are highly sensitive to how well the current token belongs (or doesn't belong) with the preceding context, i.e., the perplexity of the current token.
Thus, when the preceding context is gone (in the removal case) or devoid of meaningful information (in the replacement case), naturally, the perplexity of the current token is high, the likelihood is low, and the black box methods score it as more likely to be hallucinated.

On the other hand \cl, \glsim, and \svar \ white box methods may also be sensitive to language model biases, such as sensitivity to position \citep{jiang_peek_2024, barbero_why_2025}. 
If this language-biased sensitivity outweighs other factors, then these methods would behave more similarly to token position, as \svar \ clearly does.

\subsection{Language vs. image-sensitive partitioning} \label{app:partition}

Here we provide further elaboration on the language vs. image-sensitive hallucination partitioning, as there may be concern of its sensitivity to out of distribution images.
To summarize, we partition hallucinations by feeding the generated caption with a \textit{blank} gray image into the VLM and examining whether the new predicted token is the same object or not \citep{zhang_mathverse_2025, zhao_mitigate_2025, favero_multimodal_2024, xie_vdpo_2024}.
We used a gray image as it more closely represents the mean pixel values compared to a black or white image.
If the predicted token is still the hallucinated object when the image is blank, it belongs to the language-based partition. 
This is because the VLM still hallucinates the object, even when there is no image \citep{zhang_mathverse_2025, zhao_mitigate_2025, favero_multimodal_2024, xie_vdpo_2024}.
Otherwise, the hallucination likely arises from image-sensitive sources which would be primarily attributed to visual deficiencies, which can include ambiguous visual representations or misalignment between vision and language, based on prior work \citep{liu_survey_2024, chen_survey_2026}.
We note that prior work tended to remove the image, but in our settings this is infeasible because the detection methods require an image representation.

Let us examine each partition (language vs. vision) separately.
The language partition includes hallucinated objects that are still hallucinated when given a blank image. 
We believe the objects that fall into this partition are safe from out of distribution effects; if they are still hallucinated even with an extremely out of distribution image, then the VLM is most likely not referring to the image to generate this word.

So, if the partition function is sensitive to out of distribution images, then likely these would fall into the ``image-sensitive'' partition.
This partition then would contain hallucinations from vision sources as well as out of distribution effects.
Nevertheless, our main claim is on the language-based hallucination partition, and not this second partition. Therefore such contamination does not affect our claim that the advantage of white box methods over black box methods is apparent only in the first partition, which is language-based hallucinations. 
While this partition would include vision-based as well as other sources of hallucinations, we hypothesize that a cleaner partition function would only highlight the performance differences since more language-based hallucinations would be appropriately partitioned.

Results for all VLMs reported in the main paper can be found in \figref{fig:partition}.
Here we see white box methods consistently outperform black box methods in language-based hallucinations but perform on par---or even \emph{under}perform---black box methods on image-sensitive hallucinations.
This trend is consistent across VLMs. 
For example, while black box methods outperform all white box methods on image-sensitive hallucinations for \qwen \ and \internvl, performance is more comparable across the two categories for other VLMs. 
These results suggest that white box methods' competitive advantage comes from detecting hallucinations that persist, even when an image is removed, i.e., the language-based hallucinations, suggesting a general absence of image sensitivity.
Surprisingly, \svar \ and \pas, which do not seem to exhibit image sensitivity, can detect these image-sensitive hallucinations better than random chance, suggesting there is a non-trivial interplay between signal dependencies. 
We leave this for future work to explore. 

\subsection{Quantifying image sensitivity and visual grounding}\label{app:visual}

This experimental setup was motivated by prior work that aims to identify the role of language in VLM hallucinations \citep{zhang_mathverse_2025, zhao_mitigate_2025, favero_multimodal_2024, xie_vdpo_2024}. 
While most works compare the outputs of the image and text prompt with the text prompt only, we compare the original setting to a setting with the text prompt with a blank image, since the white box detection methods require some image representation.
If a method is sensitive to the image, its performance should drop when removing the image but evaluating the original caption over the original labels.

\begin{figure}[h!]
    \centering
    \vspace{-0.2in}
    \includegraphics[width=0.7\linewidth, trim={0 0.1in 0 0.08in}, clip]{figures/llava1.5-13b/auprc_blank_performance_diff.pdf}
    \includegraphics[width=0.7\linewidth, trim={0 0.0in 0 0.00in}, clip]{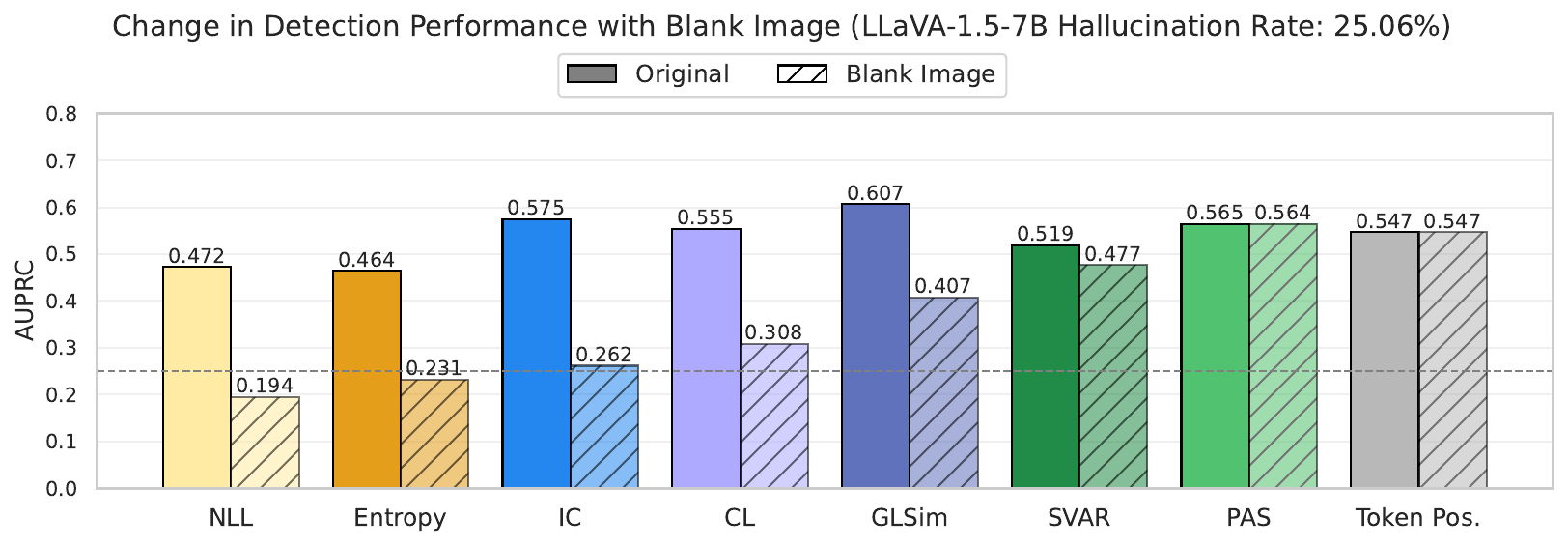}
    \includegraphics[width=0.7\linewidth, trim={0 0.0in 0 0.00in}, clip]{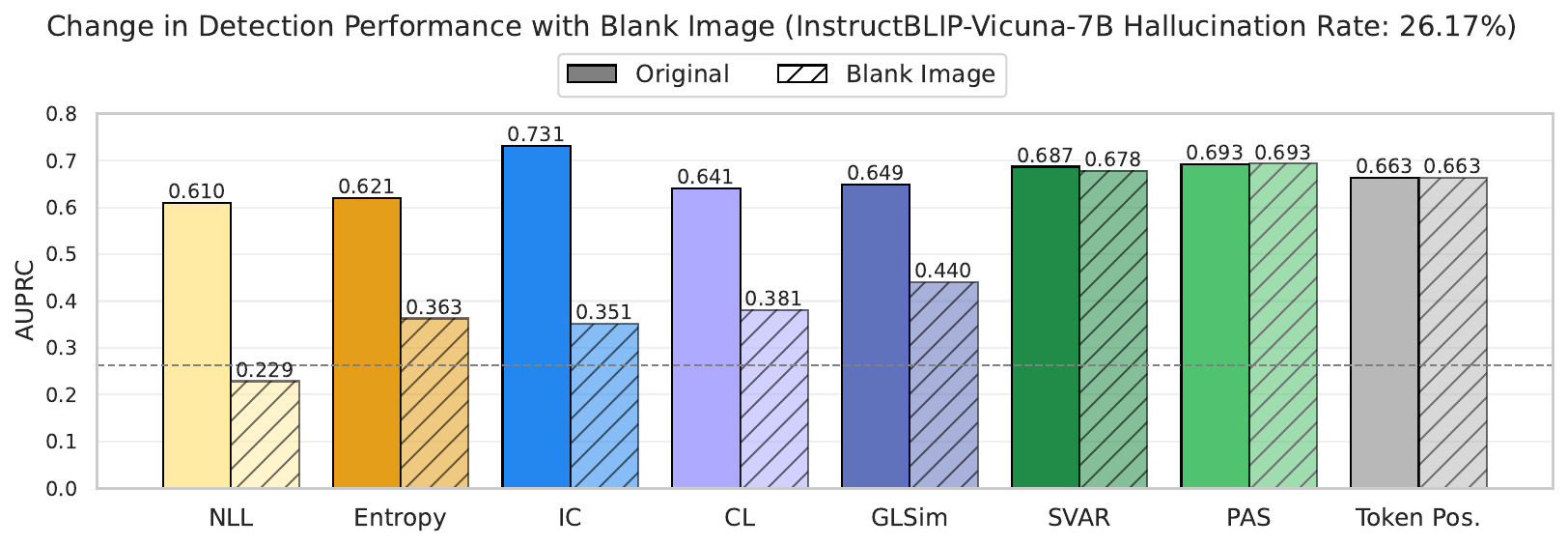}
    \includegraphics[width=0.7\linewidth, trim={0 0.0in 0 0.00in}, clip]{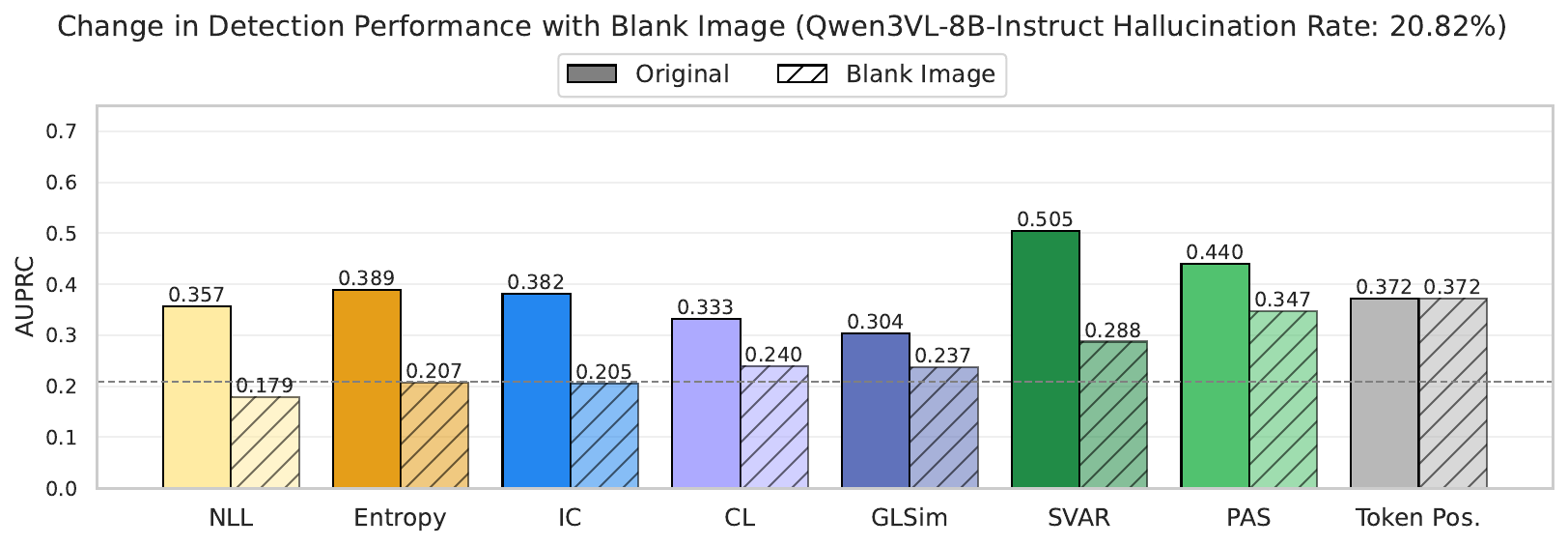}
    \includegraphics[width=0.7\linewidth, trim={0 0.0in 0 0.00in}, clip]{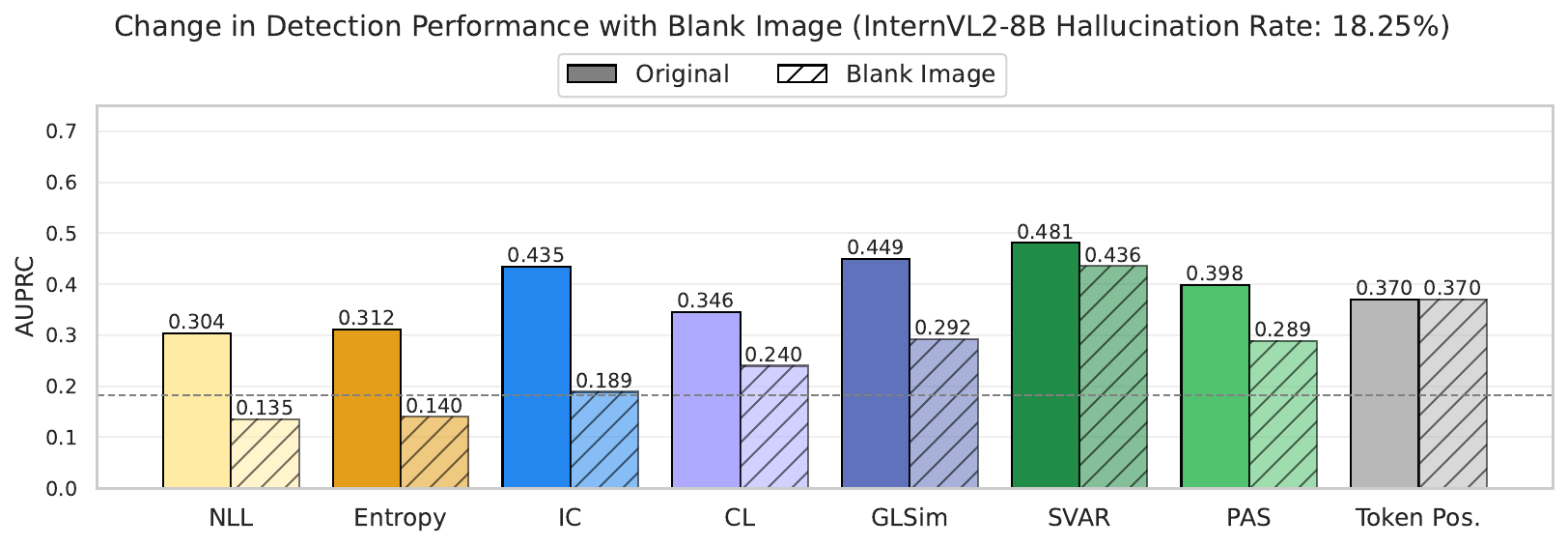}
    \includegraphics[width=0.7\linewidth, trim={0 0.0in 0 0.00in}, clip]{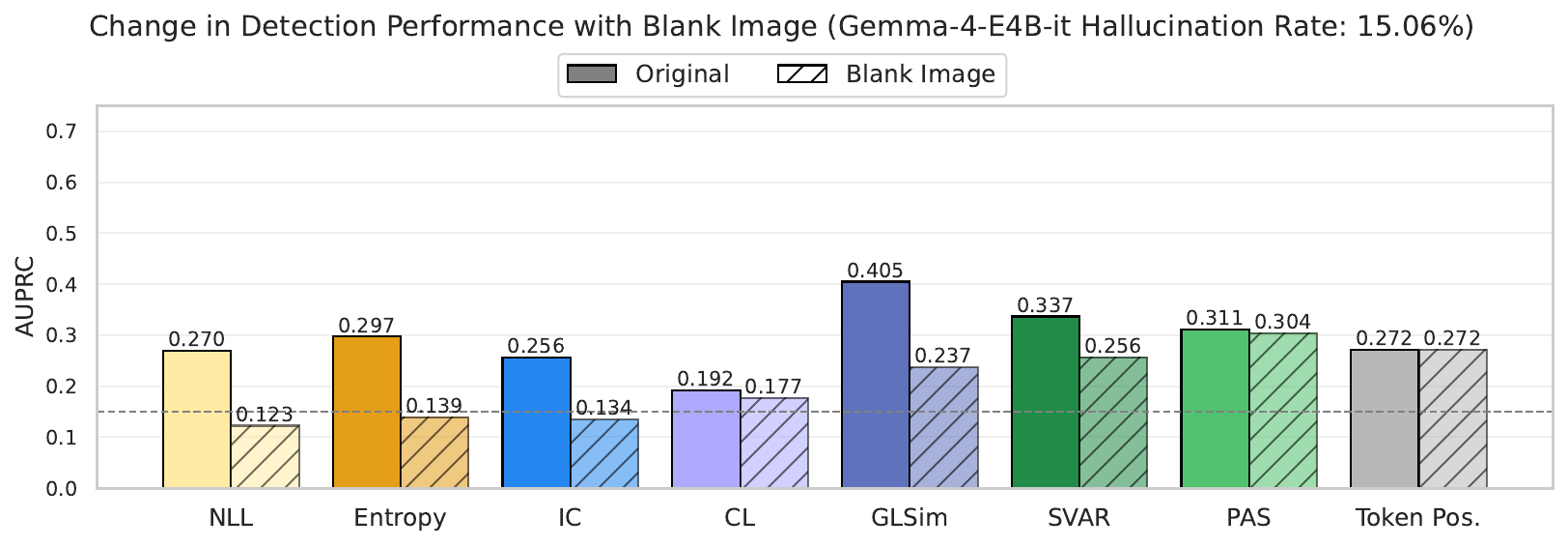}
    \caption{\textbf{Across most VLMs except \qwen \ and \internvl, \svar \ and \pas \ do not exhibit image sensitivity, suggesting they are not visually grounded.} 
    We replace the original image (solid) with a blank image (hatched) and examine the performance drop. 
    Larger performance drops, i.e., to random chance, suggest the detection method is sensitive to image-based signals. }
    \label{fig:visual_grounding}
\end{figure}

We see most methods are sensitive to the image in \figref{fig:visual_grounding}. 
Most notably, performance of \nll, \ent,\ and \ic \ across VLMs drops to near random chance. 
For \cl \ and \glsim, we see this is the case to a slightly lesser degree; across all VLMs except \gemma, the blank image performance is slightly above random chance, but significantly lower than in the original image. This suggests that \cl \ and \glsim \ exhibit a notable degree of sensitivity to the image, but not necessarily as much as \nll, \ent, and \ic.
On the other hand, it is not always the case for \svar \ or \pas.
On \llavasm, \llavalg, and \instructblip, both \pas \ and \svar \ retain high performance in the blank image setting.
This suggests that these detection methods with these VLMs tend to be dependent on token position \emph{in lieu} of visual grounding. 
However, it is not always necessarily the case that image sensitivity is sacrificed, as \svar \ with \qwen \ and \gemma \ and \pas with \internvl \ demonstrate reduced performance in the blank image setting. 
Taken together, these results suggest that the degree of image sensitivity depends on the method-VLM pair.

\subsection{Token position as an unreliable predictor}\label{app:unreliable}

\begin{figure}[h!]
    \centering
    \vspace{-0.2in}
    \begin{minipage}[t]{0.48\linewidth}\centering\small\textbf{Including only objects that appear in both captions}\end{minipage}\hfill
    \begin{minipage}[t]{0.48\linewidth}\centering\small\textbf{Including all objects}\end{minipage}\hfill
    
    \includegraphics[width=0.49\linewidth, trim={0 0.1in 0 0.08in}, clip]{figures/llava1.5-13b/small_first_shared.pdf}
    \includegraphics[width=0.49\linewidth, trim={0 0.1in 0 0.08in}, clip]{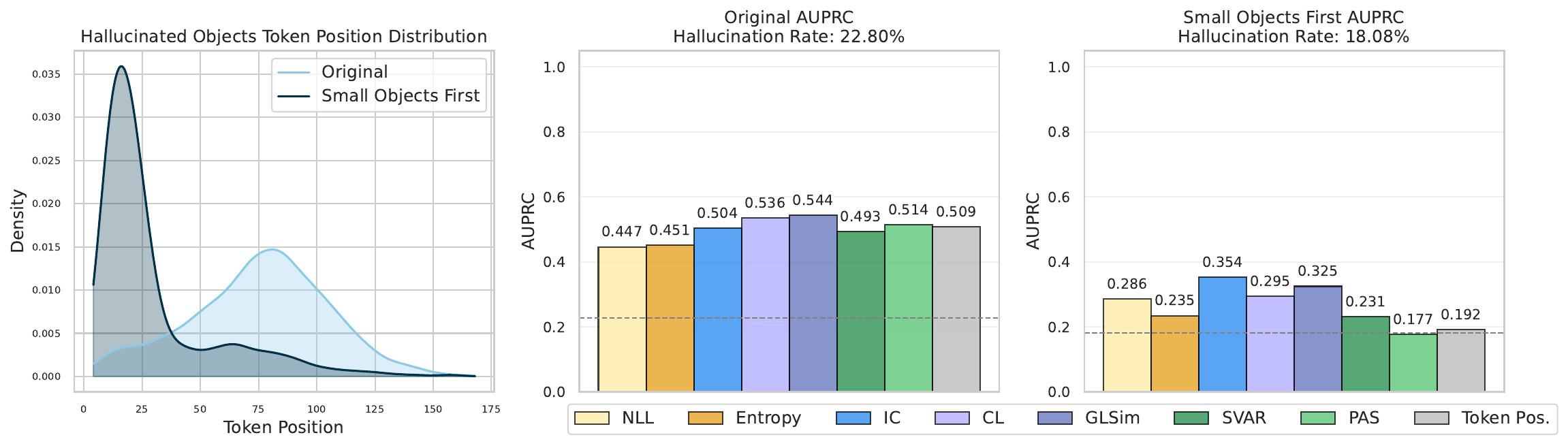}
    \includegraphics[width=0.49\linewidth, trim={0 0.1in 0 0.08in}, clip]{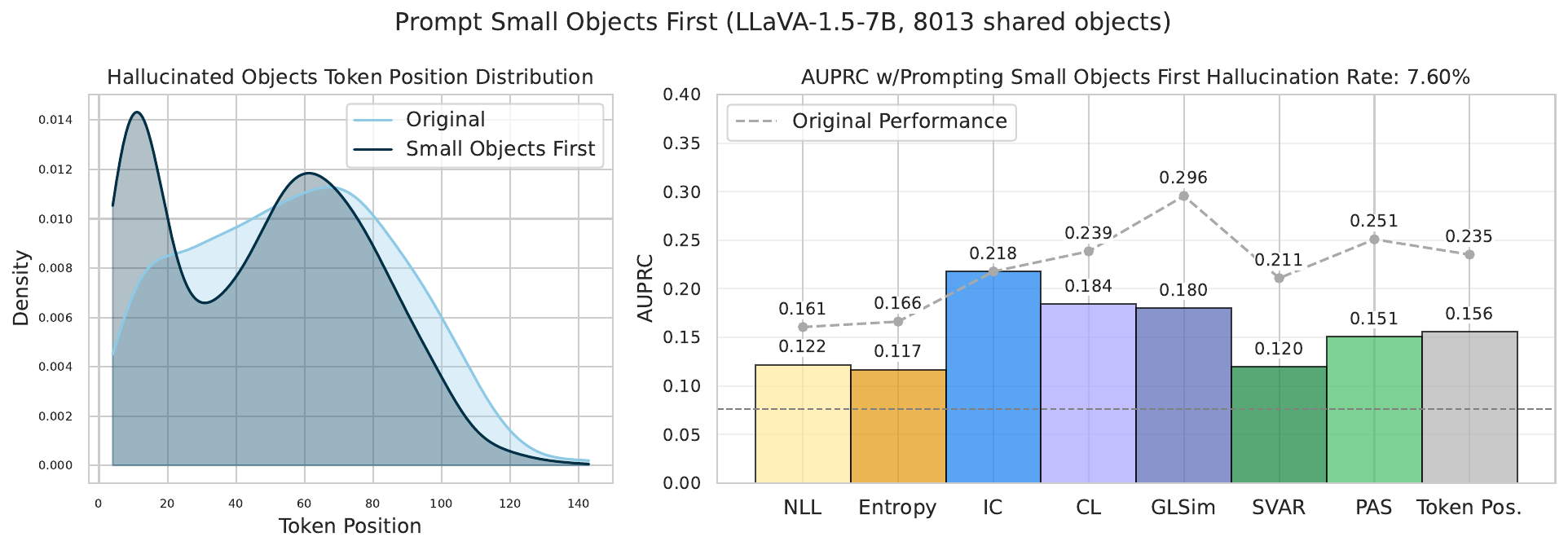}
    \includegraphics[width=0.49\linewidth, trim={0 0.1in 0 0.08in}, clip]{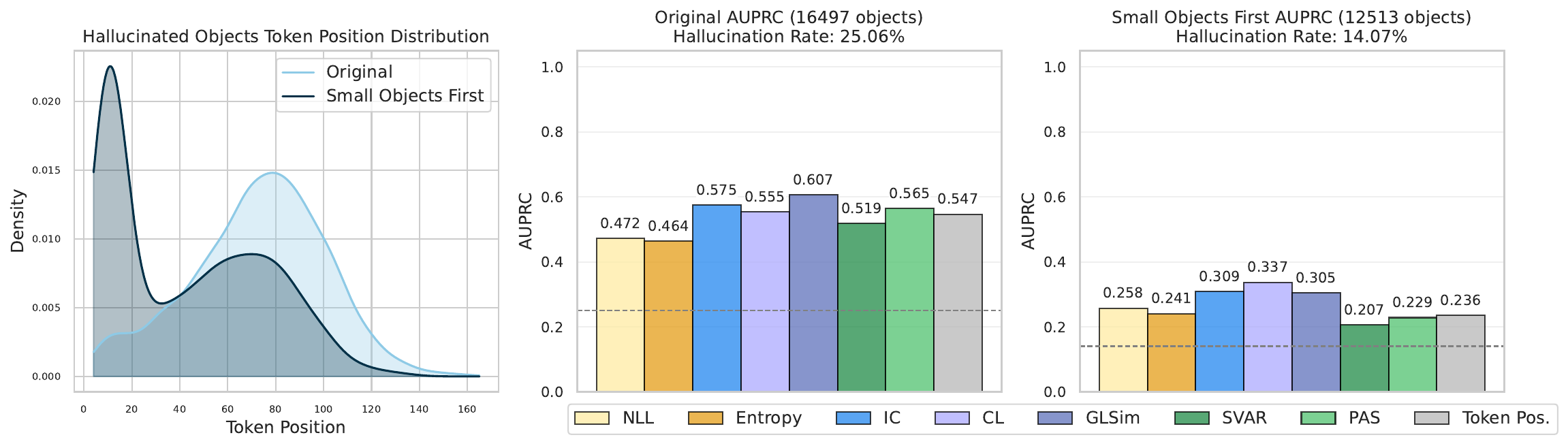}
    \includegraphics[width=0.49\linewidth, trim={0 0.1in 0 0.08in}, clip]{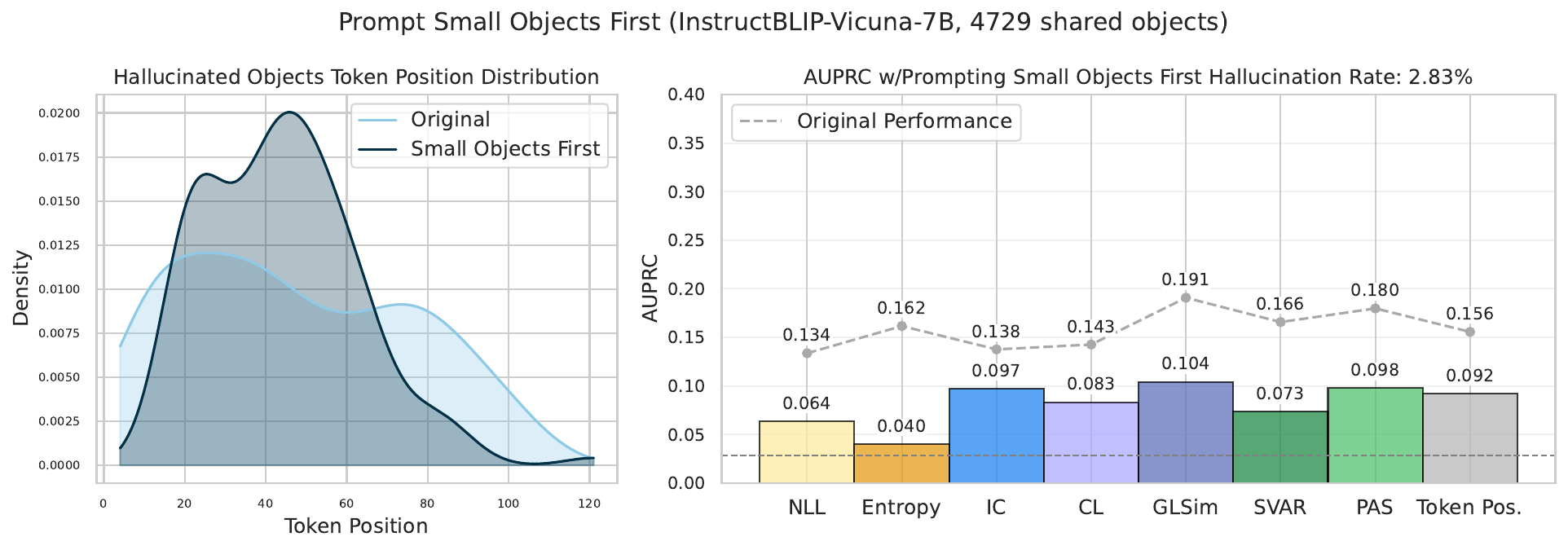}
    \includegraphics[width=0.49\linewidth, trim={0 0.1in 0 0.08in}, clip]{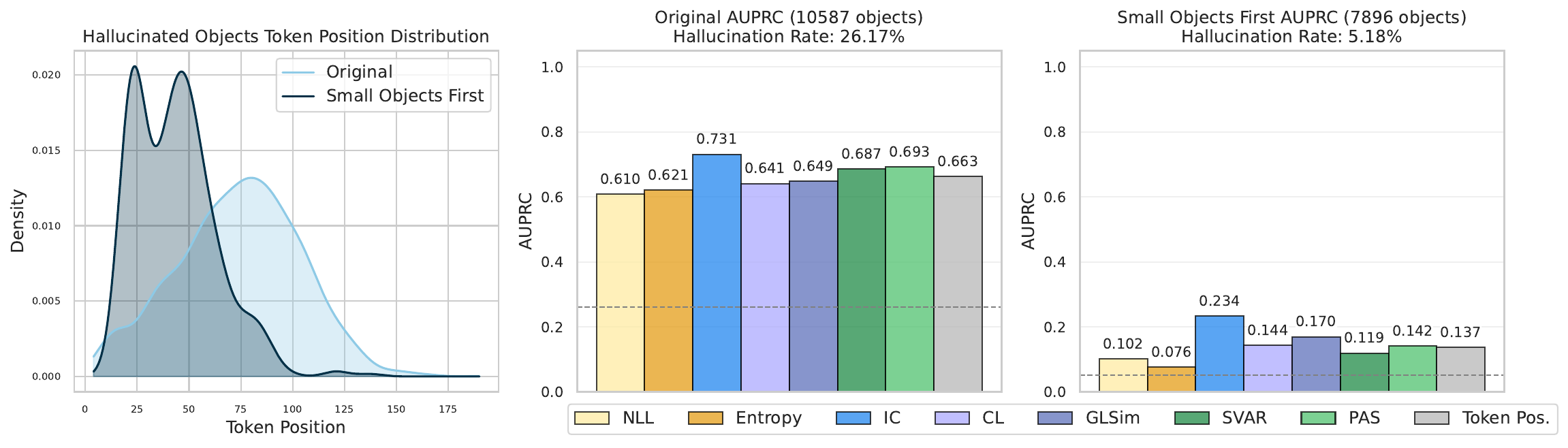}
    \includegraphics[width=0.49\linewidth, trim={0 0.1in 0 0.08in}, clip]{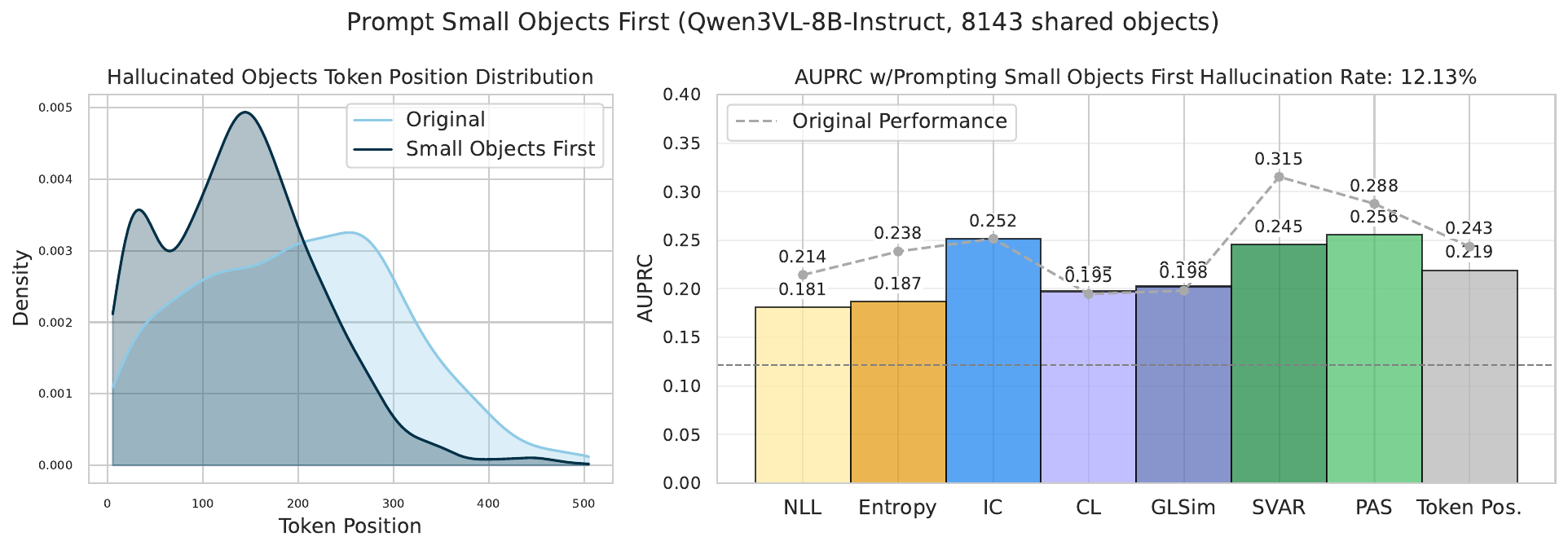}
    \includegraphics[width=0.49\linewidth, trim={0 0.1in 0 0.08in}, clip]{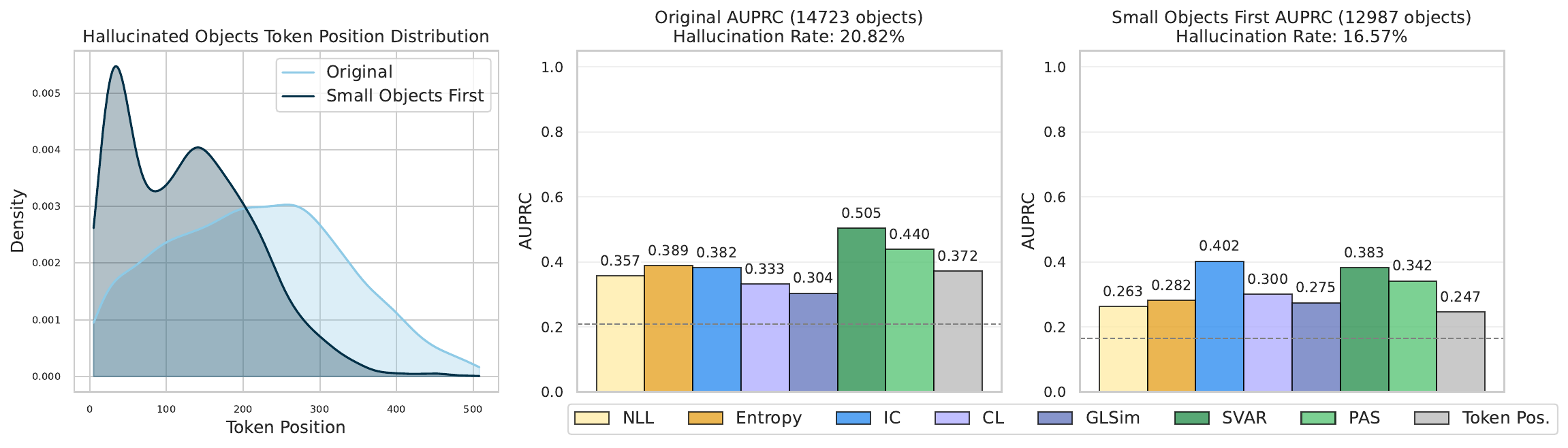}
    \includegraphics[width=0.49\linewidth, trim={0 0.1in 0 0.08in}, clip]{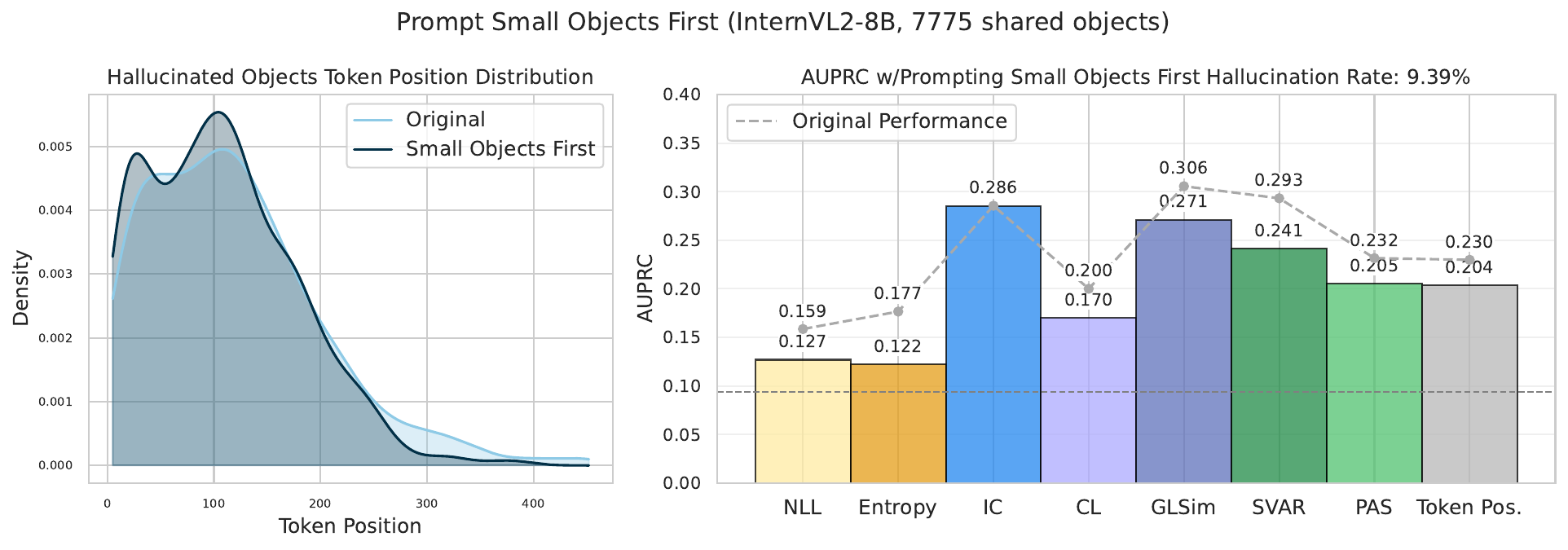}
    \includegraphics[width=0.49\linewidth, trim={0 0.1in 0 0.08in}, clip]{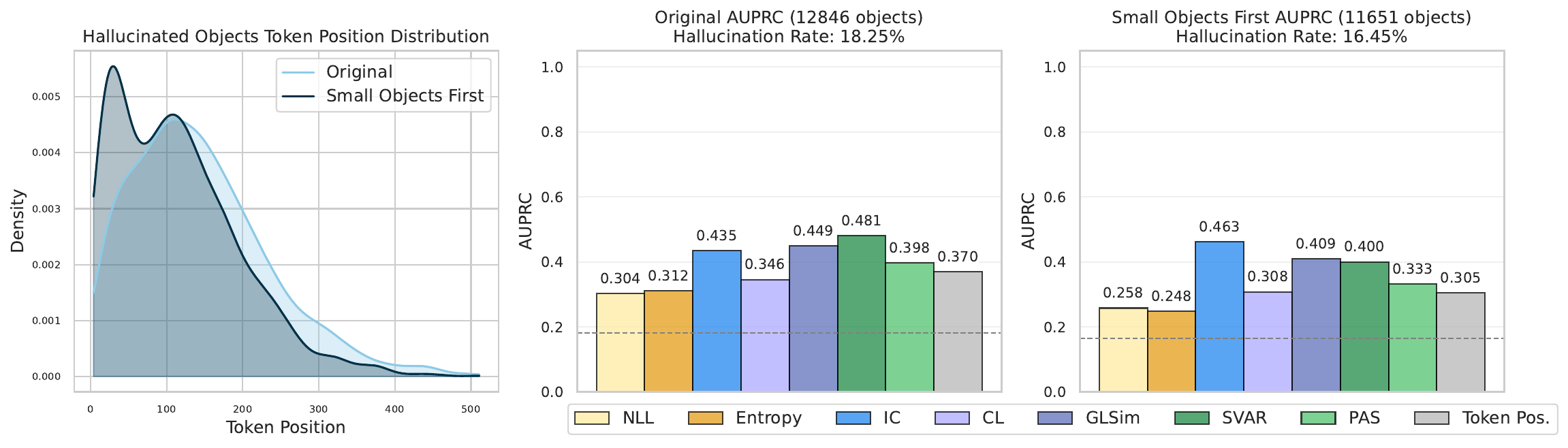}
    \includegraphics[width=0.49\linewidth, trim={0 0.1in 0 0.08in}, clip]{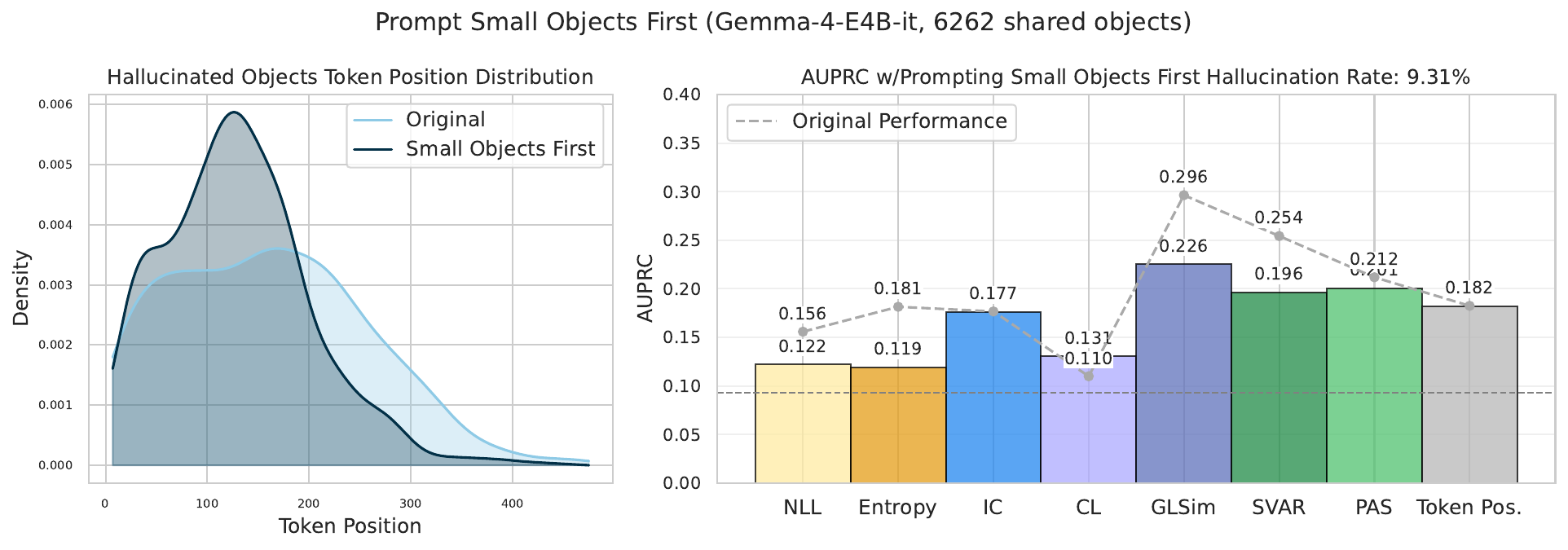}
    \includegraphics[width=0.49\linewidth, trim={0 0.1in 0 0.08in}, clip]{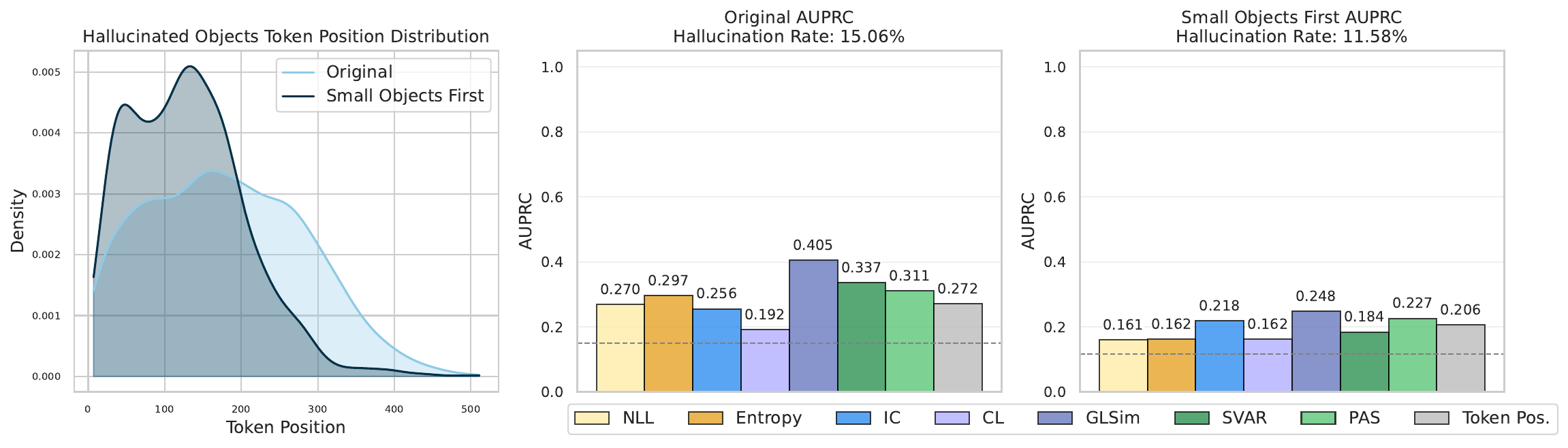}
    
    \caption{\textbf{Unreliability of token position when prompted to mention small objects first.}  \textbf{Left}: Comparing AUPRC between original (dots) and small objects first setting (bars) over \textbf{subset of objects} that appear for an image in both settings. Graphs demonstrate how much of the performance drop is due to token position changes, which varies by VLM.
    \textbf{Right}: Comparing hallucinated token position distribution and performance with AUPRC of the original prompt and the small objects first prompt over \textbf{all objects}.
    Across many VLMs, especially \llavalg, \llavasm, and \instructblip, the performance of \svar \ (dark green), \pas \ (light green), and token position reduce in their competitive advantage, dropping to on par with other methods or even below.}
    \label{fig:unreliability}
\end{figure}

In \figref{fig:unreliability}, we demonstrate results for other VLMs when prompting to describe smaller objects first. 
Consistent with the figure in the main paper, we show plots of AUPRC when only looking at the subset of objects that occur in both the original and the new setting. 
For completeness, we also show figures when evaluating over all objects.
When looking at all objects, we present results as two AUPRC graphs comparing original and small-object-first setting because the baseline AUPRC (i.e., the hallucination rate) differs across settings.

Performance drops for token position-dependent methods (\svar, \pas) in the small object setting are most notable for \llavalg, \llavasm, and \instructblip.
The performance still drops for \svar \ and \pas \ for the remaining three VLMs, but to a smaller extent, relative to other methods.
Although these methods do not perform as poorly with this prompt, there is likely to be some generation setting where these hallucinations are not as aligned to token position and these methods would fail.

\subsection{Evidence for complementary signals using ensembles}\label{app:ensemble}

Correlation analysis across scores of all methods including the baseline token position revealed that a majority of white box methods are highly correlated with token position and one another, and the black box methods are correlated with each other.
However, there is little correlation between these two subgroups. 
In order to understand what additional discriminative power white box methods have over the token position, as well as to determine if black box methods have any differential discriminative power with respect to white box methods we built ensemble models using 5 different groupings of these methods: (1) Token position (\{\texttt{TP}\}) alone; (2) \{\texttt{TP}\} + \{\svar, \pas\}; (3) \{\texttt{TP}\} + \{\svar, \pas\} + \{\glsim, \cl\}; (4) \{\texttt{TP}\} + \{\svar, \pas\} + \{\glsim, \cl\} + \{\ic\}; (5) \{\texttt{TP}\} + \{\svar, \pas\} + \{\glsim, \cl\} + \{\ic\} + \{\nll, \ent\}.

We train on $5,000$ randomly sampled images from the MSCOCO \cite{lin_microsoft_2014} 2014 training split and evaluate on the same validation split as other experiments. We report 95\% confidence intervals computed via percentile bootstrap of the validation predictions.

For each group, we split the training set of MSCOCO \citep{lin_microsoft_2014} into train and test sets and trained using logistic regression with L2 norm with the default regularization penalty of C=1.0. We used normalized token position as well as normalized scores for all methods in order to avoid imbalanced penalty due to score magnitude. 
All the models were then evaluated and reported on 5,000 images from the official validation set of MSCOCO.

\begin{figure}[h!]
    \centering
    \includegraphics[width=1.0\linewidth, trim={0 0.1in 0 0.08in}, clip]{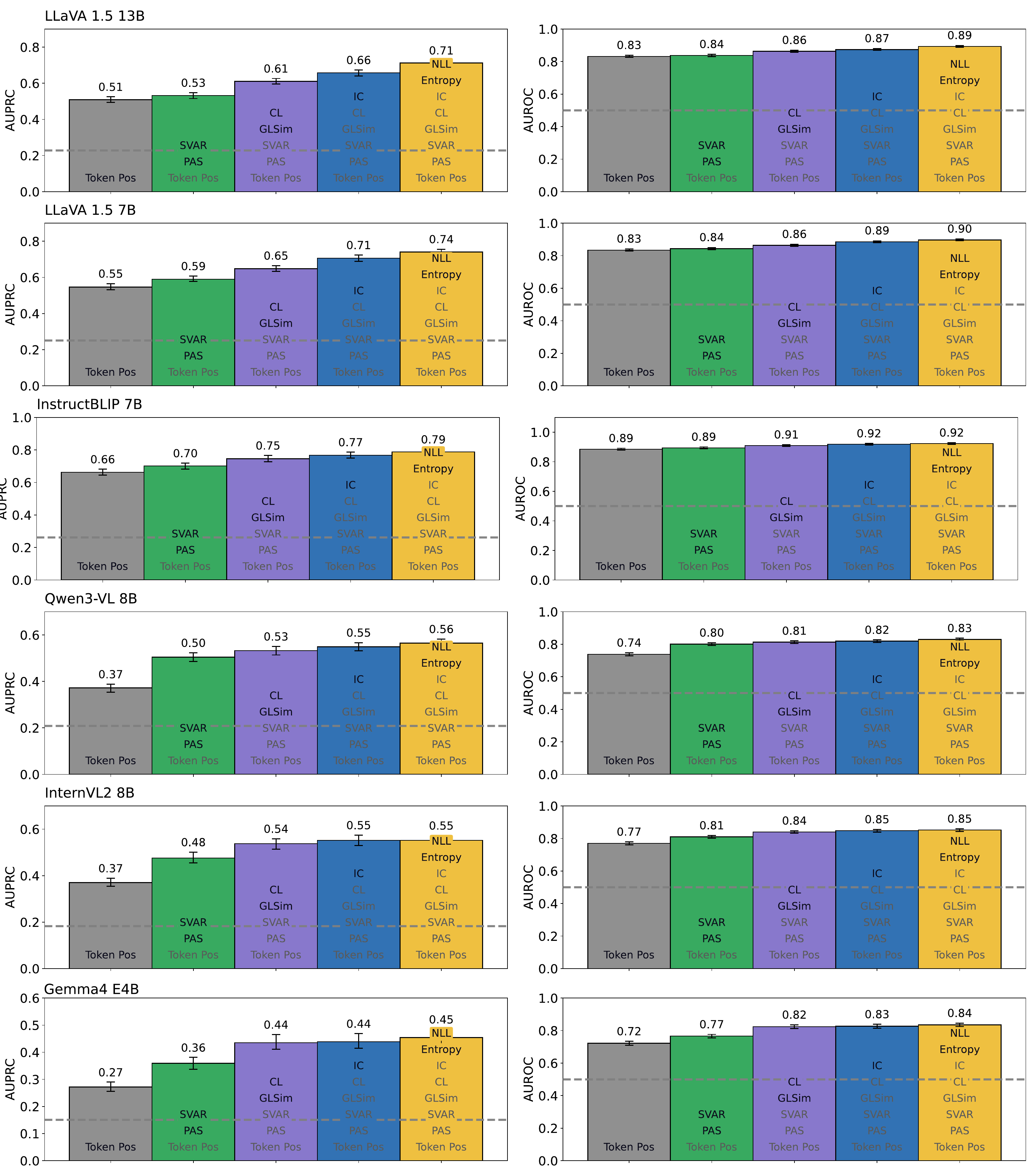}
    \caption{Ensemble method performances for each VLM, where we add \{\svar, \pas\}, \{\glsim, \cl\}, \{\ic\} and \{\nll, \ent\} incrementally on top of token position. 
    In all VLMs we observe that token position is a strong baseline. 
    Despite high correlation between token position and  \{\svar, \pas\}, this group improves the AUPRC substantially over token position for \qwen, \internvl, and \gemma. 
    Meanwhile, \{\cl, \glsim\} contribute significant signal for \llavalg, \llavasm, \instructblip, \internvl, and \gemma.
    \{\ic\} has additional discriminative power over this ensemble for \llavalg \ and \llavasm. 
    Lastly, black box methods \{\nll, \ent\} the most improvement for \llavalg \ and \llavasm, but less so for other VLMs. \looseness=-1} 
    \label{fig:all_vlms_ensemble}
\end{figure}

We report in \figref{fig:all_vlms_ensemble} the additional ensemble model performances for all VLMs: \llavalg, \llavasm, \instructblip, \qwen, \internvl, and \gemma \ which shows a similar trend to the results reported in \secref{sec:ensemble} for \llavalg.
Namely, that there are complementary signals between methods of different groups, most notably between TP, \{\svar, \pas\}, and \{\glsim, \cl\} across most VLMs and even \ic \ and \{\nll, \ent\} for certain VLMs. 
While the complementarity between groups may slightly differ across VLMs, we observe when \{\svar, \pas\} are slightly less (but still strongly) correlated with token position (i.e., \qwen, \gemma), these methods provide greater complementarity with token position, but in these cases, \{\nll, \ent\} exhibit reduced benefit.

\subsection{Experiments disentangling dependency on VLM vs. caption-generation patterns.} \label{app:cross-model}

We additionally conducted experiments to understand complementarity of signals between detection methods.
Here we try to tease apart why certain detection methods that perform poorly on certain VLMs. 
Specifically, why \cl \ and \glsim \ performance drop significantly on \qwen, \internvl, and \gemma.
One potential hypothesis is that the methods are tailored to the caption-generation processes of specific VLMs, as opposed to the VLMs themselves. 
To test this hypothesis, we evaluate these methods in a decoupled setting: one VLM generates the caption, and a second VLM detects hallucinations using its own internal or output signals, rather than from the caption-generating VLM.

\begin{figure}[h!]
    \centering
    \includegraphics[width=1.0\linewidth, trim={0 0.1in 0 0.08in}, clip]{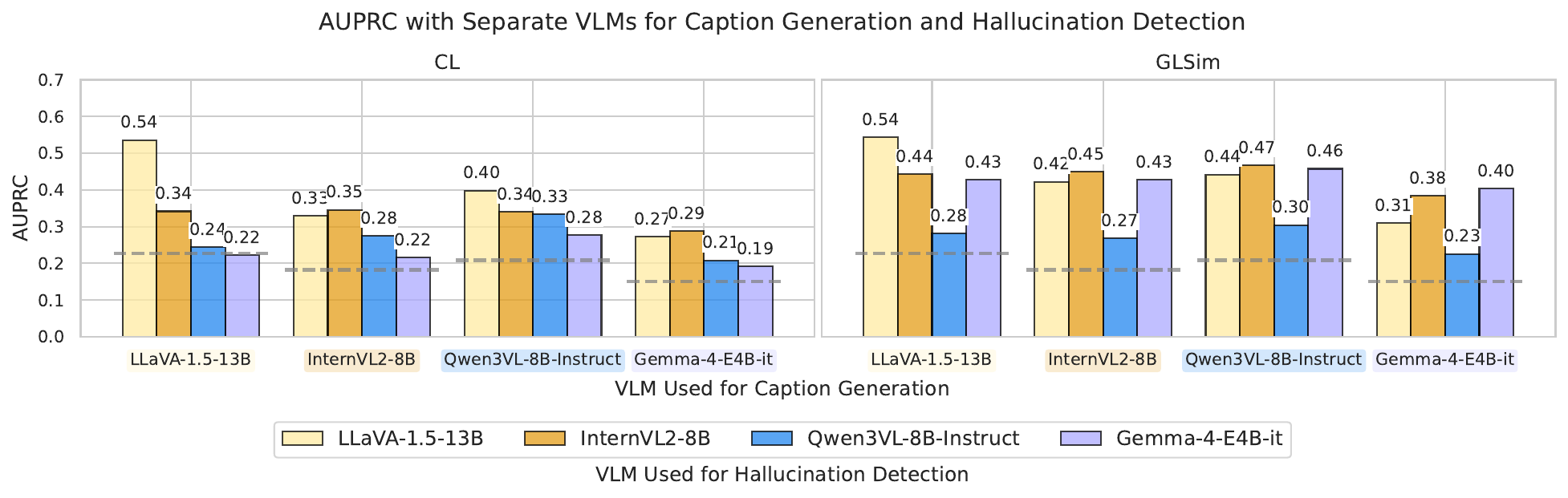}
    \caption{\textbf{Disentangling caption-generation process from VLM specificity.} 
    We observe that independent of the VLM that generated the caption, the relative performance of using \internvl, \qwen, and \gemma \ to detect hallucinations is relatively consistent.
    We use \cl \ and \glsim \ on certain VLMs (indicated by colored bars) to detect hallucinations in captions from other VLMs (indicated by each grouping). 
    We plot AUPRC and dashed horizontal lines indicated the random baseline (i.e., hallucination rate).
    \looseness=-1} 
    \label{fig:cross_model}
\end{figure}

In \figref{fig:cross_model}, we report all pairwise combinations of \llavalg, \qwen, \internvl, and \gemma.
While we cannot draw any strong conclusions, we note a few observations here. 
First, that independent of which VLM generated the caption, using \cl \ with \gemma \ always performs the worst, even for \gemma \ captions. 
Similarly, indepndent of the caption-generating VLM, \glsim \ always performs worst when it is used with \qwen \ to detect hallucinations.
These results may begin to suggest that the hallucination detection methods are more dependent on the VLM itself (e.g., the architecture or internal representational spaces), than features of the caption generation processes.

\begin{figure}[h!]
    \centering
    \vspace{-0.2in}
    \includegraphics[width=0.7\linewidth, trim={0 0.1in 0 0.08in}, clip]{figures/llava1.5-13b/hall_source_auprc.pdf}
    \includegraphics[width=0.7\linewidth, trim={0 0.0in 0 0.00in}, clip]{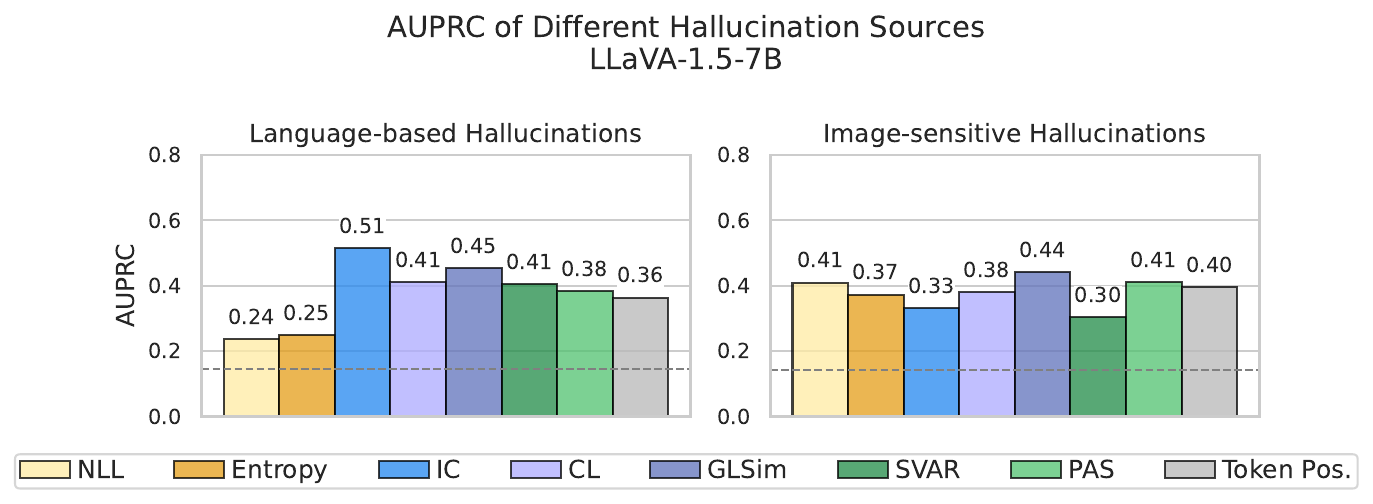}
    \includegraphics[width=0.7\linewidth, trim={0 0.0in 0 0.00in}, clip]{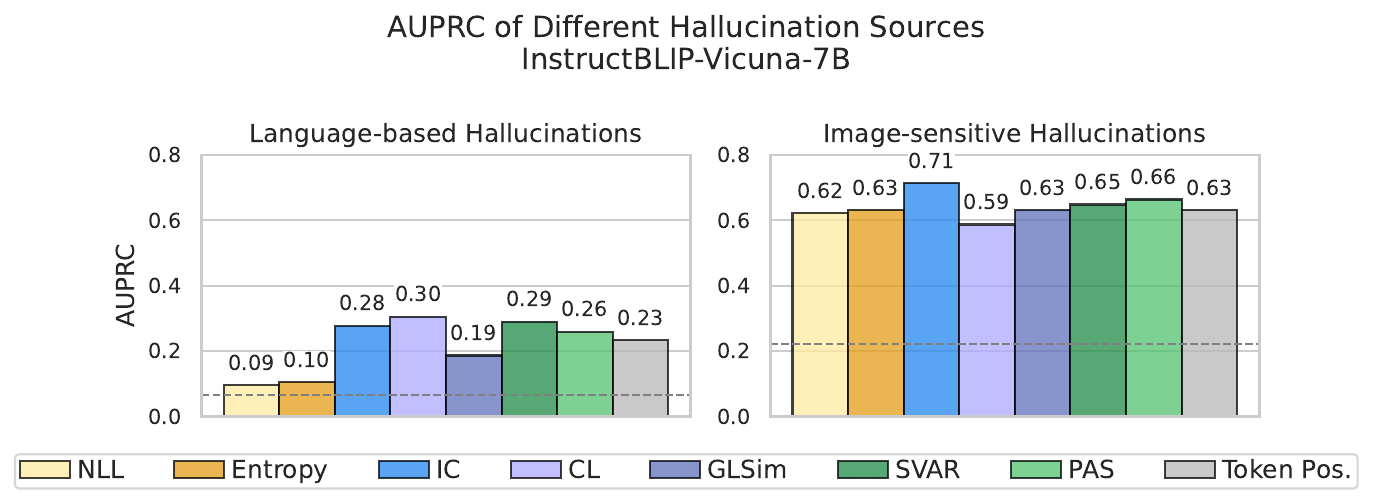}
    \includegraphics[width=0.7\linewidth, trim={0 0.0in 0 0.00in}, clip]{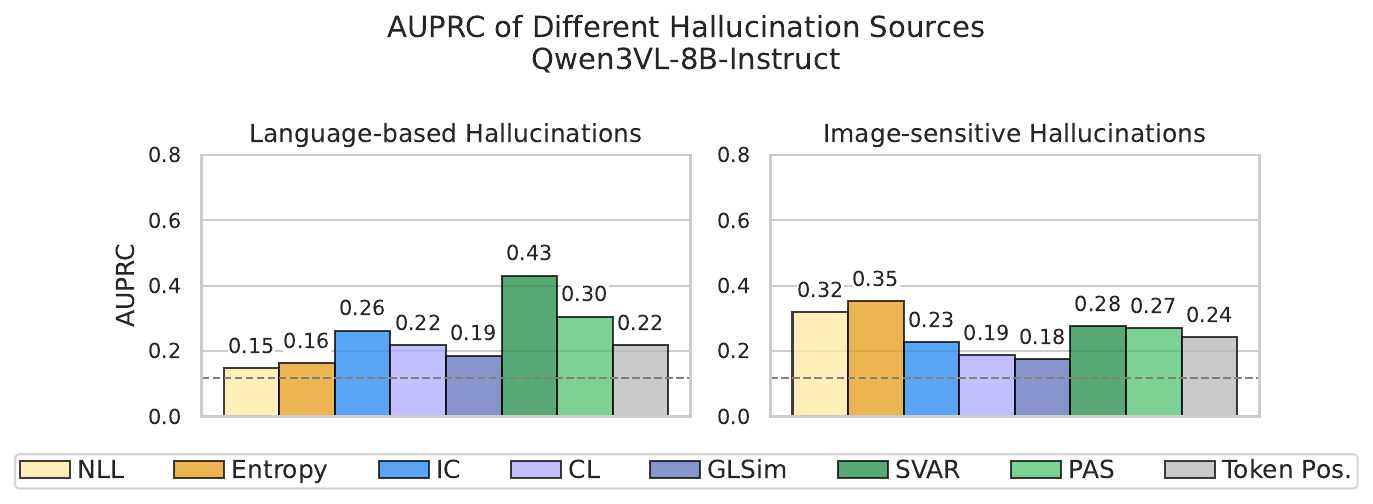}
    \includegraphics[width=0.7\linewidth, trim={0 0.0in 0 0.00in}, clip]{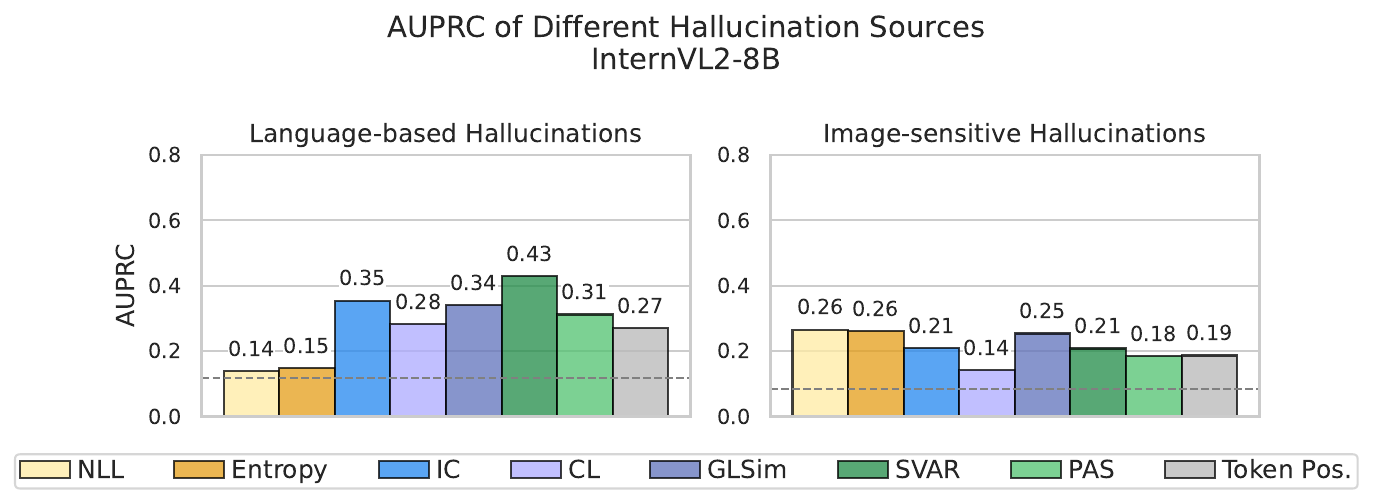}
    \includegraphics[width=0.7\linewidth, trim={0 0.0in 0 0.00in}, clip]{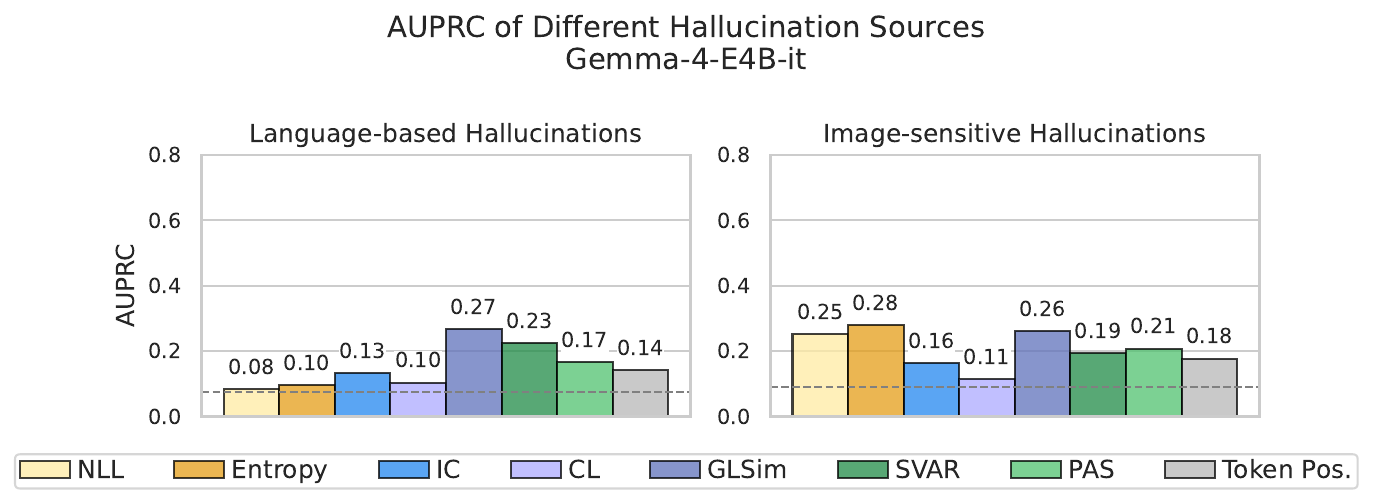}
    \caption{\textbf{White box detection methods gain their competitive performance over black box methods for language-based hallucinations but not necessarily for non-language-based hallucinations.} 
    We partition hallucinations into language and non-language-based depending on whether the generated token differs when using a blank image. 
    White box methods are strong at detecting hallucinations that persist even when images are removed, suggesting a lack of image sensitivity.}
    \label{fig:partition}
\end{figure}

\clearpage

\section{Generalization experiments}\label{app:generalization}

In this section, we demonstrate how results are consistent when using newer versions of VLMs (\secref{app:newer_vlms}). Additionally, for the VLMs reported in the main paper and \secref{app:exp_details_results} (\llavalg, \llavasm, \instructblip, \qwen, \internvl, and \gemma), we demonstrate how results hold across a different dataset (\secref{app:o365}) , using sampling with \texttt{temp > 0} and repeated trials (3 trials) (\secref{app:repeated_trials}), varying the prompt  (\secref{app:prompt_variations}), and implementing a context-aware filter to avoid evaluating non-objects (e.g., avoid ``phone'' in ``phone number'' from being flagged) (\secref{app:non_objects}). 
For generalization experiments in \secref{app:o365} - \secref{app:non_objects}, we present figures grouped by VLM to demonstrate how trends for a certain VLM are consistent in the variety of settings.
Because figures tend to be large, most are placed at the end of this section to enable continuity of the text.

\subsection{Newer VLMs}\label{app:newer_vlms}

In addition to the six main VLMs (\llavalg, \llavasm, \instructblip, \qwen, \internvl, and \gemma), we also ran experiments for the updated versions of some of the VLMs (\qwennew \ and \internvlnew).
As shown in \tabref{tab:new_vlms}, token position is still a competitive hallucination detector to other white box methods. 
Largely, \glsim \ and \svar \ are the top performing detection methods for these models, based on AUPRC. 

\begin{table}
  \caption{\textbf{Hallucination detection performance on updated versions of Qwen and InternVL on MSCOCO.}}
  \centering
  \resizebox{\textwidth}{!}{
  \begin{tabular}{ll c cc ccccc cc}
    \toprule
    \multirow{2}{*}{\textbf{Metric}} & \multirow{2}{*}{\textbf{Model}} & & \multicolumn{2}{c}{\textbf{Black Box}} & \multicolumn{5}{c}{\textbf{White Box}} & \multicolumn{2}{c}{\textbf{Heuristic}} \\
    \cmidrule(lr){4-5} \cmidrule(lr){6-10} \cmidrule(lr){11-12}
    & & Random & NLL & Entropy & IC & CL & GLSim & SVAR & PAS & Token Pos. & \makecell{Token Pos.\\(norm.)} \\

    \midrule
    \multirow{3}{*}{AUROC} 
    & \qwennew  & \multirow{2}{*}{50.0} & 68.2 & 70.8 & 74.2 & 69.8 & \textbf{79.7} & 78.7 & \underline{79.1} & 78.3 & 78.3 \\
    & \internvlnew        &  & 66.9 & 69.0 & 62.8 & 65.2 & 74.6 & \underline{77.4} & \textbf{78.3} & 76.3 & 77.2 \\
    
    \cmidrule{2-12}
    & \textit{Average} & \textit{50.0} & \textit{67.6} & \textit{69.9} & \textit{68.5} & \textit{67.5} & \textit{77.2} & \textit{\underline{78.1}} & \textit{\textbf{78.7}} & \textit{77.3} & \textit{77.8}\\
    \midrule
    
    \multirow{3}{*}{AUPRC} 
    & \qwennew        &  21.3 & 37.4 & 42.9 & 47.5 & 38.9 & \textbf{52.6} & \underline{49.5} & 48.7 & 43.5 & 43.5 \\
    & \internvlnew        & 15.0 & 28.9 & 31.3 & 22.2 & 24.7 & \textbf{35.5} & \underline{35.5} & 33.5 & 30.0 & 33.7 \\
    \cmidrule{2-12}
    & \textit{Average} & \textit{18.1} & \textit{33.1} & \textit{37.1} & \textit{34.9} & \textit{31.8} & \textit{\textbf{44.0}} & \textit{\underline{42.5}} & \textit{41.1} & \textit{36.8} & \textit{38.6}\\
    \bottomrule
  \end{tabular}
  }
  \label{tab:new_vlms}
  \vspace{0.1in}
\end{table}

\begin{figure}[h!]
    \centering
    \begin{minipage}[t]{0.48\linewidth}\centering\small\textbf{\qwen}\end{minipage}\hfill
    \begin{minipage}[t]{0.48\linewidth}\centering\small\textbf{\qwennew}\end{minipage}\hfill
    \includegraphics[width=0.48\linewidth, trim={0 0.1in 0 0.08in}, clip]{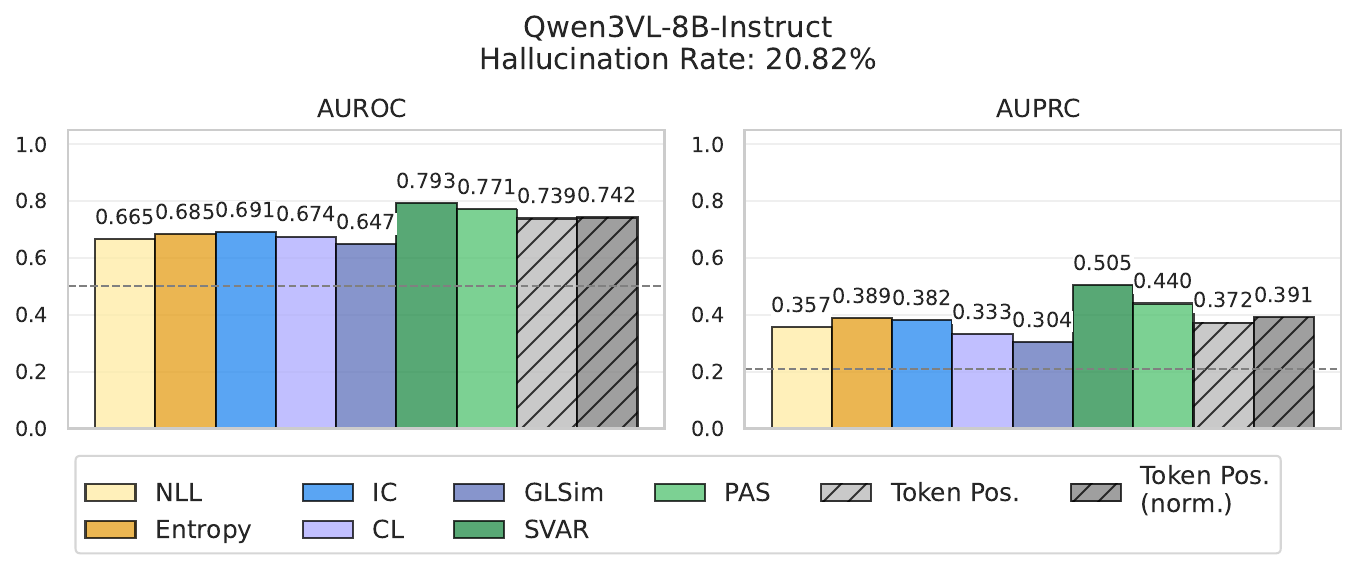}
    \includegraphics[width=0.48\linewidth, trim={0 0.1in 0 0.08in}, clip]{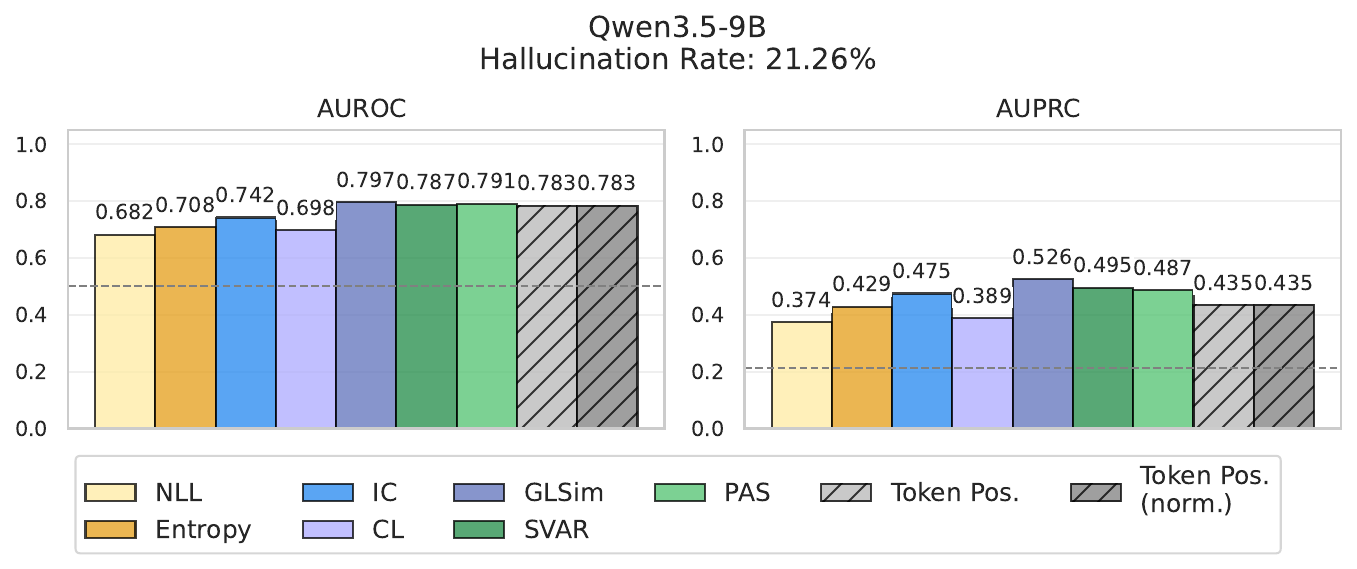}
    
    \includegraphics[width=0.48\linewidth, trim={0 0.1in 0 0.08in}, clip]{figures/qwen3vl-8b/remove_replace.pdf}
    \includegraphics[width=0.48\linewidth, trim={0 0.1in 0 0.08in}, clip]{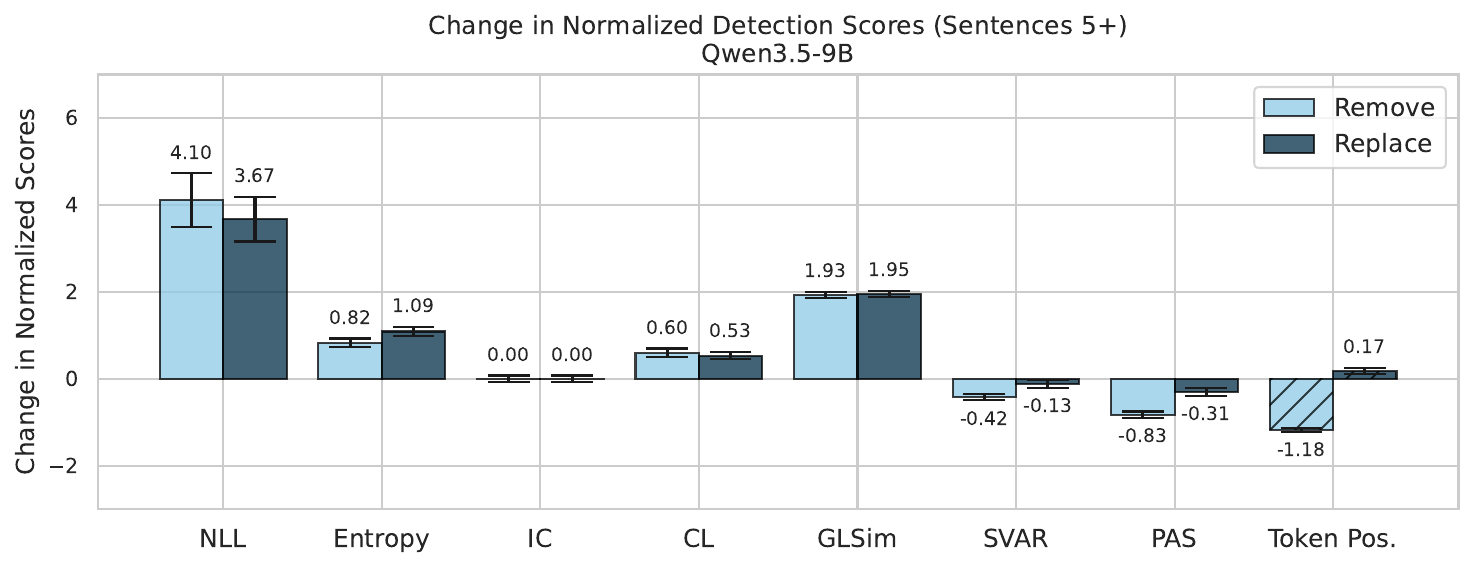}

    \includegraphics[width=0.48\linewidth, trim={0 0.1in 0 0.08in}, clip]{figures/qwen3vl-8b/auprc_blank_performance_diff.pdf}
    \includegraphics[width=0.48\linewidth, trim={0 0.1in 0 0.08in}, clip]{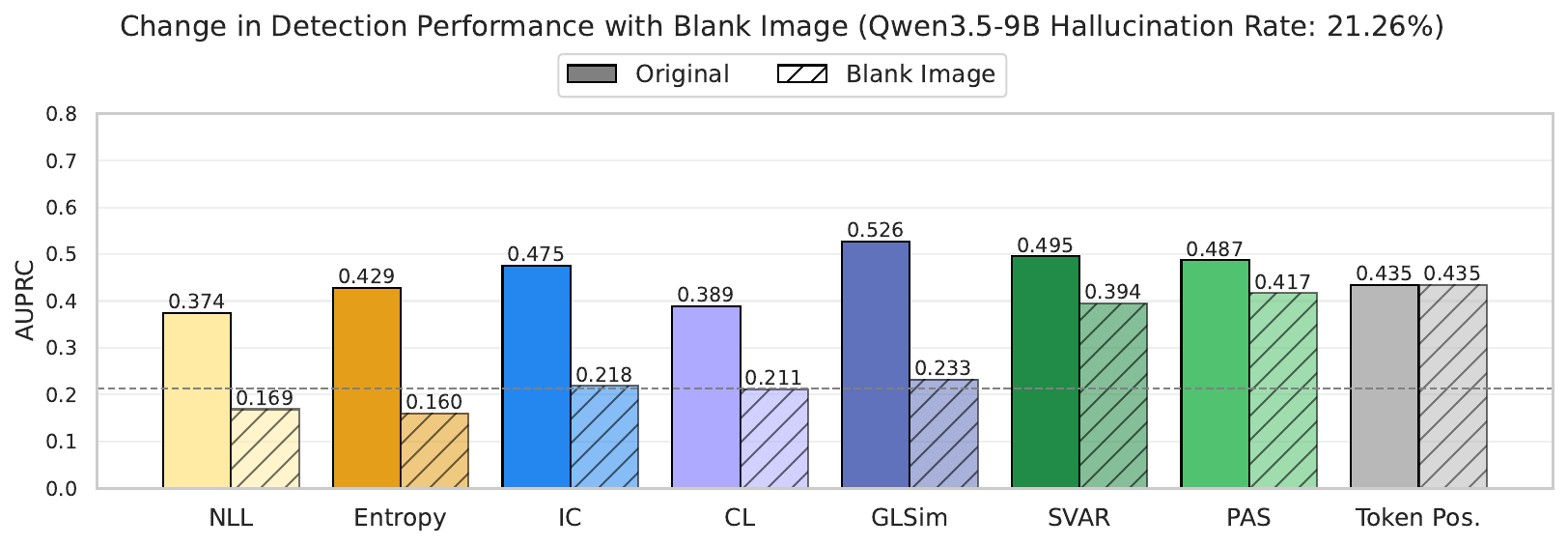}\par 
    
    \vspace{0.3in}
    \begin{minipage}[t]{0.48\linewidth}\centering\small\textbf{\internvl}\end{minipage}\hfill
    \begin{minipage}[t]{0.48\linewidth}\centering\small\textbf{\internvlnew}\end{minipage}\hfill
    \includegraphics[width=0.48\linewidth, trim={0 0.1in 0 0.08in}, clip]{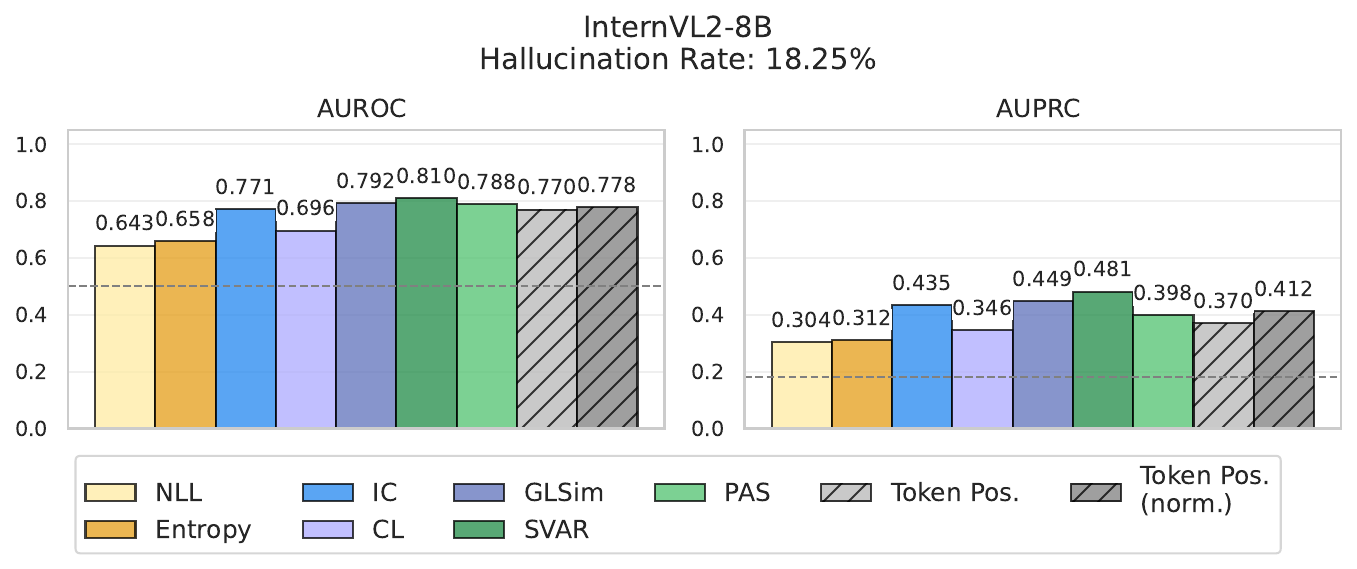}
    \includegraphics[width=0.48\linewidth, trim={0 0.1in 0 0.08in}, clip]{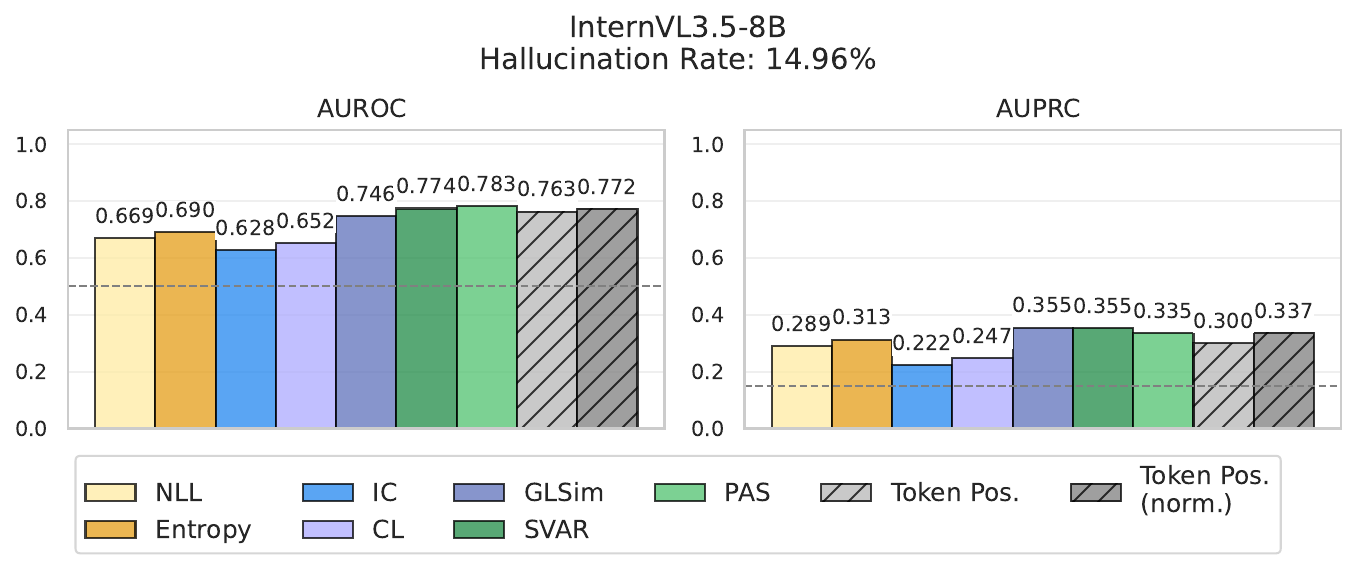}
    
    \includegraphics[width=0.48\linewidth, trim={0 0.1in 0 0.08in}, clip]{figures/internvl2-8b/remove_replace.pdf}
    \includegraphics[width=0.48\linewidth, trim={0 0.1in 0 0.08in}, clip]{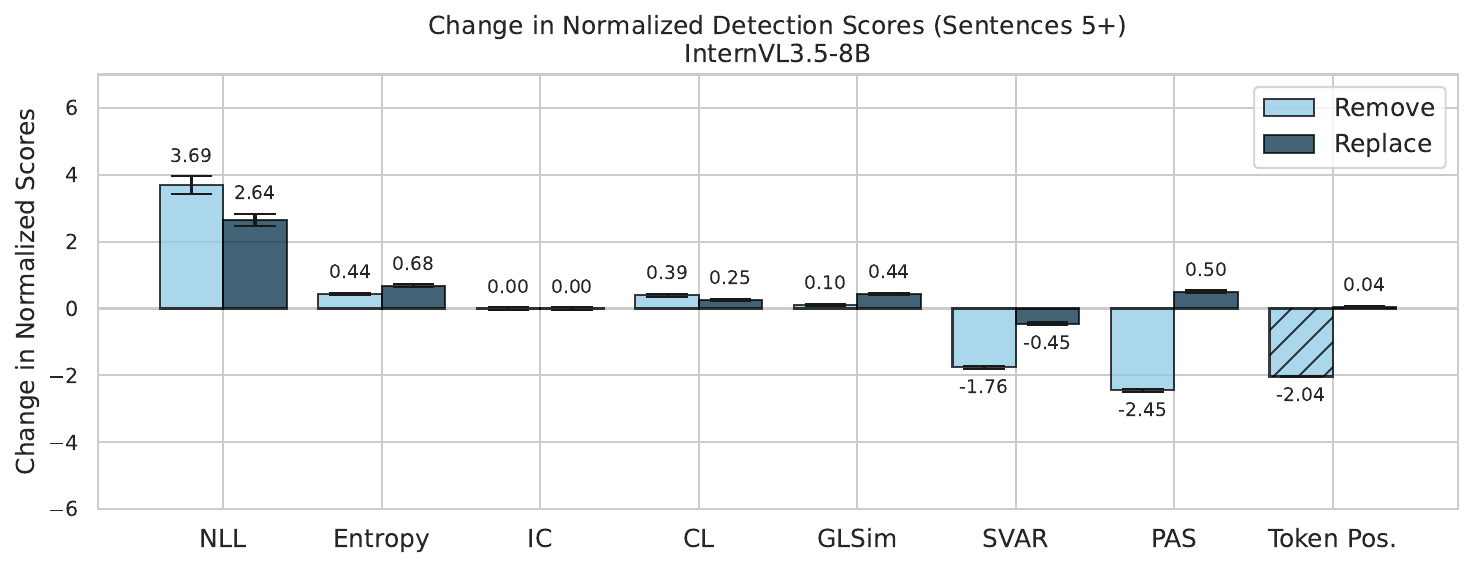}
    
    \includegraphics[width=0.48\linewidth, trim={0 0.1in 0 0.08in}, clip]{figures/internvl2-8b/auprc_blank_performance_diff.pdf}
    \includegraphics[width=0.48\linewidth, trim={0 0.1in 0 0.08in}, clip]{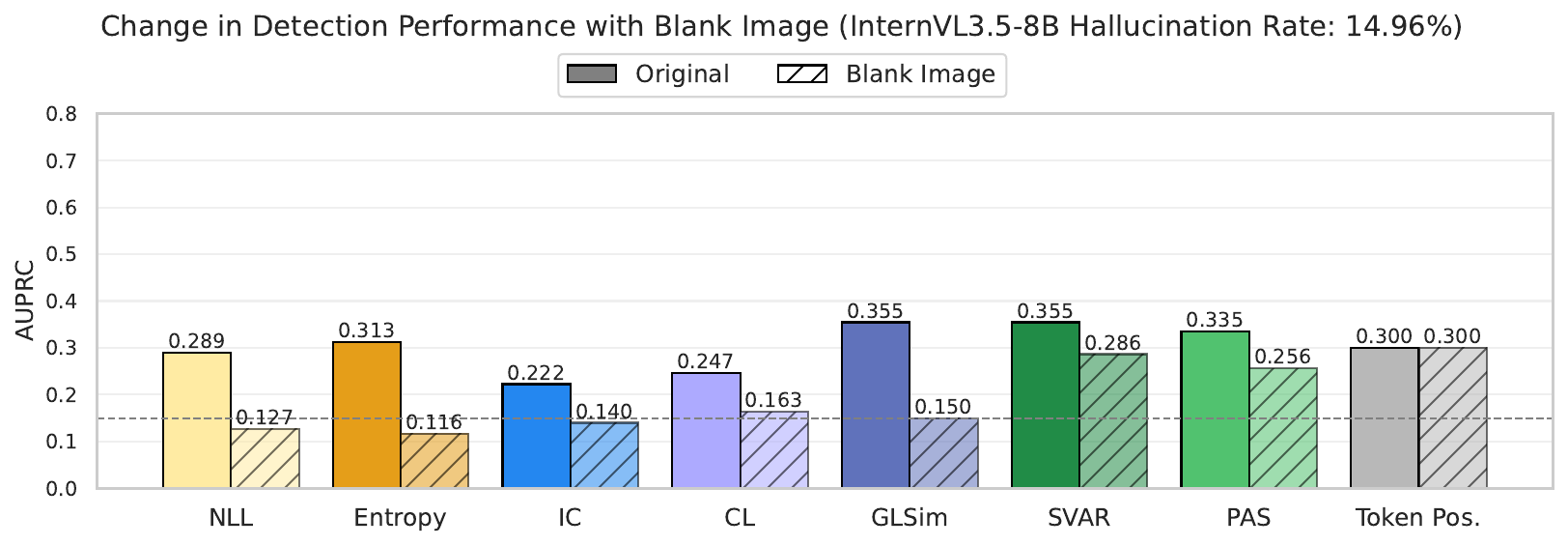}

    \caption{\textbf{Evaluation performance, token position dependency, and image-signal sensitivity are similar for newer versions of \qwennew \ and \internvlnew.} 
    Overall performance of detection methods (1st and 3rd rows) on newer versions of VLMs  are are similar. 
    The blue remove/replace graphs (2nd and 5th rows) show that \pas \ on \qwennew \ exhibits less context dependency than with \qwen.
    However the image sensitivity graphs (3rd and 6th rows) show \svar \ on \qwennew \ exhibits even less image sensitivity than with \qwen.
    On the other hand, \pas \ with \internvlnew \ seems much more sensitive to position and context than with \internvl \ while other trends are similar.}
    \label{fig:new_vlm_graphs}
\end{figure}

We additionally present results for their token position dependency and visual sensitivity in \figref{fig:new_vlm_graphs}. 
Here we demonstrate that trends of token position dependency and image sensitivity are relatively consistent across versions of the same VLM family.
We observe the largest differences with \pas's token position dependency for both VLM families and \svar's image sensitivity across Qwen.
Specifically, with \qwennew, \pas \ is less sensitive to token position and contextual changes of the caption compared to \qwen. On the other hand, \pas \ is more sensitive to both with \internvlnew \ compared to \internvl. 
However, across all VLMs, experiments with \pas \ and \svar \ demonstrate that compared to the mean score in the replace setting, the mean score in the remove setting is lower. This suggests that when holding semantic information constant, these methods are still sensitive to token position to some degree.

Regarding image sensitivity, trends across all methods are stable, with an exception of \svar \ exhibiting even \emph{less} image sensitivity on the newer \qwennew \ compared to \qwen. 
\svar \ is consistently dependent on token position and \pas \ sometimes is as well. Further, both methods exhibit little sensitivity to the image, suggesting they lack visual grounding. We note that although the change in scores for \svar \ with \qwennew \ is small, the change in score of token position itself is low and roughly 80\% of \svar's change is still accounted for by solely changing token position.

\begin{figure}[h!]
    \centering
    \includegraphics[width=1.0\linewidth, trim={0 0.1in 0 0.08in}, clip]{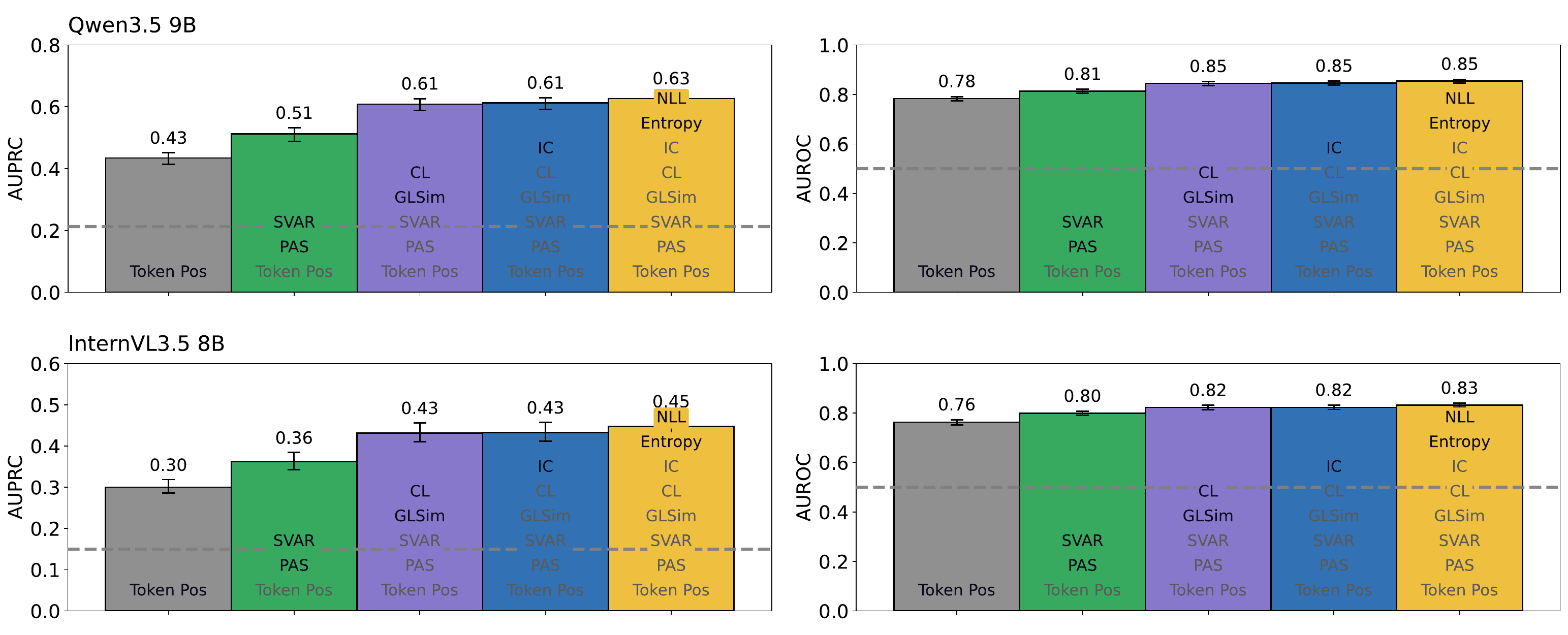}
    \caption{Ensemble method performances for \qwennew \ and \internvlnew, where we add \{\svar, \pas\}, \{\glsim, \cl\}, \{\ic\} and \{\nll, \ent\} incrementally on top of token position. For these models, beyond adding \cl \ and \glsim, additional methods do not seem to contribute much additional signal.}
    \label{fig:new_vlms_ensemble}
\end{figure}

Lastly, we present ensemble results in \figref{fig:new_vlms_ensemble} which demonstrate more notable complementary strengths between \{\svar, \pas\} and \{\cl, \glsim\} but not so much when adding \ic, \nll, \ent. 
This suggests for newer models, perhaps \svar, \pas, \cl \ and \glsim \ encompass signals of weaker methods.

\subsection{Dataset generalization}\label{app:o365}

\begin{figure}[h!]
    \centering
    \includegraphics[width=0.48\linewidth, trim={0 0.1in 0 0.08in}, clip]{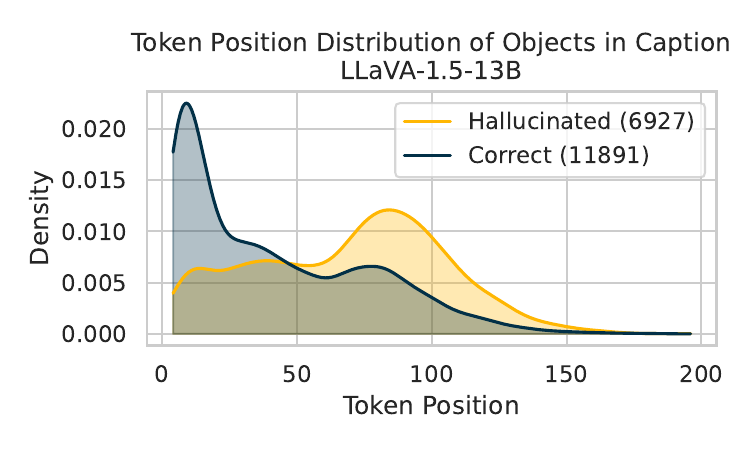}
    \includegraphics[width=0.48\linewidth, trim={0 0.1in 0 0.08in}, clip]{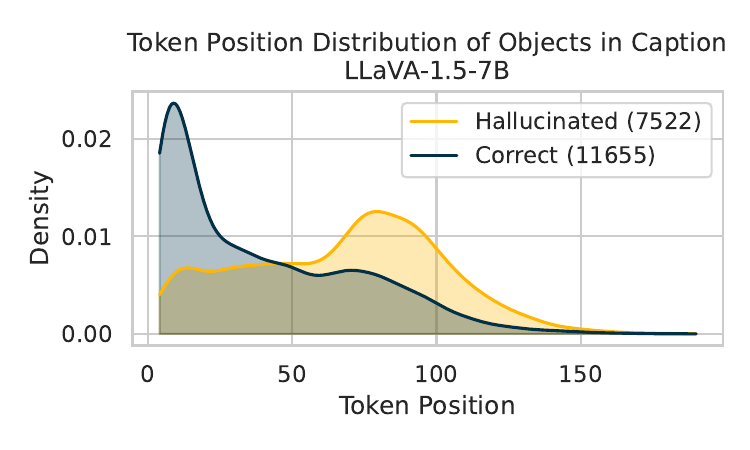}
    \includegraphics[width=0.48\linewidth, trim={0 0.1in 0 0.08in}, clip]{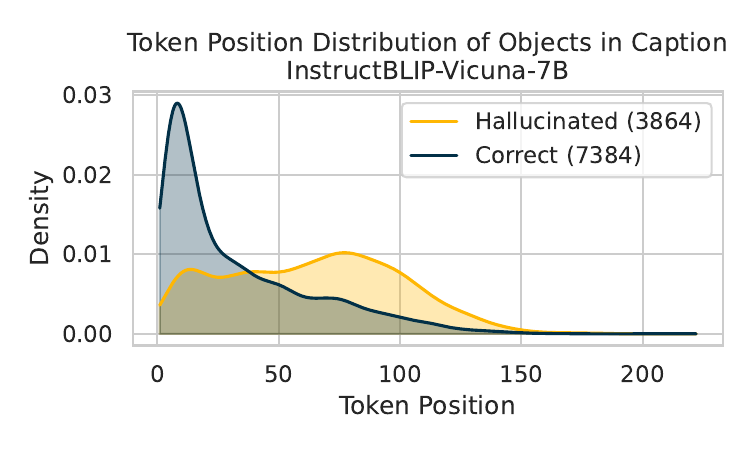}
    \includegraphics[width=0.48\linewidth, trim={0 0.1in 0 0.08in}, clip]{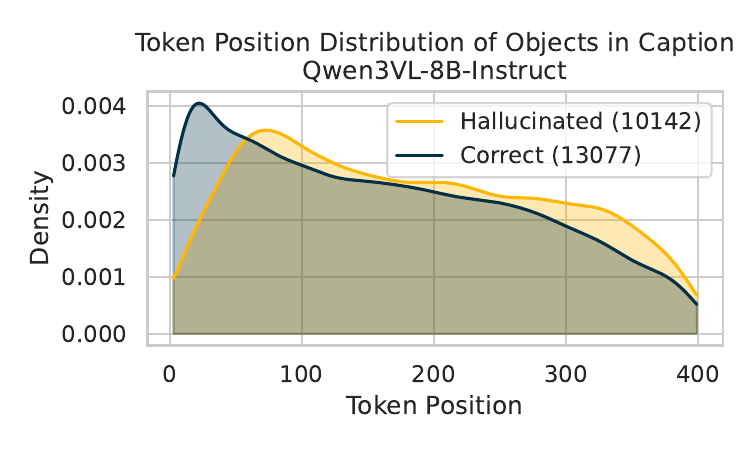}
    \includegraphics[width=0.48\linewidth, trim={0 0.1in 0 0.08in}, clip]{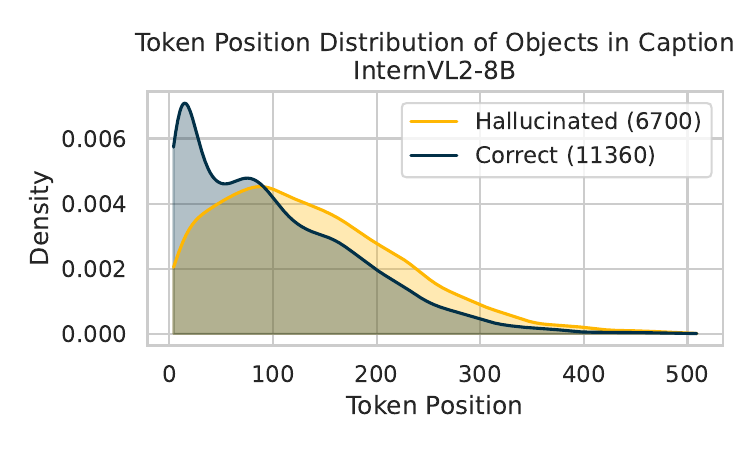}
    \includegraphics[width=0.48\linewidth, trim={0 0.1in 0 0.08in}, clip]{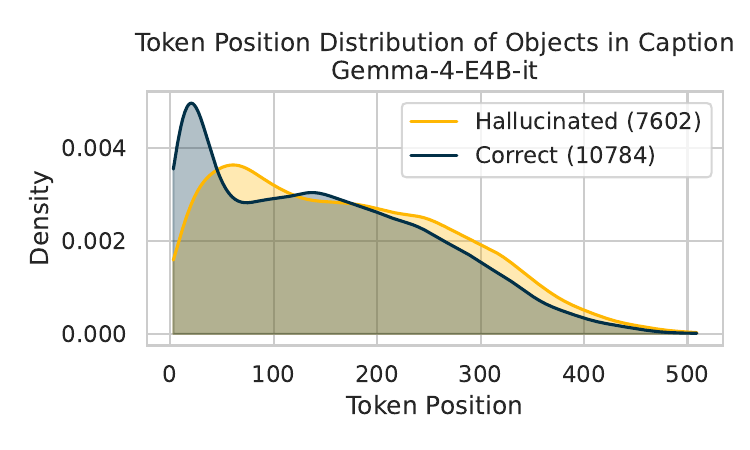}
    \caption{\textbf{Hallucination patterns on Objects365 slightly differ from MSCOCO.} While \llavalg, \llavasm, and \instructblip \ still hallucinate objects later than real objects, the distinction is not as pronounced for \qwen \ and \internvl. We plot token position distributions of real and hallucinated objects on Objects365 dataset.}
    \label{fig:o365_token_pos_dist}
\end{figure}

\begin{figure}[h!]
    \centering
    \small MSCOCO  \par
    \includegraphics[width=0.49\linewidth, trim={0 0.1in 0 0.08in}, clip]{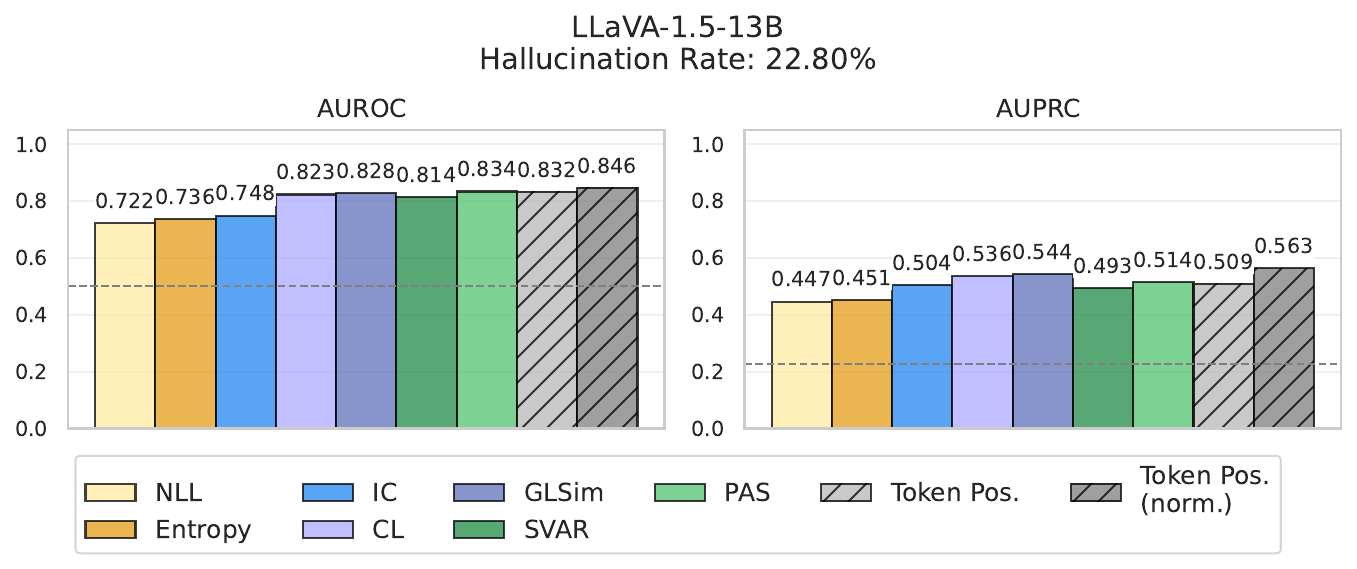}
    \includegraphics[width=0.49\linewidth, trim={0 0.1in 0 0.08in}, clip]{figures/llava1.5-13b/remove_replace.pdf}
    \\
    \small Objects365 \par
    \includegraphics[width=0.49\linewidth, trim={0 0.1in 0 0.08in}, clip]{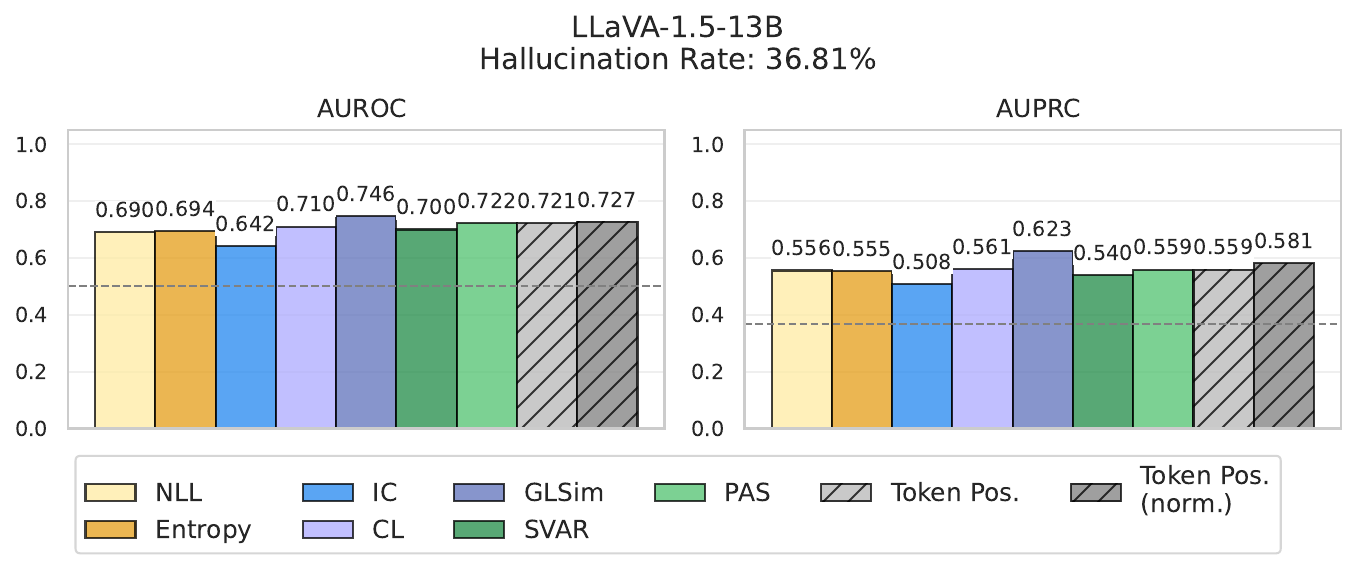}
    \includegraphics[width=0.49\linewidth, trim={0 0.1in 0 0.08in}, clip]{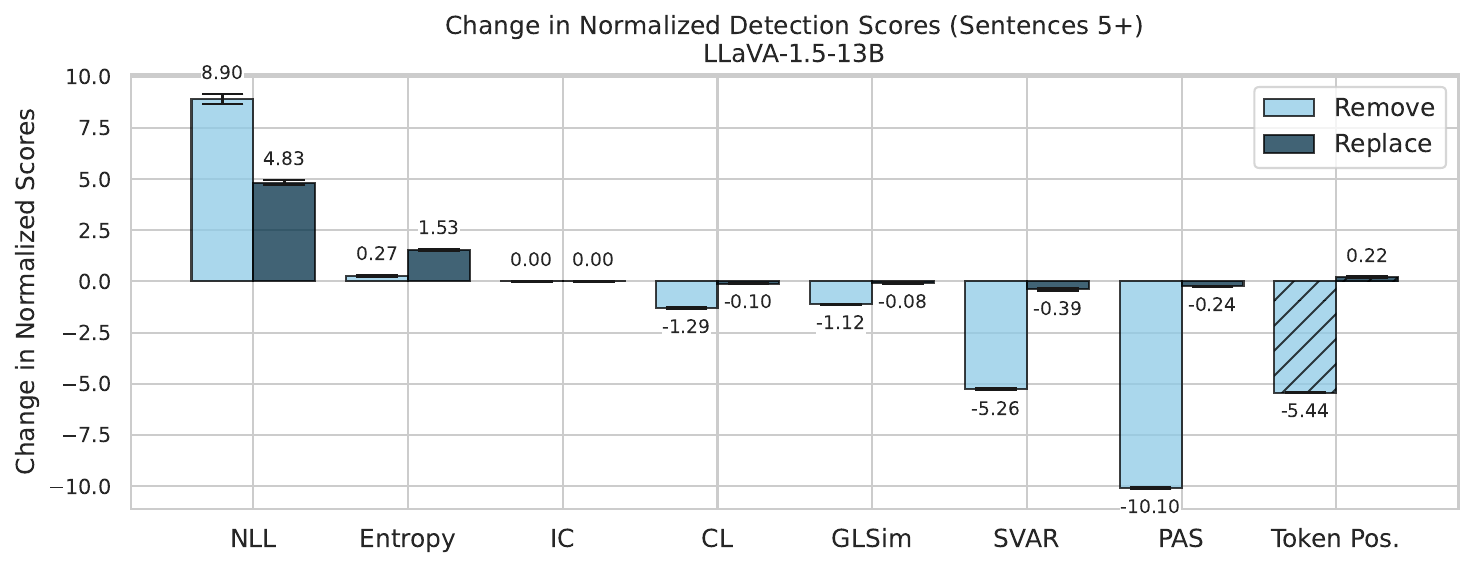}
    \\
    \small Prompt Variation \par
    \includegraphics[width=0.49\linewidth, trim={0 0.1in 0 0.08in}, clip]{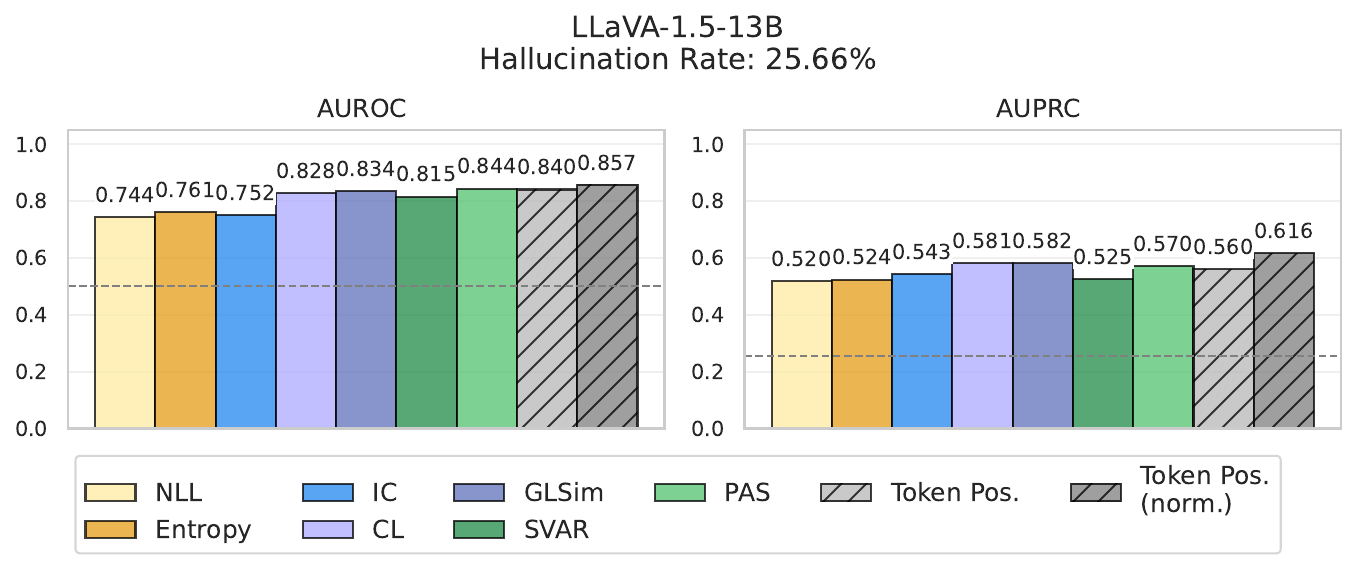}
    \includegraphics[width=0.49\linewidth, trim={0 0.1in 0 0.08in}, clip]{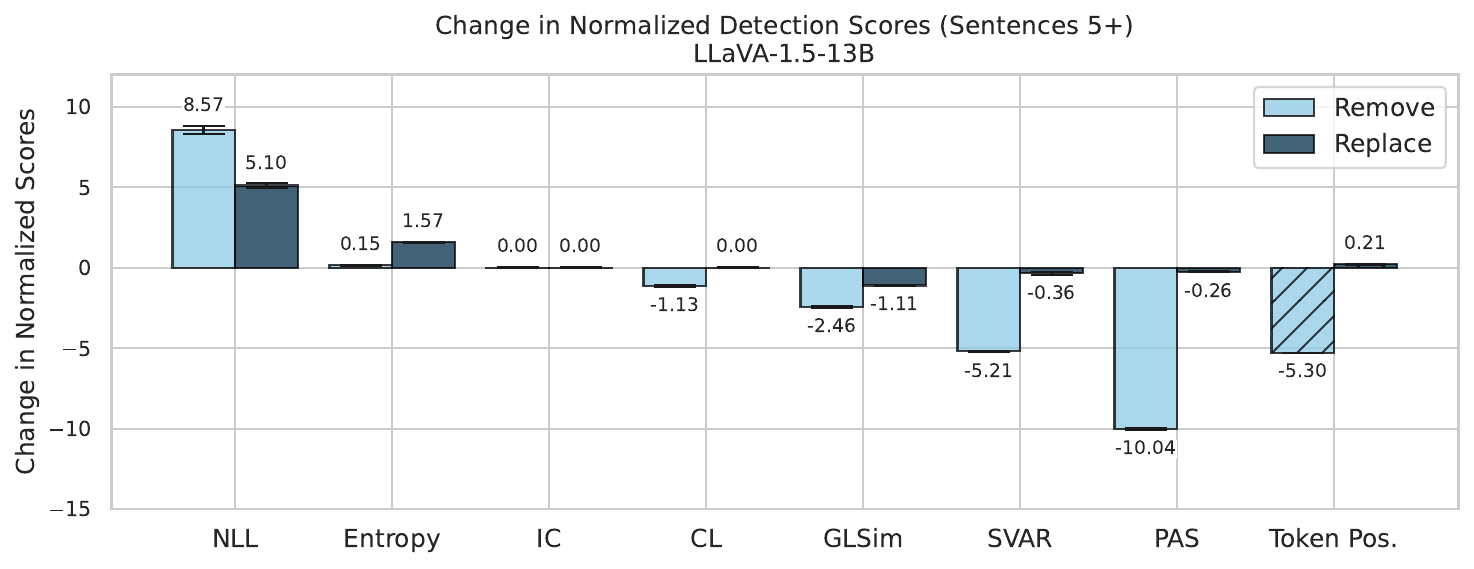}
    \\
    \small Repeated Trials \par
    \includegraphics[width=0.49\linewidth, trim={0 0.1in 0 0.08in}, clip]{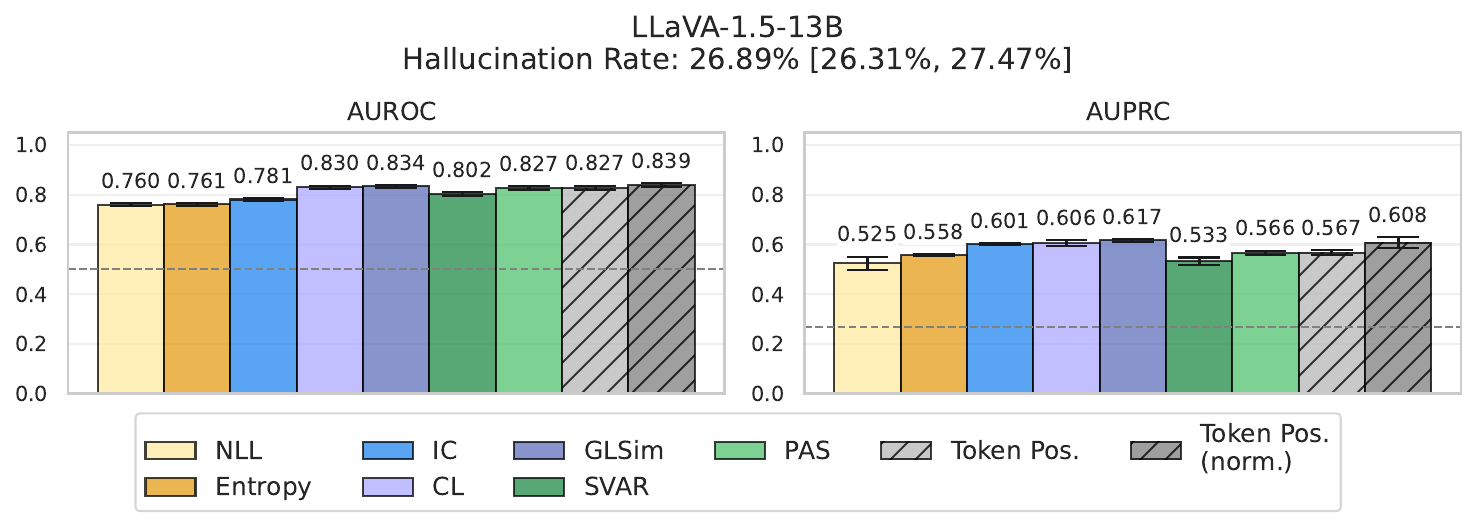}
    \includegraphics[width=0.49\linewidth, trim={0 0.1in 0 0.08in}, clip]{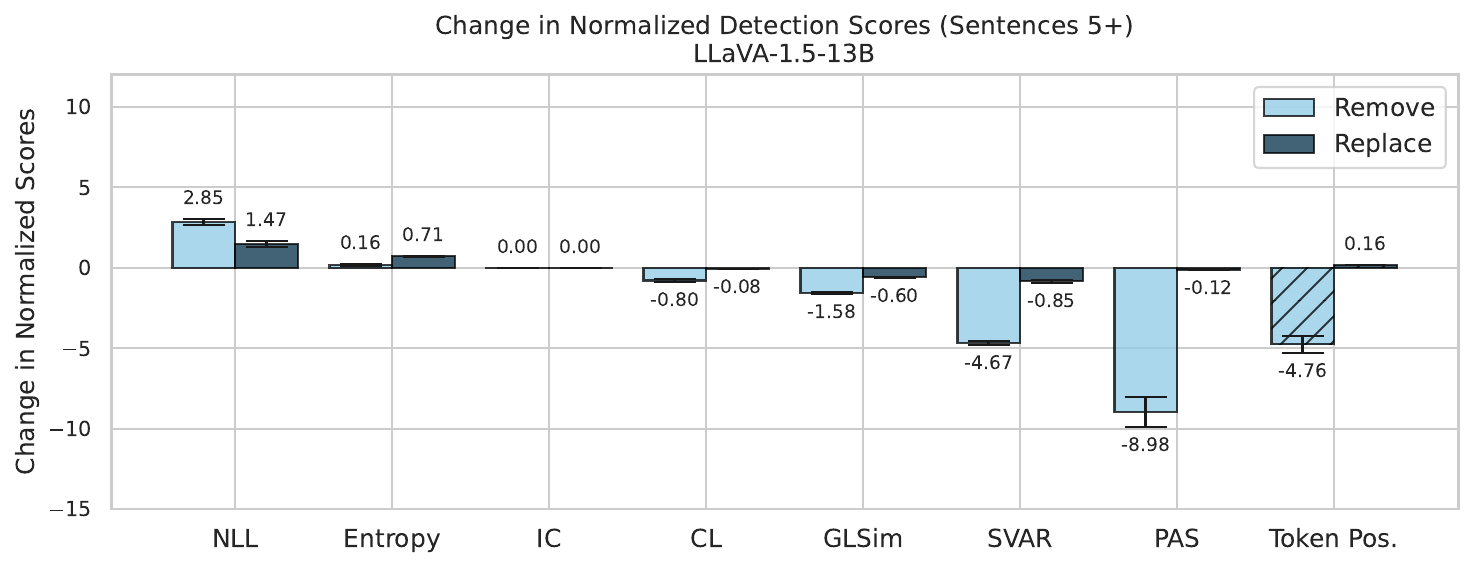}
    \\
    \small Context-aware Filtering \par
    \includegraphics[width=0.49\linewidth, trim={0 0.1in 0 0.08in}, clip]{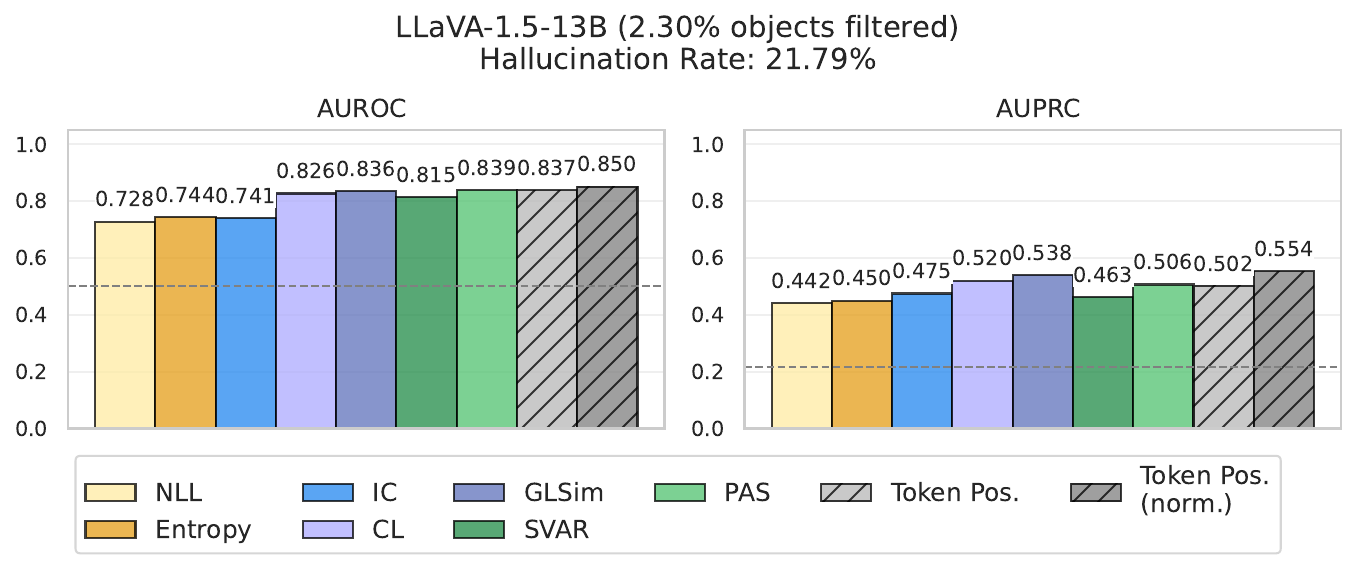}
    \includegraphics[width=0.49\linewidth, trim={0 0.1in 0 0.08in}, clip]{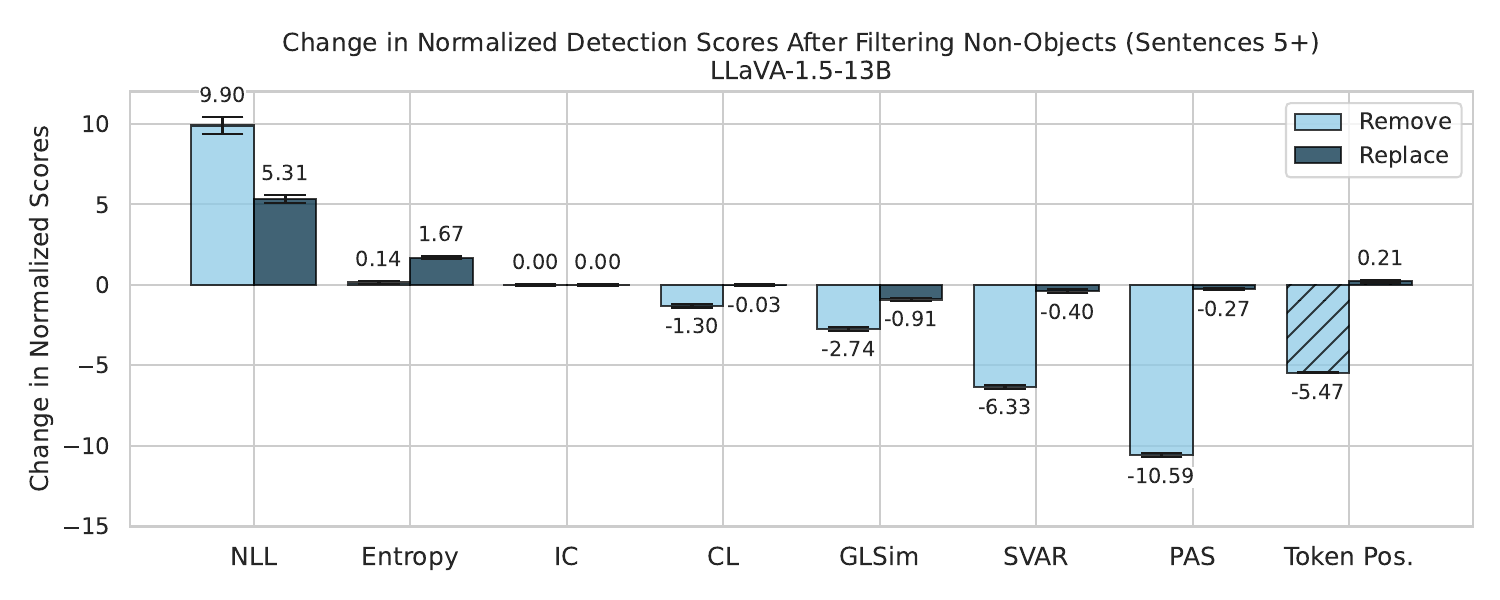}
    \\
    \small No Filtering ``orange'' \par
    \includegraphics[width=0.49\linewidth, trim={0 0.1in 0 0.08in}, clip]{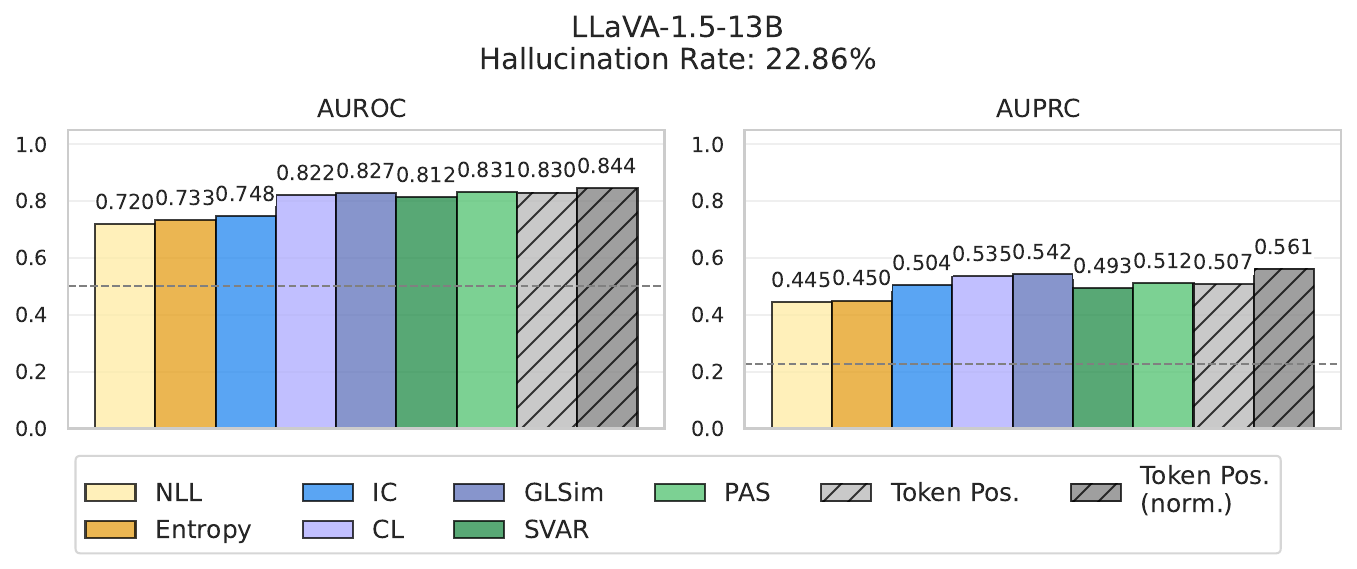}
    \includegraphics[width=0.49\linewidth, trim={0 0.1in 0 0.08in}, clip]{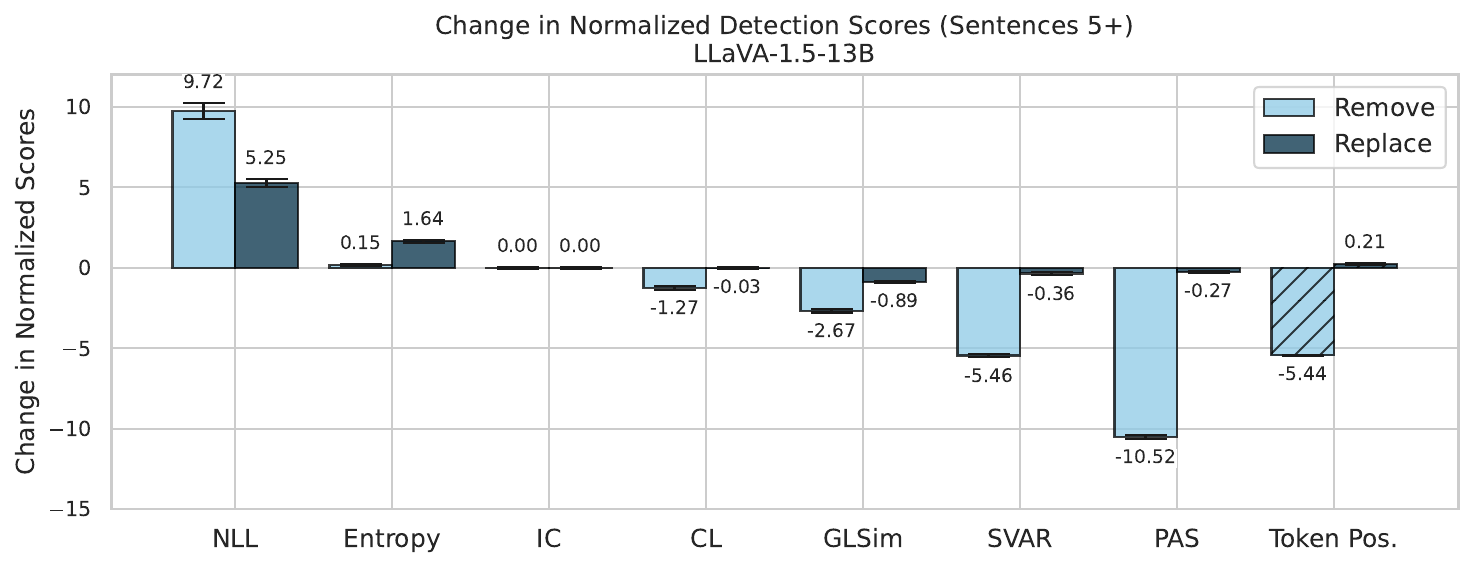}
    \caption{\textbf{Left:} Performance of hallucination detection methods on \llavalg \ generalizes across various data and prompt settings. \textbf{Right:} Method behavior with remove/replace preceding sentences \llavalg \ generalizes across various data and prompt settings.}
    \label{fig:llava13b_perf_generalize1}
\end{figure}

\begin{figure}[h!]
    \centering
    \small MSCOCO  \par
    \includegraphics[width=0.49\linewidth, trim={0 0.1in 0 0.08in}, clip]{figures/llava1.5-13b/hall_source_auprc.pdf}
    \includegraphics[width=0.49\linewidth, trim={0 0.1in 0 0.08in}, clip]{figures/llava1.5-13b/auprc_blank_performance_diff.pdf}
    \\
    \small Objects365 \par
    \includegraphics[width=0.49\linewidth, trim={0 0.1in 0 0.08in}, clip]{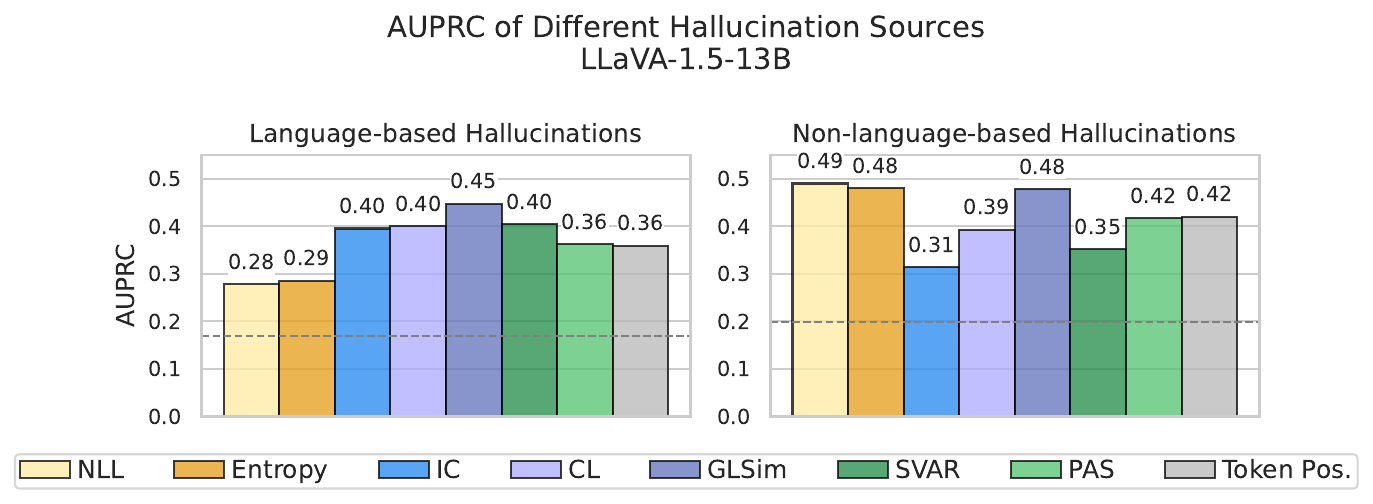}
    \includegraphics[width=0.49\linewidth, trim={0 0.1in 0 0.08in}, clip]{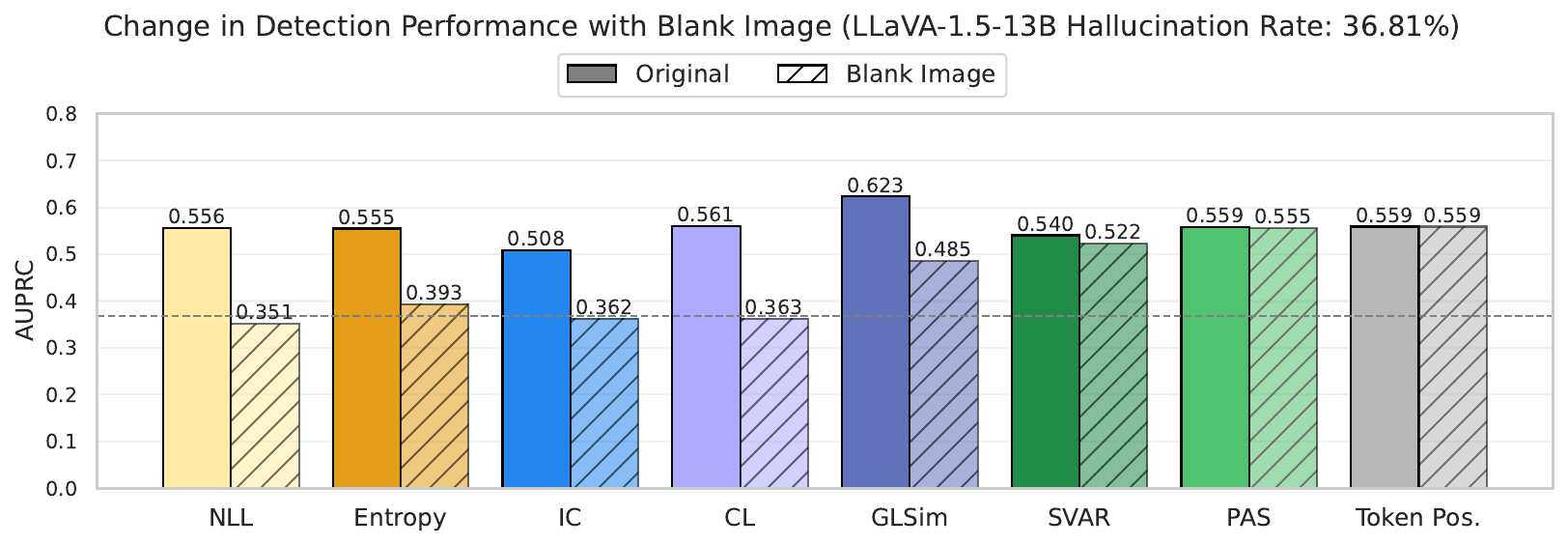}
    \\
    \small Prompt Variation \par
    \includegraphics[width=0.49\linewidth, trim={0 0.1in 0 0.08in}, clip]{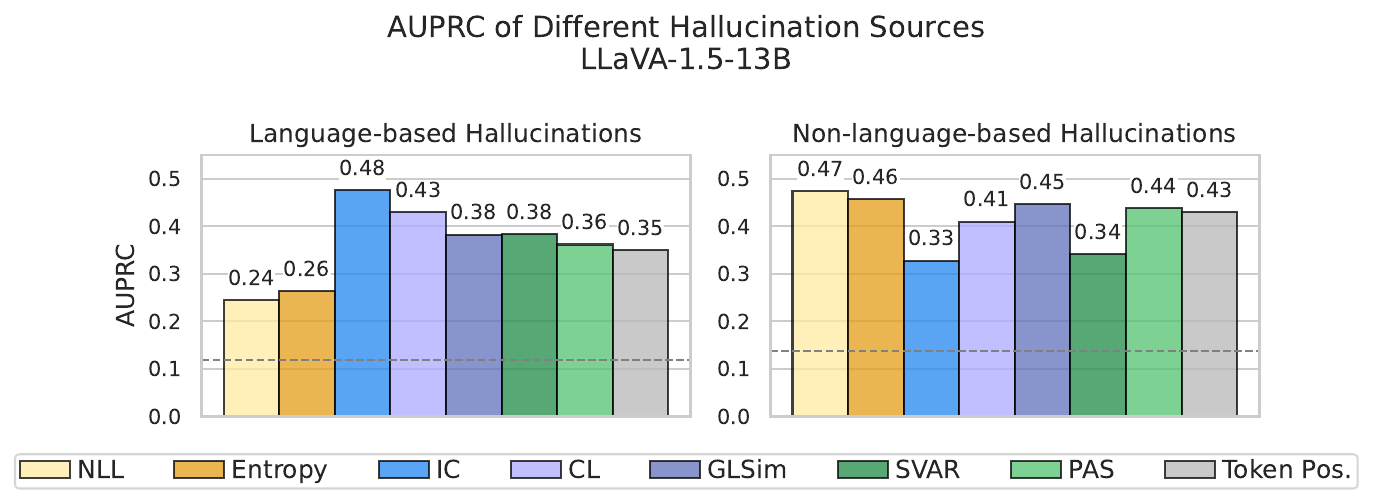}
    \includegraphics[width=0.49\linewidth, trim={0 0.1in 0 0.08in}, clip]{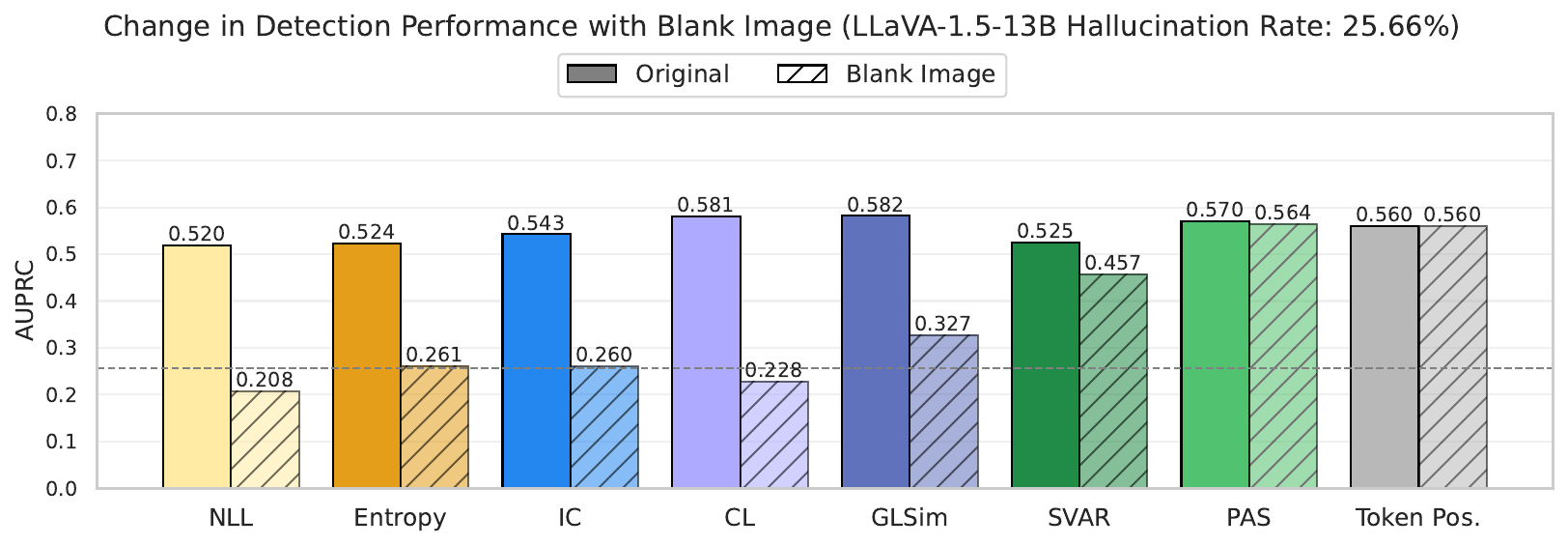}
    \\
    \small Repeated Trials \par
    \includegraphics[width=0.49\linewidth]{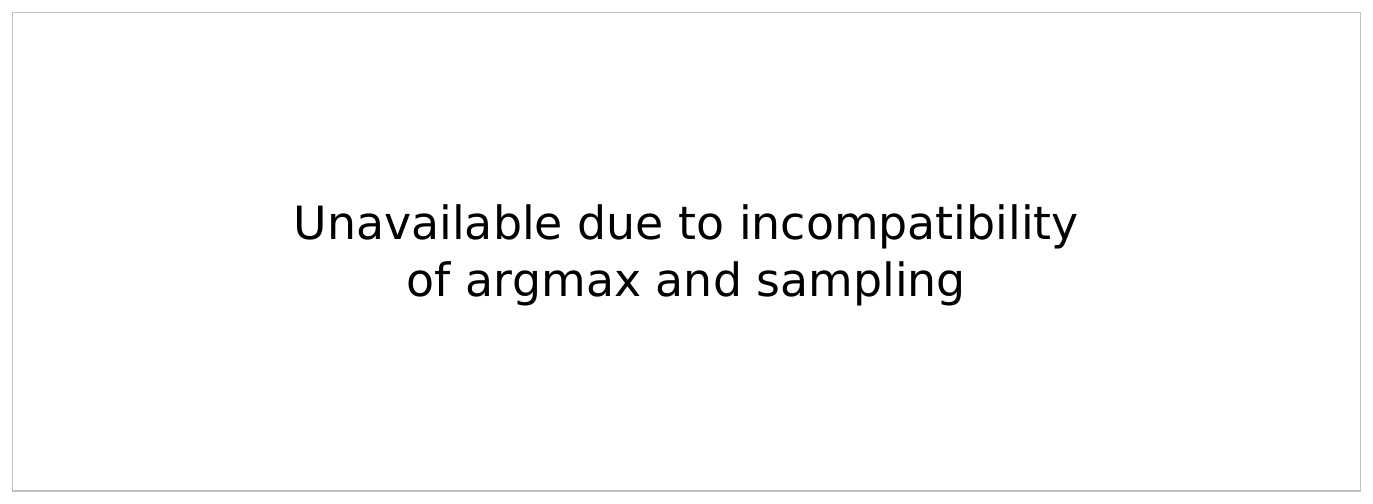}
    \includegraphics[width=0.49\linewidth, trim={0 0.1in 0 0.08in}, clip]{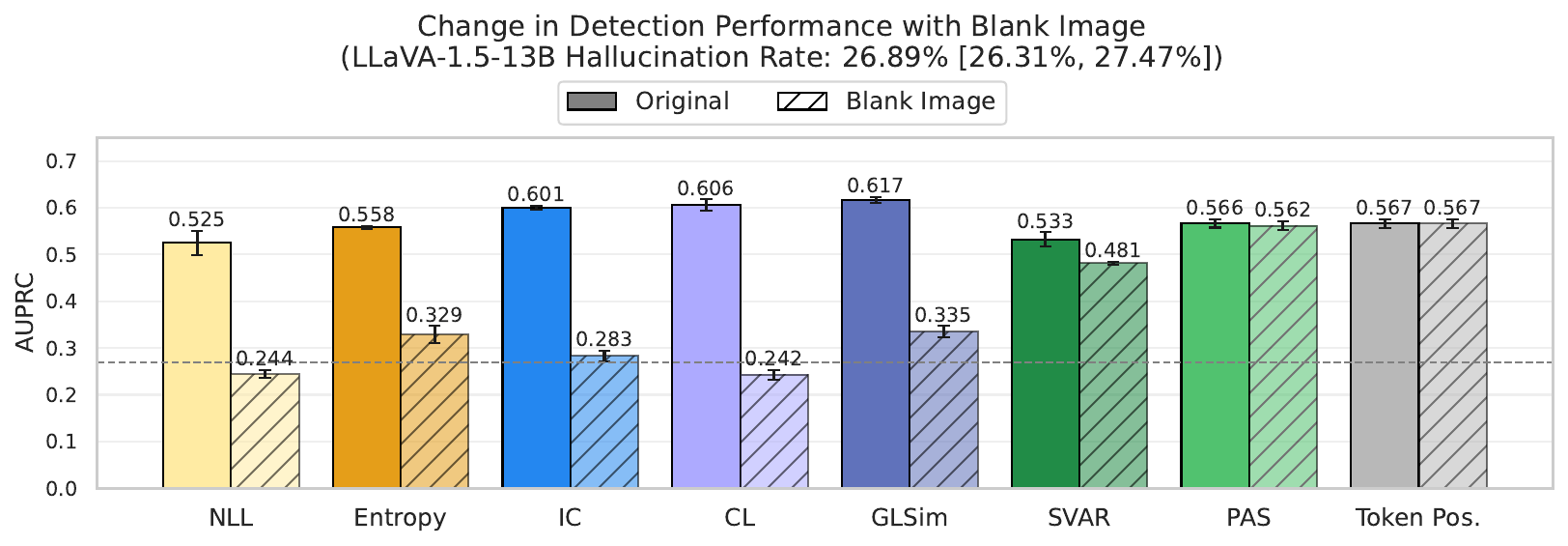}
    \\
    \small Context-aware Filtering \par
    \includegraphics[width=0.49\linewidth, trim={0 0.1in 0 0.08in}, clip]{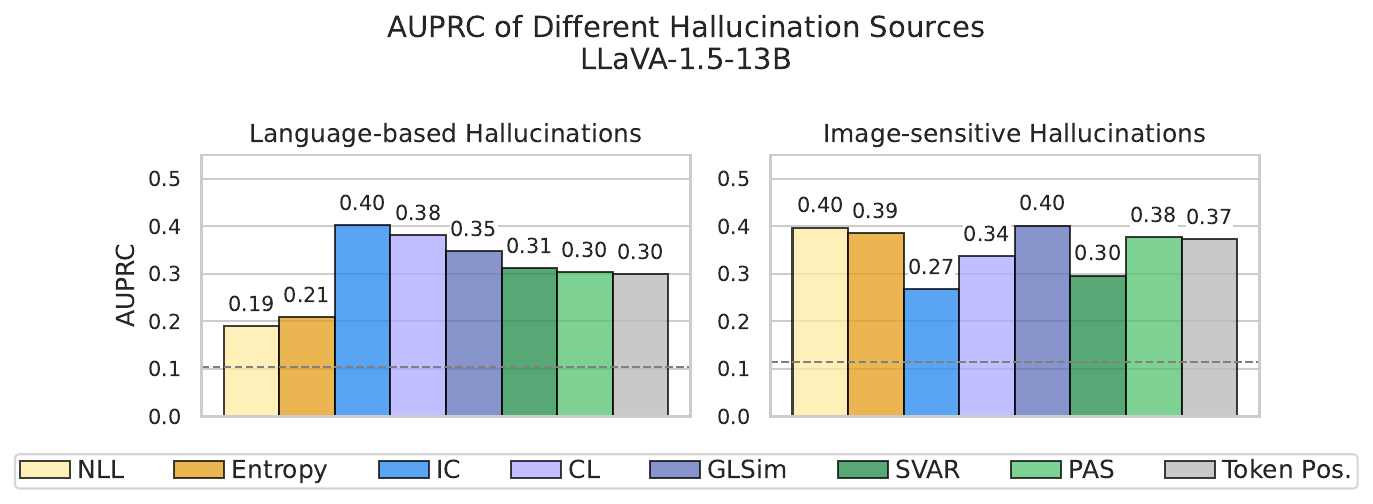}
    \includegraphics[width=0.49\linewidth, trim={0 0.1in 0 0.08in}, clip]{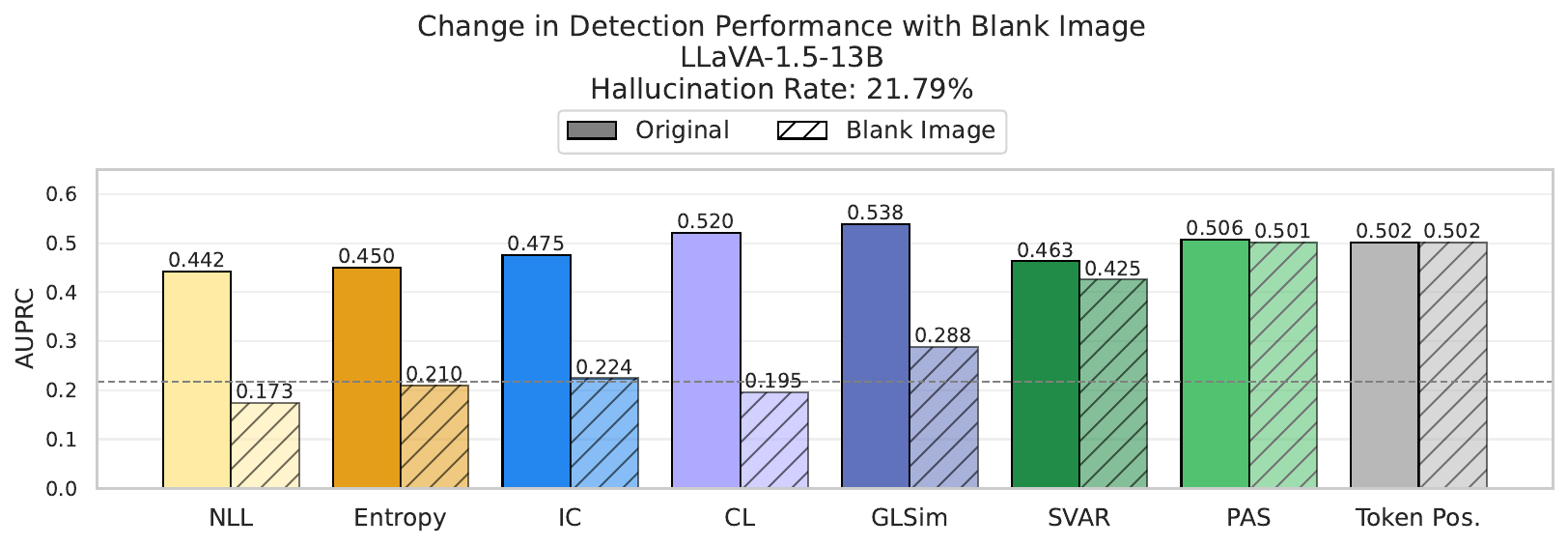}
    \\
    \small No Filtering ``orange'' \par
    \includegraphics[width=0.49\linewidth, trim={0 0.1in 0 0.08in}, clip]{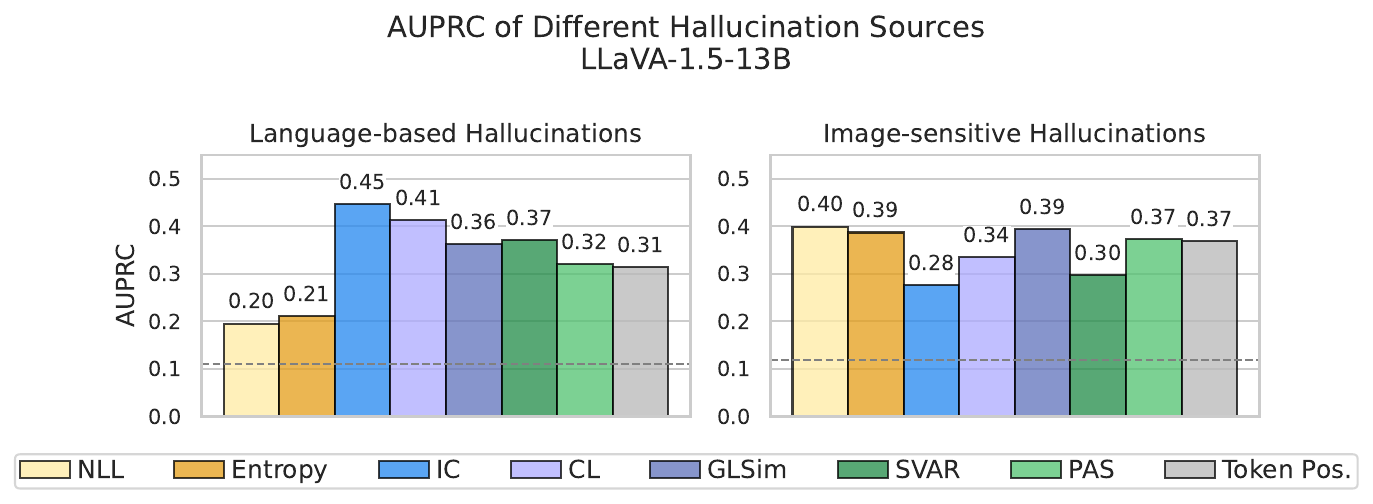}
    \includegraphics[width=0.49\linewidth, trim={0 0.1in 0 0.08in}, clip]{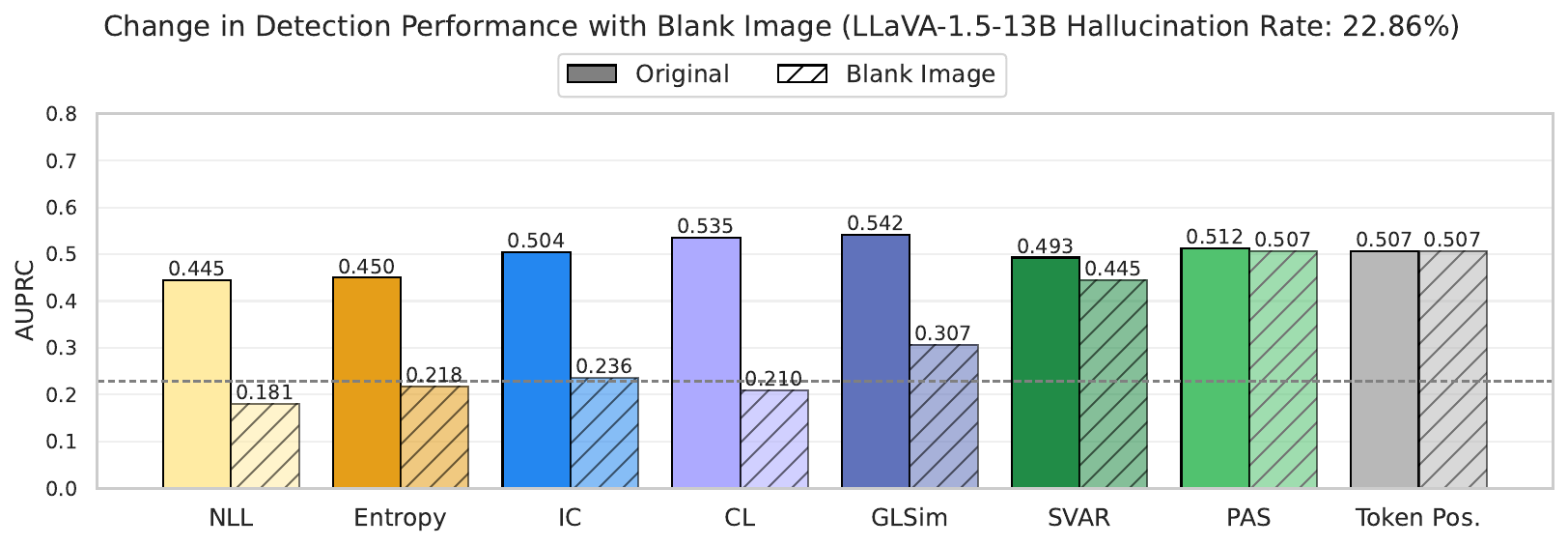}
    \caption{Method behavior partitioning by hallucination modality and visual groundedness on \llavalg \ generalizes across various data and prompt settings.}
    \label{fig:llava13b_perf_generalize2}
\end{figure}

\begin{figure}[h!]
    \centering
    \small MSCOCO  \par
    \includegraphics[width=0.49\linewidth, trim={0 0.1in 0 0.08in}, clip]{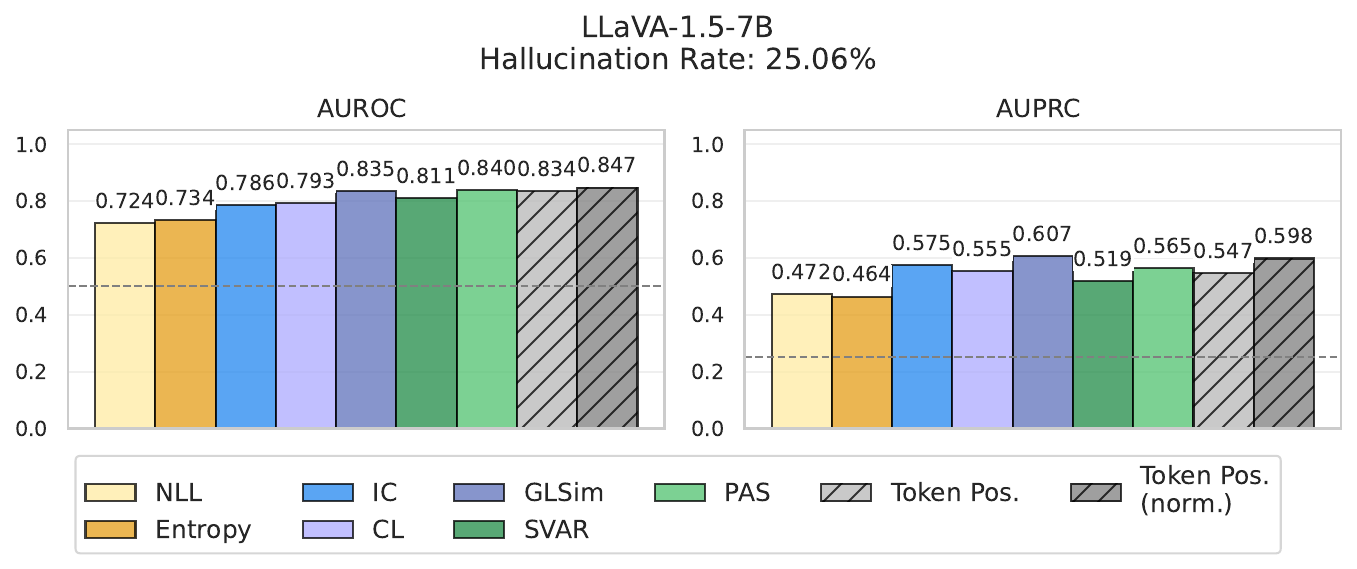}
    \includegraphics[width=0.49\linewidth, trim={0 0.1in 0 0.08in}, clip]{figures/llava1.5-7b/remove_replace.pdf}
    \\
    \small Objects365 \par
    \includegraphics[width=0.49\linewidth, trim={0 0.1in 0 0.08in}, clip]{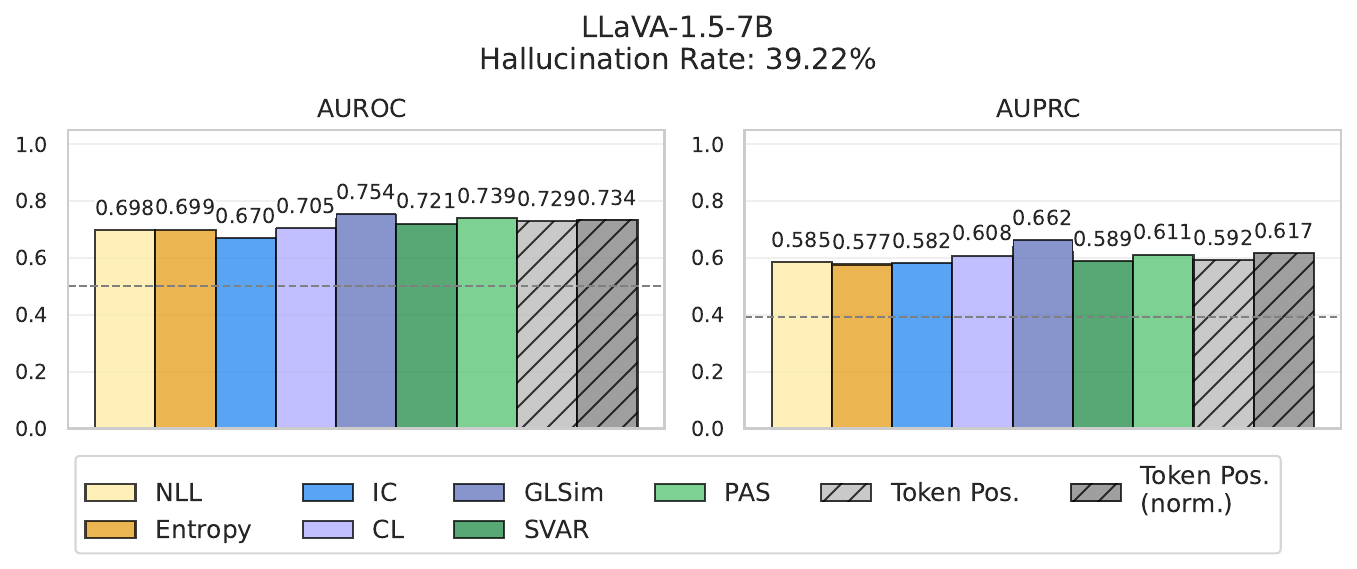}
    \includegraphics[width=0.49\linewidth, trim={0 0.1in 0 0.08in}, clip]{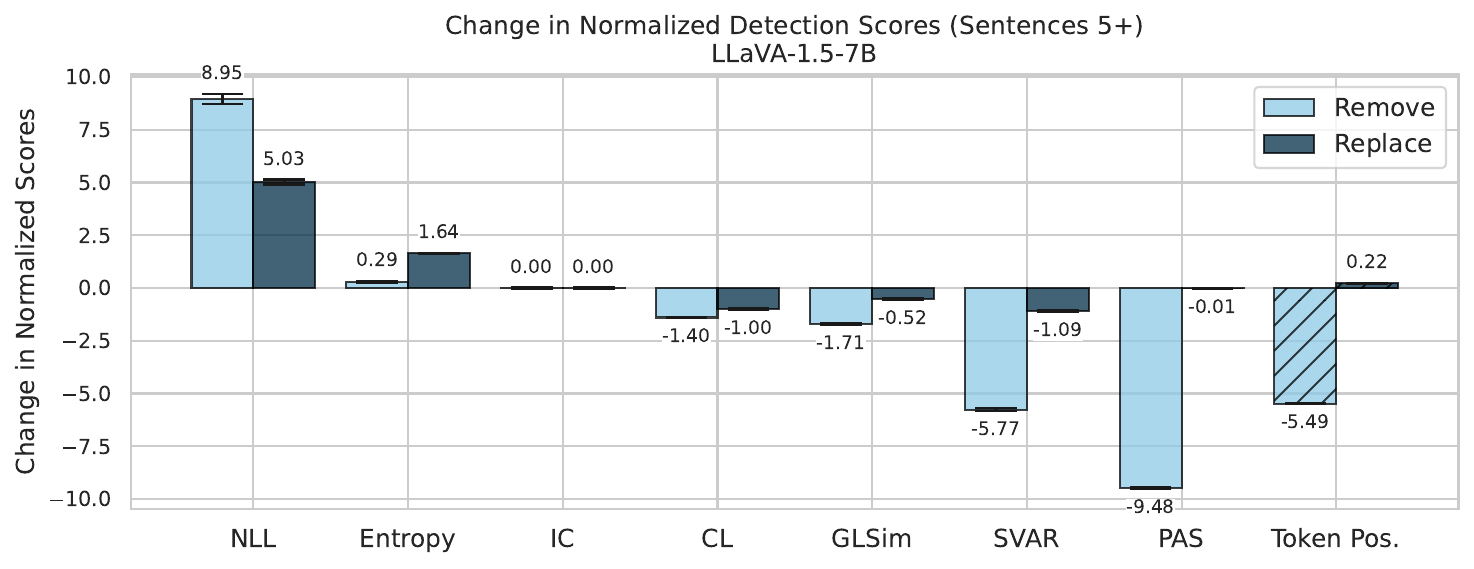}
    \\
    \small Prompt Variation \par
    \includegraphics[width=0.49\linewidth, trim={0 0.1in 0 0.08in}, clip]{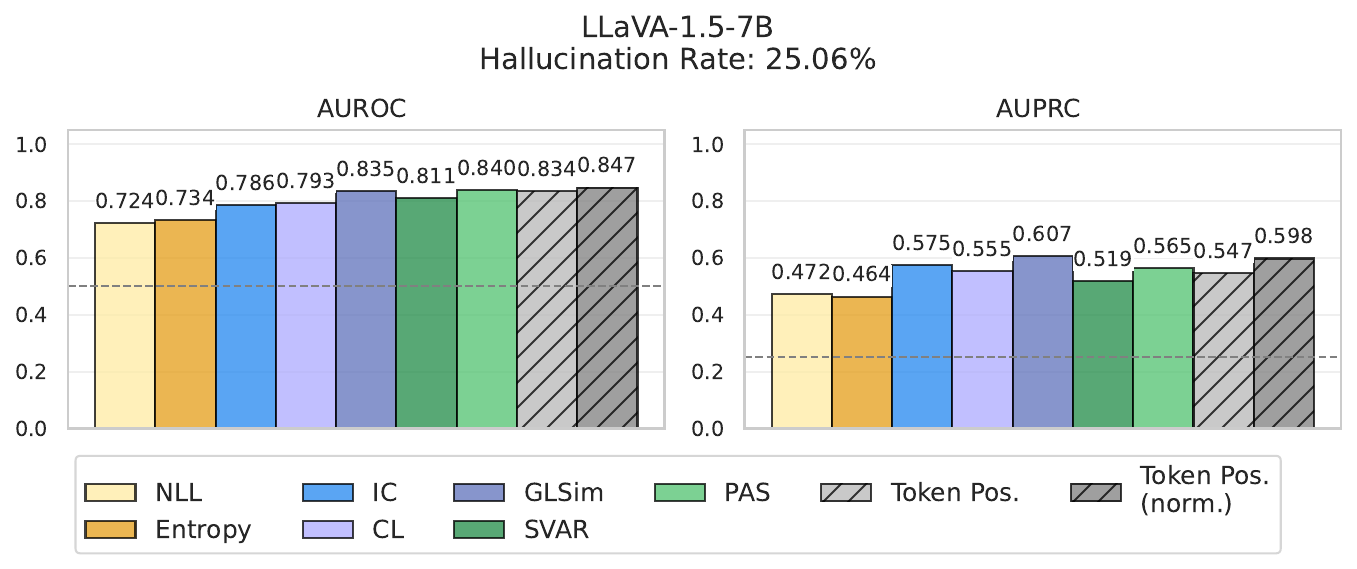}
    \includegraphics[width=0.49\linewidth, trim={0 0.1in 0 0.08in}, clip]{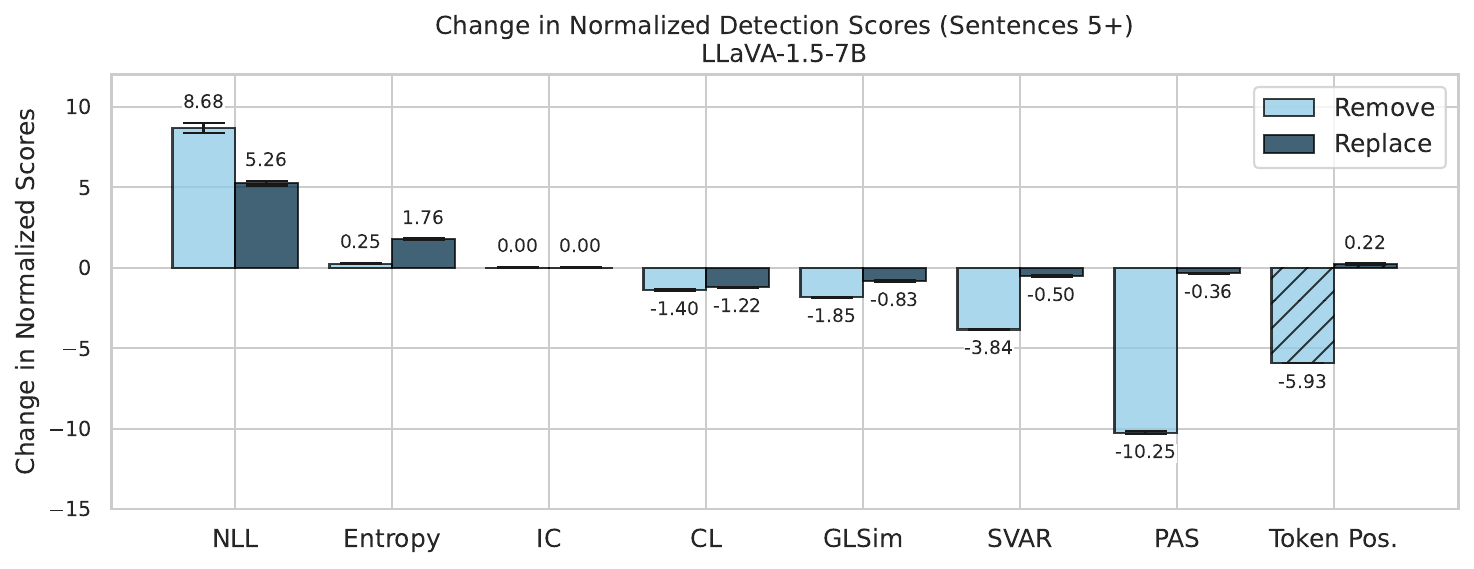}
    \\
    \small Repeated Trials \par
    \includegraphics[width=0.49\linewidth, trim={0 0.1in 0 0.08in}, clip]{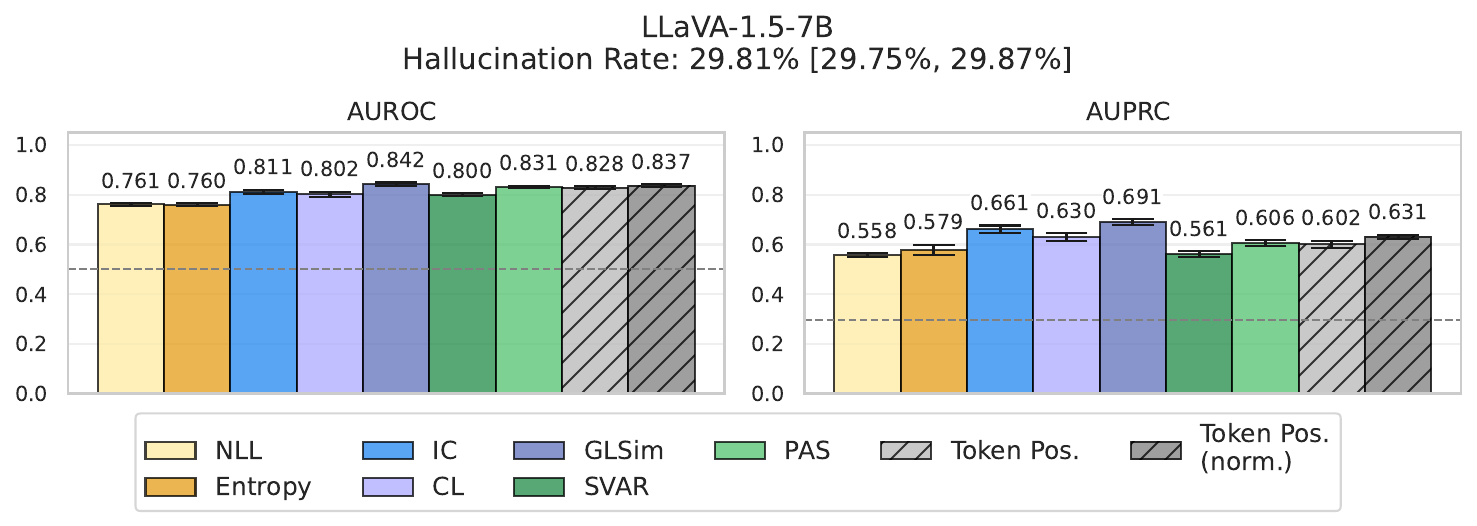}
    \includegraphics[width=0.49\linewidth, trim={0 0.1in 0 0.08in}, clip]{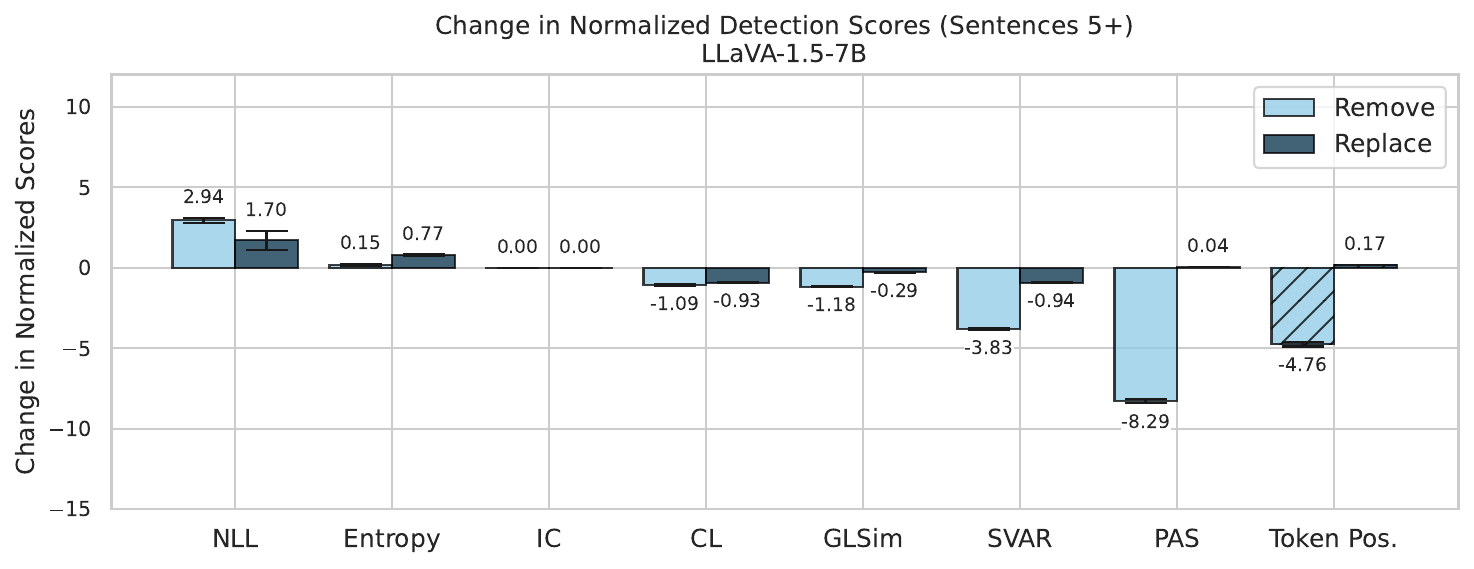}
    \\
    \small Context-aware Filtering \par
    \includegraphics[width=0.49\linewidth, trim={0 0.1in 0 0.08in}, clip]{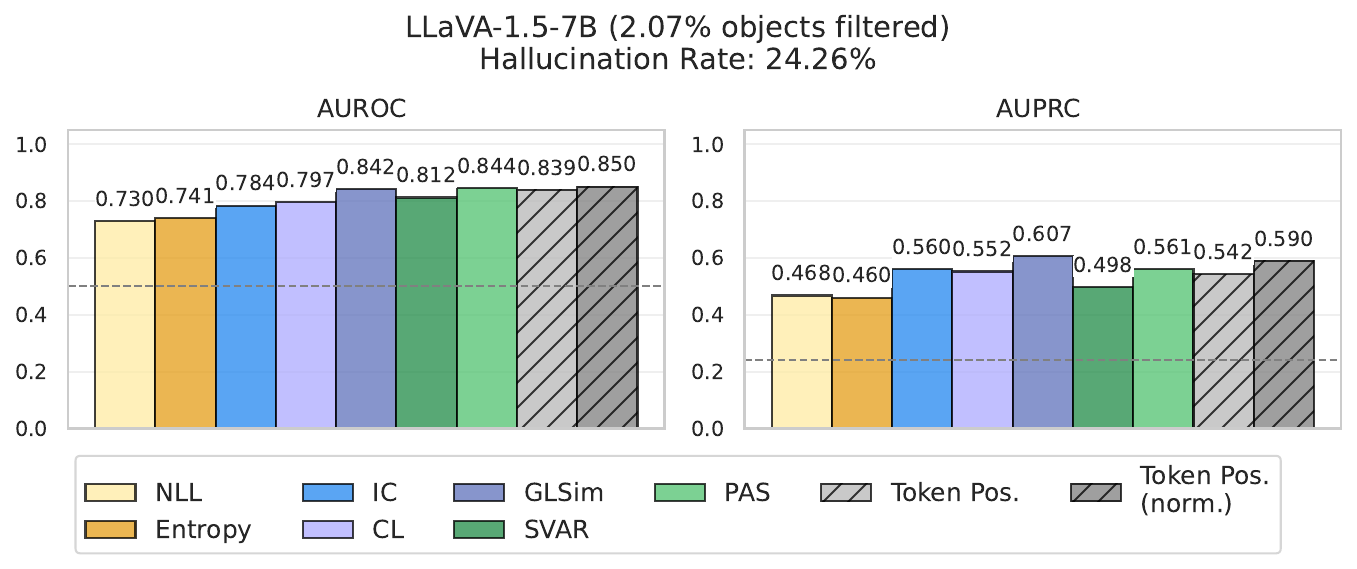}
    \includegraphics[width=0.49\linewidth, trim={0 0.1in 0 0.08in}, clip]{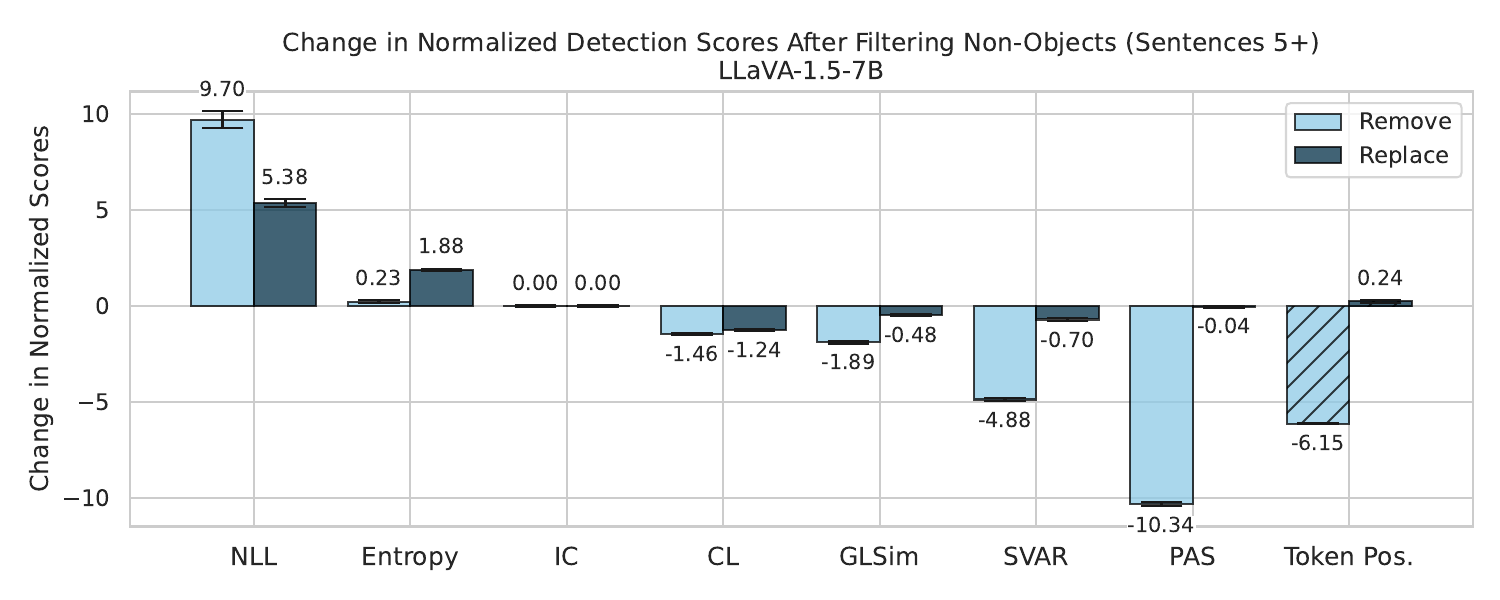}
    \\
    \small No Filtering ``orange'' \par
    \includegraphics[width=0.49\linewidth, trim={0 0.1in 0 0.08in}, clip]{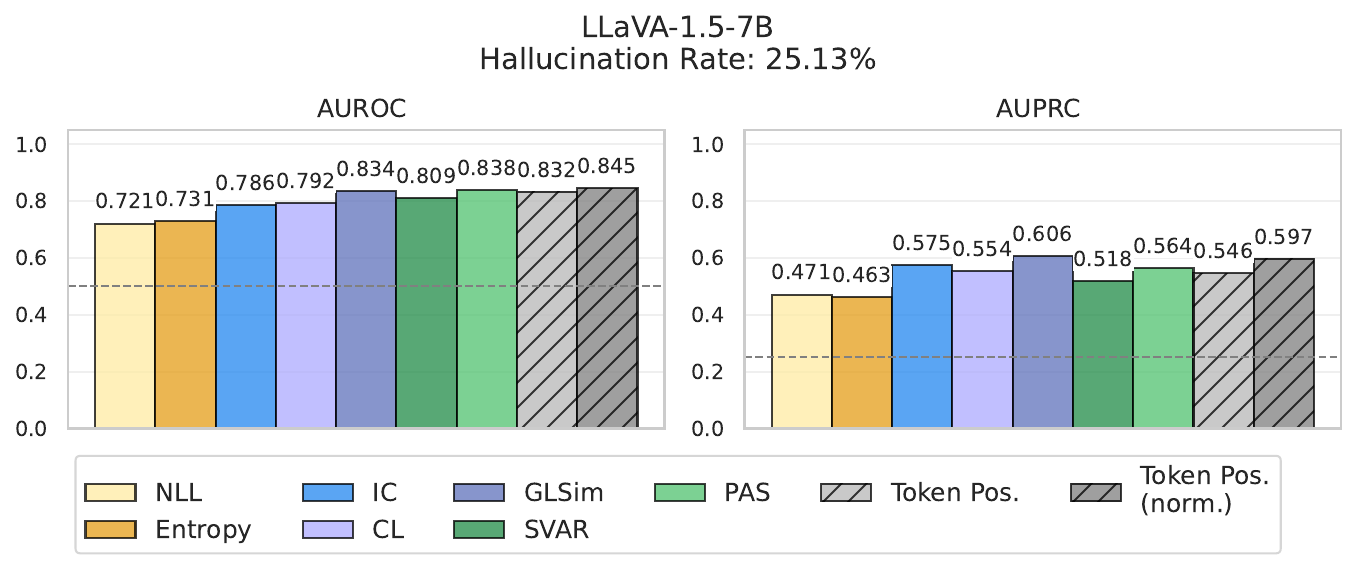}
    \includegraphics[width=0.49\linewidth, trim={0 0.1in 0 0.08in}, clip]{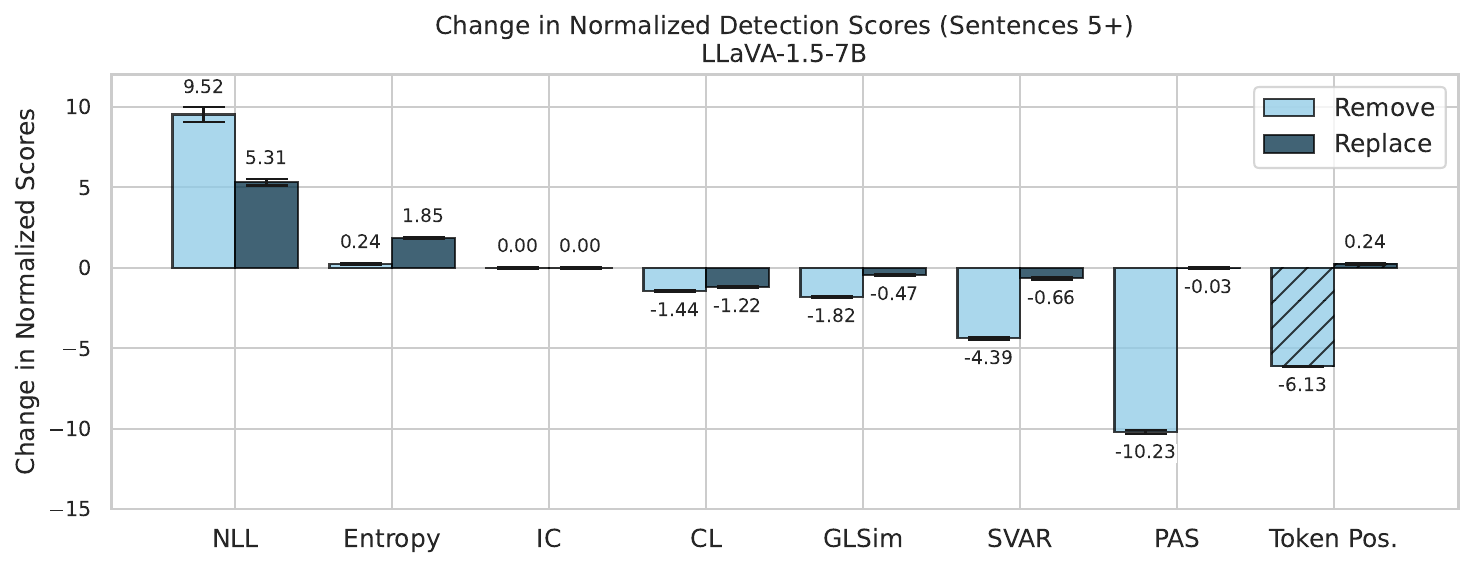}
    \caption{\textbf{Left:} Performance of hallucination detection methods on \llavasm \ generalizes across various data and prompt settings. \textbf{Right:} Method behavior with remove/replace preceding sentences \llavasm \ generalizes across various data and prompt settings.}
    \label{fig:llava7b_perf_generalize1}
\end{figure}

\begin{figure}[h!]
    \centering
    \small MSCOCO  \par
    \includegraphics[width=0.49\linewidth, trim={0 0.1in 0 0.08in}, clip]{figures/llava1.5-7b/hall_source_auprc.pdf}
    \includegraphics[width=0.49\linewidth, trim={0 0.1in 0 0.08in}, clip]{figures/llava1.5-7b/auprc_blank_performance_diff.pdf}
    \\
    \small Objects365 \par
    \includegraphics[width=0.49\linewidth, trim={0 0.1in 0 0.08in}, clip]{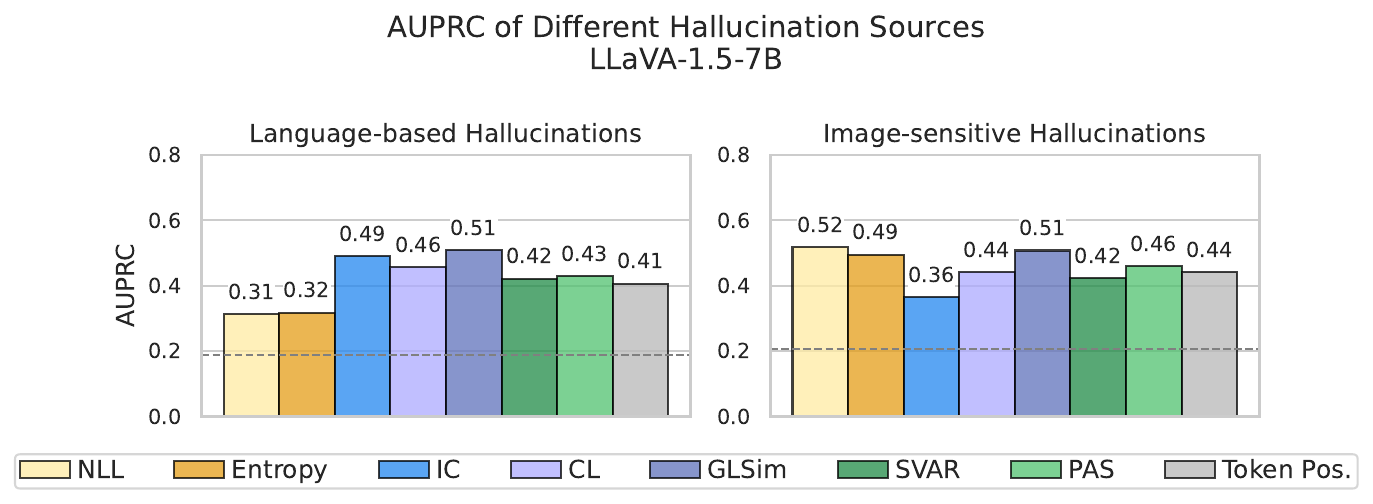}
    \includegraphics[width=0.49\linewidth, trim={0 0.1in 0 0.08in}, clip]{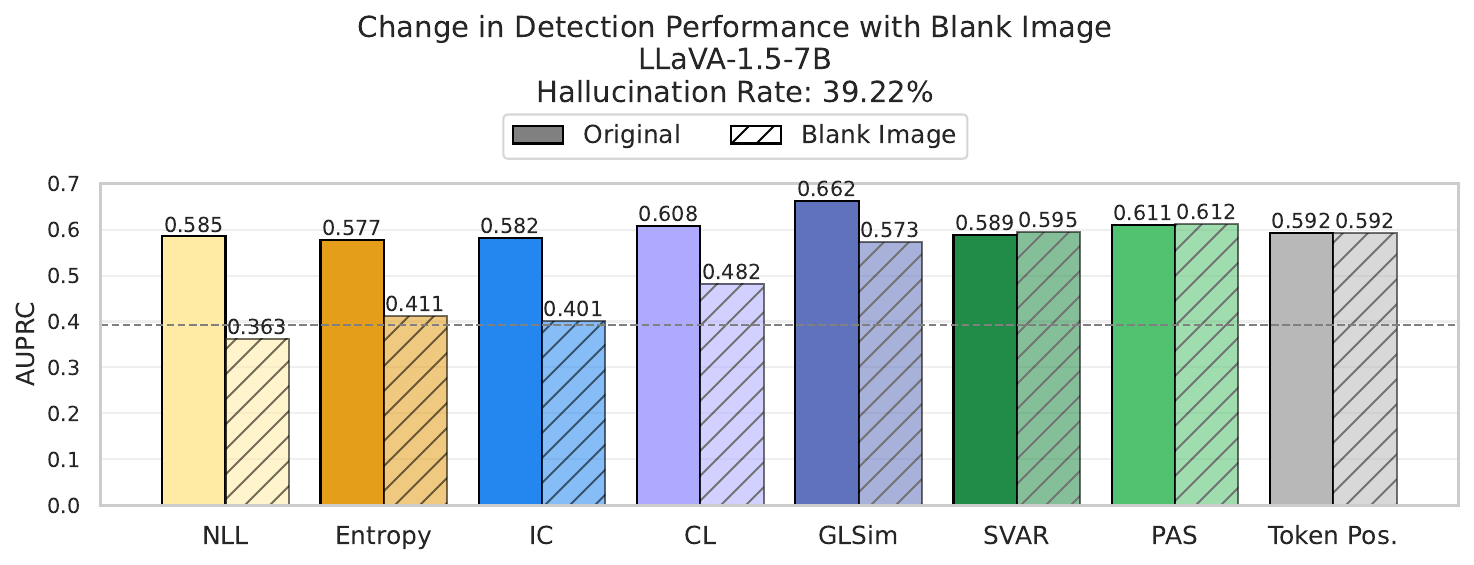}
    \\
    \small Prompt Variation \par
    \includegraphics[width=0.49\linewidth, trim={0 0.1in 0 0.08in}, clip]{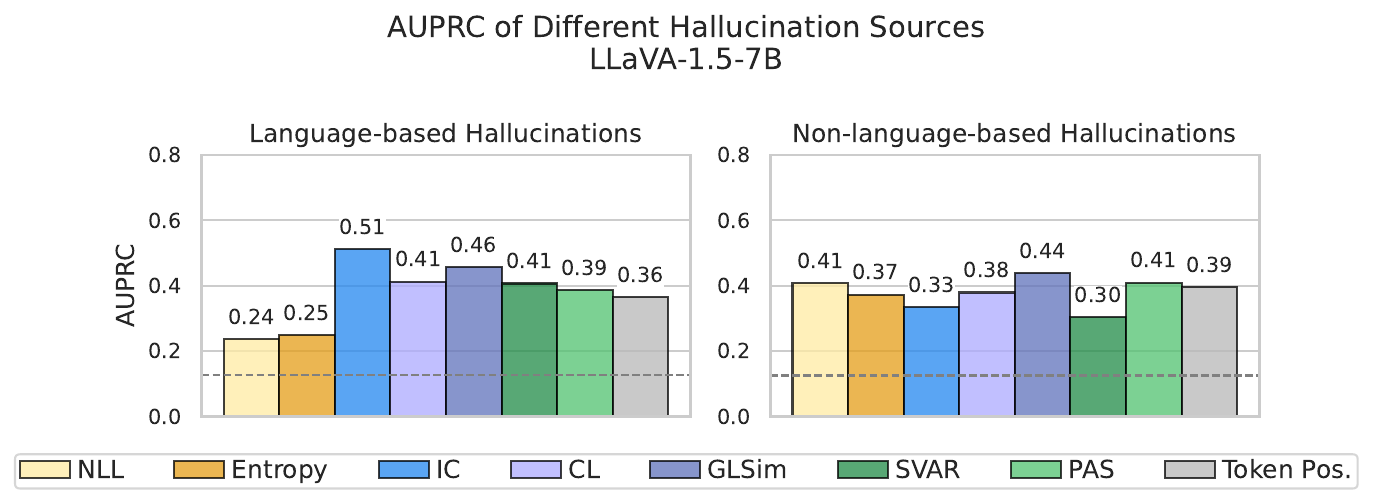}
    \includegraphics[width=0.49\linewidth, trim={0 0.1in 0 0.08in}, clip]{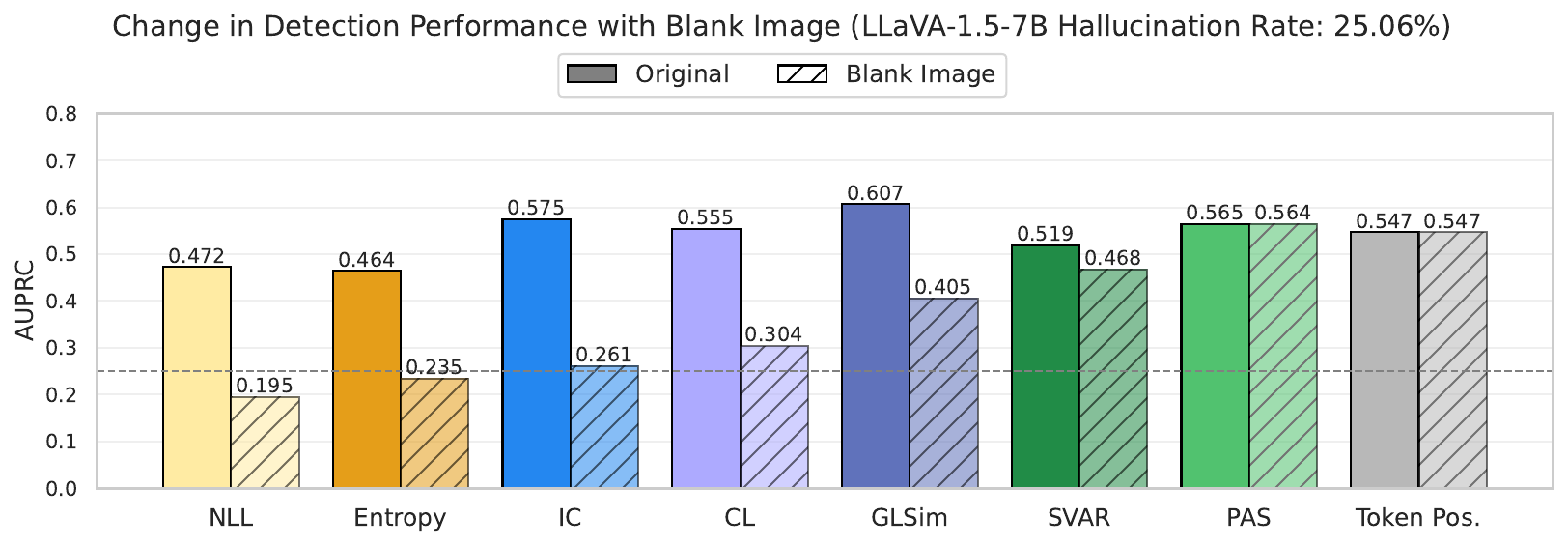}
    \\
    \small Repeated Trials \par
    \includegraphics[width=0.49\linewidth]{figures/coco_sampling/unavailable_figure.pdf}
    \includegraphics[width=0.49\linewidth, trim={0 0.1in 0 0.08in}, clip]{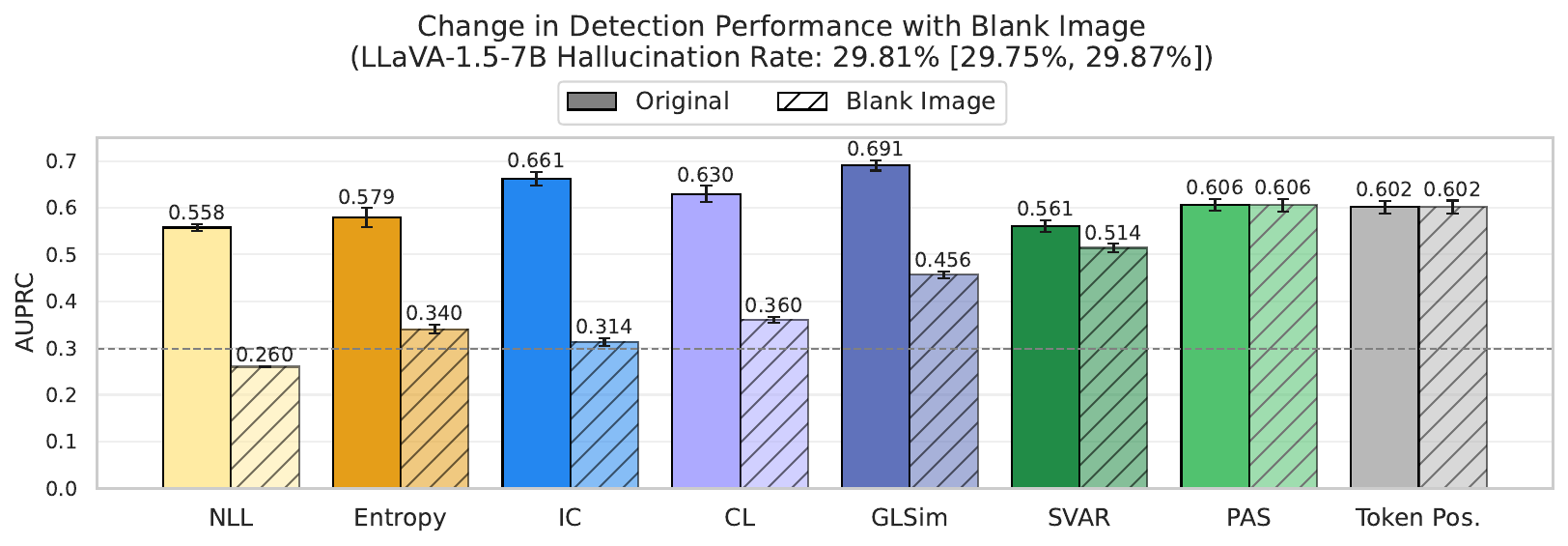}
    \\
    \small Context-aware Filtering \par
    \includegraphics[width=0.49\linewidth, trim={0 0.1in 0 0.08in}, clip]{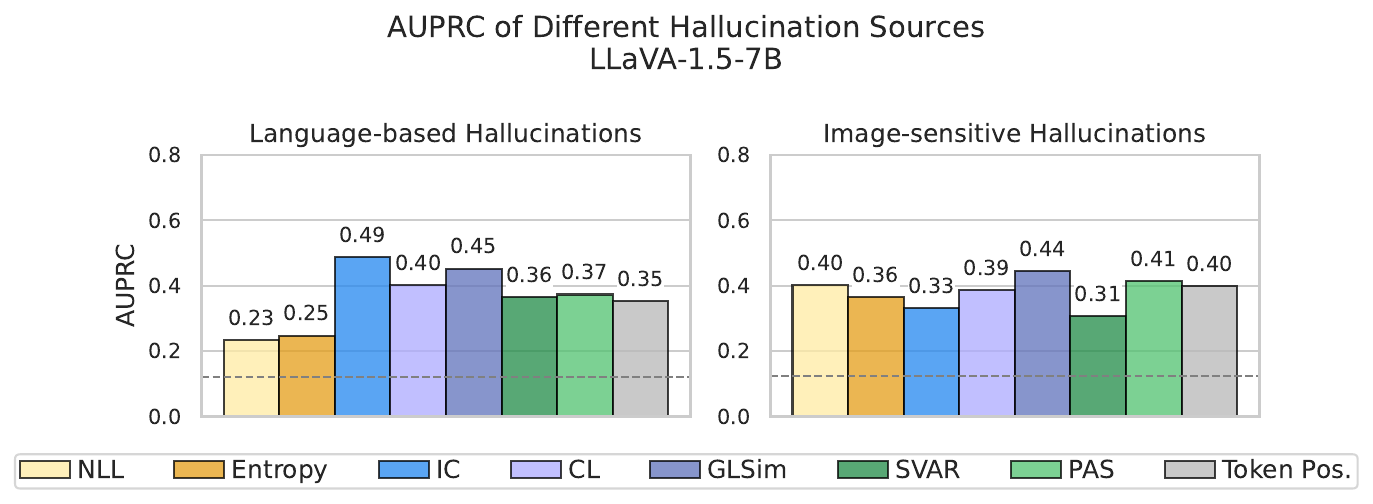}
    \includegraphics[width=0.49\linewidth, trim={0 0.1in 0 0.08in}, clip]{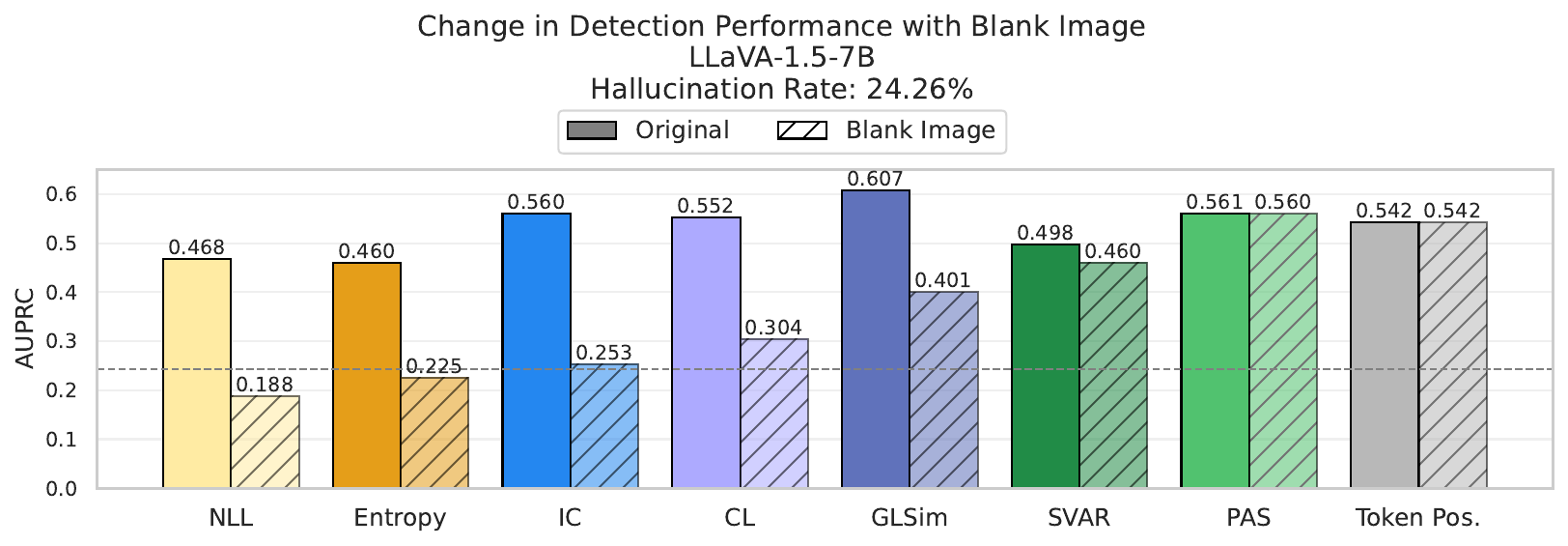}
    \\
    \small No Filtering ``orange''\par
    \includegraphics[width=0.49\linewidth, trim={0 0.1in 0 0.08in}, clip]{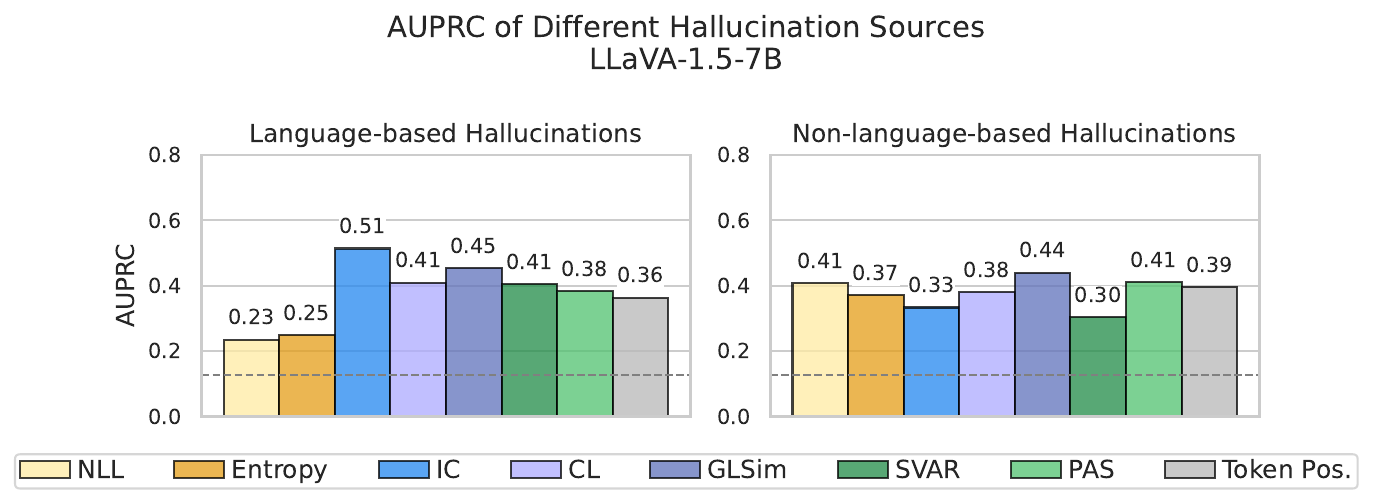}
    \includegraphics[width=0.49\linewidth, trim={0 0.1in 0 0.08in}, clip]{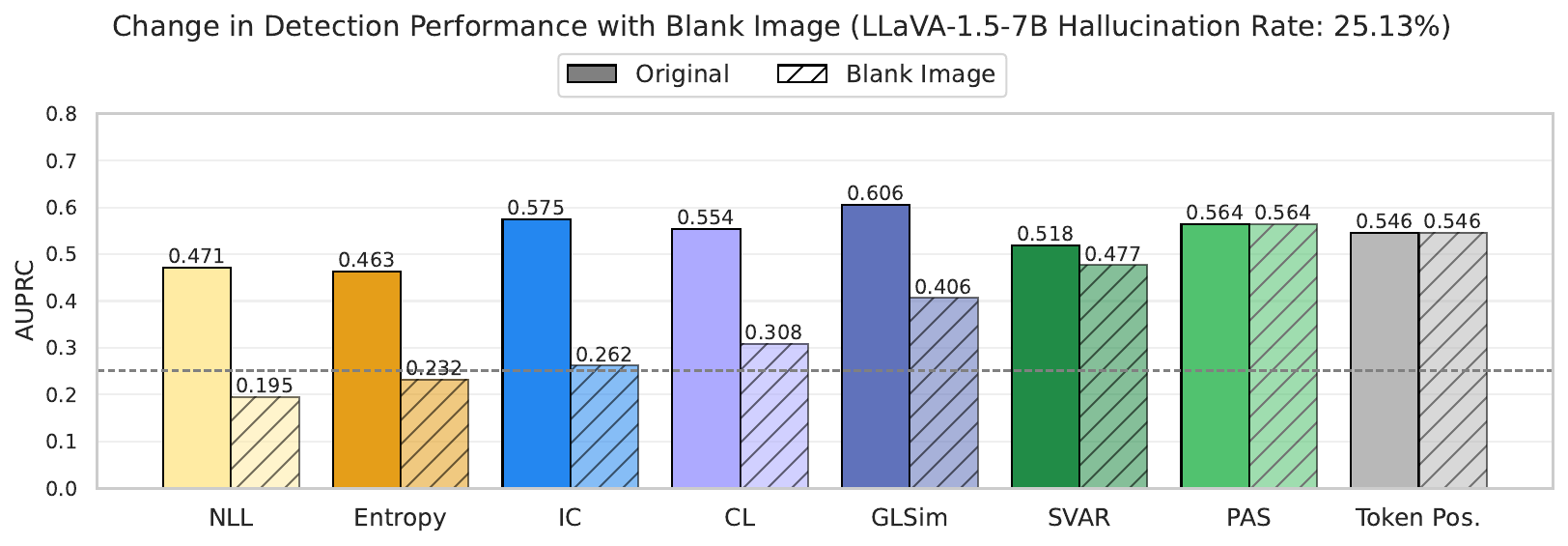}
    \caption{Method behavior partitioning by hallucination modality and visual groundedness on \llavasm \ generalizes across various data and prompt settings.}
    \label{fig:llava7b_perf_generalize2}
\end{figure}

\begin{figure}[h!]
    \centering
    \small MSCOCO  \par
    \includegraphics[width=0.49\linewidth, trim={0 0.1in 0 0.08in}, clip]{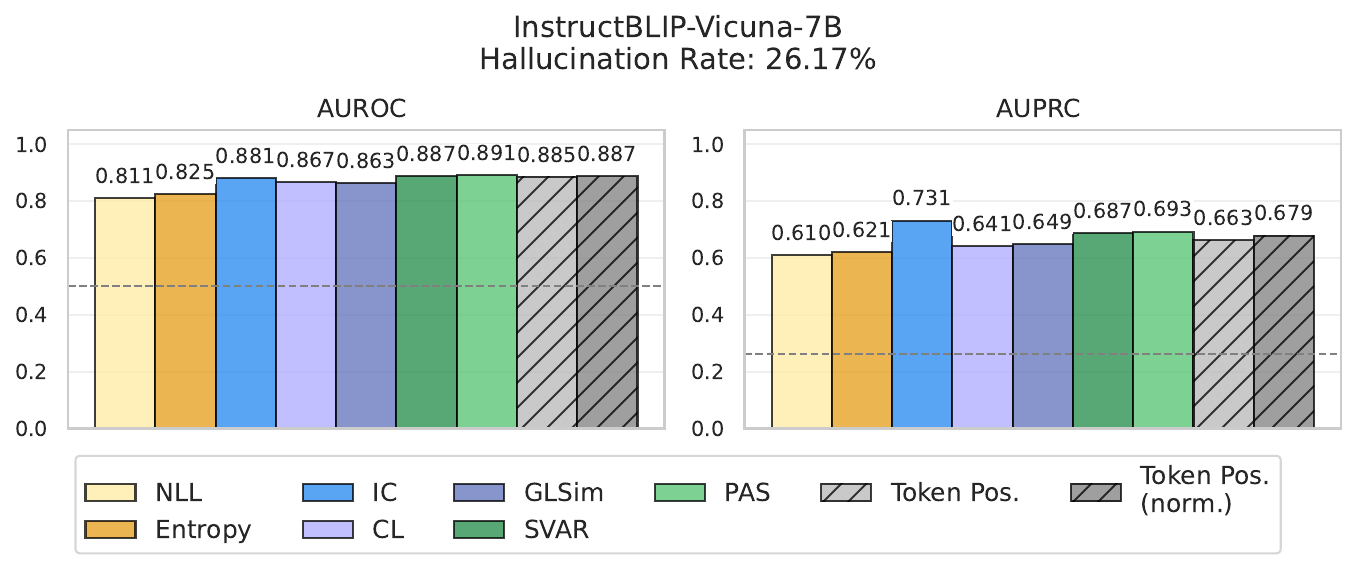}
    \includegraphics[width=0.49\linewidth, trim={0 0.1in 0 0.08in}, clip]{figures/instructblip-7b/remove_replace.pdf}
    \\
    \small Objects365 \par
    \includegraphics[width=0.49\linewidth, trim={0 0.1in 0 0.08in}, clip]{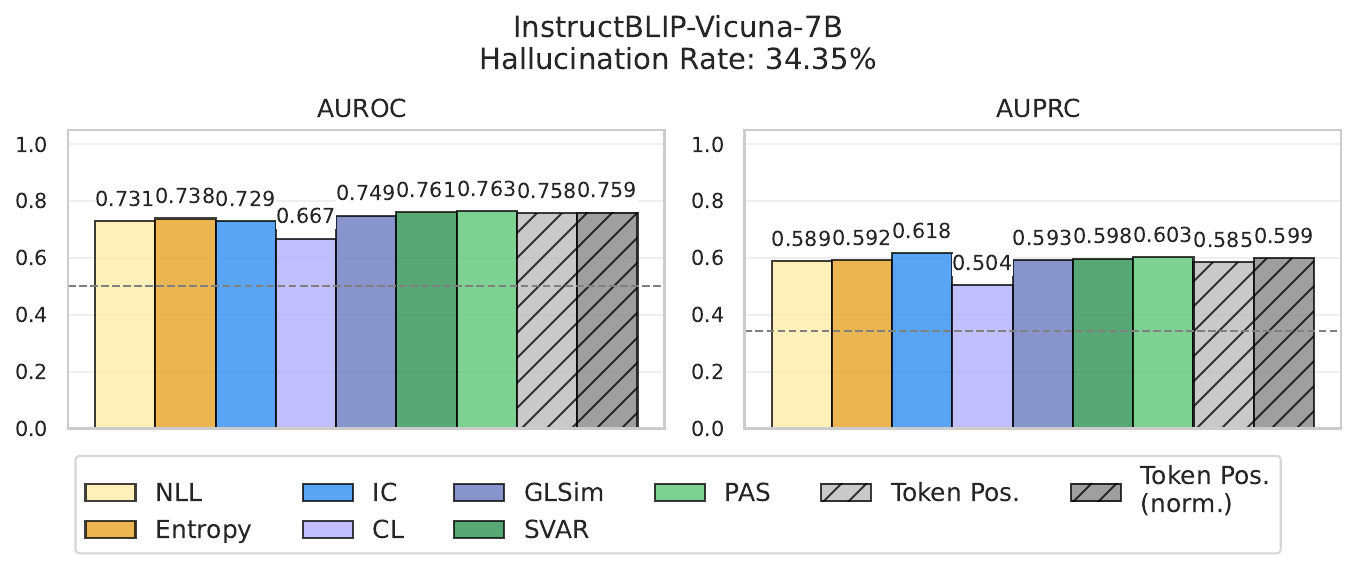}
    \includegraphics[width=0.49\linewidth, trim={0 0.1in 0 0.08in}, clip]{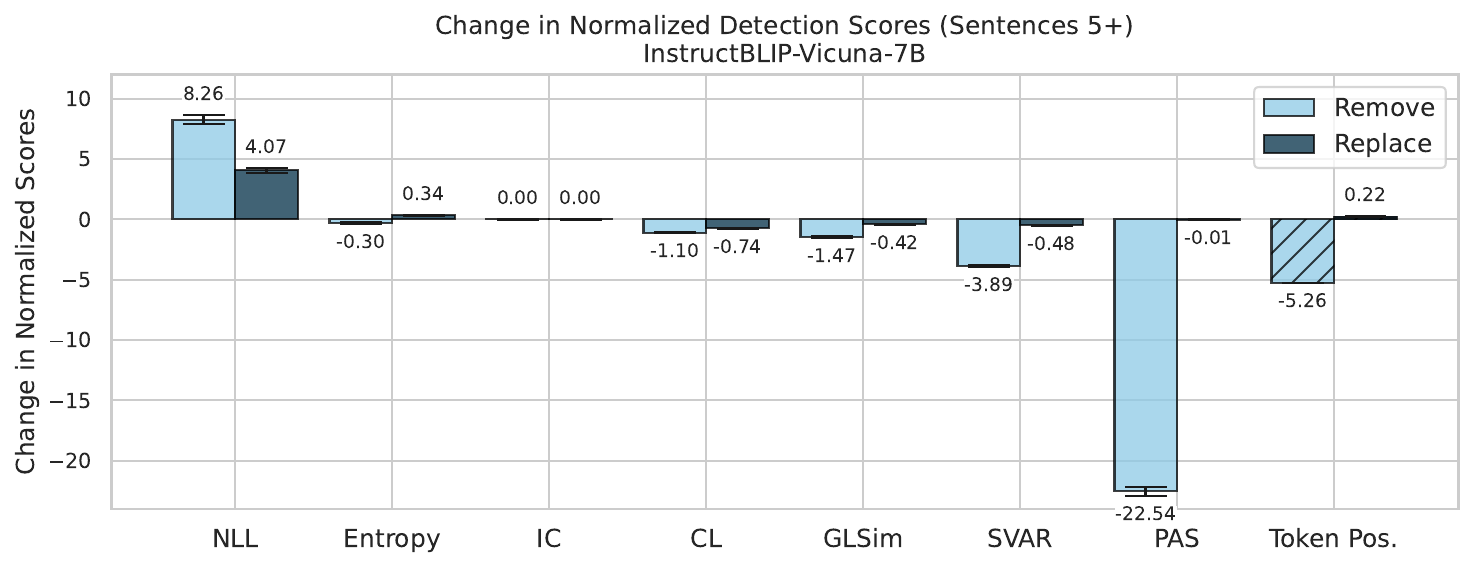}
    \\
    \small Prompt Variation \par
    \includegraphics[width=0.49\linewidth, trim={0 0.1in 0 0.08in}, clip]{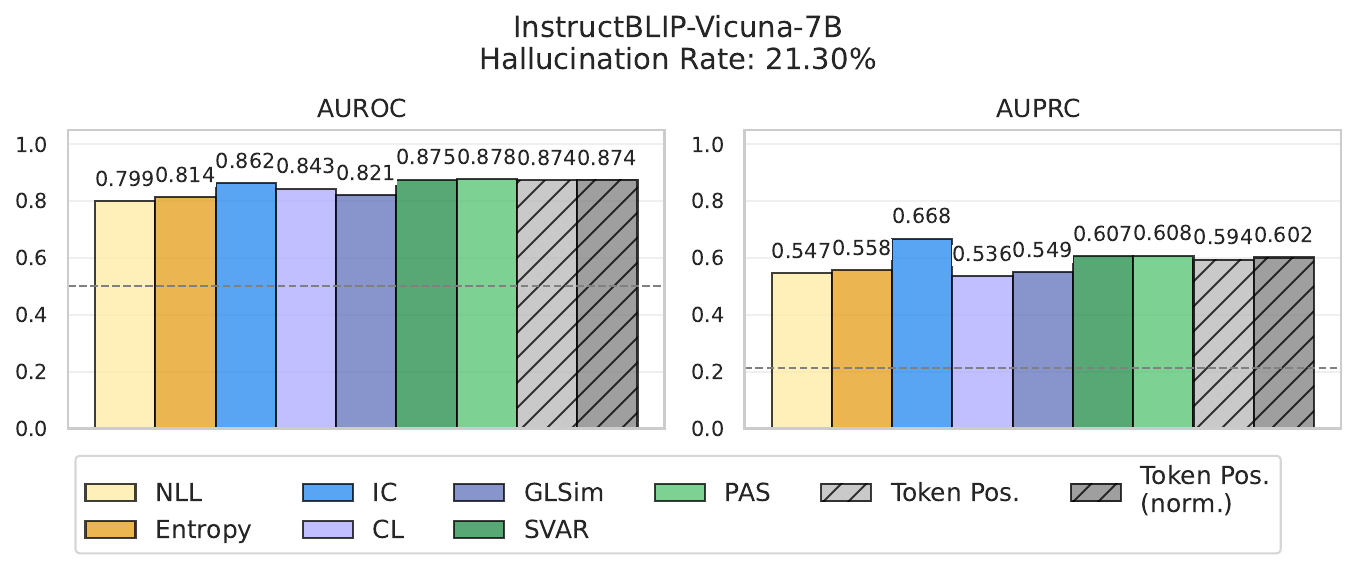}
    \includegraphics[width=0.49\linewidth, trim={0 0.1in 0 0.08in}, clip]{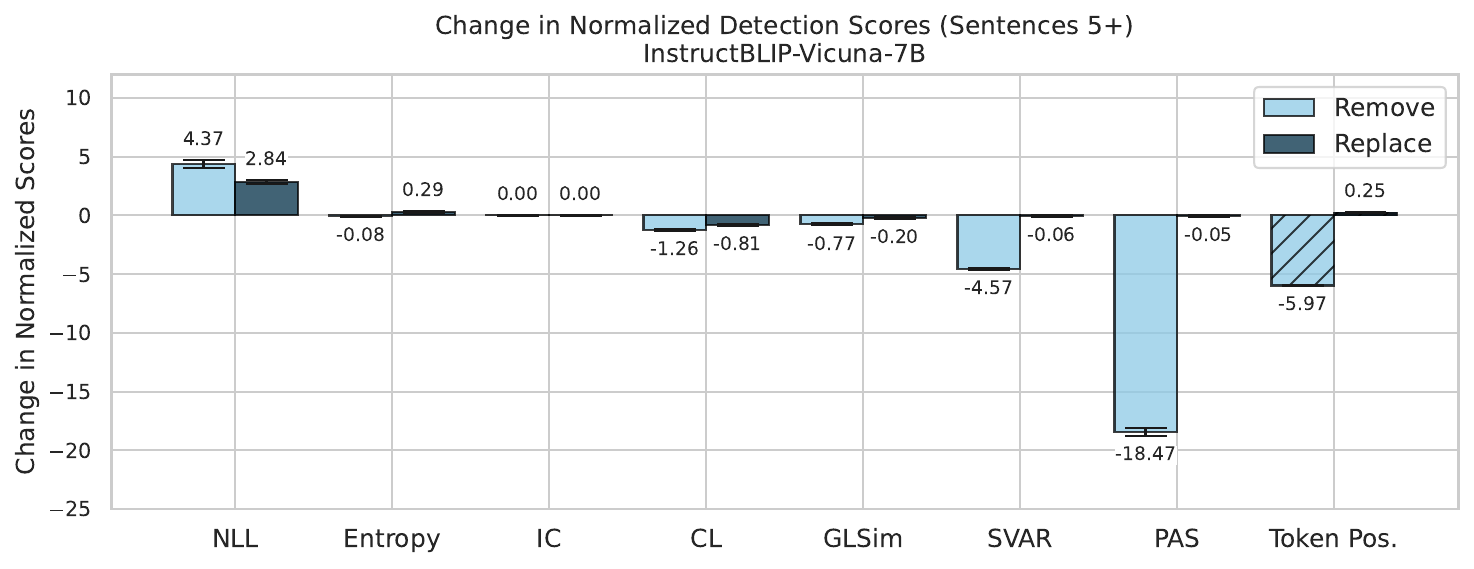}
    \\
    \small Repeated Trials \par
    \includegraphics[width=0.49\linewidth, trim={0 0.1in 0 0.08in}, clip]{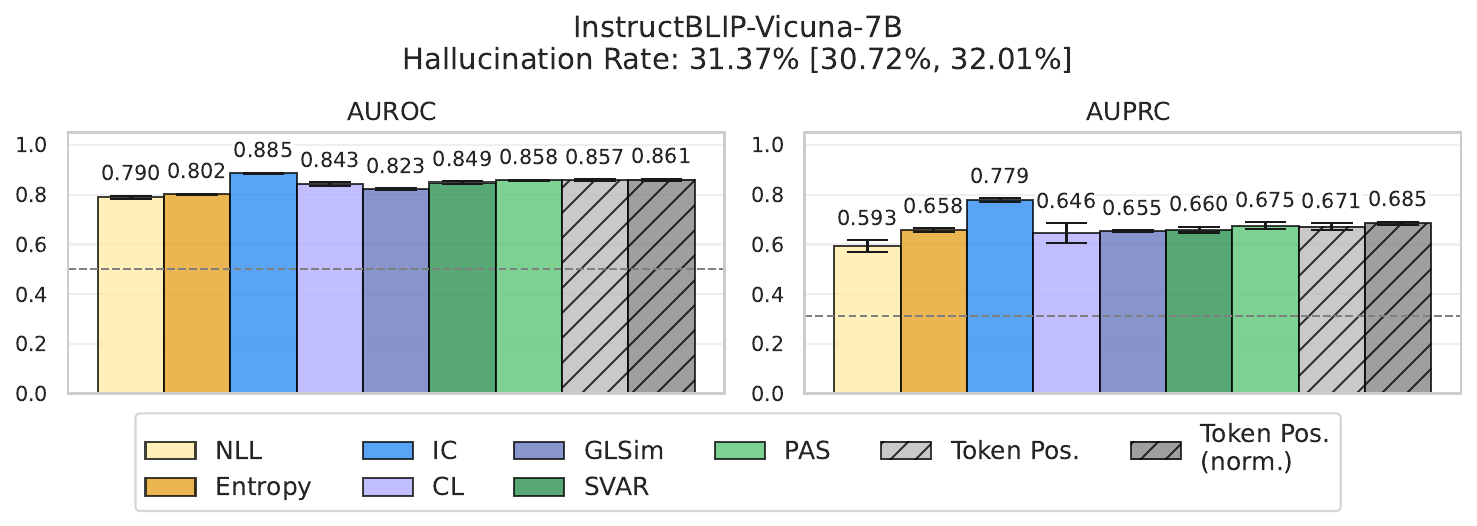}
    \includegraphics[width=0.49\linewidth, trim={0 0.1in 0 0.08in}, clip]{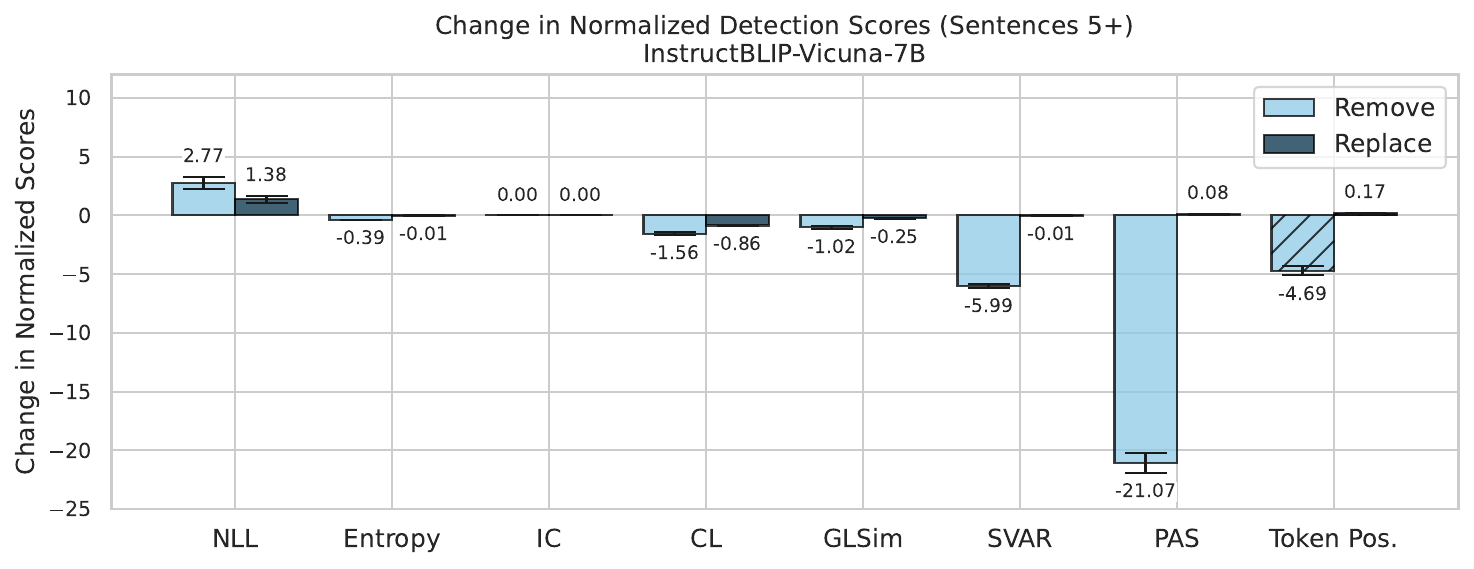}
    \\
    \small Context-aware Filtering \par
    \includegraphics[width=0.49\linewidth, trim={0 0.1in 0 0.08in}, clip]{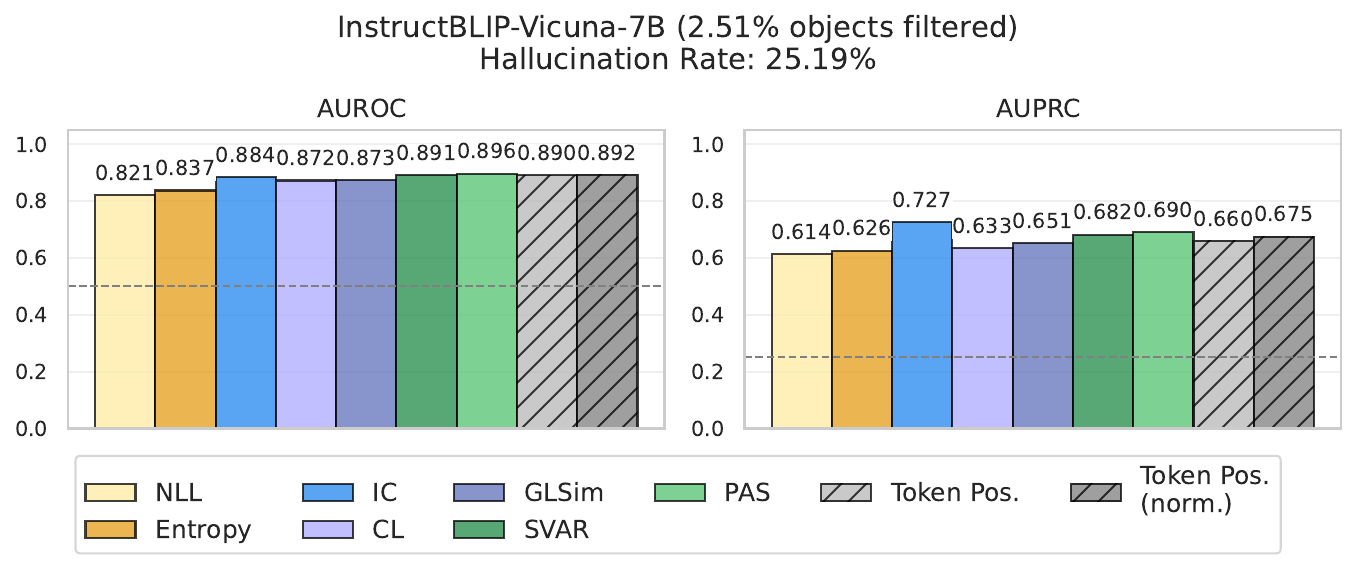}
    \includegraphics[width=0.49\linewidth, trim={0 0.1in 0 0.08in}, clip]{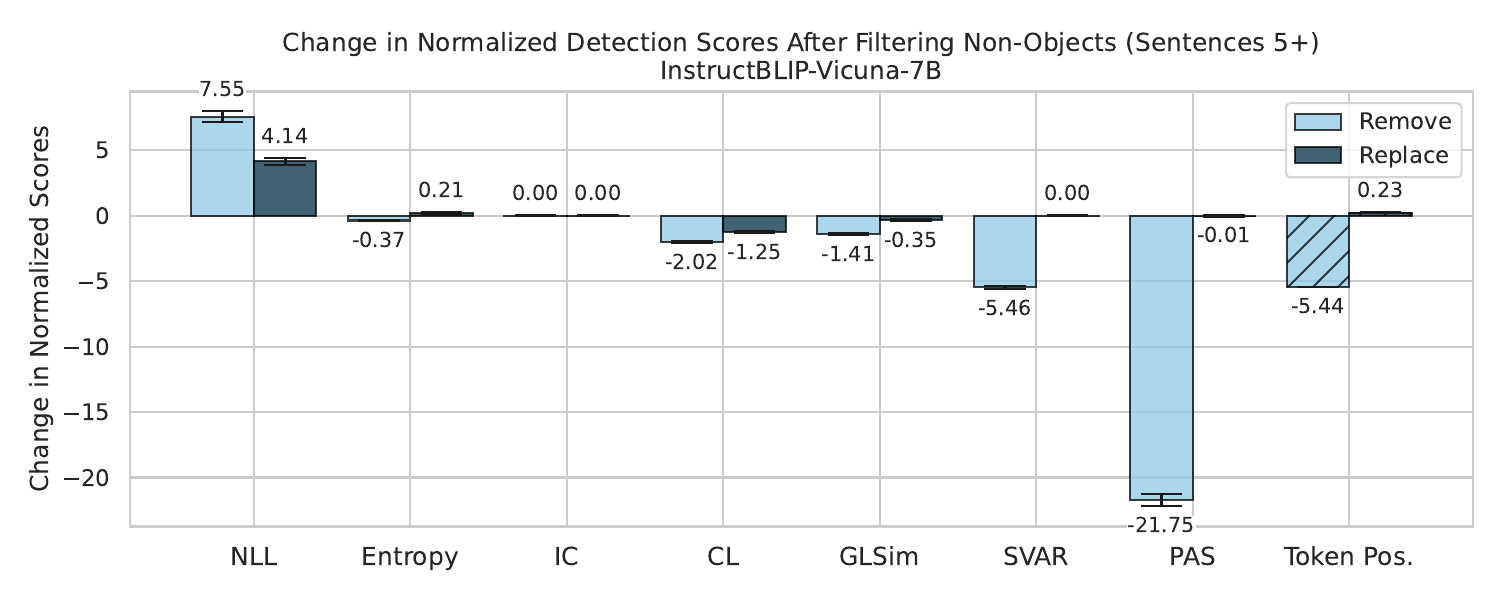}
    \\
    \small No Filtering ``orange''\par
    \includegraphics[width=0.49\linewidth, trim={0 0.1in 0 0.08in}, clip]{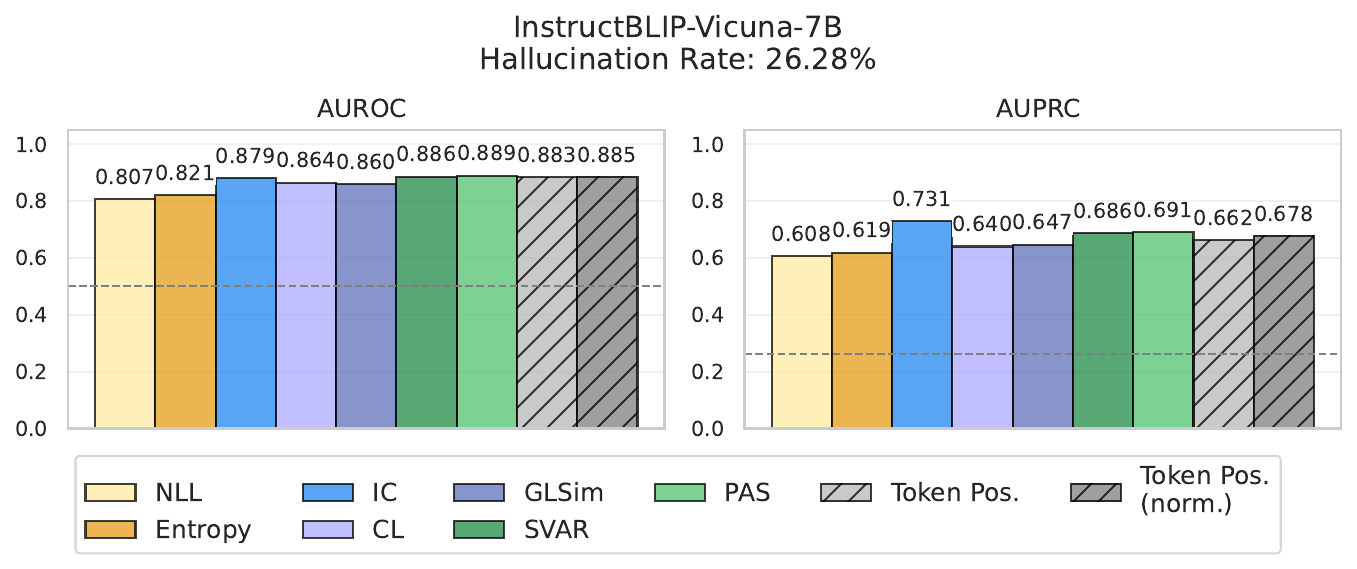}
    \includegraphics[width=0.49\linewidth, trim={0 0.1in 0 0.08in}, clip]{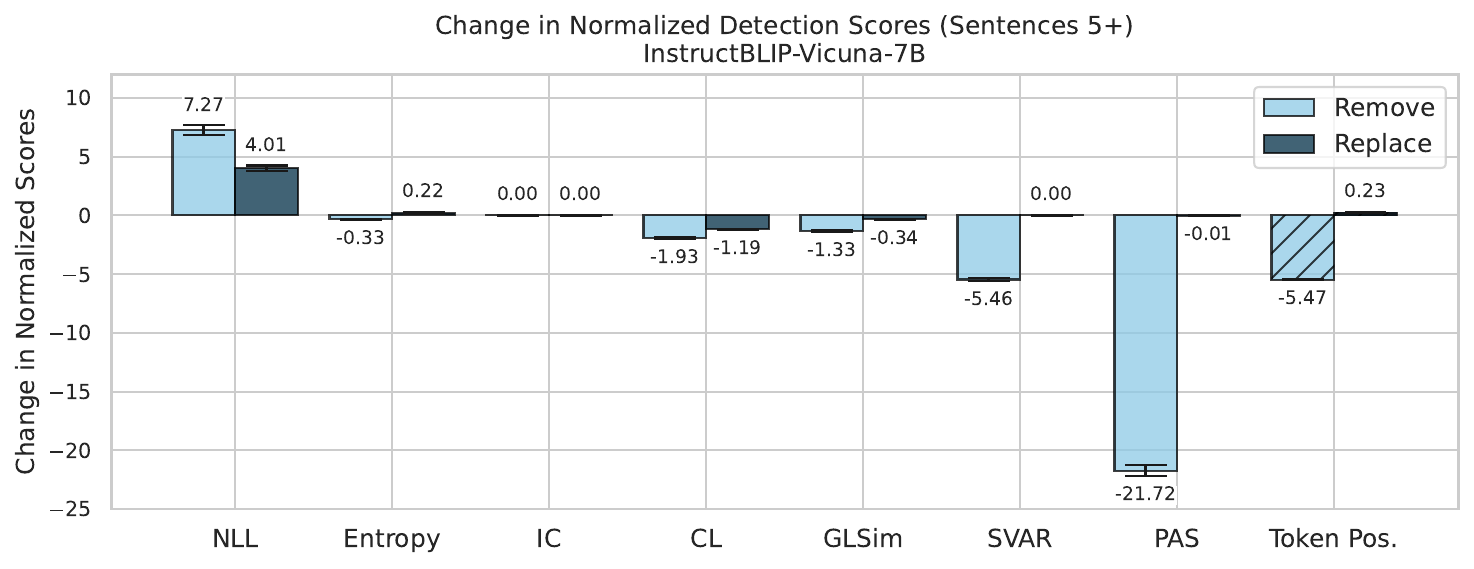}
    \caption{\textbf{Left:} Performance of hallucination detection methods on \instructblip \ generalizes across various data and prompt settings. \textbf{Right:} Method behavior with remove/replace preceding sentences \instructblip \ generalizes across various data and prompt settings.}
    \label{fig:instructblip_perf_generalize1}
\end{figure}

\begin{figure}[h!]
    \centering
    \small MSCOCO  \par
    \includegraphics[width=0.49\linewidth, trim={0 0.1in 0 0.08in}, clip]{figures/instructblip-7b/hall_source_auprc.pdf}
    \includegraphics[width=0.49\linewidth, trim={0 0.1in 0 0.08in}, clip]{figures/instructblip-7b/auprc_blank_performance_diff.pdf}
    \\
    \small Objects365 \par
    \includegraphics[width=0.49\linewidth, trim={0 0.1in 0 0.08in}, clip]{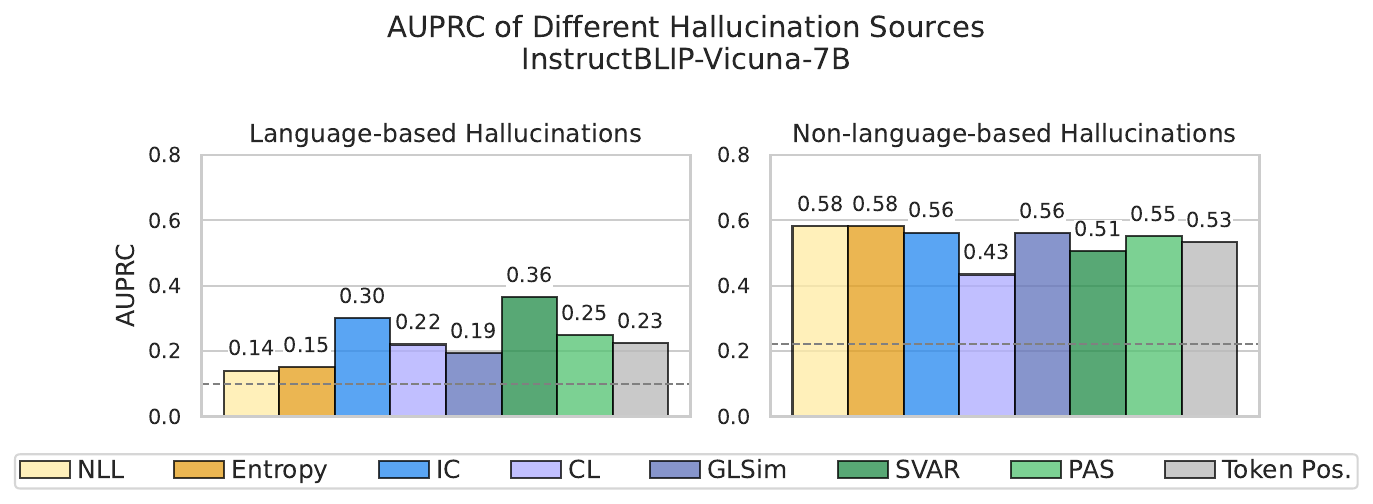}
    \includegraphics[width=0.49\linewidth, trim={0 0.1in 0 0.08in}, clip]{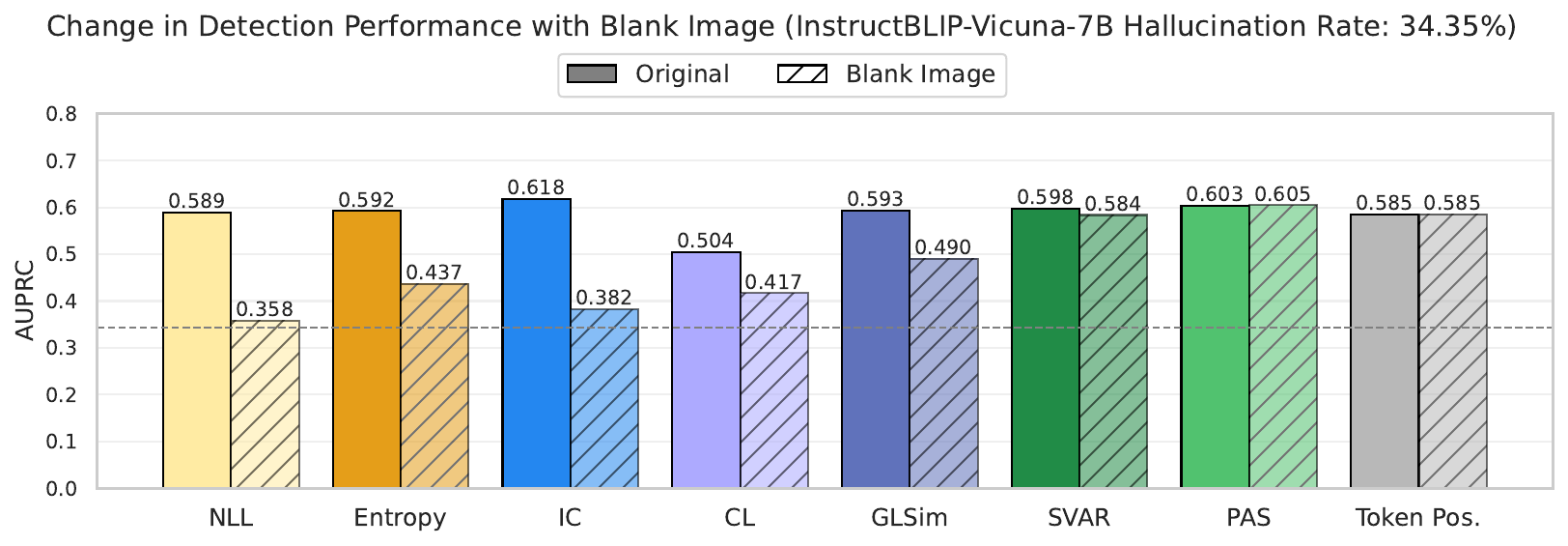}
    \\
    \small Prompt Variation \par
    \includegraphics[width=0.49\linewidth, trim={0 0.1in 0 0.08in}, clip]{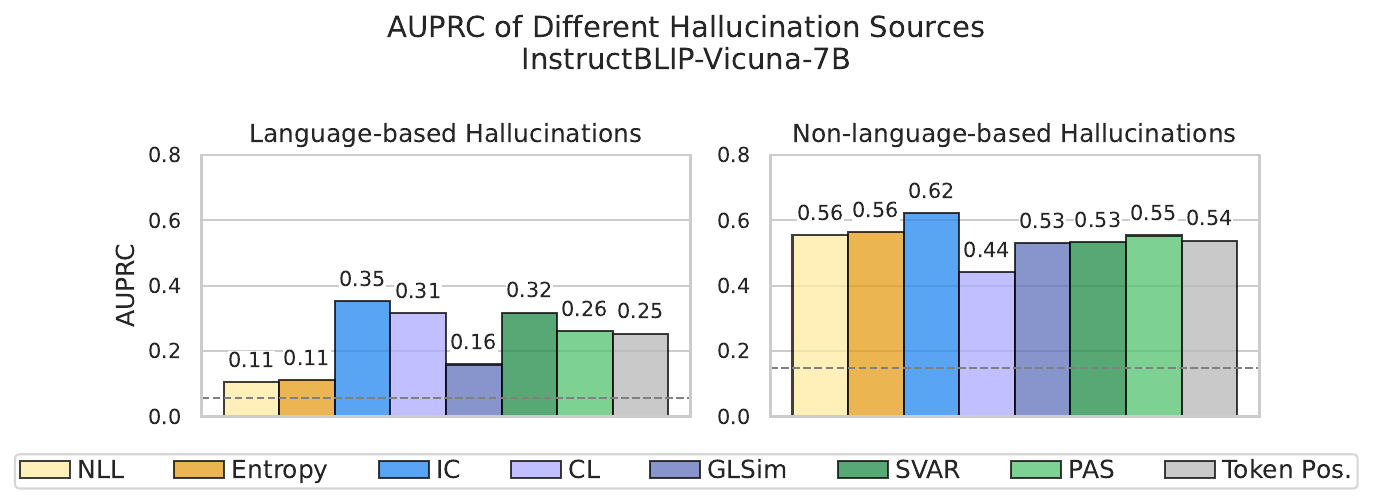}
    \includegraphics[width=0.49\linewidth, trim={0 0.1in 0 0.08in}, clip]{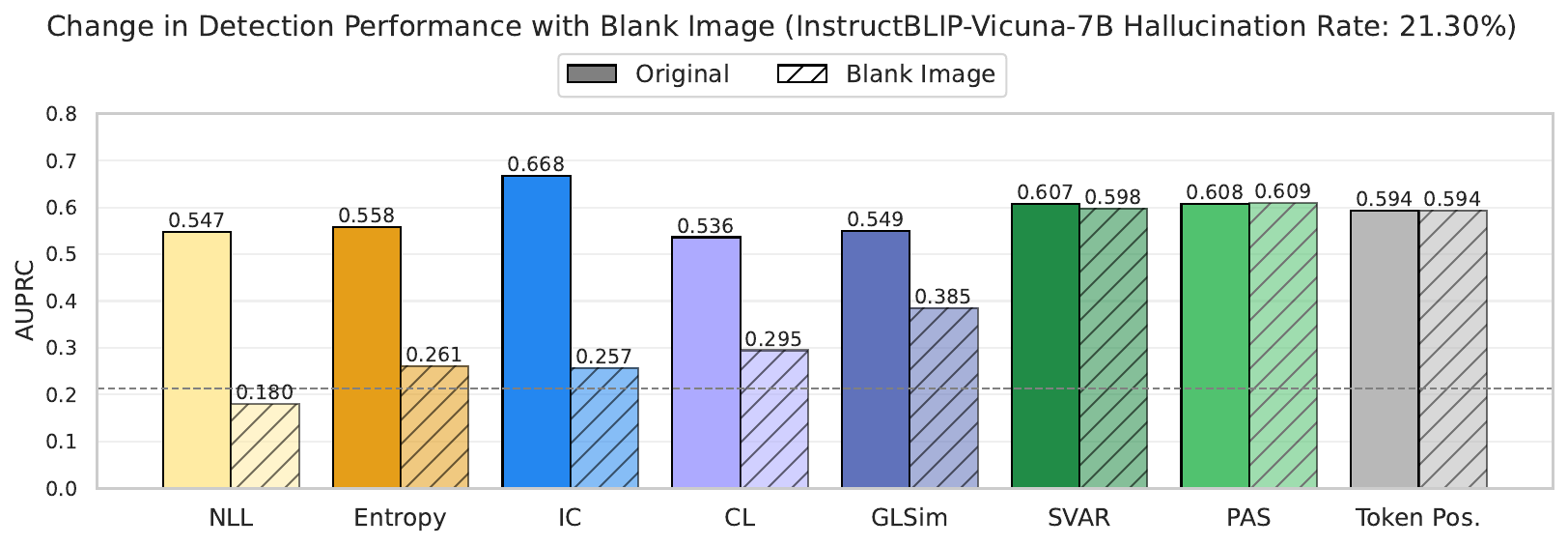}
    \\
    \small Repeated Trials \par
    \includegraphics[width=0.49\linewidth]{figures/coco_sampling/unavailable_figure.pdf}
    \includegraphics[width=0.49\linewidth, trim={0 0.1in 0 0.08in}, clip]{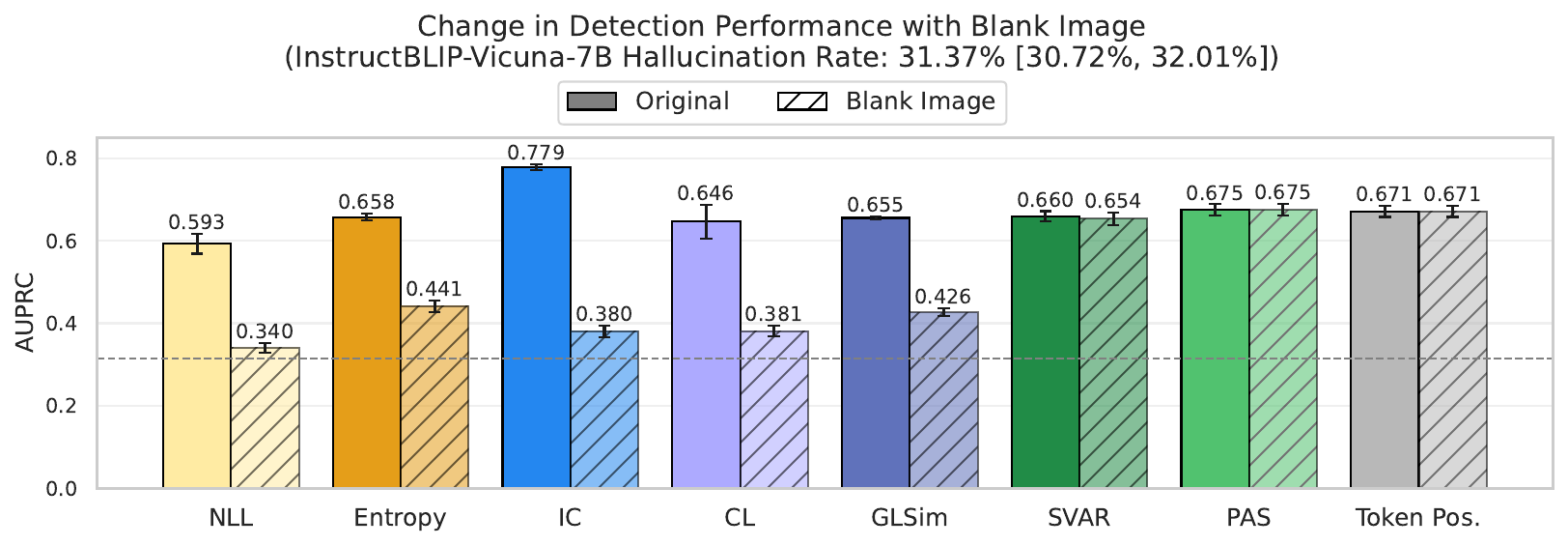}
    \\
    \small Context-aware Filtering \par
    \includegraphics[width=0.49\linewidth, trim={0 0.1in 0 0.08in}, clip]{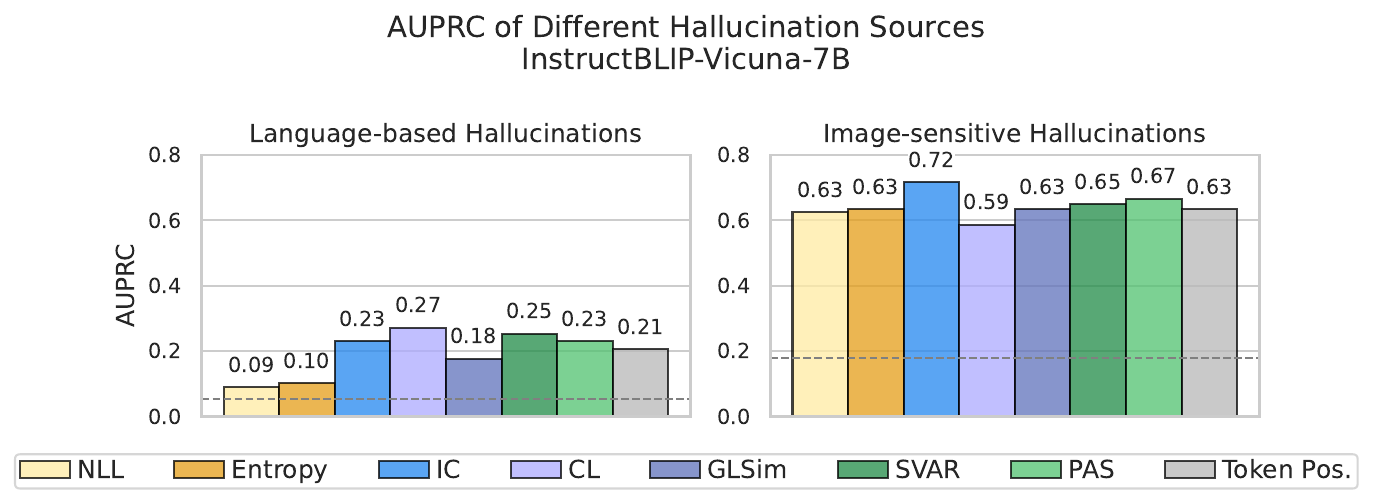}
    \includegraphics[width=0.49\linewidth, trim={0 0.1in 0 0.08in}, clip]{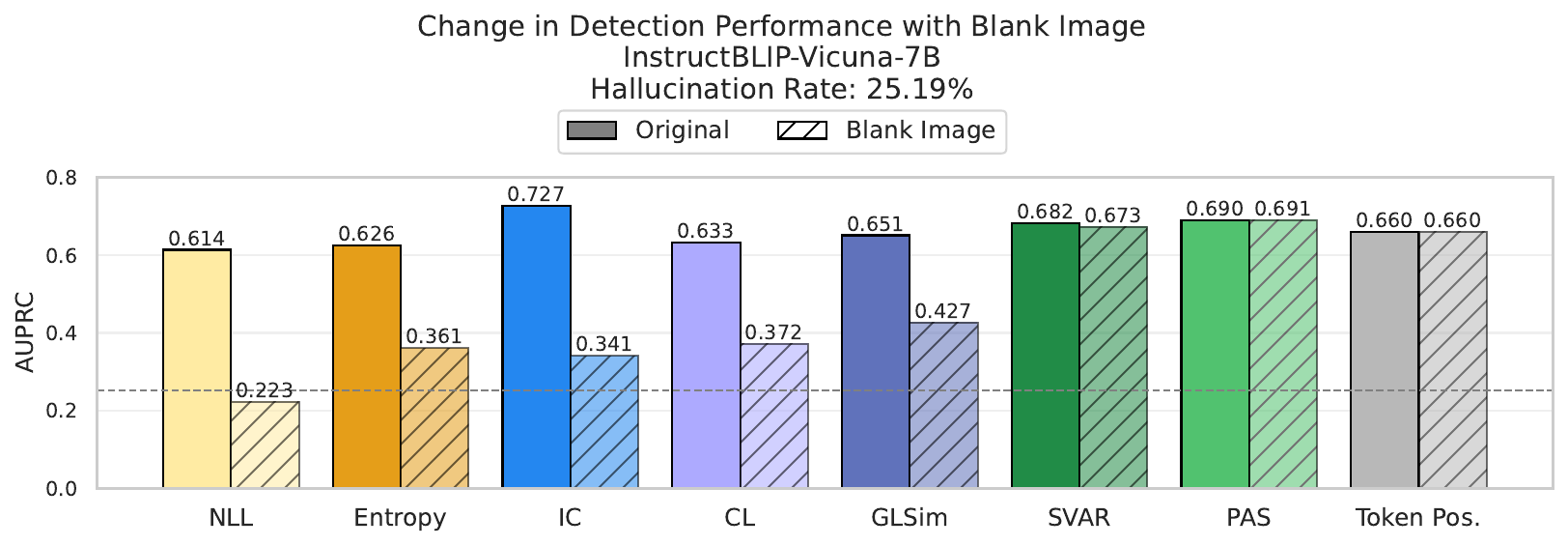}
    \\
    \small No Filtering ``orange''\par
    \includegraphics[width=0.49\linewidth, trim={0 0.1in 0 0.08in}, clip]{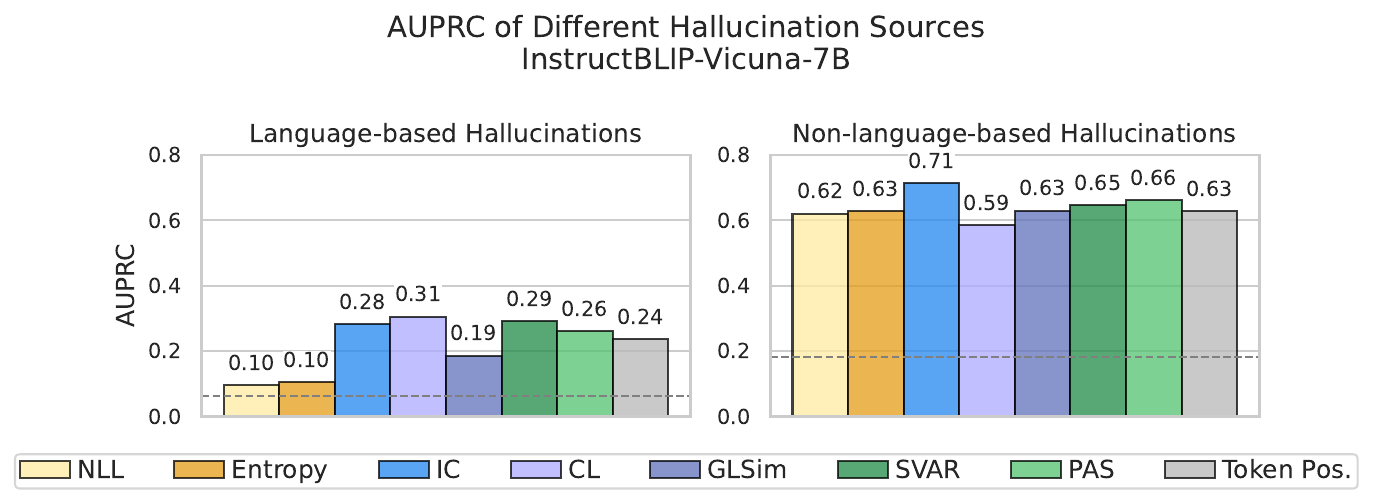}
    \includegraphics[width=0.49\linewidth, trim={0 0.1in 0 0.08in}, clip]{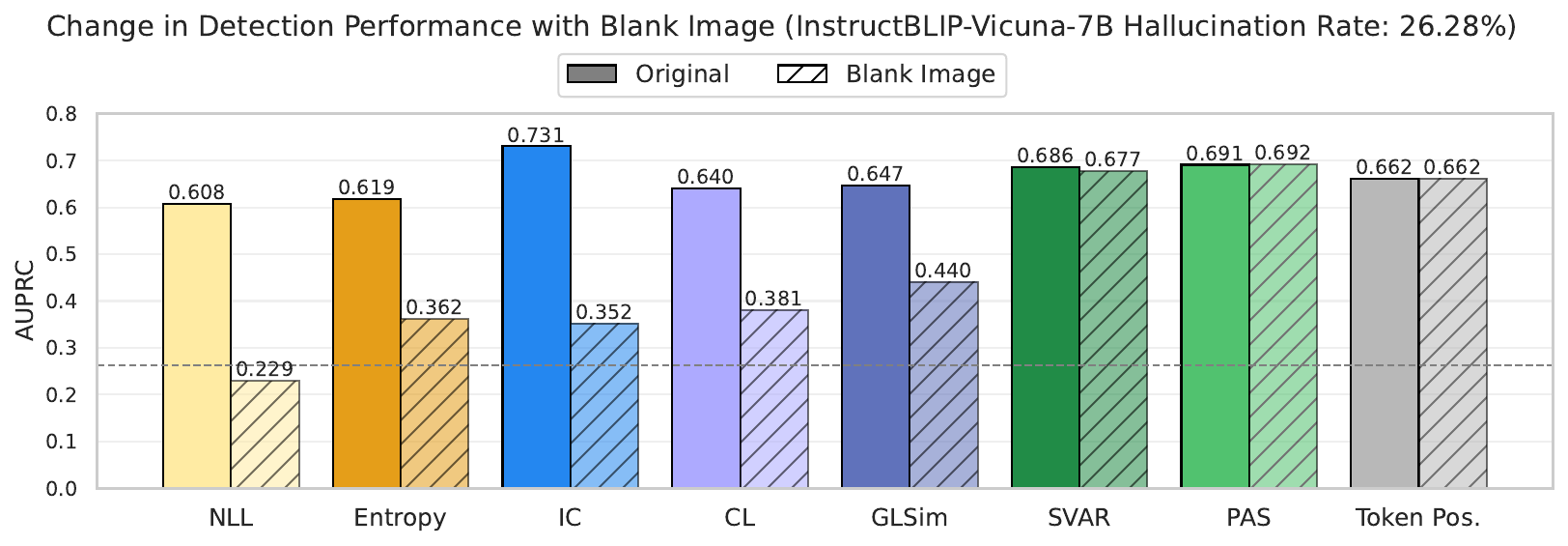}
    \caption{Method behavior partitioning by hallucination modality and visual groundedness on \instructblip \ generalizes across various data and prompt settings.}
    \label{fig:instructblip7b_perf_generalize2}
\end{figure}

\begin{figure}[h!]
    \centering
    \small MSCOCO  \par
    \includegraphics[width=0.49\linewidth, trim={0 0.1in 0 0.08in}, clip]{figures/qwen3vl-8b/compare_auroc_auprc_legend.pdf}
    \includegraphics[width=0.49\linewidth, trim={0 0.1in 0 0.08in}, clip]{figures/qwen3vl-8b/remove_replace.pdf}
    \\
    \small Objects365 \par
    \includegraphics[width=0.49\linewidth, trim={0 0.1in 0 0.08in}, clip]{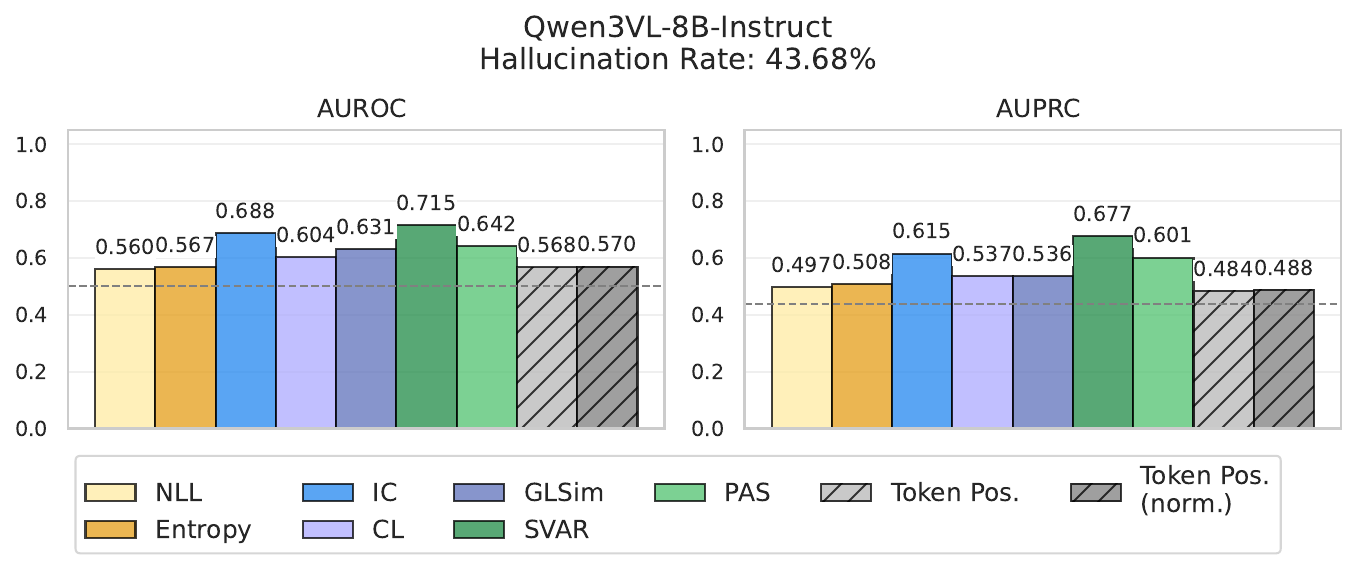}
    \includegraphics[width=0.49\linewidth, trim={0 0.1in 0 0.08in}, clip]{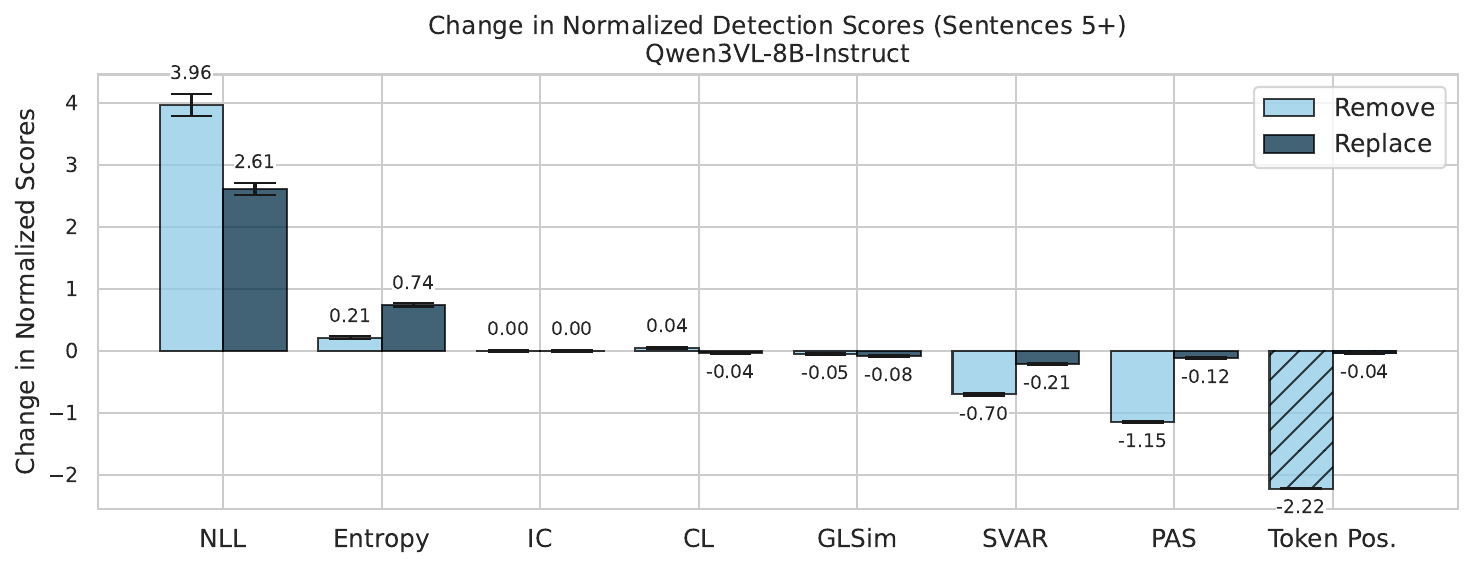}
    \\
    \small Prompt Variation \par
    \includegraphics[width=0.49\linewidth, trim={0 0.1in 0 0.08in}, clip]{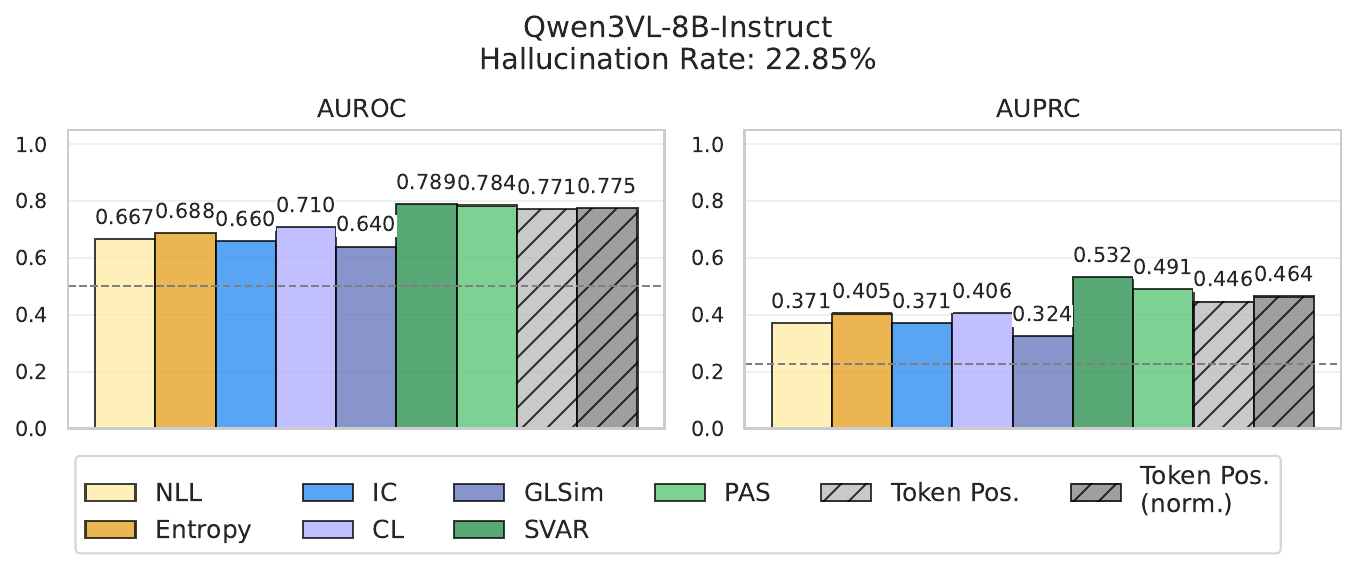}
    \includegraphics[width=0.49\linewidth, trim={0 0.1in 0 0.08in}, clip]{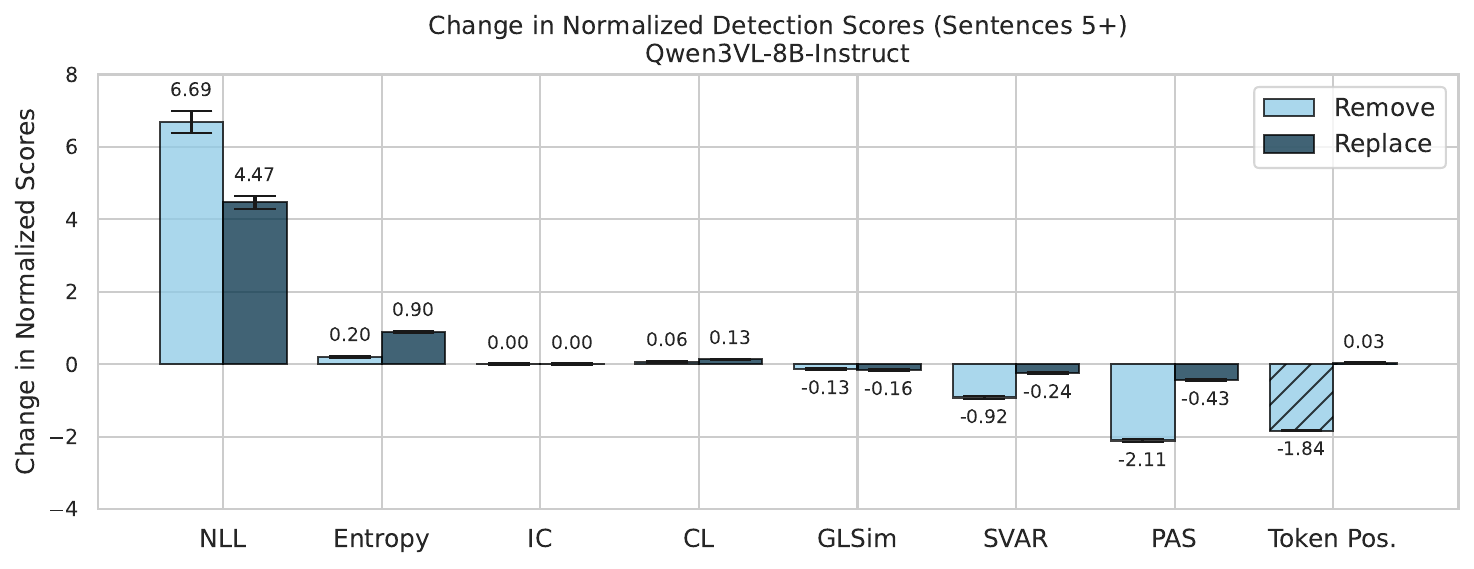}
    \\
    \small Repeated Trials \par
    \includegraphics[width=0.49\linewidth, trim={0 0.1in 0 0.08in}, clip]{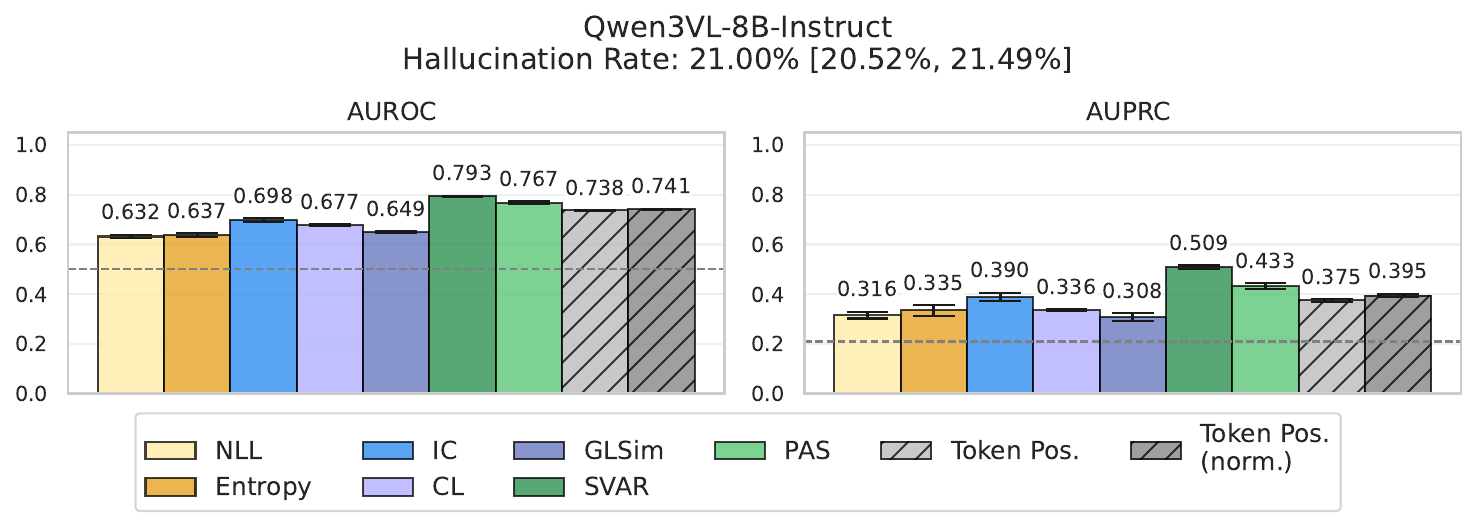}
    \includegraphics[width=0.49\linewidth, trim={0 0.1in 0 0.08in}, clip]{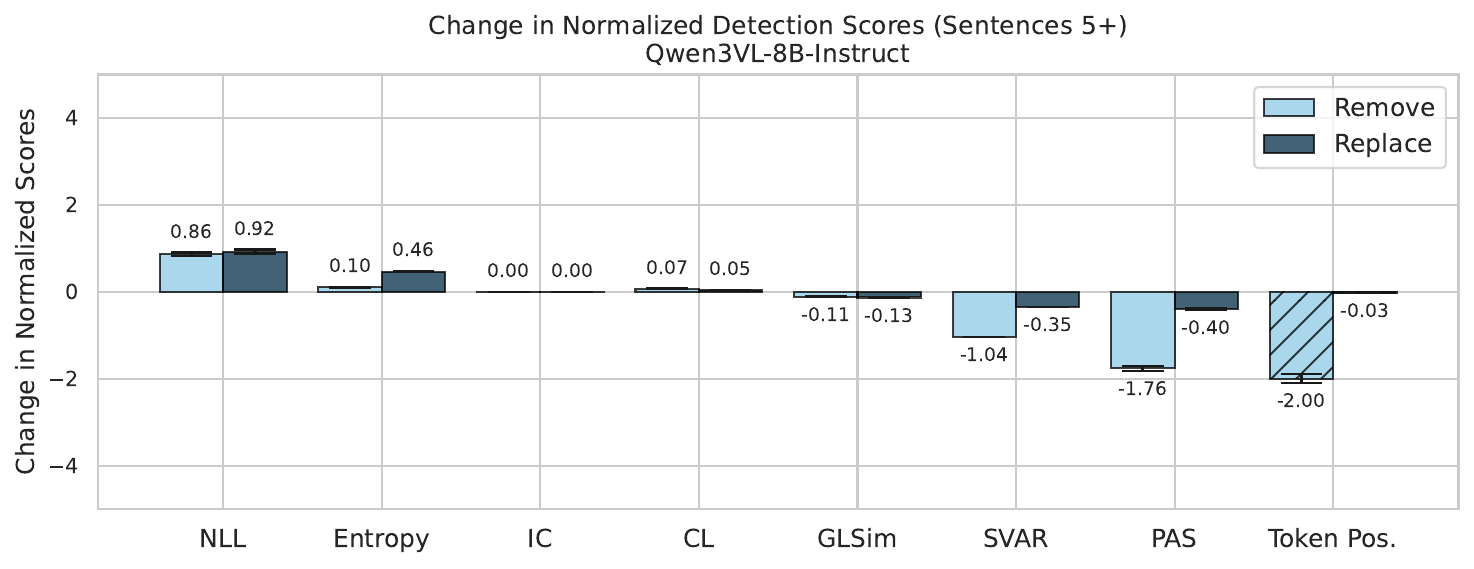}
    \\
    \small Context-aware Filtering \par
    \includegraphics[width=0.49\linewidth, trim={0 0.1in 0 0.08in}, clip]{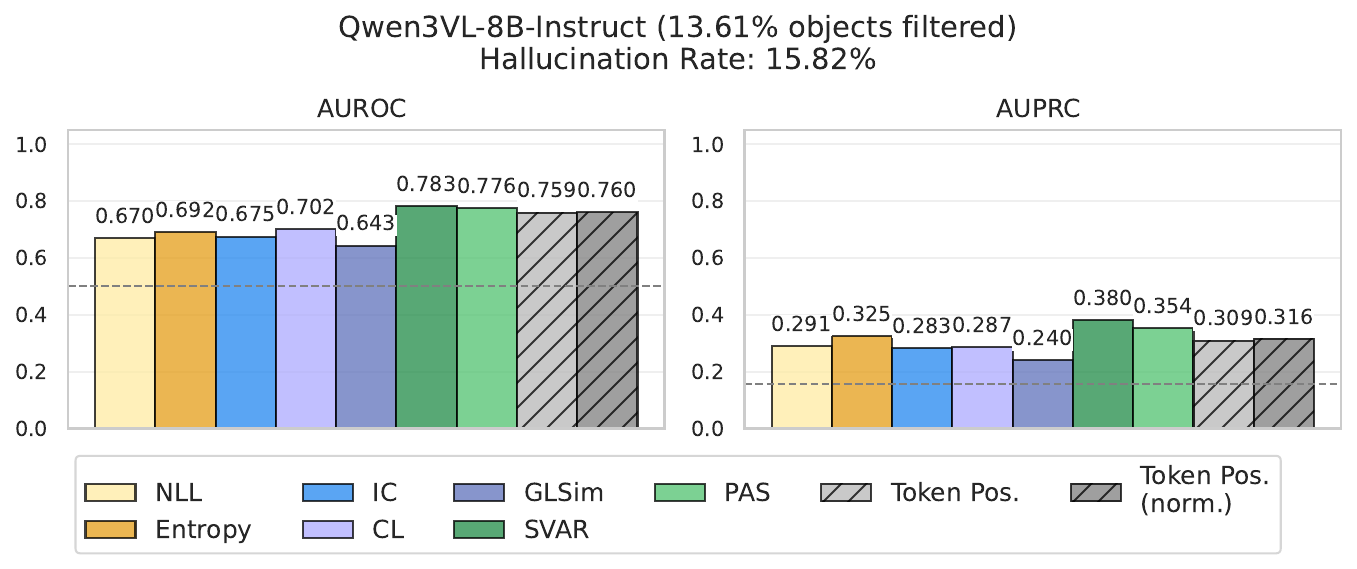}
    \includegraphics[width=0.49\linewidth, trim={0 0.1in 0 0.08in}, clip]{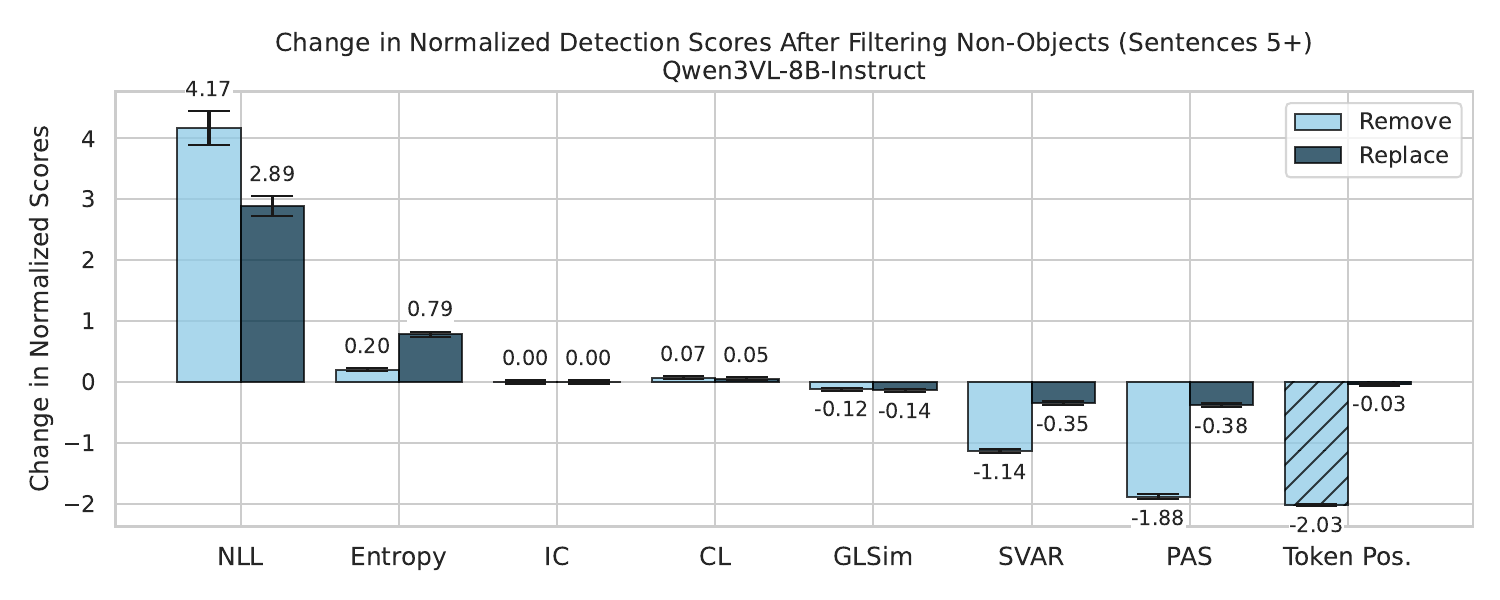}
    \\
    \small No Filtering ``orange''\par
    \includegraphics[width=0.49\linewidth, trim={0 0.1in 0 0.08in}, clip]{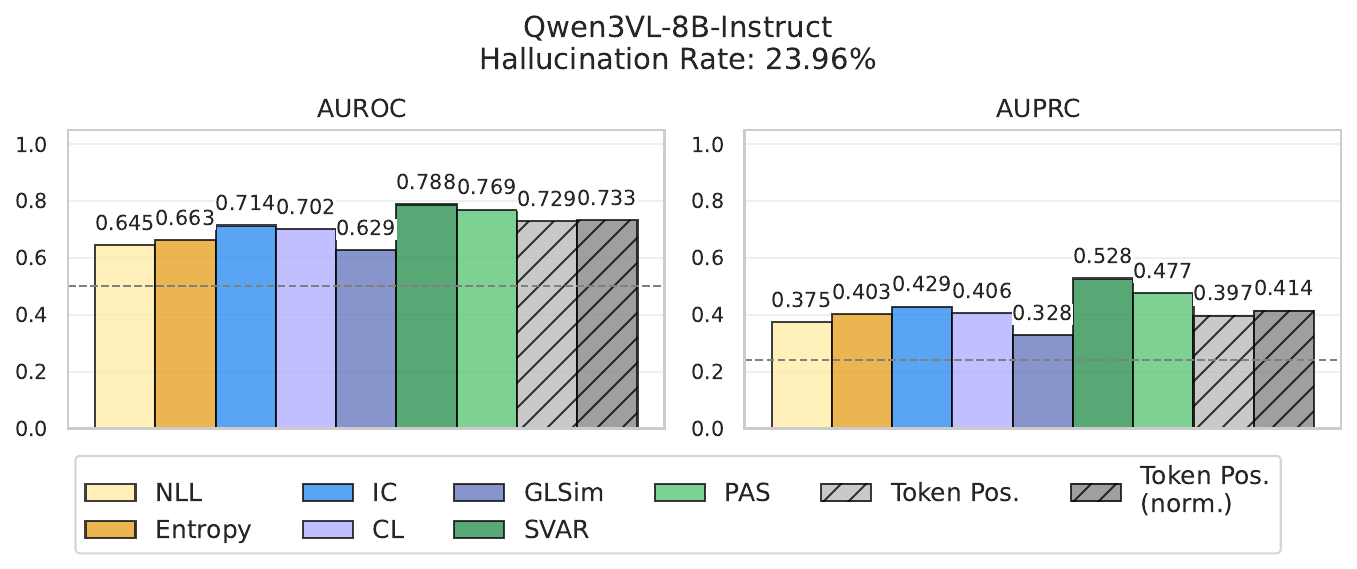}
    \includegraphics[width=0.49\linewidth, trim={0 0.1in 0 0.08in}, clip]{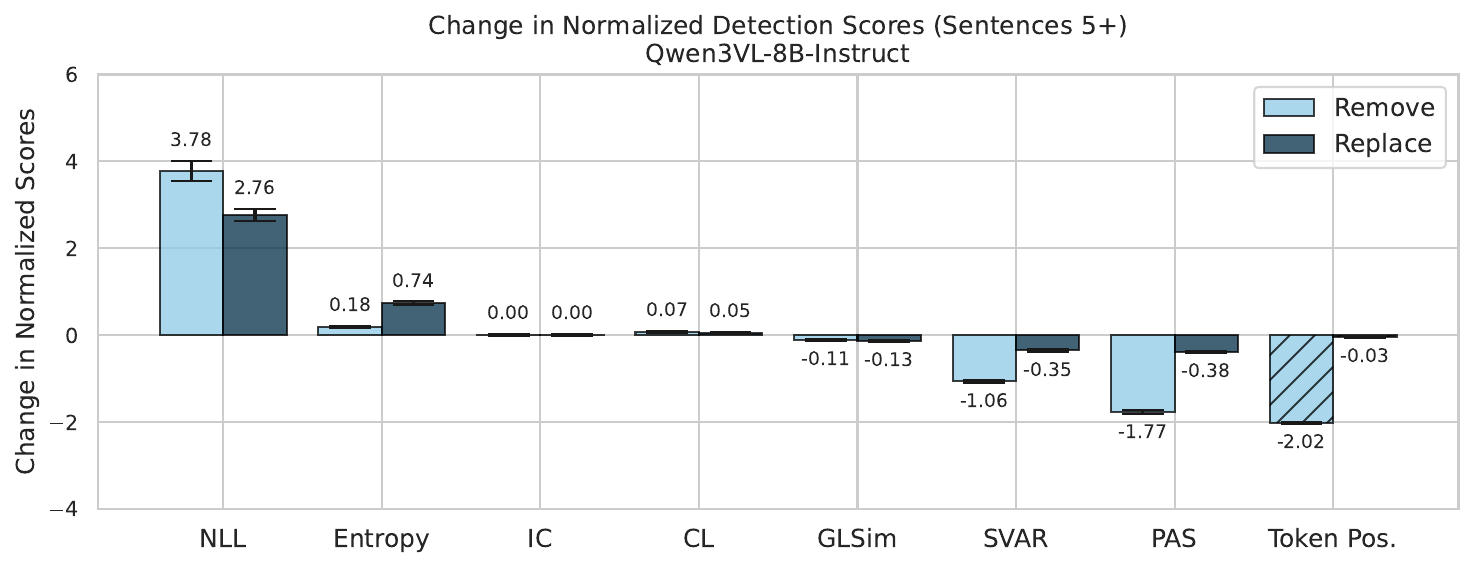}
    \caption{\textbf{Left:} Performance of hallucination detection methods on \qwen \ generalizes across various data and prompt settings. \textbf{Right:} Method behavior with remove/replace preceding sentences \qwen \ generalizes across various data and prompt settings.}
    \label{fig:qwen3vl8bperf_generalize1}
\end{figure}

\begin{figure}[h!]
    \centering
    \small MSCOCO  \par
    \includegraphics[width=0.49\linewidth, trim={0 0.1in 0 0.08in}, clip]{figures/qwen3vl-8b/hall_source_auprc.pdf}
    \includegraphics[width=0.49\linewidth, trim={0 0.1in 0 0.08in}, clip]{figures/qwen3vl-8b/auprc_blank_performance_diff.pdf}
    \\
    \small Objects365 \par
    \includegraphics[width=0.49\linewidth, trim={0 0.1in 0 0.08in}, clip]{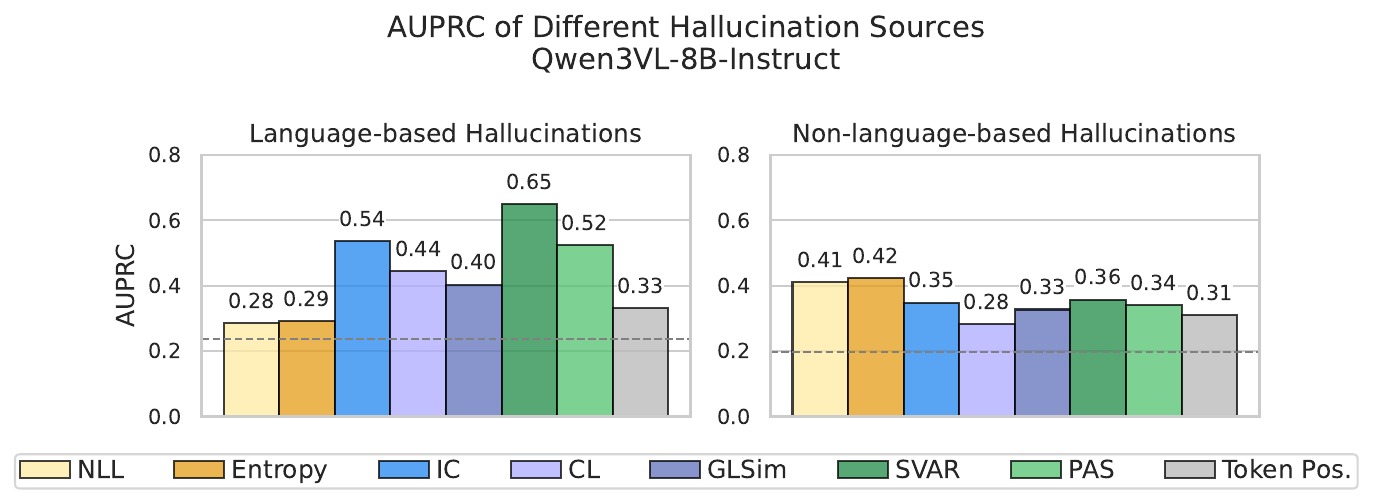}
    \includegraphics[width=0.49\linewidth, trim={0 0.1in 0 0.08in}, clip]{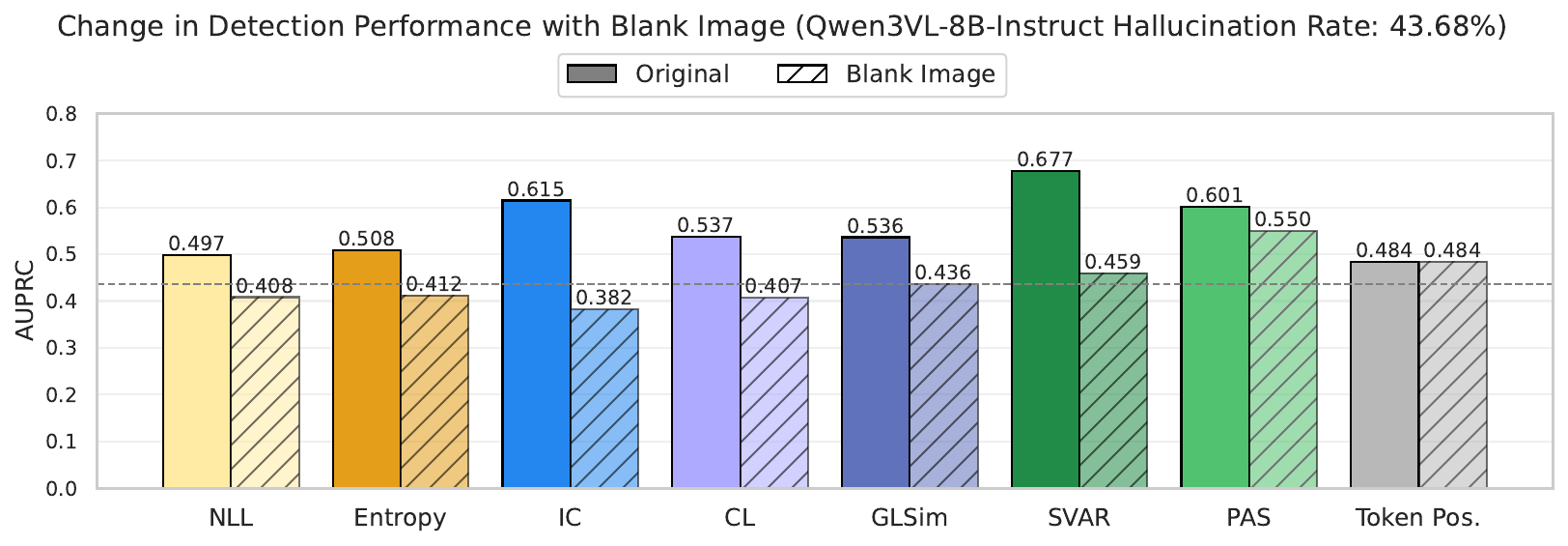}
    \\
    \small Prompt Variation \par
    \includegraphics[width=0.49\linewidth, trim={0 0.1in 0 0.08in}, clip]{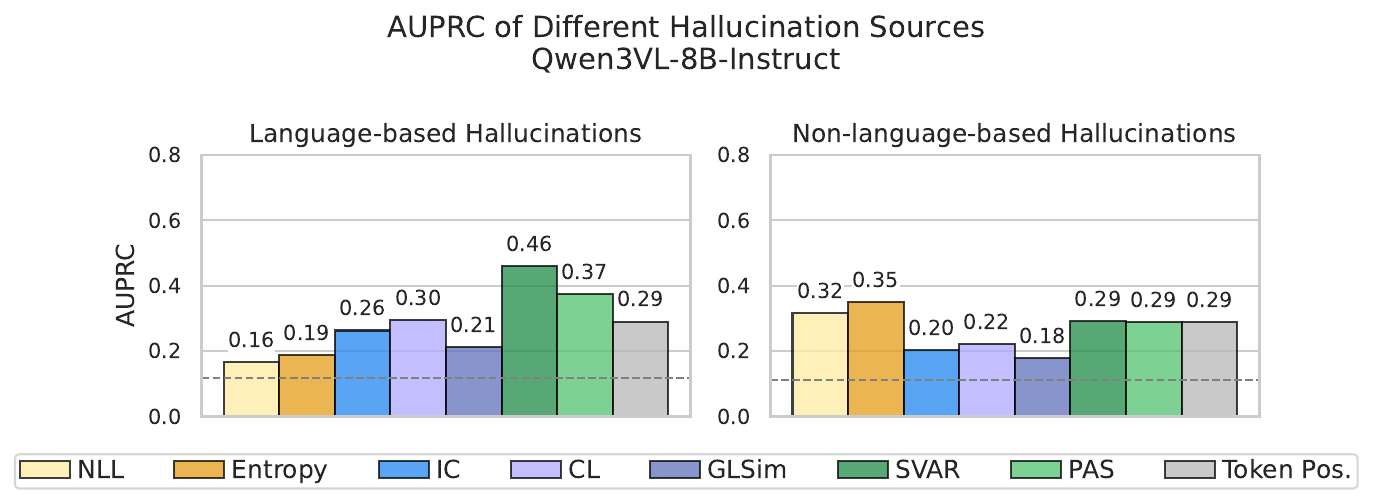}
    \includegraphics[width=0.49\linewidth, trim={0 0.1in 0 0.08in}, clip]{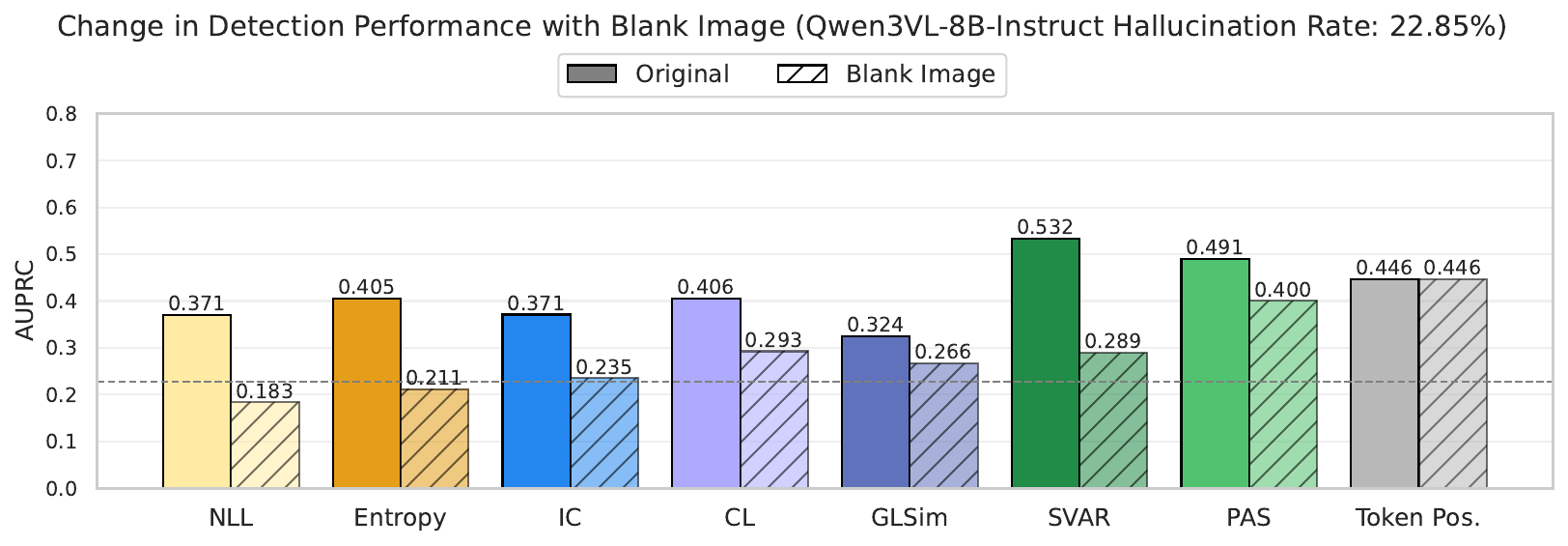}
    \\
    \small Repeated Trials \par
    \includegraphics[width=0.49\linewidth]{figures/coco_sampling/unavailable_figure.pdf}    \includegraphics[width=0.49\linewidth, trim={0 0.1in 0 0.08in}, clip]{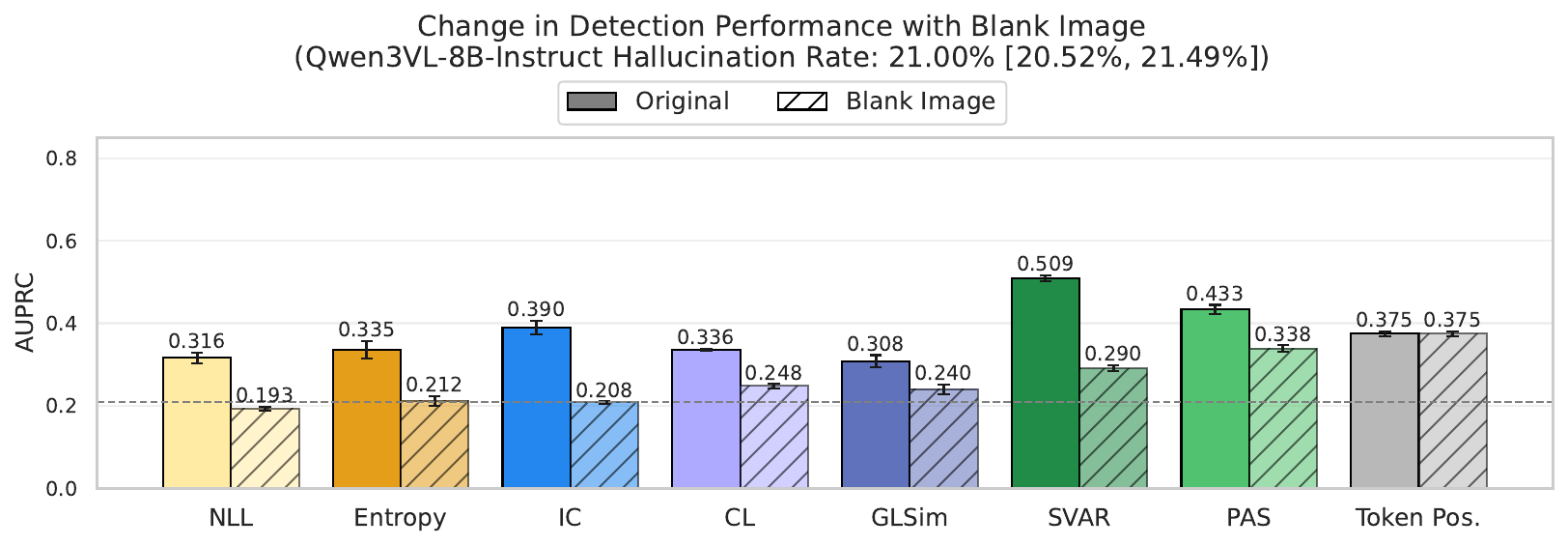}
    \\
    \small Context-aware Filtering \par
    \includegraphics[width=0.49\linewidth, trim={0 0.1in 0 0.08in}, clip]{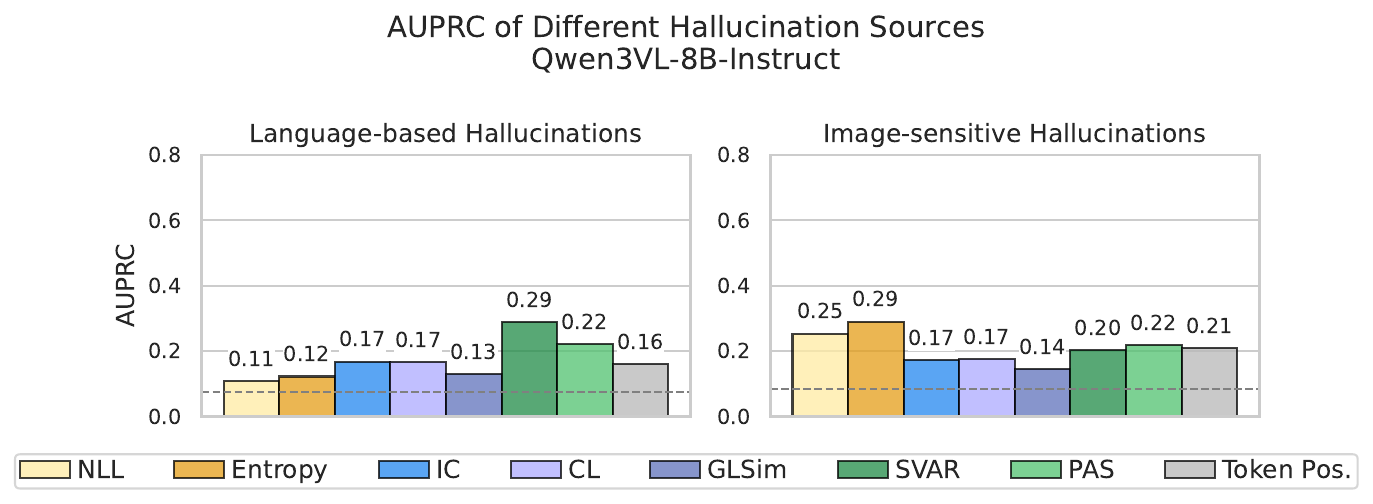}
    \includegraphics[width=0.49\linewidth, trim={0 0.1in 0 0.08in}, clip]{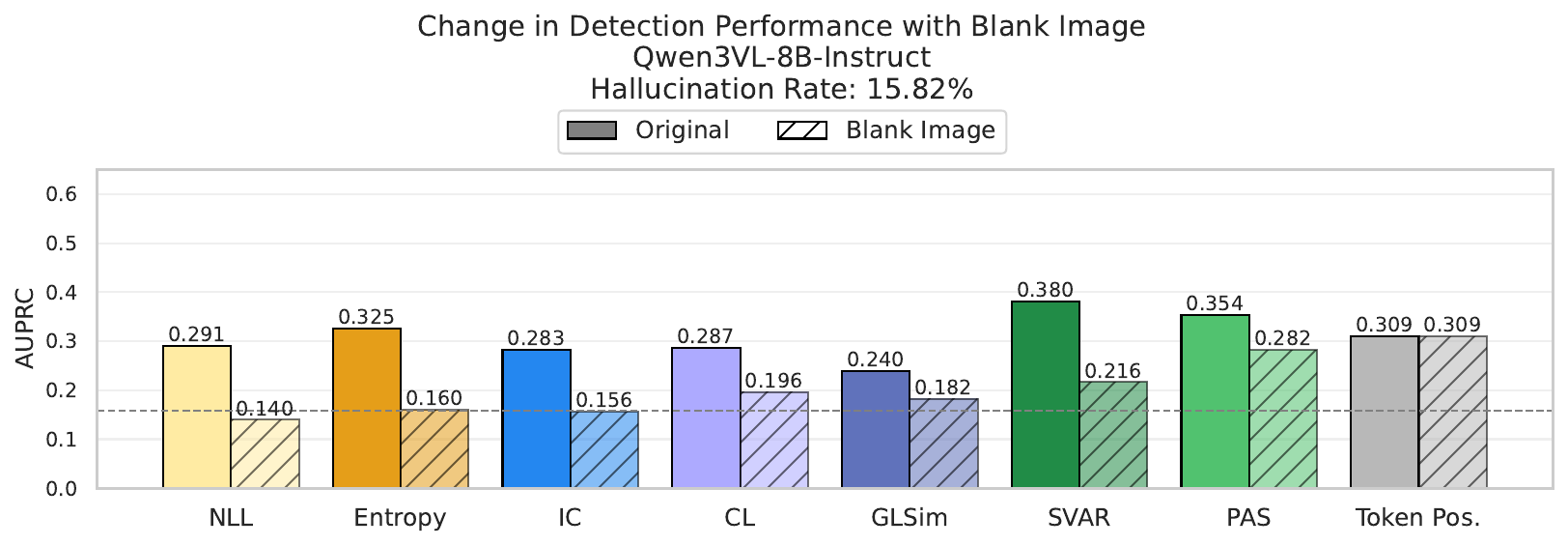}
    \\
    \small No Filtering ``orange''\par
    \includegraphics[width=0.49\linewidth, trim={0 0.1in 0 0.08in}, clip]{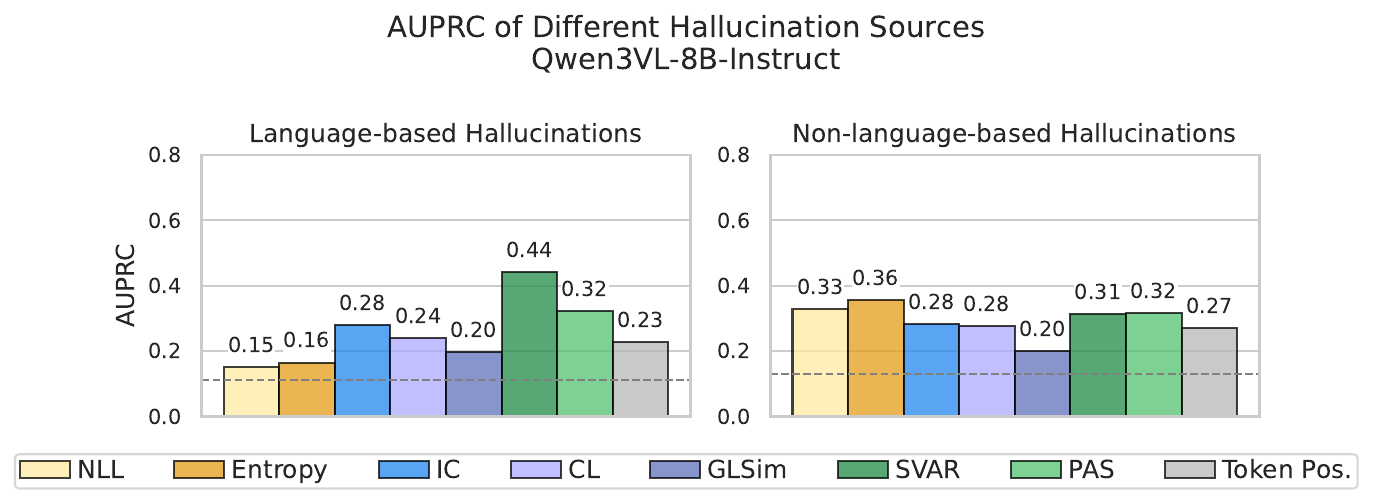}
    \includegraphics[width=0.49\linewidth, trim={0 0.1in 0 0.08in}, clip]{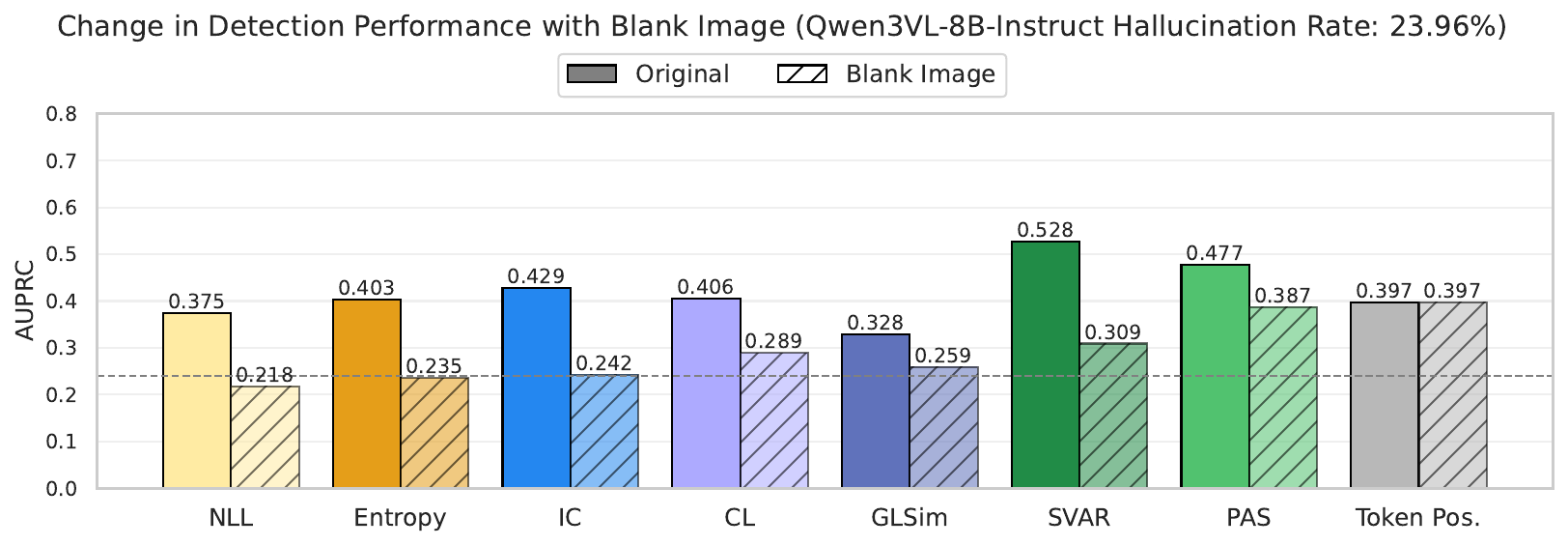}
    
    \caption{Method behavior partitioning by hallucination modality and visual groundedness on \qwen \ generalizes across various data and prompt settings.}
    \label{fig:qwen3vl8b_perf_generalize2}
\end{figure}

\begin{figure}[h!]
    \centering
    \small MSCOCO  \par
    \includegraphics[width=0.49\linewidth, trim={0 0.1in 0 0.08in}, clip]{figures/internvl2-8b/compare_auroc_auprc_legend.pdf}
    \includegraphics[width=0.49\linewidth, trim={0 0.1in 0 0.08in}, clip]{figures/internvl2-8b/remove_replace.pdf}
    \\
    \small Objects365 \par
    \includegraphics[width=0.49\linewidth, trim={0 0.1in 0 0.08in}, clip]{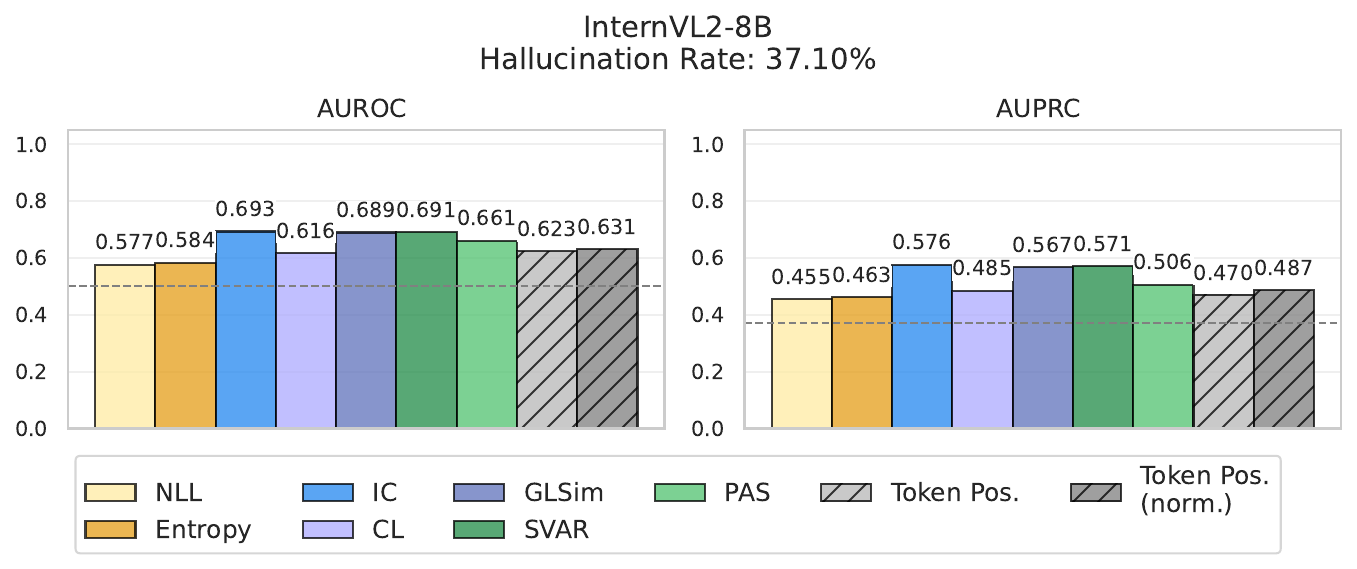}
    \includegraphics[width=0.49\linewidth, trim={0 0.1in 0 0.08in}, clip]{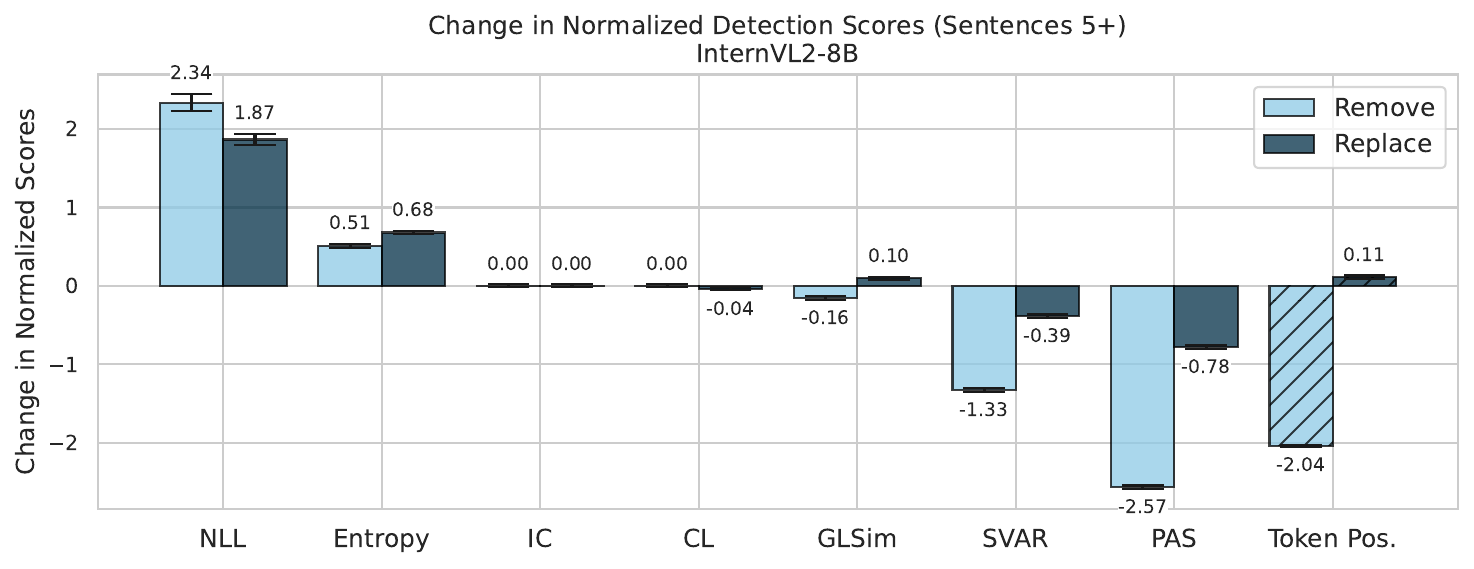}
    \\
    \small Prompt Variation \par
    \includegraphics[width=0.49\linewidth, trim={0 0.1in 0 0.08in}, clip]{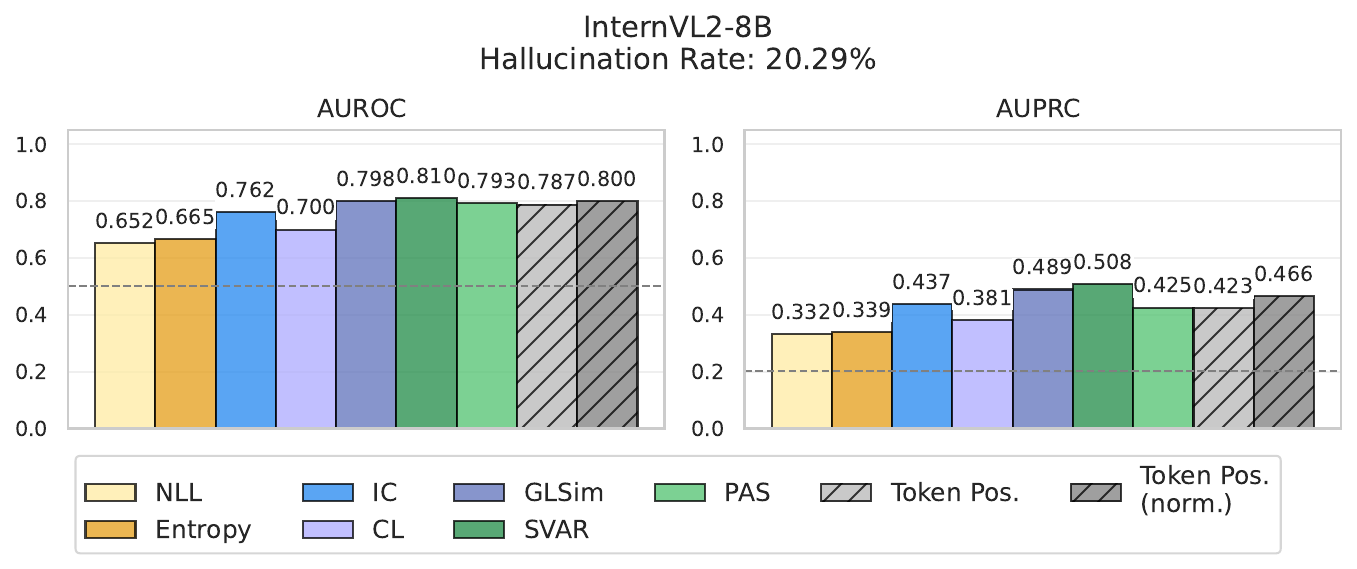}
    \includegraphics[width=0.49\linewidth, trim={0 0.1in 0 0.08in}, clip]{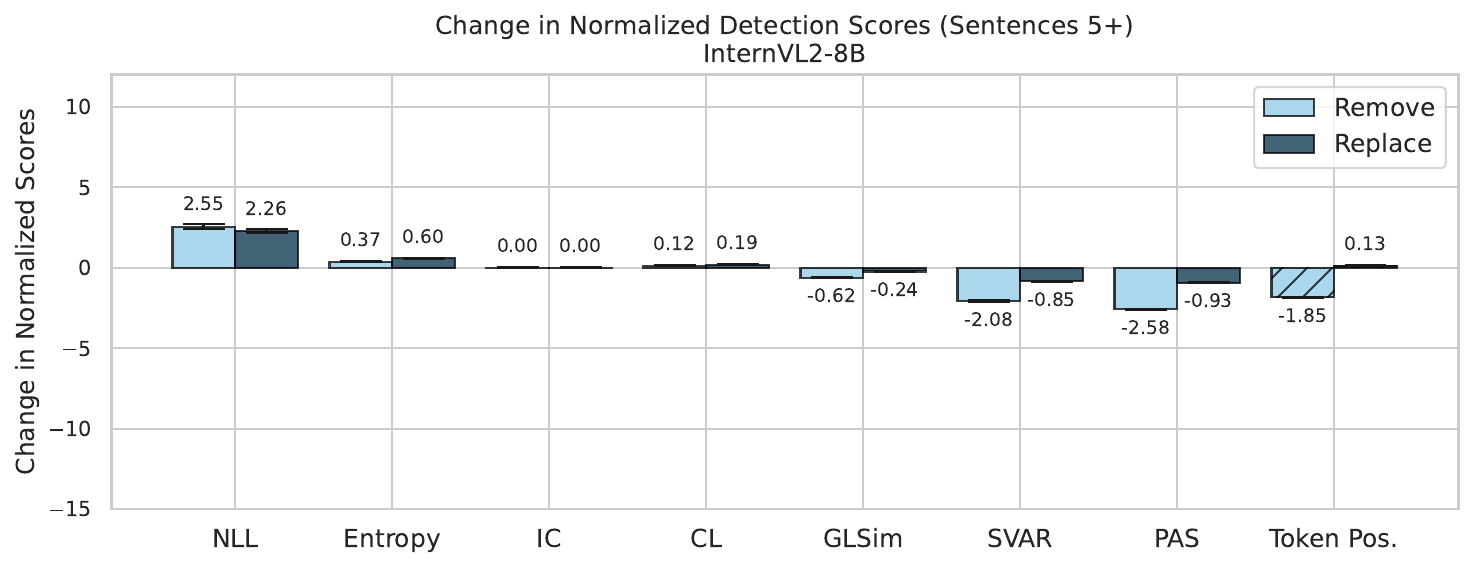}
    \\
    \small Repeated Trials \par
    \includegraphics[width=0.49\linewidth, trim={0 0.1in 0 0.08in}, clip]{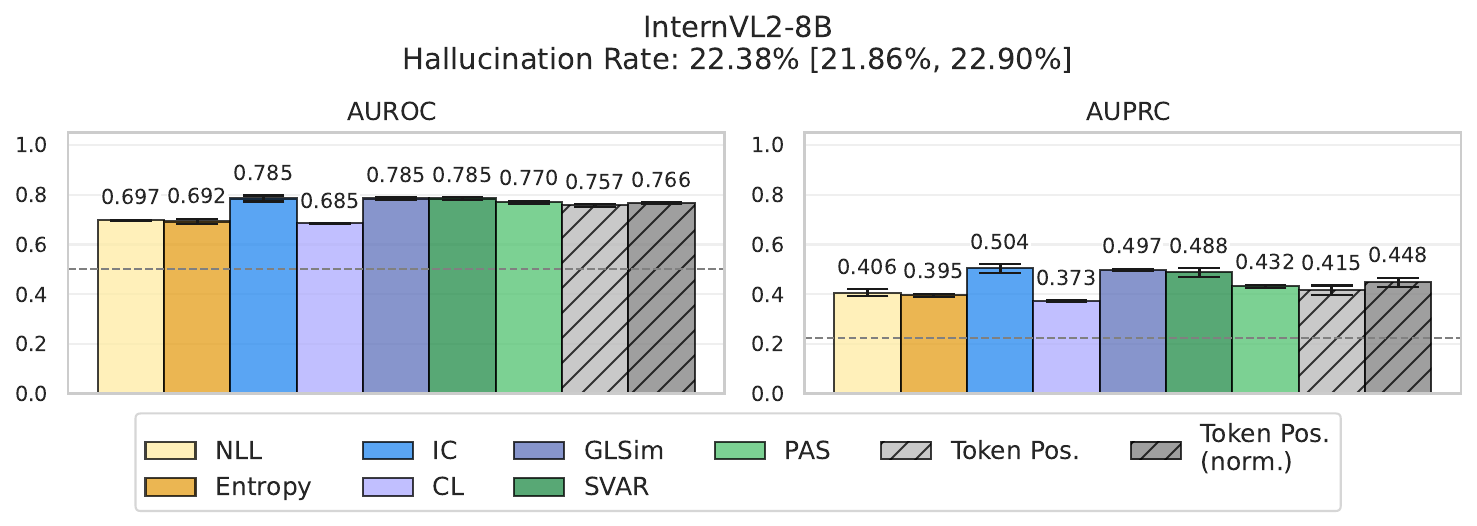}
    \includegraphics[width=0.49\linewidth, trim={0 0.1in 0 0.08in}, clip]{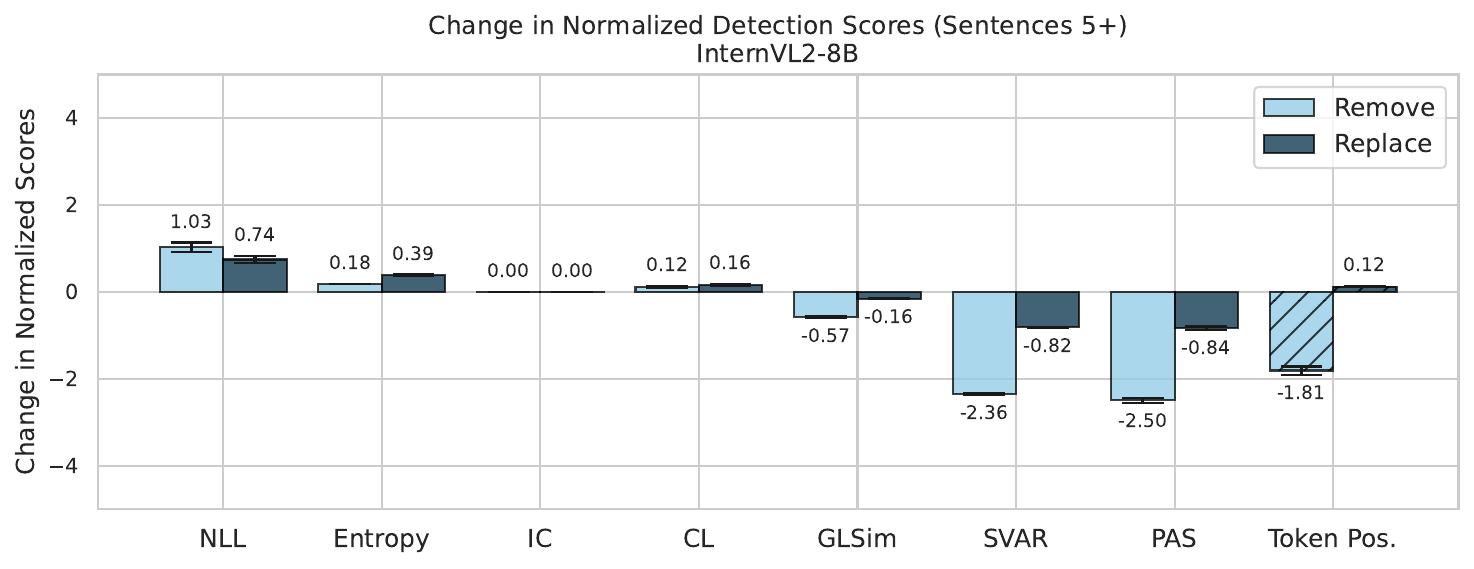}
    \\
    \small Context-aware Filtering \par
    \includegraphics[width=0.49\linewidth, trim={0 0.1in 0 0.08in}, clip]{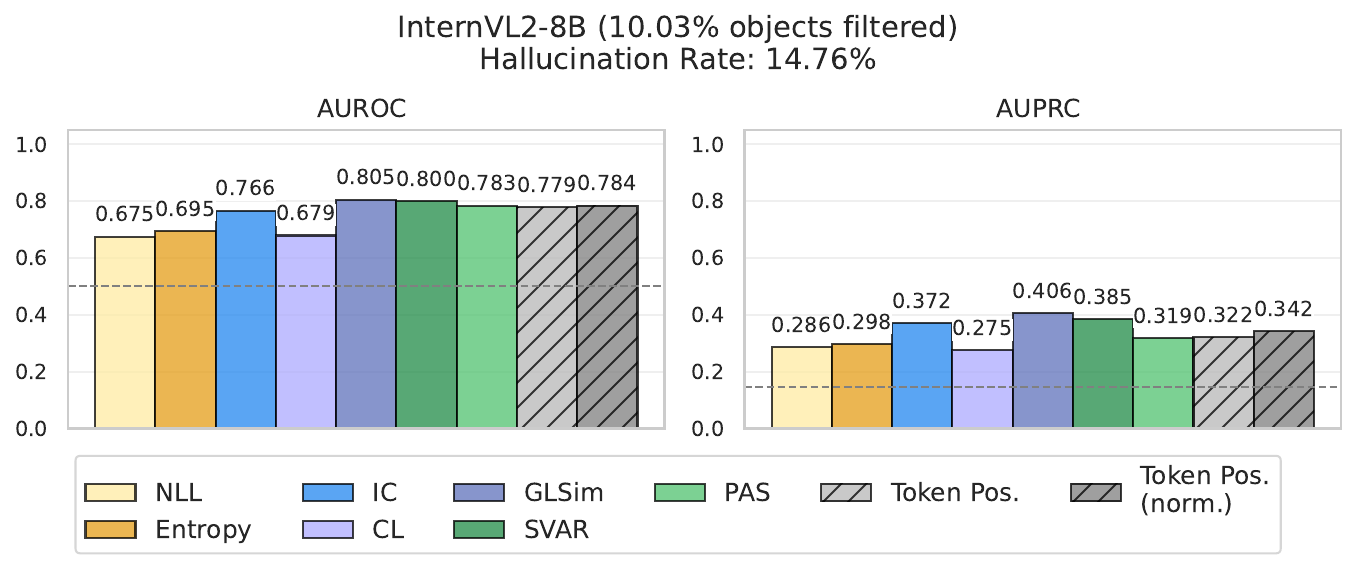}
    \includegraphics[width=0.49\linewidth, trim={0 0.1in 0 0.08in}, clip]{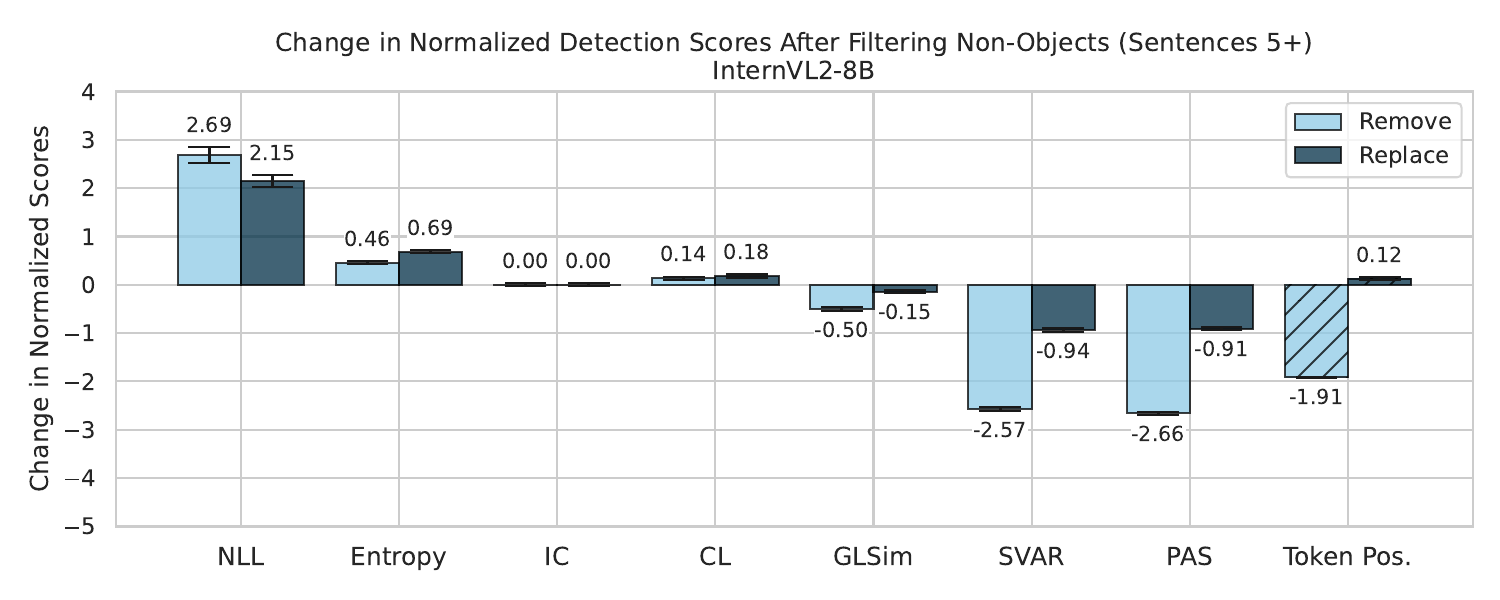}
    \\
    \small No Filtering ``orange''\par
    \includegraphics[width=0.49\linewidth, trim={0 0.1in 0 0.08in}, clip]{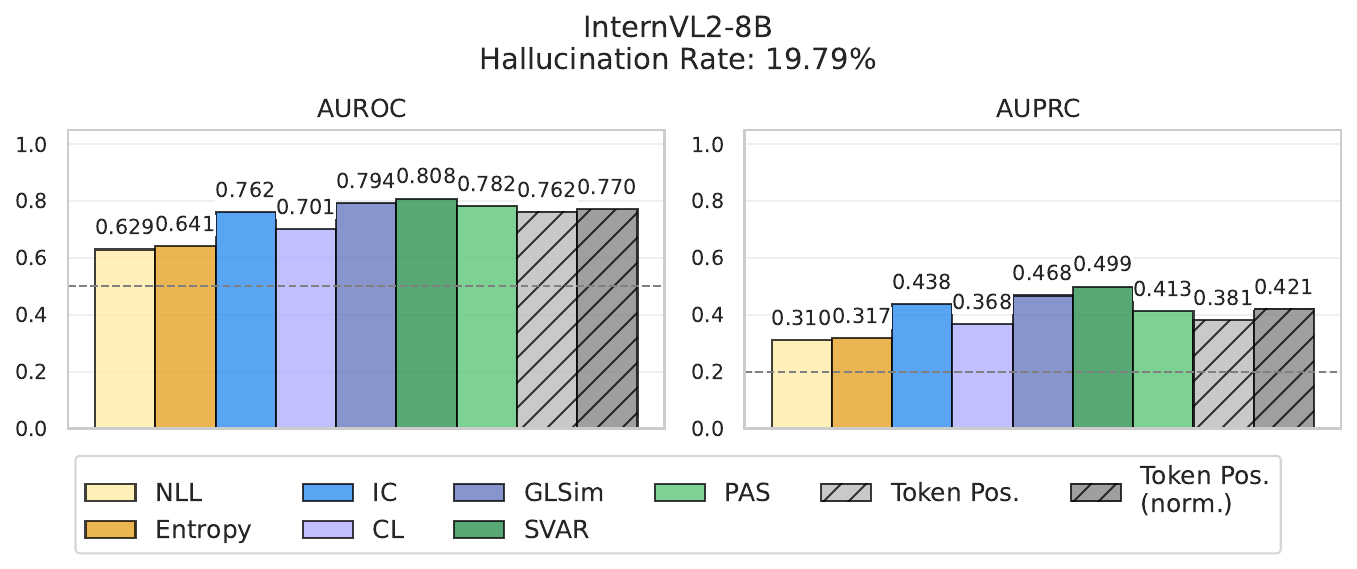}
    \includegraphics[width=0.49\linewidth, trim={0 0.1in 0 0.08in}, clip]{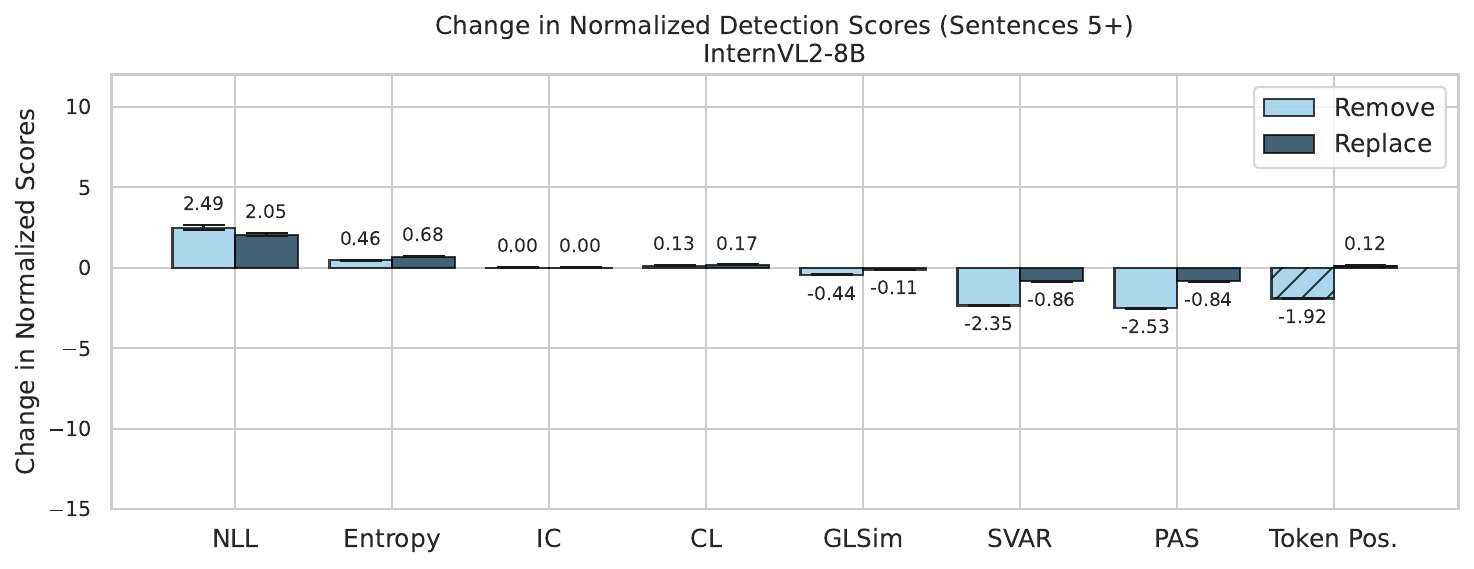}
    \caption{\textbf{Left:} Performance of hallucination detection methods on \internvl \ generalizes across various data and prompt settings. \textbf{Right:} Method behavior with remove/replace preceding sentences \internvl \ generalizes across various data and prompt settings.}

    \label{fig:internvl8b_perf_generalize1}
\end{figure}

\begin{figure}[h!]
    \centering
    \small MSCOCO  \par
    \includegraphics[width=0.49\linewidth, trim={0 0.1in 0 0.08in}, clip]{figures/internvl2-8b/hall_source_auprc.pdf}
    \includegraphics[width=0.49\linewidth, trim={0 0.1in 0 0.08in}, clip]{figures/internvl2-8b/auprc_blank_performance_diff.pdf}
    \\
    \small Objects365 \par
    \includegraphics[width=0.49\linewidth, trim={0 0.1in 0 0.08in}, clip]{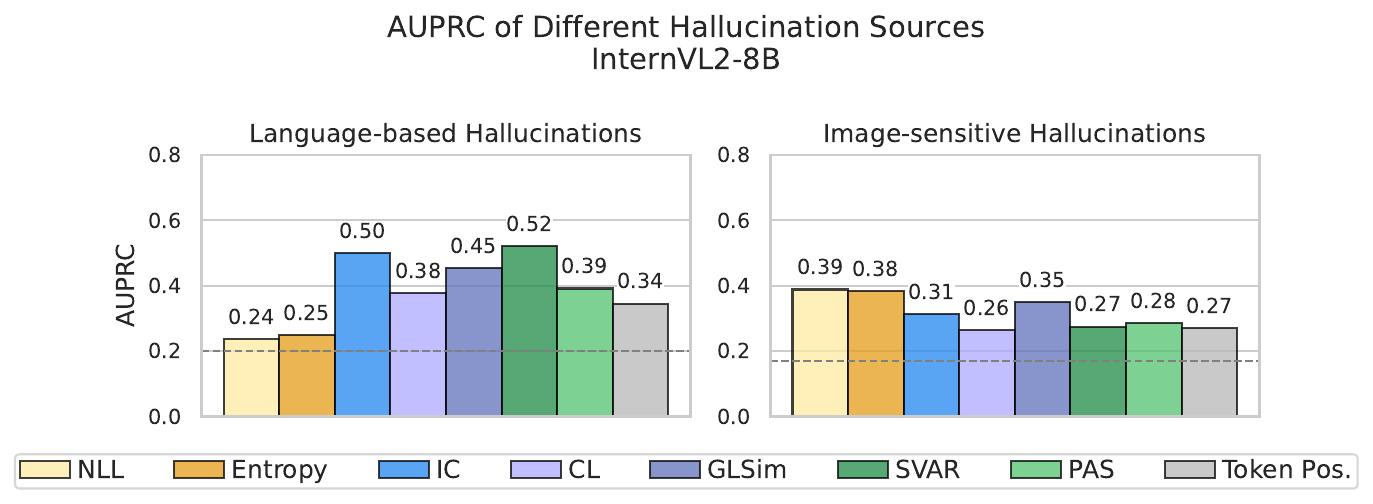}
    \includegraphics[width=0.49\linewidth, trim={0 0.1in 0 0.08in}, clip]{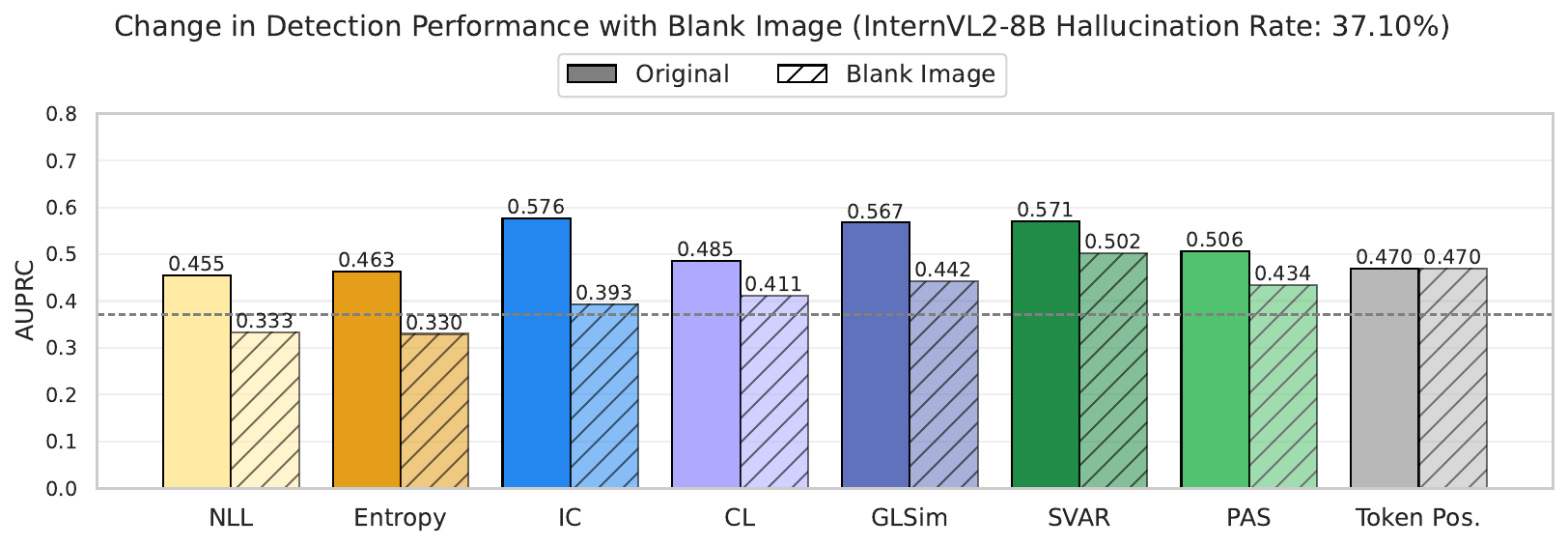}
    \\
    \small Prompt Variation \par
    \includegraphics[width=0.49\linewidth, trim={0 0.1in 0 0.08in}, clip]{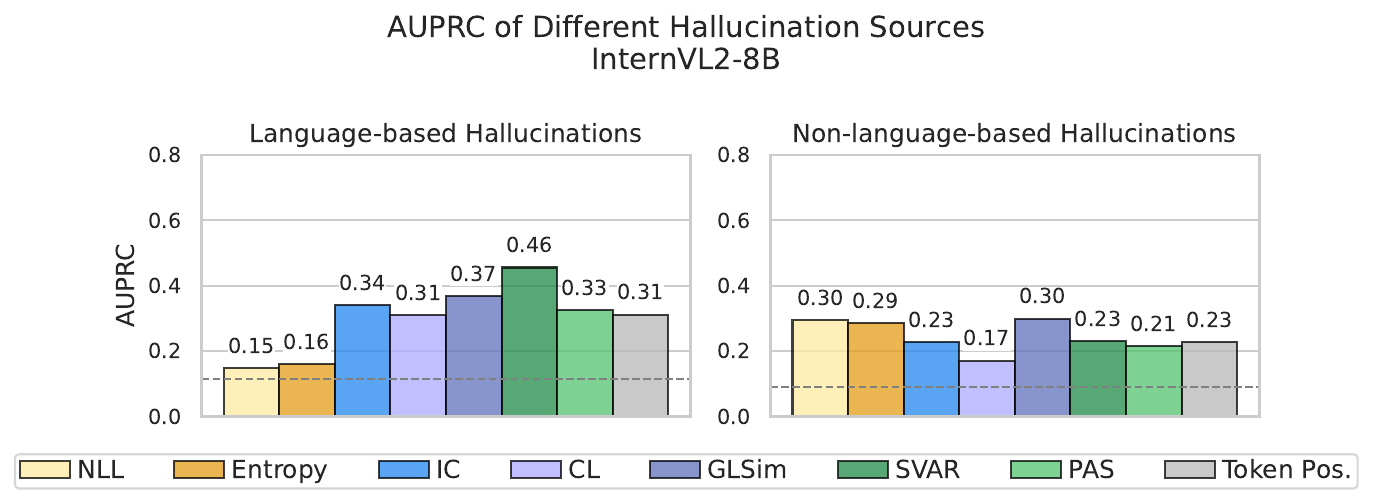}
    \includegraphics[width=0.49\linewidth, trim={0 0.1in 0 0.08in}, clip]{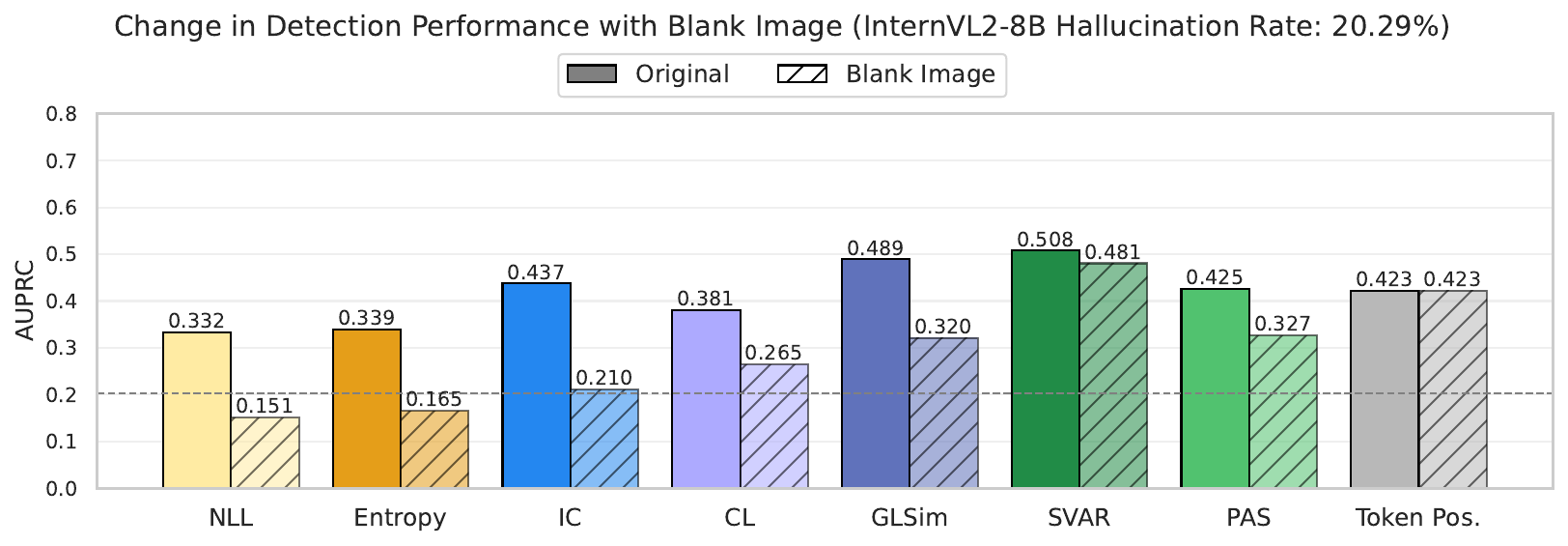}
    \\
    \small Repeated Trials \par
    \includegraphics[width=0.49\linewidth]  {figures/coco_sampling/unavailable_figure.pdf}    \includegraphics[width=0.49\linewidth, trim={0 0.1in 0 0.08in}, clip]{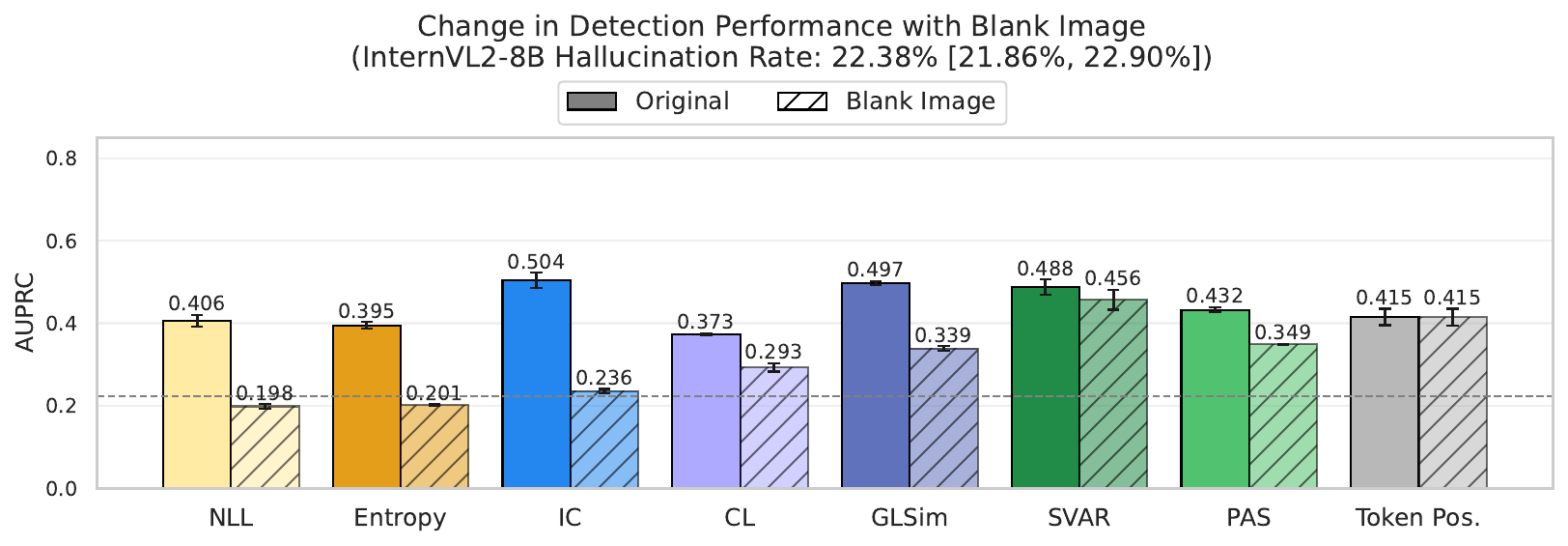}
    \\
    \small Context-aware Filtering \par
    \includegraphics[width=0.49\linewidth, trim={0 0.1in 0 0.08in}, clip]{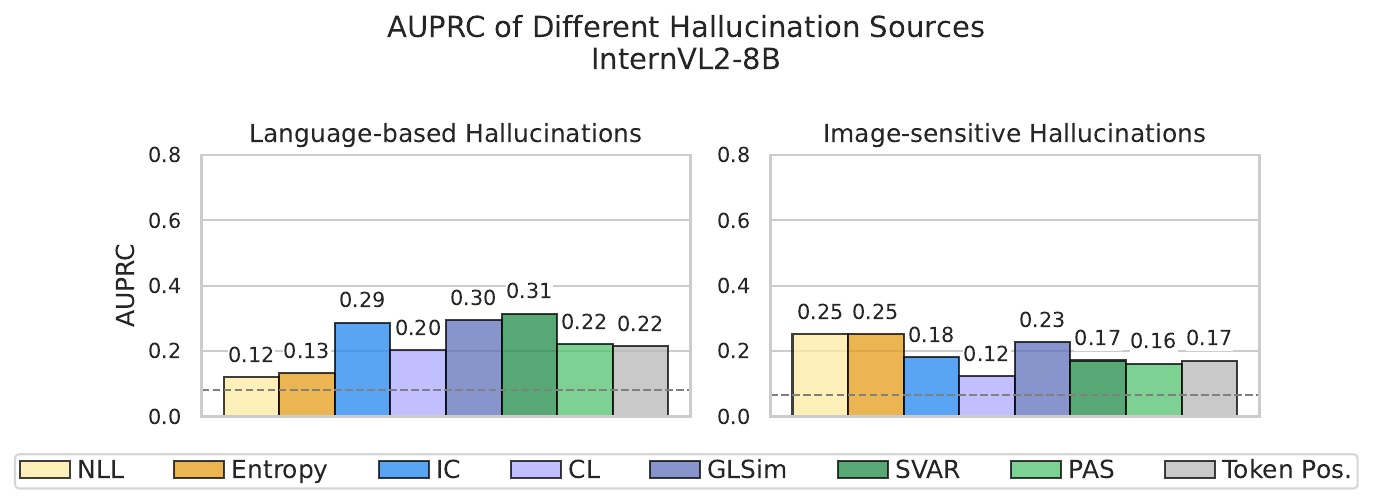}
    \includegraphics[width=0.49\linewidth, trim={0 0.1in 0 0.08in}, clip]{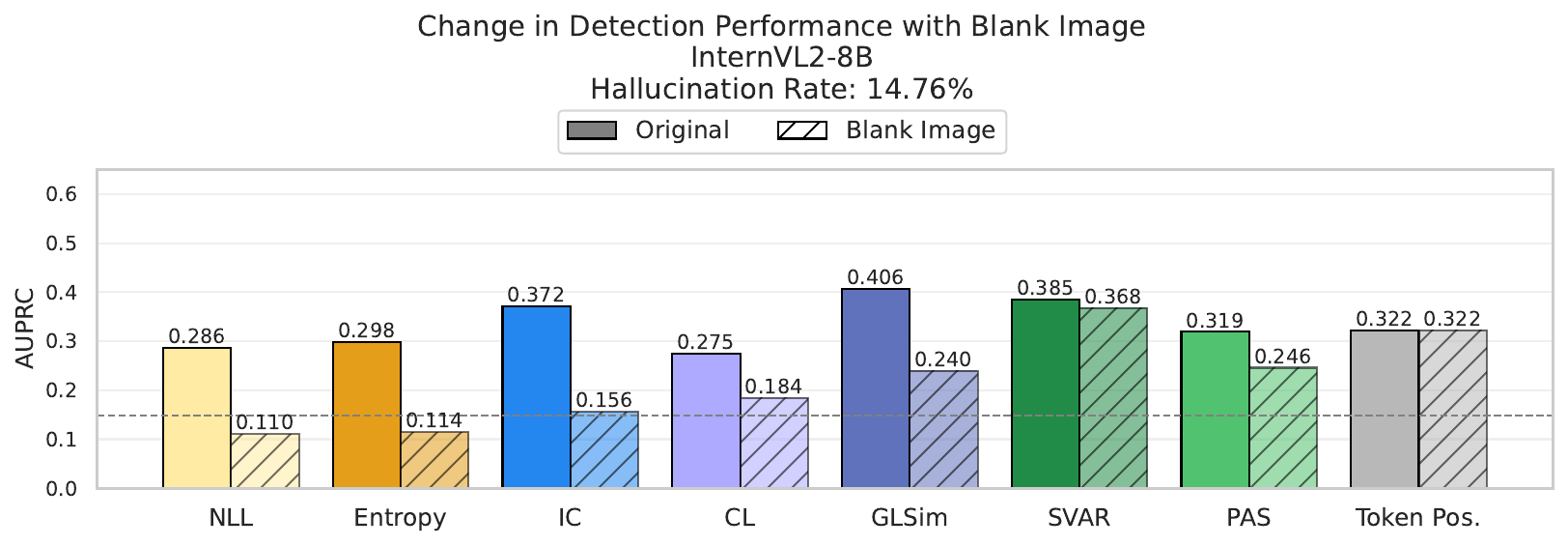}
    \\
    \small No Filtering ``orange''\par
    \includegraphics[width=0.49\linewidth, trim={0 0.1in 0 0.08in}, clip]{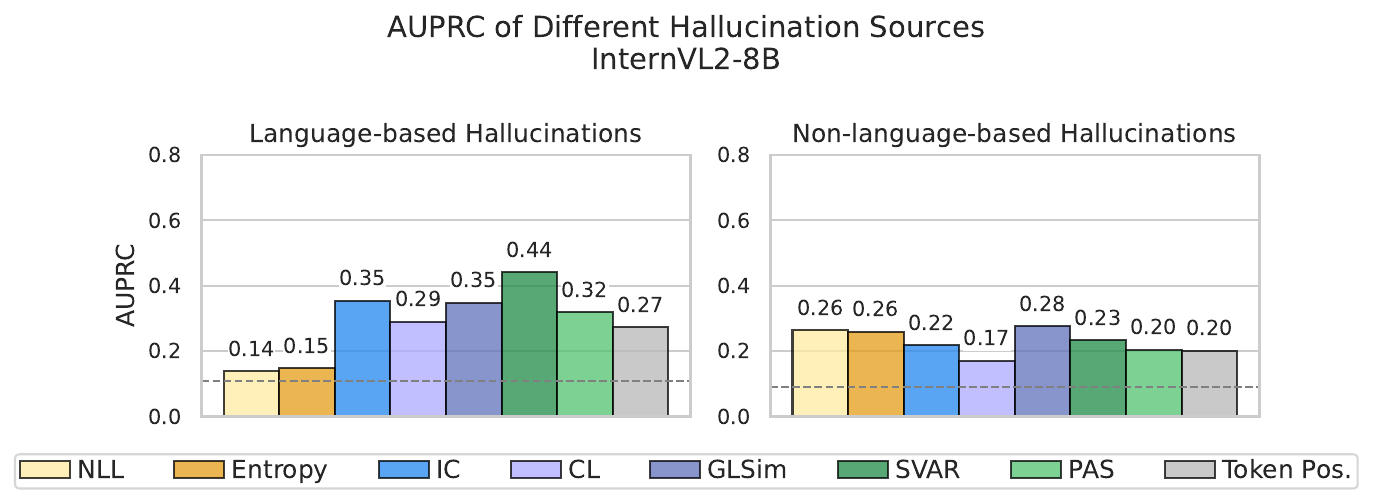}
    \includegraphics[width=0.49\linewidth, trim={0 0.1in 0 0.08in}, clip]{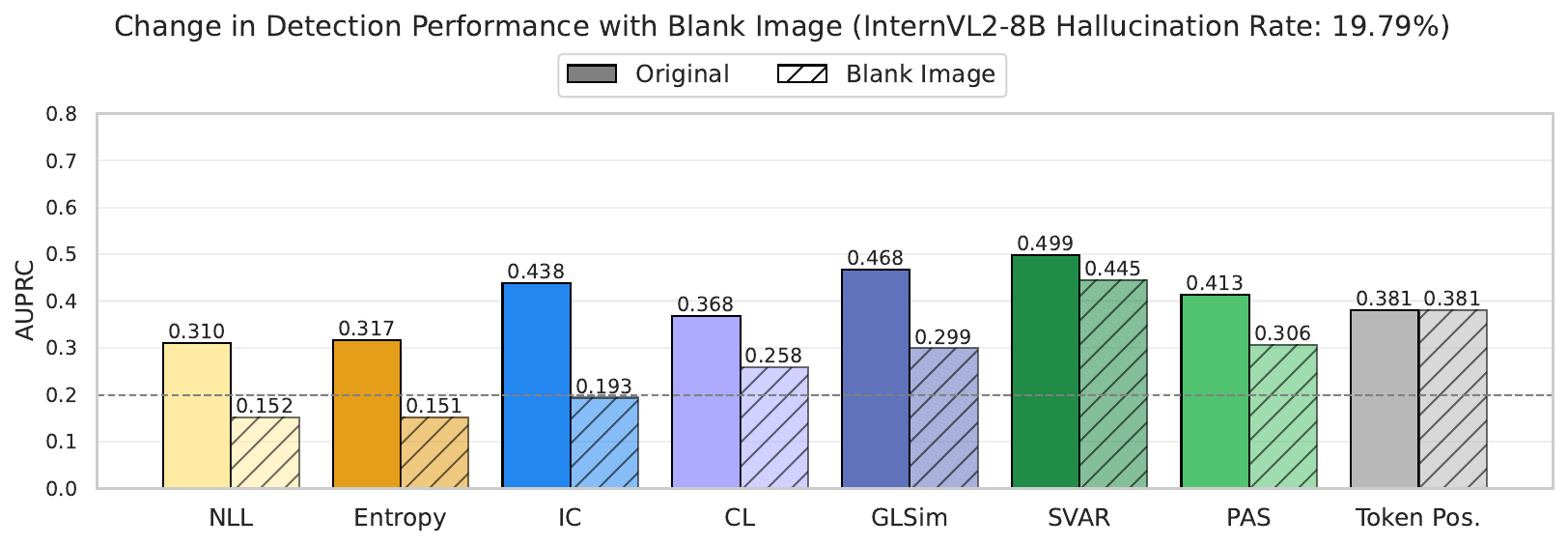}
    \caption{Method behavior partitioning by hallucination modality and visual groundedness on \internvl \ generalizes across various data and prompt settings.}
    \label{fig:internvl8b_perf_generalize2}
\end{figure}

\begin{figure}[h!]
    \centering
    \small MSCOCO  \par
    \includegraphics[width=0.49\linewidth, trim={0 0.1in 0 0.08in}, clip]{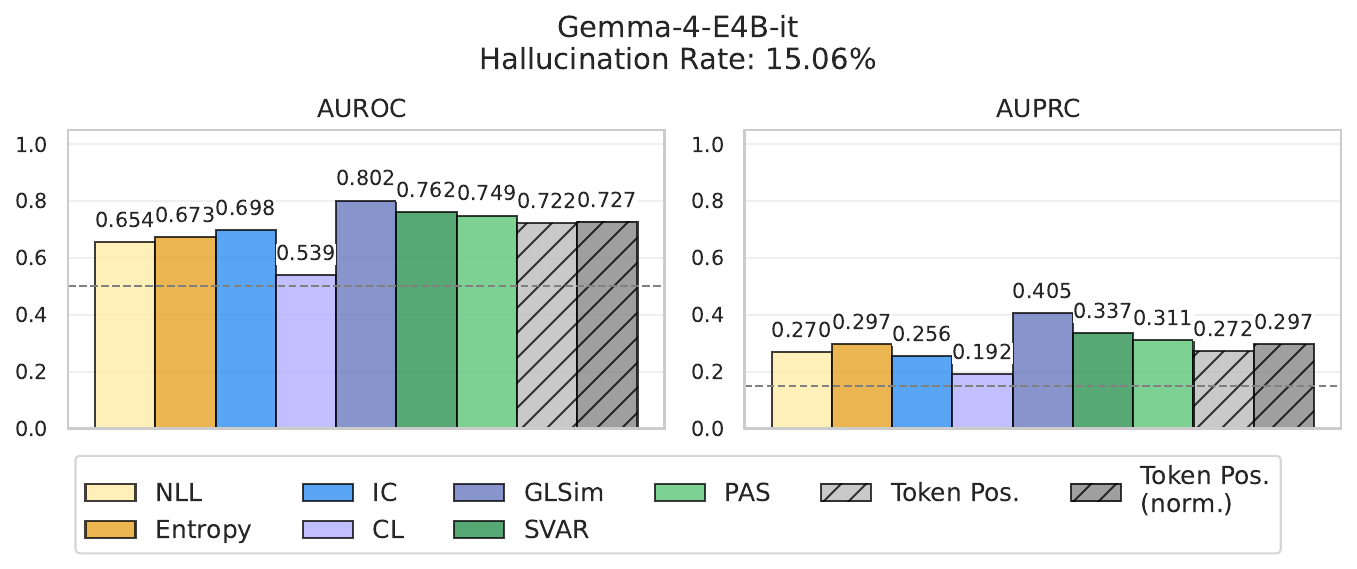}
    \includegraphics[width=0.49\linewidth, trim={0 0.1in 0 0.08in}, clip]{figures/gemma4-e4b-it/remove_replace.pdf}
    \\
    \small Objects365 \par
    \includegraphics[width=0.49\linewidth, trim={0 0.1in 0 0.08in}, clip]{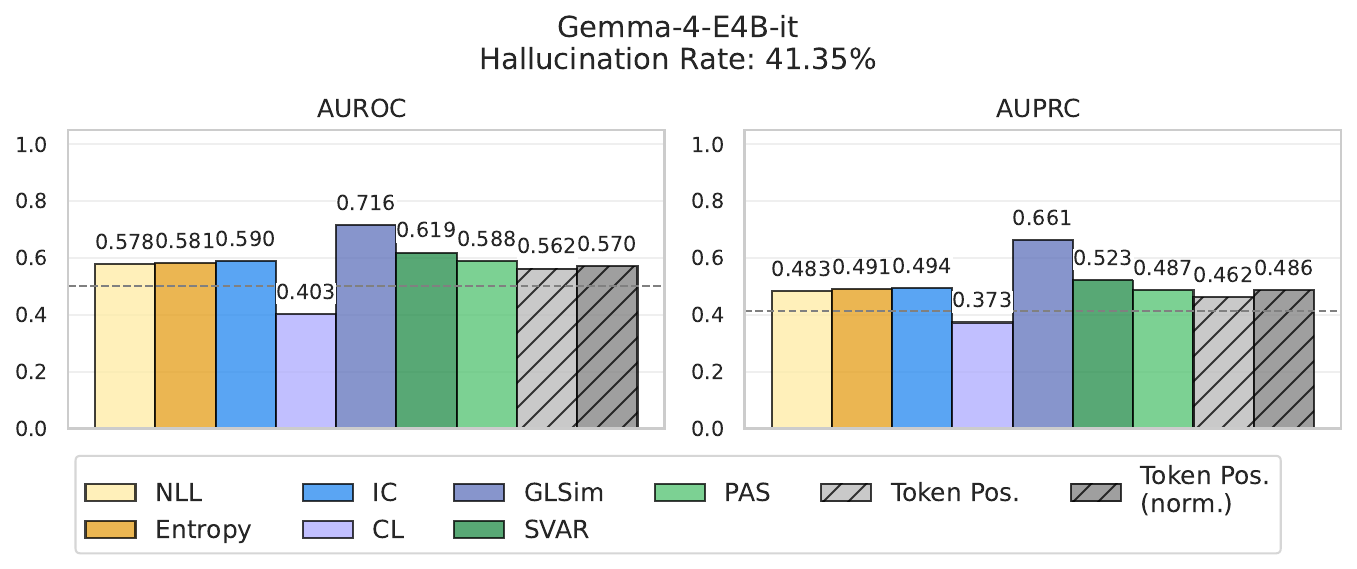}
    \includegraphics[width=0.49\linewidth, trim={0 0.1in 0 0.08in}, clip]{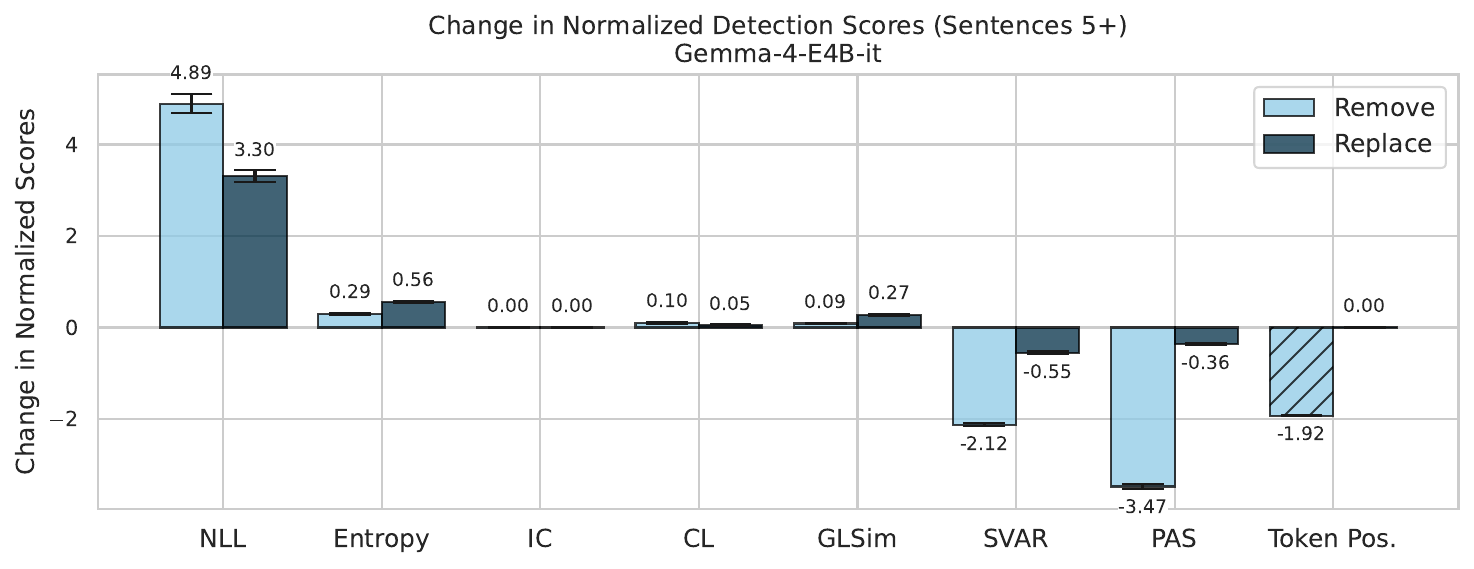}
    \\
    \small Prompt Variation \par
    \includegraphics[width=0.49\linewidth, trim={0 0.1in 0 0.08in}, clip]{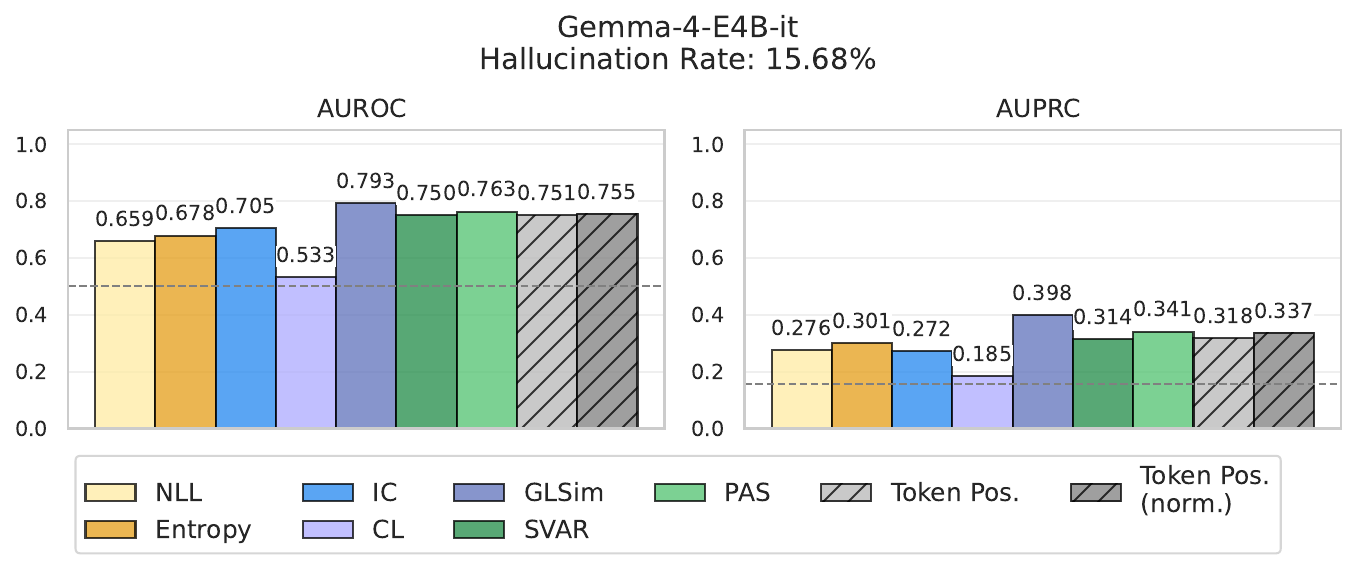}
    \includegraphics[width=0.49\linewidth, trim={0 0.1in 0 0.08in}, clip]{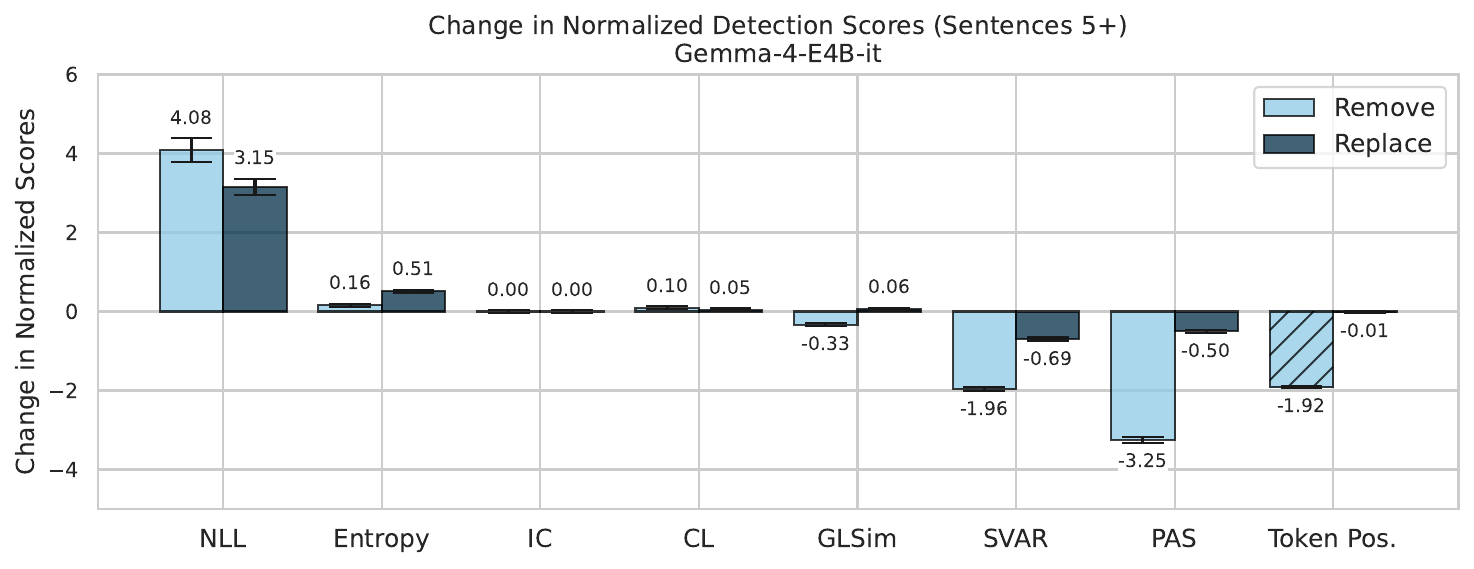}
    \\
    \small Repeated Trials \par
    \includegraphics[width=0.49\linewidth, trim={0 0.1in 0 0.08in}, clip]{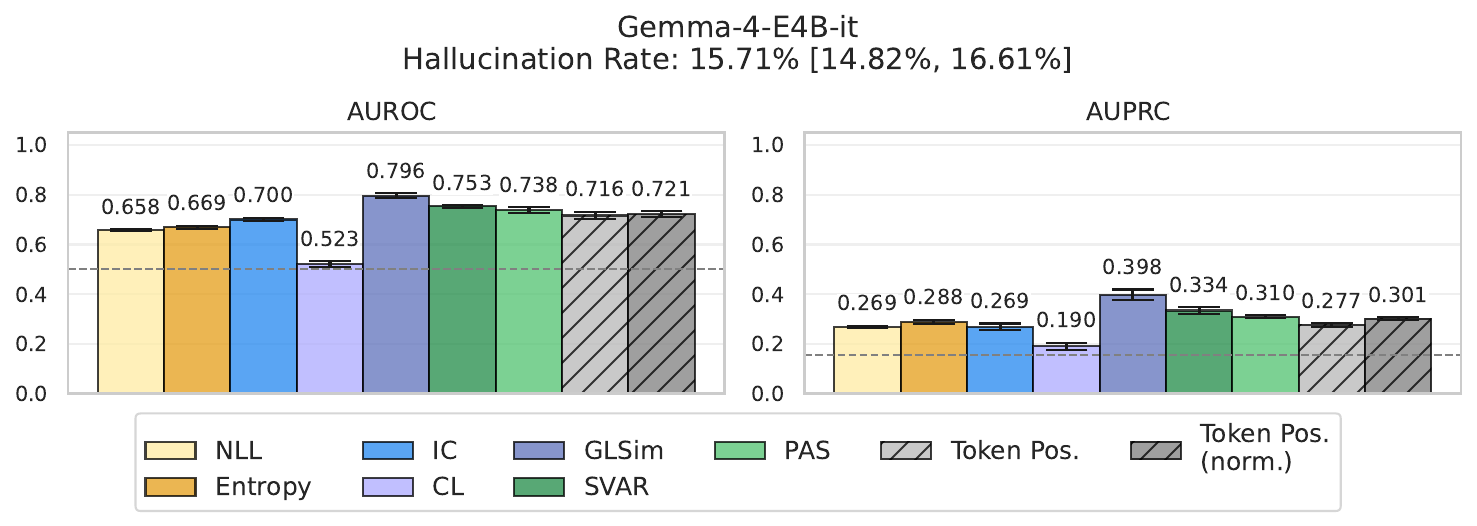}
    \includegraphics[width=0.49\linewidth, trim={0 0.1in 0 0.08in}, clip]{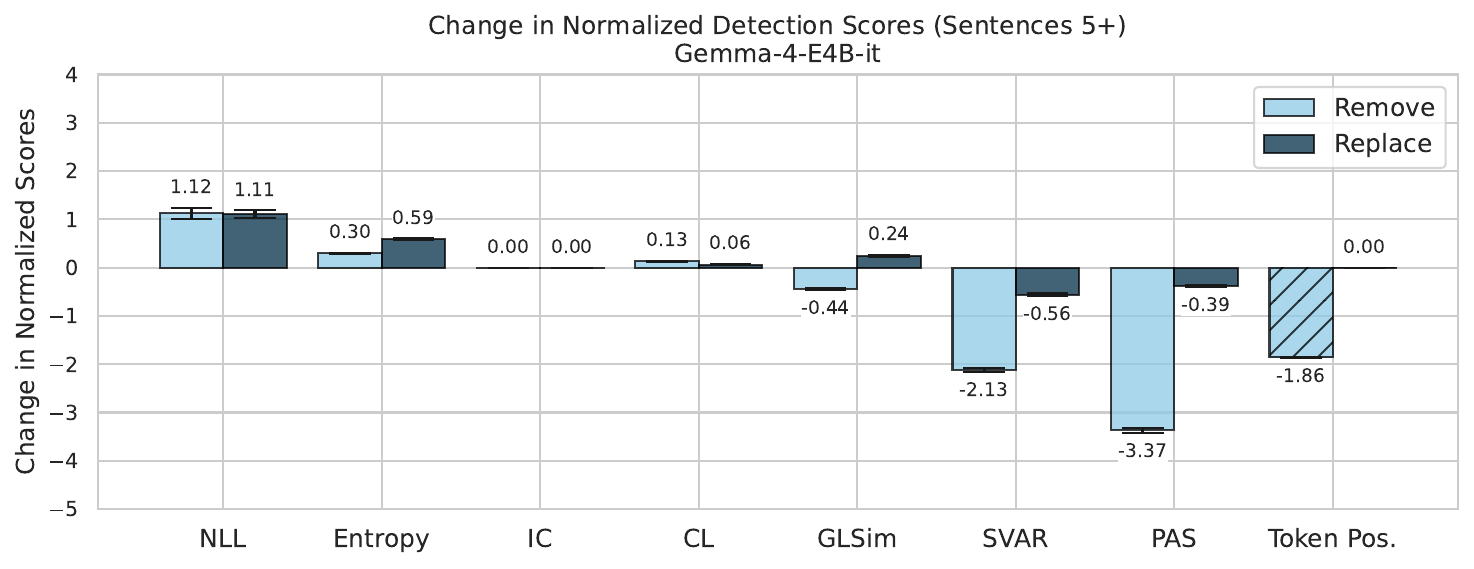}
    \\
    \small Context-aware Filtering \par
    \includegraphics[width=0.49\linewidth, trim={0 0.1in 0 0.08in}, clip]{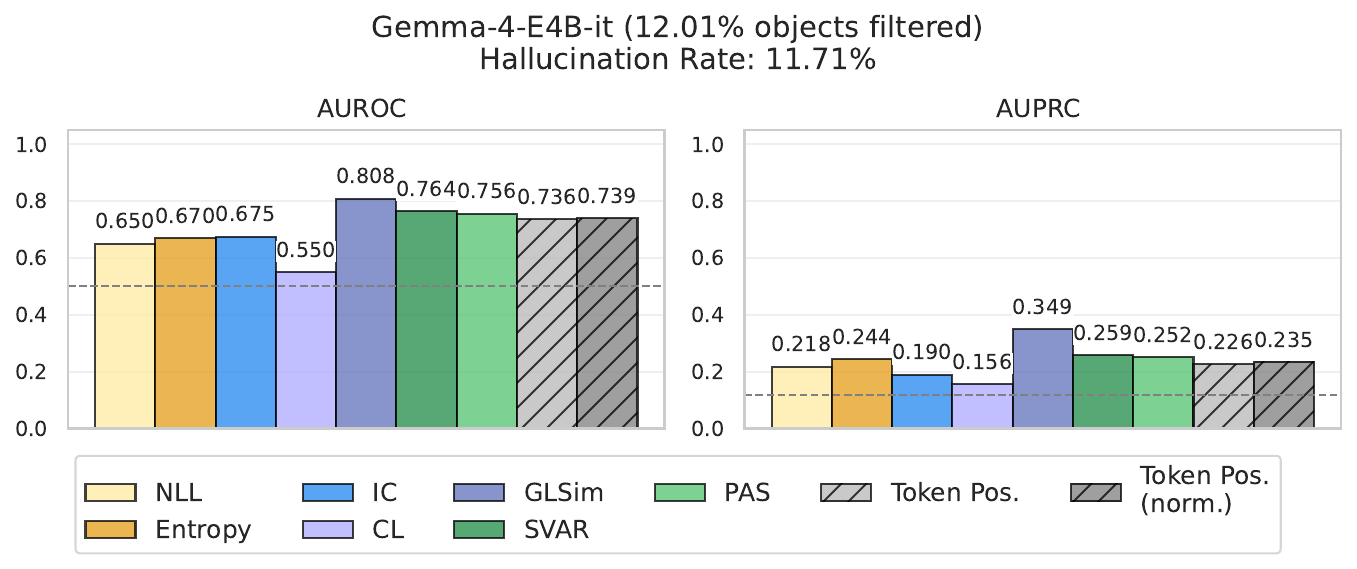}
    \includegraphics[width=0.49\linewidth, trim={0 0.1in 0 0.08in}, clip]{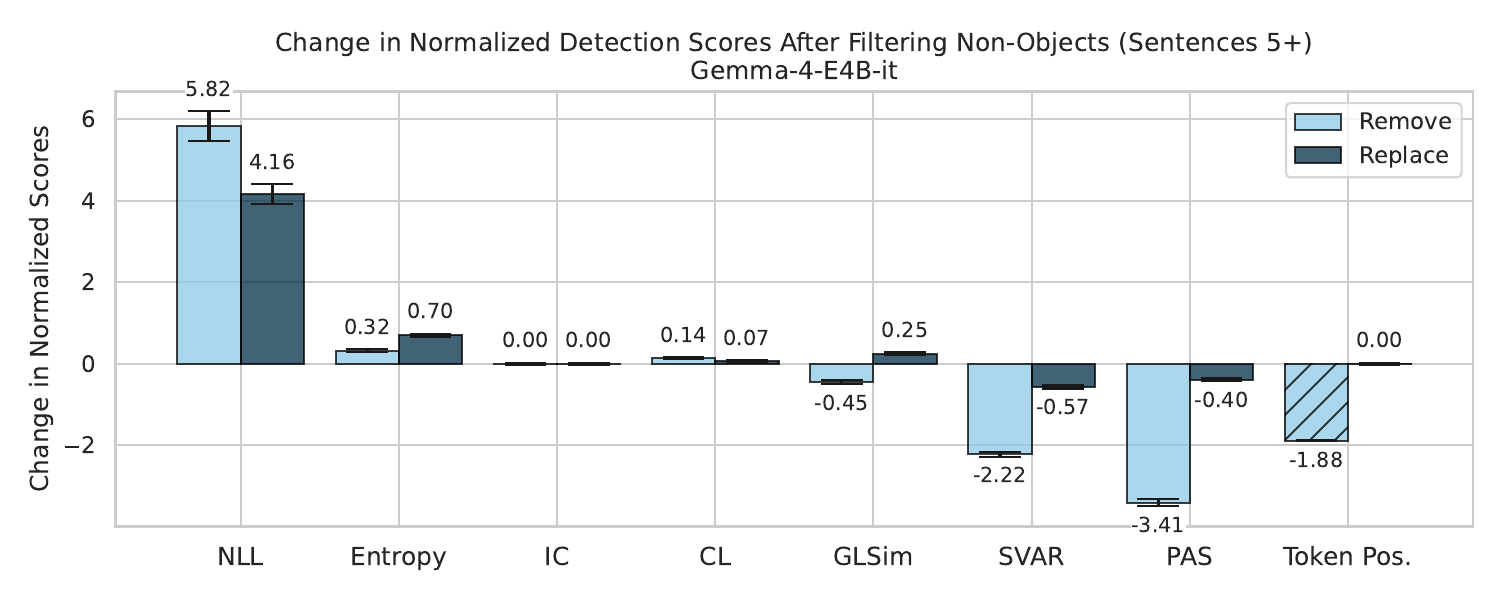}
    \\
    \small No Filtering ``orange''\par
    \includegraphics[width=0.49\linewidth, trim={0 0.1in 0 0.08in}, clip]{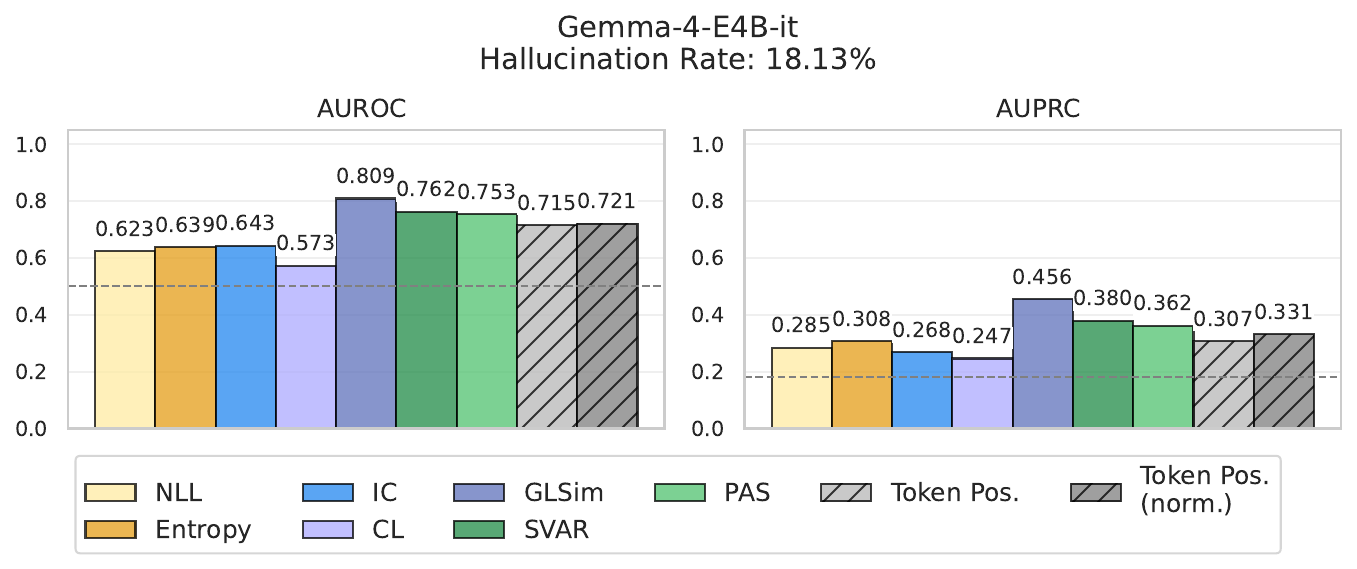}
    \includegraphics[width=0.49\linewidth, trim={0 0.1in 0 0.08in}, clip]{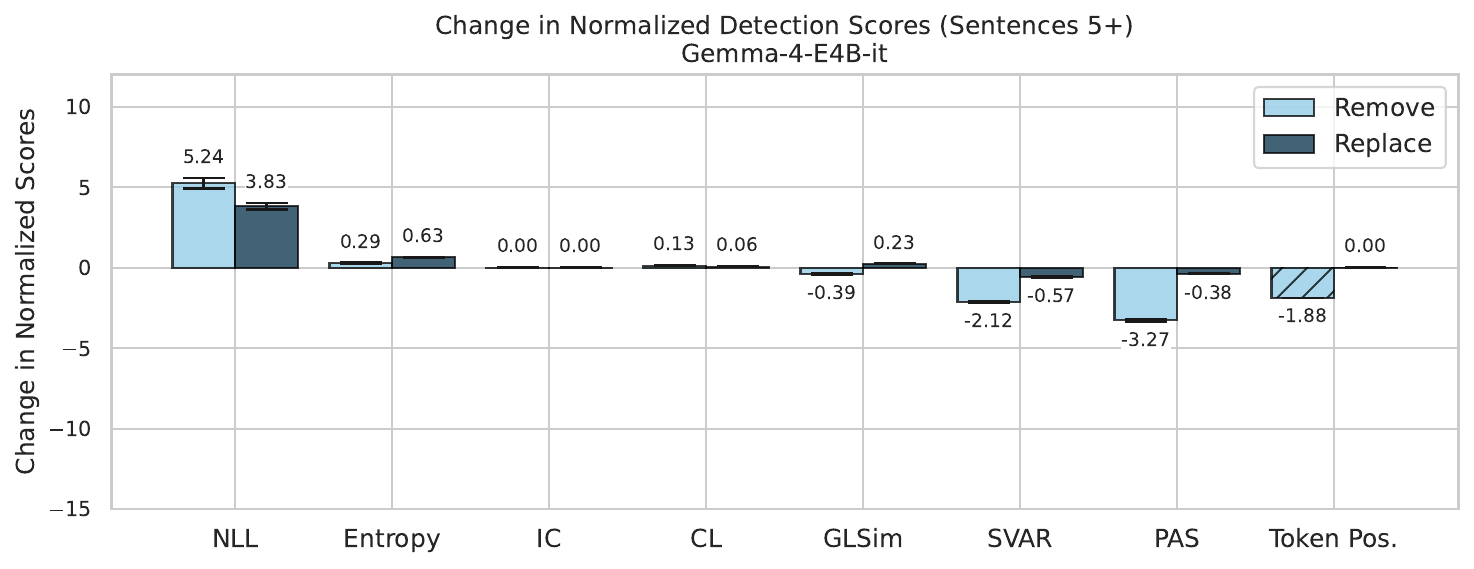}
    \caption{\textbf{Left:} Performance of hallucination detection methods on \gemma \ generalizes across various data and prompt settings. \textbf{Right:} Method behavior with remove/replace preceding sentences \gemma \ generalizes across various data and prompt settings.}
    \label{fig:gemma4_perf_generalize1}
\end{figure}

\begin{figure}[h!]
    \centering
    \small MSCOCO  \par
    \includegraphics[width=0.49\linewidth, trim={0 0.1in 0 0.08in}, clip]{figures/gemma4-e4b-it/hall_source_auprc.pdf}
    \includegraphics[width=0.49\linewidth, trim={0 0.1in 0 0.08in}, clip]{figures/gemma4-e4b-it/auprc_blank_performance_diff.pdf}
    \\
    \small Objects365 \par
    \includegraphics[width=0.49\linewidth, trim={0 0.1in 0 0.08in}, clip]{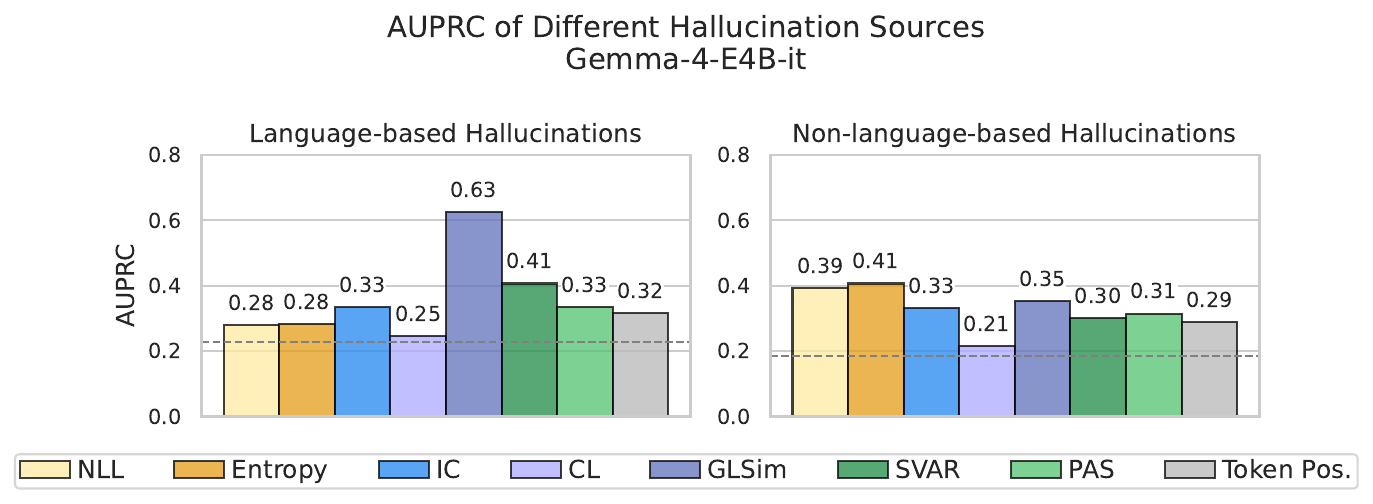}
    \includegraphics[width=0.49\linewidth, trim={0 0.1in 0 0.08in}, clip]{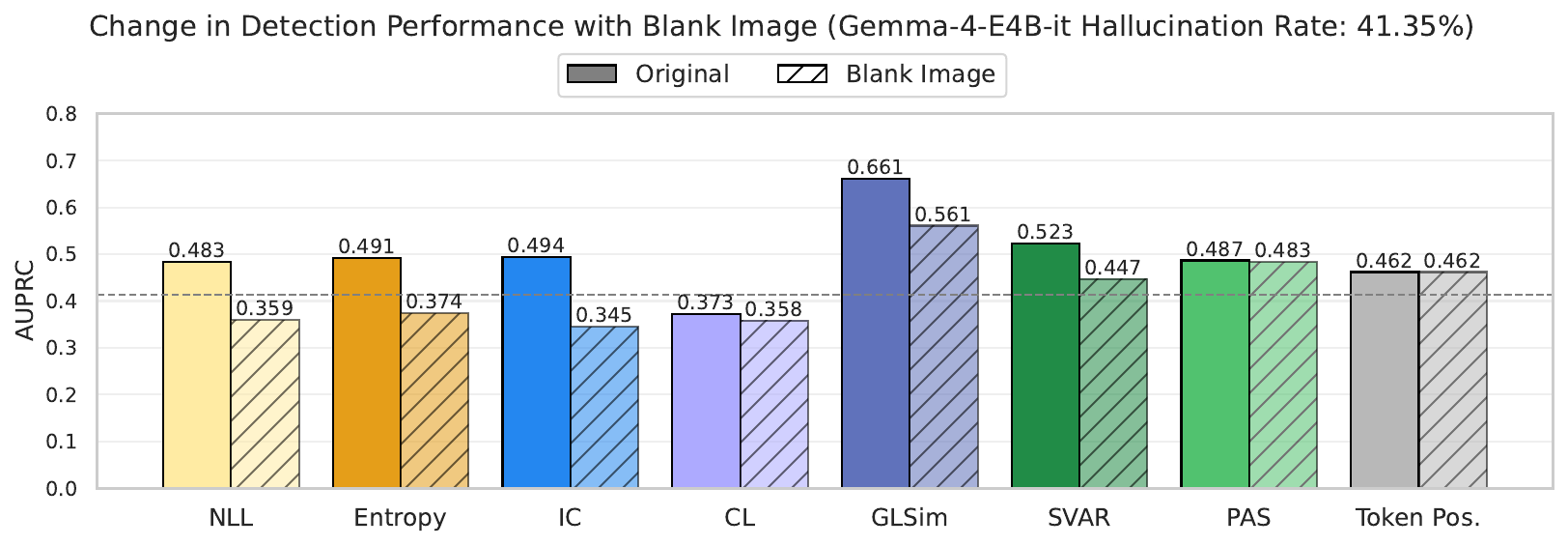}
    \\
    \small Prompt Variation \par
    \includegraphics[width=0.49\linewidth, trim={0 0.1in 0 0.08in}, clip]{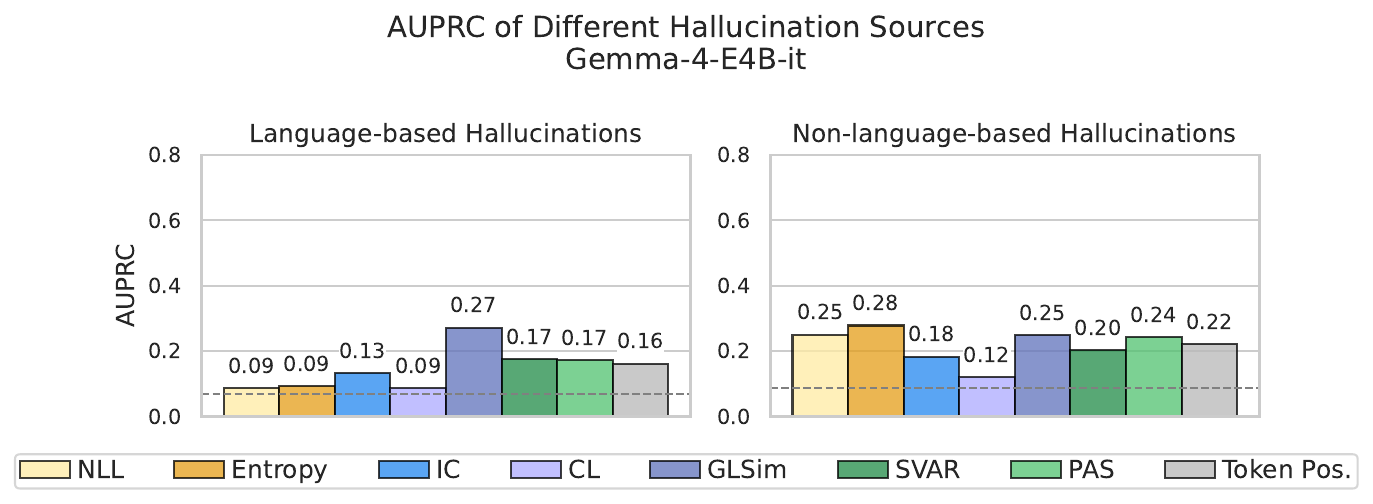}
    \includegraphics[width=0.49\linewidth, trim={0 0.1in 0 0.08in}, clip]{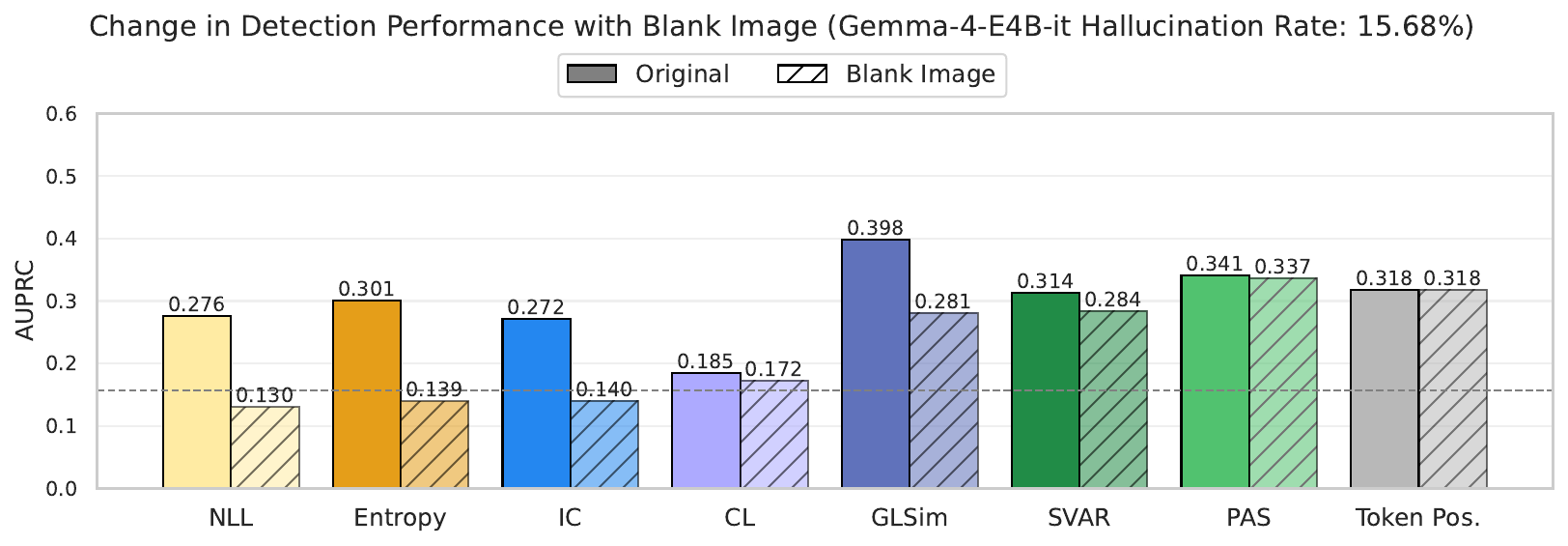}
    \\
    \small Repeated Trials \par
    \includegraphics[width=0.49\linewidth]{figures/coco_sampling/unavailable_figure.pdf}
    \includegraphics[width=0.49\linewidth, trim={0 0.1in 0 0.08in}, clip]{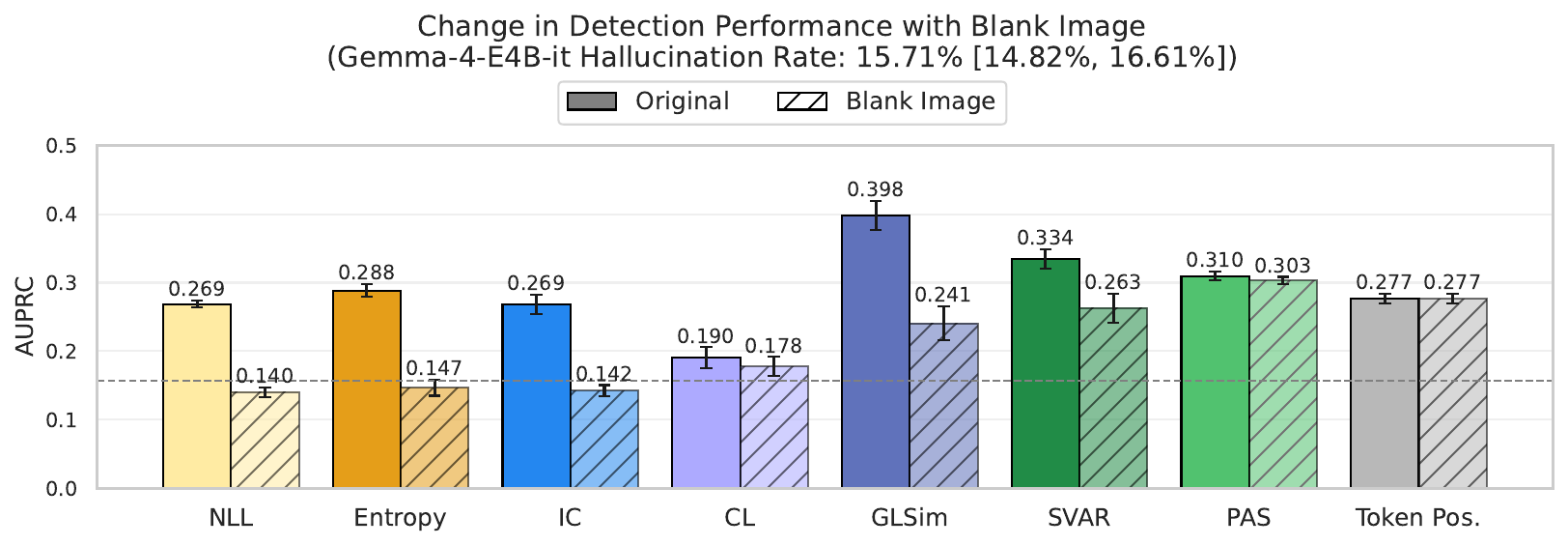}
    \\
    \small Context-aware Filtering \par
    \includegraphics[width=0.49\linewidth, trim={0 0.1in 0 0.08in}, clip]{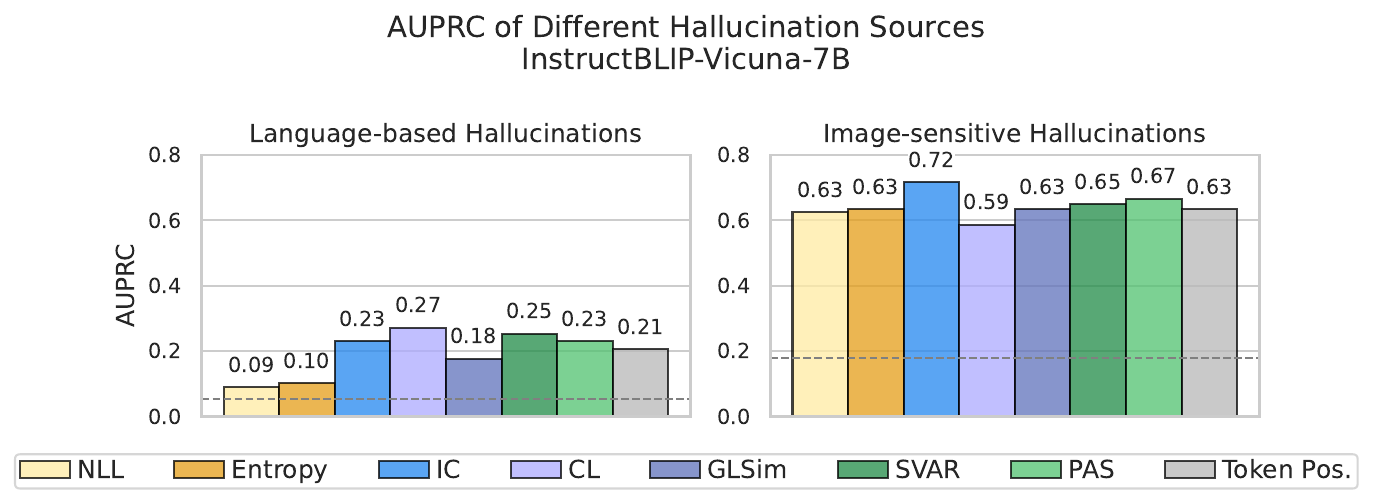}
    \includegraphics[width=0.49\linewidth, trim={0 0.1in 0 0.08in}, clip]{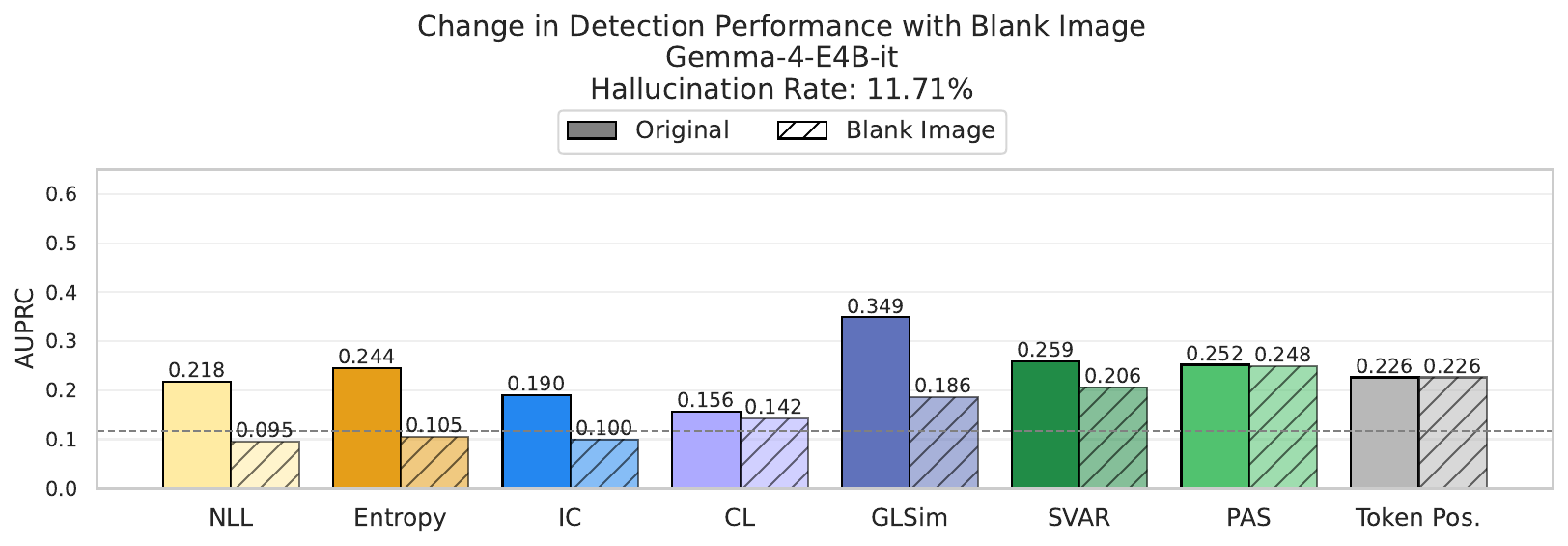}
    \\
    \small No Filtering ``orange''\par
    \includegraphics[width=0.49\linewidth, trim={0 0.1in 0 0.08in}, clip]{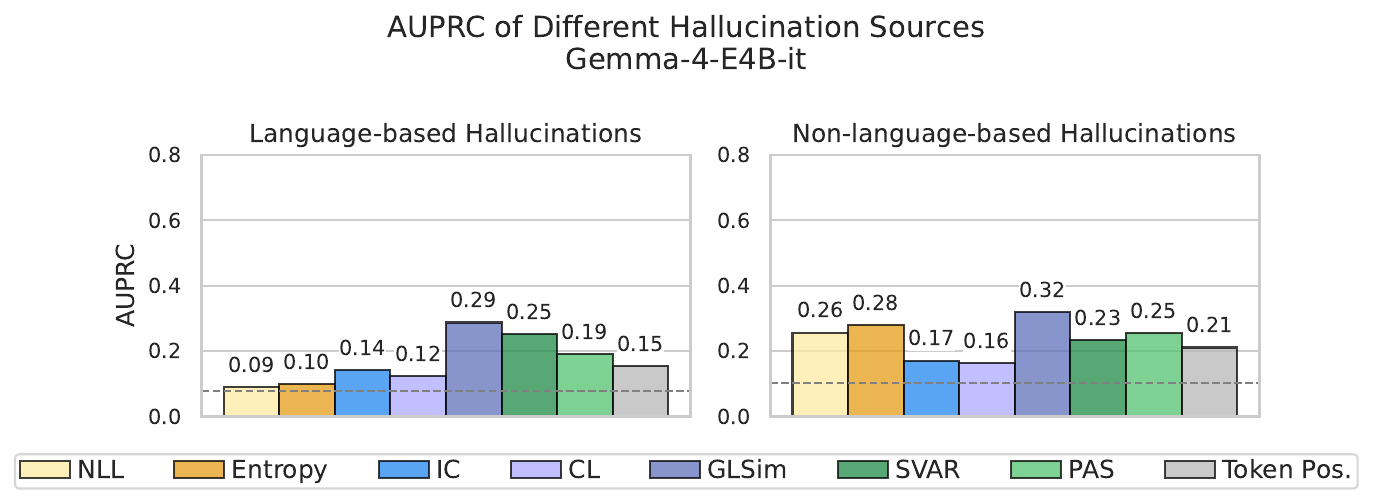}
    \includegraphics[width=0.49\linewidth, trim={0 0.1in 0 0.08in}, clip]{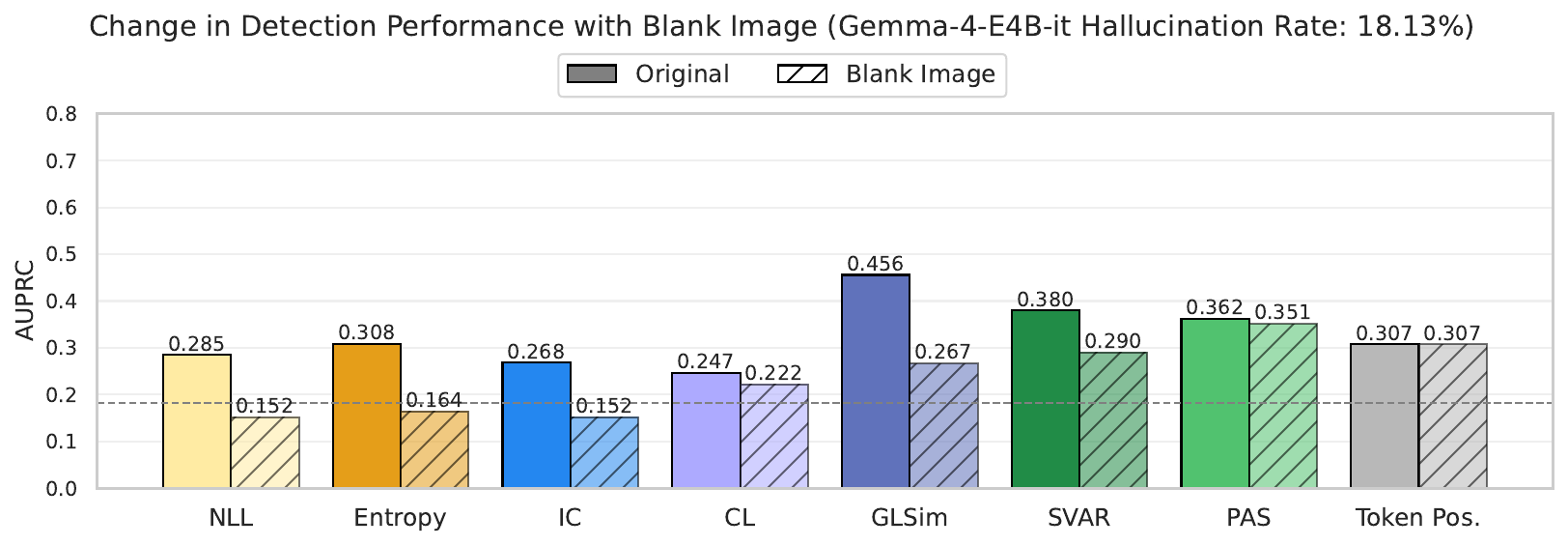}
    \caption{Method behavior partitioning by hallucination modality and visual groundedness on \gemma \ generalizes across various data and prompt settings.}
    \label{fig:gemma4_perf_generalize2}
\end{figure}

In order to demonstrate the robustness of our results, we also run our primary experiments on a different dataset, Objects365 \citep{shao_objects365_2019}. 
Similarly to the experiments in the main paper, we evaluate hallucination detection methods across 5,000 randomly selected images in the validation set. 
Standard evaluation metrics can be found in \tabref{tab:o365_aucs} and token position distributions in \figref{fig:o365_token_pos_dist}.
Results comparing overall performance, token position dependency, and image sensitivity between evaluating on MSCOCO and Objects365 can be found on a per-model basis in \figref{fig:llava13b_perf_generalize1}-\figref{fig:gemma4_perf_generalize2}.

We observe that with Objects365 data, the pattern of hallucinations slightly differs from that in MSCOCO. 
For example, \tabref{tab:o365_aucs} demonstrates that VLMs tend to hallucinate at a higher rate and \figref{fig:o365_token_pos_dist} demonstrates for some VLMs, primarily \qwen \ and \internvl, hallucinations do not occur as reliably at later positions. 
However, despite these differences, we see that methods exhibit similar relative performances between MSCOCO and Objects365. 
Across datasets, methods also exhibit similar reliance to token position and image sensitivity, as demonstrated for each VLM separately in \figref{fig:llava13b_perf_generalize1}-\figref{fig:gemma4_perf_generalize2}. \looseness=-1

\begin{table}
  \caption{\textbf{Token position remains competitive on a different dataset (Objects365).} 
  Compared to MSCOCO, hallucination rates tend to be higher and black box methods (\nll, \ent) tend to perform relatively better on Objects365. 
  Token position (or normalized token position) is still competitive, outperforming at least one other white box method for all VLMs on AUPRC, except for \qwen. 
  We report AUROC and AUPRC of hallucination detection methods across the six main VLMs on 5,000 images from Objects365. \looseness=-1
  }
  \centering
  \resizebox{\textwidth}{!}{
  \begin{tabular}{ll c cc ccccc cc}
    \toprule
    \multirow{2}{*}{\textbf{Metric}} & \multirow{2}{*}{\textbf{Model}} & & \multicolumn{2}{c}{\textbf{Black Box}} & \multicolumn{5}{c}{\textbf{White Box}} & \multicolumn{2}{c}{\textbf{Heuristic}} \\
    \cmidrule(lr){4-5} \cmidrule(lr){6-10} \cmidrule(lr){11-12}
    & & Random & NLL & Entropy & IC & CL & GLSim & SVAR & PAS & Token Pos. & \makecell{Token Pos.\\(norm.)} \\

    \midrule
    \multirow{7}{*}{AUROC} 
    & \llavalg        & \multirow{6}{*}{50.0} & 69.0 & 69.4 & 64.2 & 71.0 & \textbf{74.6} & 70.0 & 72.2 & 72.1 & \underline{72.7} \\
    & \llavasm        & & 69.8 & 69.9 & 67.0 & 70.5 & \textbf{75.4} & 72.1 & \underline{73.9} & 72.9 & 73.4 \\
    & InstructBLIP-7B & & 73.1 & 73.8 & 72.9 & 66.7 & 74.9 & \underline{76.1} & \textbf{76.3} & 75.8 & 75.9 \\
    & Qwen3VL-8B      & & 56.0 & 56.7 & \underline{68.8} & 60.4 & 63.1 & \textbf{71.5} & 64.2 & 56.8 & 57.0 \\
    & \internvl       & & 57.7 & 58.4 & \textbf{69.3} & 61.6 & 68.9 & \underline{69.1} & 66.1 & 62.3 & 63.1 \\
    & \gemma & & 57.8 & 58.1 & 59.0 & 40.3 & \textbf{71.6} & \underline{61.9} & 58.8 & 56.2 & 57.0 \\

    \cmidrule{2-12}
    & \textit{Average} & \textit{50.0} & \textit{63.9} & \textit{64.4} & \textit{66.9} & \textit{61.8} & \textit{\textbf{71.4}} & \textit{\underline{70.1}} & \textit{68.6} & \textit{66.0} & \textit{66.5}\\

    \midrule
    \multirow{6}{*}{AUPRC} 
    & \llavalg        & 36.8 & 55.6 & 55.5 & 50.8 & 56.1 & \textbf{62.3} & 54.0 & 55.9 & 55.9 & \underline{58.1} \\
    & \llavasm        & 39.2 & 58.5 & 57.7 & 58.2 & 60.8 & \textbf{66.2} & 58.9 & 61.1 & 59.2 & \underline{61.7} \\
    & InstructBLIP-7B & 34.4 & 58.9 & 59.2 & \textbf{61.8} & 50.4 & 59.3 & 59.8 & \underline{60.3} & 58.5 & 59.9 \\
    & Qwen3VL-8B      & 43.7 & 49.7 & 50.8 & \underline{61.5} & 53.7 & 53.6 & \textbf{67.7} & 60.1 & 48.4 & 48.8 \\
    & \internvl       & 37.1 & 45.5 & 46.3 & \textbf{57.6} & 48.5 & 56.7 & \underline{57.1} & 50.6 & 47.0 & 48.7 \\
    & \gemma        & 41.3 & 48.3 & 49.1 & 49.4 & 37.3 & \textbf{66.1} & \underline{52.3} & 48.7 & 46.2 & 48.6 \\

    \cmidrule{2-12}
    & \textit{Average} & \textit{38.8} & \textit{52.8} & \textit{53.1} & \textit{56.6} & \textit{51.1} & \textit{\textbf{60.7}} & \textit{\underline{58.3}} & \textit{56.1} & \textit{52.5} & \textit{54.3}\\
    \bottomrule
  \end{tabular}
  }
  \label{tab:o365_aucs}
  \vspace{0.1in}
\end{table}


\subsection{Prompt variation}\label{app:prompt_variations}

To ensure our results are not dependent on specific prompts for image captioning, we additionally prompt the model with a variation to describe the image. 
Concretely, we prompt the model to ``Describe this image in detail, beginning with the most obvious objects first, then smaller background details.''
As shown in \tabref{tab:prompt_var_aucs} and \figref{fig:llava13b_perf_generalize1}-\figref{fig:gemma4_perf_generalize2}, token position is still a competitive baseline, performing on par with or even outperforming at least one white box method on AUPRC. 
We note that performance of \llavasm \ is very similar in this setting and the original, indicating that this VLM is not very sensitive to our prompt differences.
Further, as demonstrated in these figures, trends of token position dependency and image sensitivity are consistent for a given VLM.
This suggests our results are not due to specific prompt wording, but similar prompts for this task would produce similar results.

\begin{table}
  \caption{Token position remains competitive with prompt variations.}
  \centering
  \resizebox{\textwidth}{!}{
  \begin{tabular}{ll c cc ccccc cc}
    \toprule
    \multirow{2}{*}{\textbf{Metric}} & \multirow{2}{*}{\textbf{Model}} & & \multicolumn{2}{c}{\textbf{Black Box}} & \multicolumn{5}{c}{\textbf{White Box}} & \multicolumn{2}{c}{\textbf{Heuristic}} \\
    \cmidrule(lr){4-5} \cmidrule(lr){6-10} \cmidrule(lr){11-12}
    & & Random & NLL & Entropy & IC & CL & GLSim & SVAR & PAS & Token Pos. & \makecell{Token Pos.\\(norm.)} \\

    \midrule
    \multirow{6}{*}{AUROC} 
    & \llavalg        & \multirow{5}{*}{50.0} & 74.4 & 76.1 & 75.2 & 82.8 & 83.4 & 81.5 & \underline{84.4} & 84.0 & \textbf{85.7} \\
    & \llavasm        & & 72.4 & 73.4 & 78.6 & 79.3 & 83.5 & 81.1 & \underline{84.0} & 83.4 & \textbf{84.7} \\
    & InstructBLIP-7B & & 79.9 & 81.4 & 86.2 & 84.3 & 82.1 & \underline{87.5} & \textbf{87.8} & 87.4 & 87.4 \\
    & Qwen3VL-8B      & & 66.7 & 68.8 & 66.0 & 71.0 & 64.0 & \textbf{78.9} & \underline{78.4} & 77.1 & 77.5 \\
    & \internvl       & & 65.2 & 66.5 & 76.2 & 70.0 & 79.8 & \textbf{81.0} & 79.3 & 78.7 & \underline{80.0} \\
    & \gemma       & & 65.9 & 67.8 & 70.5 & 53.3 & \textbf{79.3} & 75.0 & \underline{76.3} & 75.1 & 75.5 \\

    \cmidrule{2-12}
    & \textit{Average} & \textit{50.0} & \textit{70.8} & \textit{72.3} & \textit{75.5} & \textit{73.5} & \textit{78.7} & \textit{80.8} & \textit{\underline{81.7}} & \textit{80.9} & \textit{\textbf{81.8}}\\  

    \midrule
    \multirow{6}{*}{AUPRC} 
    & \llavalg        & 25.7 & 52.0 & 52.4 & 54.3 & 58.1 & \underline{58.2} & 52.5 & 57.0 & 56.0 & \textbf{61.6} \\
    & \llavasm        & 25.1 & 47.2 & 46.4 & 57.5 & 55.5 & \textbf{60.7} & 51.9 & 56.5 & 54.7 & \underline{59.8} \\
    & InstructBLIP-7B & 21.3 & 54.7 & 55.8 & \textbf{66.8} & 53.6 & 54.9 & 60.7 & \underline{60.8} & 59.4 & 60.2 \\
    & Qwen3VL-8B      & 22.8 & 37.1 & 40.5 & 37.1 & 40.6 & 32.4 & \textbf{53.2} & \underline{49.1} & 44.6 & 46.4 \\
    & \internvl       & 20.3 & 33.2 & 33.9 & 43.7 & 38.1 & \underline{48.9} & \textbf{50.8} & 42.5 & 42.3 & 46.6 \\
    & \gemma       & 15.7 & 27.6 & 30.1 & 27.2 & 18.5 & \textbf{39.8} & 31.4 & \underline{34.1} & 31.8 & 33.7 \\
    
    \cmidrule{2-12}
    & \textit{Average} & \textit{21.8} & \textit{42.0} & \textit{43.2} & \textit{47.8} & \textit{44.1} & \textit{49.2} & \textit{\underline{50.1}} & \textit{50.0} & \textit{48.1} & \textit{\textbf{51.4}}\\    \bottomrule
  \end{tabular}
  }
  \label{tab:prompt_var_aucs}
  \vspace{0.1in}
\end{table}

\subsection{Repeated trials}\label{app:repeated_trials}

In our main paper, we use the convention of using greedy decoding, i.e., no sampling, when evaluating hallucination detection methods, such as in \citet{park_glsim_2025}. 
However, this prevents us from being able to run repeated trials, since outputs are deterministic with greedy decoding. 
To understand the robustness of our results to different generation outputs, we enable sampling with \texttt{temperature=1.0} and re-run our experiments with 3 random seeds. Here, we present the results with 95\% confidence intervals. Results for detection performance, dependence on token position, and sensitivity can be found in \figref{fig:llava13b_perf_generalize1}-\figref{fig:gemma4_perf_generalize2}, enabling comparisons between all generalization settings for a specific experiment. 
We note that we are not able to reproduce \figref{fig:partition} with repeated trials and sampling. This is because the partitioning function requires looking at the change in predicted token using argmax, but when sampling is enabled the argmax is no longer indicative of the predicted token.

First, our results demonstrate insensitivity to sampling with a moderate temperature, as trends between the MSCOCO results without repeated trials and experiments with repeated trials exhibit the same trends.
Second, our results also demonstrate robustness to different random seeds for the VLM, demonstrated by the small 95\% confidence intervals.

\subsection{Filtering non-objects}\label{app:non_objects}

Given the limitations of CHAIR outlined in \secref{app:chair_lims}, it could be a concern that the findings are primarily driven by the noise caused by the primitive labeling technique. 
Although our main results filter systematic issues we observed (i.e., filtering the word ``orange''), there can still be noisy labels.
To ensure our results are not dependent on the label noise, we conduct an additional experiment using a context-aware language model to help label objects as hallucinated or real.
Concretely, given a caption and the objects labeled by CHAIR, we pass this information to a sophisticated language model \texttt{Claude-sonnet-4.6} to filter out any non-objects. 
Concretely, we prompt Claude with the following:

\noindent 
\fbox{
    \parbox{\textwidth}{
        Claude prompt for filtering out non-objects: \\
        \texttt{Given a list of words and a caption to an image, your task is to identify whether the words are being used to refer to objects in the image or not. You do not have access to the image. You will output a structured JSON where the keys are the words and the values are boolean (true/false). Do not output any other text so that the outputs can be directly converted into Python dictionaries. Make sure the true/false always start with LOWERCASE letters. You should ensure that the length of the JSON output is the same length as the list of words. \\ \\
        Here is an example of an input:
        \\
        Caption: "There is an apple in the image. There is an orange car with a phone number on it. It could be used by a person because there is a window."
        \\
        Words: [``apple”, “orange”, ``car'', “phone”, “person”, “window”]
        \\ \\
        In this case, orange is not referring to an object, but the color, phone is used in the context of phone number, and person is referring to a hypothetical person, not one in the image. \\
        The output should then be: \\
        \{``apple”: true, “orange”: false, “car”: true, “phone”: false, “person”: false, “window”: true\} \\ \\
        Caption: [new caption] \\
        Word: [list of words]
        }
    }
}

Then we re-evaluate all hallucination detection methods on the objects that were \emph{not} filtered out, and compare how our results on overall performance and token position dependencies differ. 
As demonstrated in \figref{fig:llava13b_perf_generalize1}-\figref{fig:gemma4_perf_generalize2}, for each VLM, even when we filter out non-objects using Claude, the relative performance across methods and the behavior in response to token position interventions  and image interventions remains consistent.
This is the case even when 10-15\% of objects are filtered, as is the case for \qwen, \internvl, and \gemma.
This suggests although CHAIR is limited in its labeling noise, our findings are not contingent on this noise.

\subsection{Extended results without filtering for ``orange''}\label{app:orange_results}

As mentioned in \secref{app:chair_lims}, one limitation of CHAIR is that it does not take into account contextual information, so that there are some objects that are falsely labeled as hallucinations. 
For example, \qwen \ tends to describe the colors in many images, so it frequently mentions the color ``orange'' in generated captions. 
However, these instances are flagged as hallucinations by CHAIR, because oftentimes there is no orange (the fruit) in the image.

Our main results omit the object ``orange'' from all analyses, but here we demonstrate that even with ``orange'' included, the main takeaways do not change. 
We reproduce overall performance, token position dependency (i.e., remove/replace experiment), and image sensitivity (i.e., blank image) reported in the main paper with all VLMs. 
We can compare performance and behavior to main setting with MSCOCO as well as all other generalization settings in \figref{fig:llava13b_perf_generalize1}-\figref{fig:gemma4_perf_generalize2}.

\section{Hallucination mitigation} \label{app:mitigation}

A complementary approach to detecting hallucinations is mitigating them.
Common mitigation techniques include fine-tuning on optimized data \citep{liu_mitigating_2024, you_ferret_2024, yang_understanding_2024, gunjal_detecting_2024, lyu_alleviating_nodate}, improving vision representations \citep{chen_internvl_2024, jain_vcoder_2024, wang_diving_nodate}, intervening during generation \citep{huang_opera_2024, leng_mitigating_2024, wang_mitigating_2024, yang_understanding_2024, favero_multimodal_2024, seo_epistemic_2025}, or leveraging more sophisticated models to correct mistakes \citep{zhou_analyzing_2024, yin_woodpecker_2024}. 
However, because hallucinations are unlikely to ever be completely mitigated \citep{banerjee_llms_2025}, it will always be important to develop effective hallucination detection methods to filter or flag responses that are likely to be hallucinated.

\section{LLM usage declaration} \label{app:llm_usage}

In this work, LLMs contribute a small component of the core methods in this research. 
As stated in the Appendix, we used Gemini to create sentences for the replacement experiment and Claude to conduct context-aware filtering. 
These tasks could have been completed by the researchers, but using LLMs made it quicker and allowed for larger scale experiments. 

LLMs were additionally used to help write code for figure making, data processing, and implementing experiments, which the researchers manually verified for correctness and alignment of intentions. 
We additionally used it to help with grammar, and rewriting small amounts (no more than a few sentences) for clarity of thought.
LLMs were also used as a more sophisticated search engine to find related works, which we manually checked and verified were real. We did not use LLMs to directly input citations.

\newpage
\clearpage 
\section*{NeurIPS Paper Checklist}

\begin{enumerate}

\item {\bf Claims}
    \item[] Question: Do the main claims made in the abstract and introduction accurately reflect the paper's contributions and scope?
    \item[] Answer: \answerYes{} 
    \item[] Justification: The abstract and introduction (last paragraph) clearly state the three contributions of our analyses, and each contribution maps to a specific analysis. The introduction clearly describes the scope of our experiments, such as which detection methods we evaluate.  

\item {\bf Limitations}
    \item[] Question: Does the paper discuss the limitations of the work performed by the authors?
    \item[] Answer: \answerYes{} 
    \item[] Justification: The limitations of our work are expressed explicitly in \secref{app:limitations}. We acknowledge that our analyses only considers a subset of hallucination detection methods, a handful of VLMs, and a single dataset. 

\item {\bf Theory assumptions and proofs}
    \item[] Question: For each theoretical result, does the paper provide the full set of assumptions and a complete (and correct) proof?
    \item[] Answer: \answerNA{} 
    \item[] Justification: Our work does not include theoretical results. 

    \item {\bf Experimental result reproducibility}
    \item[] Question: Does the paper fully disclose all the information needed to reproduce the main experimental results of the paper to the extent that it affects the main claims and/or conclusions of the paper (regardless of whether the code and data are provided or not)?
    \item[] Answer: \answerYes{} 
    \item[] Justification: Our paper includes general information needed to reproduce experimental results in \secref{sec:exp_setting} and \secref{app:hparams}. We include experiment-specific details in \secref{app:exp_details_results}. Code is publicly available at \url{https://github.com/princetonvisualai/hallucination-detection-robustness} 

\item {\bf Open access to data and code}
    \item[] Question: Does the paper provide open access to the data and code, with sufficient instructions to faithfully reproduce the main experimental results, as described in supplemental material?
    \item[] Answer: \answerYes{} 
    \item[] Justification: Code will be publicly available at \url{https://github.com/princetonvisualai/hallucination-detection-robustness}. 

\item {\bf Experimental setting/details}
    \item[] Question: Does the paper specify all the training and test details (e.g., data splits, hyperparameters, how they were chosen, type of optimizer) necessary to understand the results?
    \item[] Answer: \answerYes{} 
    \item[] Justification: We include all analysis details in \secref{app:hparams} and \secref{app:exp_details_results}, such as hyperparameters for caption generation and detection methods and splits for the logistic regression experiments.

\item {\bf Experiment statistical significance}
    \item[] Question: Does the paper report error bars suitably and correctly defined or other appropriate information about the statistical significance of the experiments?
    \item[] Answer: \answerYes{} 
    \item[] Justification: We report error and statistical significance when available. For example, we report statistical significance for all reported correlations (\secref{sec:token_pos_baseline}), error as 95\% confidence intervals (e.g., \figref{fig:remove_replace}) for mean score values, and bootstrapped 95\% confidence intervals for ensemble results. 
    Lastly, we demonstrate the robustness of our resuls in the Appendix, including sampling with multiple seeds (3).

\item {\bf Experiments compute resources}
    \item[] Question: For each experiment, does the paper provide sufficient information on the computer resources (type of compute workers, memory, time of execution) needed to reproduce the experiments?
    \item[] Answer: \answerYes{} 
    \item[] Justification: We report all compute details in \secref{app:hparams} including GPU and CPU usage, execution time, and cluster details. 
    
\item {\bf Code of ethics}
    \item[] Question: Does the research conducted in the paper conform, in every respect, with the NeurIPS Code of Ethics \url{https://neurips.cc/public/EthicsGuidelines}?
    \item[] Answer: \answerYes{} 
    \item[] Justification: Our work conforms with the NeurIPS Code of Ethics in all respects. We do not involve human subjects or participants, and we use only publicly available datasets. 
    We do not foresee any negative societal impact concerns of our results as it aims to reveal important limitations that current evaluation measures do not consider, and it is not a system to be deployed.  

\item {\bf Broader impacts}
    \item[] Question: Does the paper discuss both potential positive societal impacts and negative societal impacts of the work performed?
    \item[] Answer: \answerYes{} 
    \item[] Justification: We discuss the potential positive societal impacts of our analyses throughout the paper, such as encouraging more robust and in-depth evaluations of hallucination detection methods. We also discuss potential negative societal impacts in \secref{app:potential_impacts}, but we do not foresee large negative societal impacts as we are not releasing public models, data, or other artifacts. 
    
\item {\bf Safeguards}
    \item[] Question: Does the paper describe safeguards that have been put in place for responsible release of data or models that have a high risk for misuse (e.g., pre-trained language models, image generators, or scraped datasets)?
    \item[] Answer: \answerNA{} 
    \item[] Justification: The work does not release data or models that require safeguards against misuse.

\item {\bf Licenses for existing assets}
    \item[] Question: Are the creators or original owners of assets (e.g., code, data, models), used in the paper, properly credited and are the license and terms of use explicitly mentioned and properly respected?
    \item[] Answer: \answerYes{} 
    \item[] Justification: We use existing frameworks, models, and datasets which we appropriately credit in \secref{sec:exp_setting}. We include URLs for all resources accessed in \secref{app:hparams}.

\item {\bf New assets}
    \item[] Question: Are new assets introduced in the paper well documented and is the documentation provided alongside the assets?
    \item[] Answer: \answerNA{} 
    \item[] Justification:  We do not introduce any new assets. 

\item {\bf Crowdsourcing and research with human subjects}
    \item[] Question: For crowdsourcing experiments and research with human subjects, does the paper include the full text of instructions given to participants and screenshots, if applicable, as well as details about compensation (if any)? 
    \item[] Answer: \answerNA{} 
    \item[] Justification:  We do not perform any crowdsourcing or experiments with human subjects. 

\item {\bf Institutional review board (IRB) approvals or equivalent for research with human subjects}
    \item[] Question: Does the paper describe potential risks incurred by study participants, whether such risks were disclosed to the subjects, and whether Institutional Review Board (IRB) approvals (or an equivalent approval/review based on the requirements of your country or institution) were obtained?
    \item[] Answer: \answerNA{} 
    \item[] Justification: We do not perform any crowdsourcing or experiments with human subjects. 

\item {\bf Declaration of LLM usage}
    \item[] Question: Does the paper describe the usage of LLMs if it is an important, original, or non-standard component of the core methods in this research? Note that if the LLM is used only for writing, editing, or formatting purposes and does \emph{not} impact the core methodology, scientific rigor, or originality of the research, declaration is not required.
    \item[] Answer: \answerYes{} 
    \item[] Justification: Our work minimally uses LLMs for small tasks, but it does not involve LLMs as important, original, or non-standard components. Regardless, a further description of how LLMs were used can be found in \secref{app:llm_usage}.

\end{enumerate}
\end{document}